\documentclass[final]{cvpr}

\usepackage{times}
\usepackage{epsfig}
\usepackage{graphicx}
\usepackage{amsmath}
\usepackage{amssymb}
\usepackage{booktabs}
\usepackage{subcaption}
\usepackage{float}

\usepackage[pagebackref=true,breaklinks=true,colorlinks,bookmarks=false]{hyperref}

\newcommand{\bC}{\mathbf{C}}

\newcommand{\bF}{\mathbf{F}}

\newcommand{\bn}{\mathbf{n}}

\newcommand{\bP}{\mathbf{P}}

\newcommand{\bv}{\mathbf{v}}

\newcommand{\bx}{\mathbf{x}}
\newcommand{\by}{\mathbf{y}}

 \newcommand{\btheta}{\boldsymbol{\theta}}

\newcommand{\cF}{\mathcal{F}}

\newcommand{\cL}{\mathcal{L}}

\newcommand{\cN}{\mathcal{N}}

\newcommand{\cV}{\mathcal{V}}

\newcommand{\cX}{\mathcal{X}}
\newcommand{\cY}{\mathcal{Y}}

\usepackage{amsfonts}

\makeatletter
\DeclareRobustCommand\onedot{\futurelet\@let@token\@onedot}
\def\@onedot{\ifx\@let@token.\else.\null\fi\xspace}
\def\eg{e.g\onedot} 
\def\ie{i.e\onedot} 
 
\def\etc{etc\onedot}

\def\wrt{wrt\onedot}

\def\etal{et~al\onedot}

\makeatother

\newcommand{\boldparagraph}[1]{\vspace{0.2cm}\noindent{\bf #1:} }

\usepackage{xcolor}
\definecolor{darkred}{rgb}{0.7,0.2,0.1}
\definecolor{darkgreen}{rgb}{0,0.7,0}
\definecolor{orange}{RGB}{255,127,0}
\definecolor{ourpurple}{RGB}{127,127,204}
\definecolor{palgreen}{RGB}{51,179,179}
\definecolor{magenta}{RGB}{199,21,133}

\begin{document}

\title{Neural Parts: Learning Expressive 3D Shape Abstractions \\
with Invertible Neural Networks}

\author{Despoina Paschalidou$^{1,5,6}$ \quad Angelos Katharopoulos$^{3,4}$ \quad Andreas Geiger$^{1,2,5}$ \quad Sanja Fidler$^{6,7,8}$\\
$^1$Max Planck Institute for Intelligent Systems T{\"u}bingen\quad
$^2$University of T{\"u}bingen\\
$^3$Idiap Research Institute, Switzerland\quad
$^4${\'E}cole Polytechique F{\'e}d{\'e}rale de Lausanne (EPFL)\\
$^5$Max Planck ETH Center for Learning Systems\quad
$^6$NVIDIA\quad
$^7$University of Toronto\quad
$^8$Vector Institute \\
{\tt\small \{firstname.lastname\}@tue.mpg.de \quad angelos.katharopoulos@idiap.ch \quad sfidler@nvidia.com}}

\maketitle

\begin{abstract}
Impressive progress in 3D shape extraction led to representations that can capture object geometries with high fidelity.
In parallel, primitive-based methods seek to represent objects as semantically consistent part arrangements.
However, due to the simplicity of existing primitive representations, these methods fail to accurately reconstruct 3D shapes using a small number of primitives/parts.
We address the trade-off between reconstruction quality and number of parts with Neural Parts, a novel 3D primitive representation that defines primitives using an Invertible Neural Network (INN) which implements homeomorphic mappings between a sphere and the target object.
The INN allows us to compute the inverse mapping of the homeomorphism, which in turn, enables the efficient computation of both the implicit surface function of a primitive and its mesh, without any additional post-processing.
Our model learns to parse 3D objects into semantically consistent part arrangements without any part-level supervision.
Evaluations on ShapeNet, D-FAUST and FreiHAND demonstrate that our primitives can capture complex geometries and thus simultaneously achieve geometrically accurate as well as interpretable reconstructions using an order of magnitude fewer primitives than state-of-the-art shape abstraction methods.
\end{abstract}

\section{Introduction}
\label{sec:intro}
\begin{figure}
    \begin{subfigure}[t]{\columnwidth}
        \includegraphics[width=\textwidth]{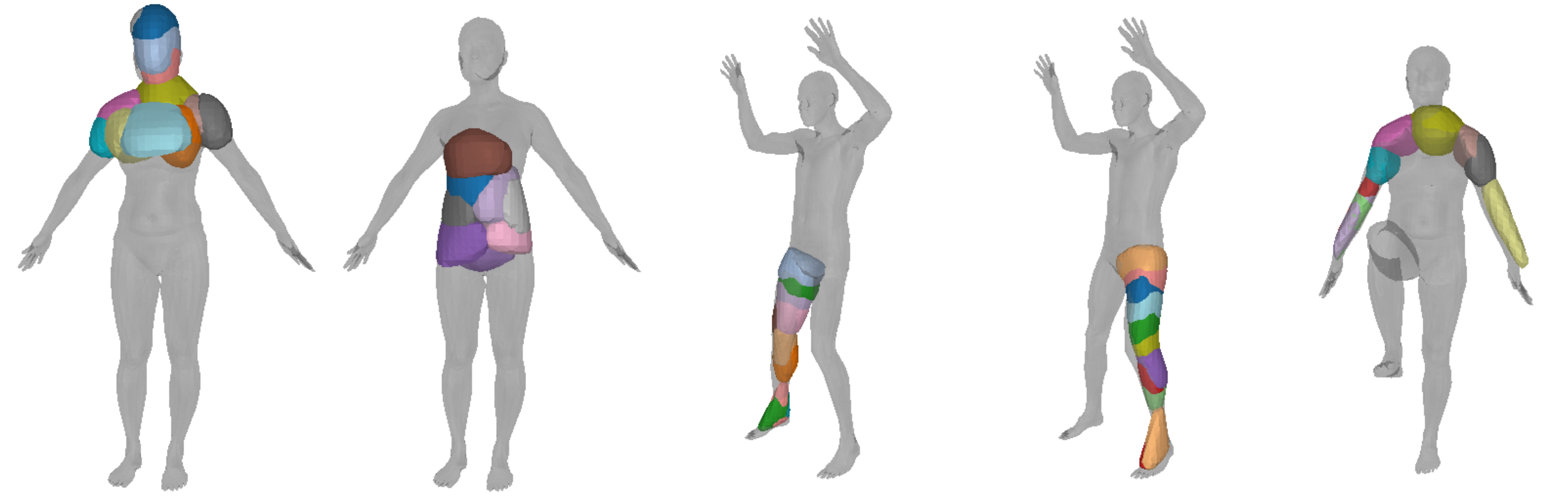}%
        \caption{Convexes}
    \end{subfigure}
    \begin{subfigure}[t]{\columnwidth}
        \includegraphics[width=\textwidth]{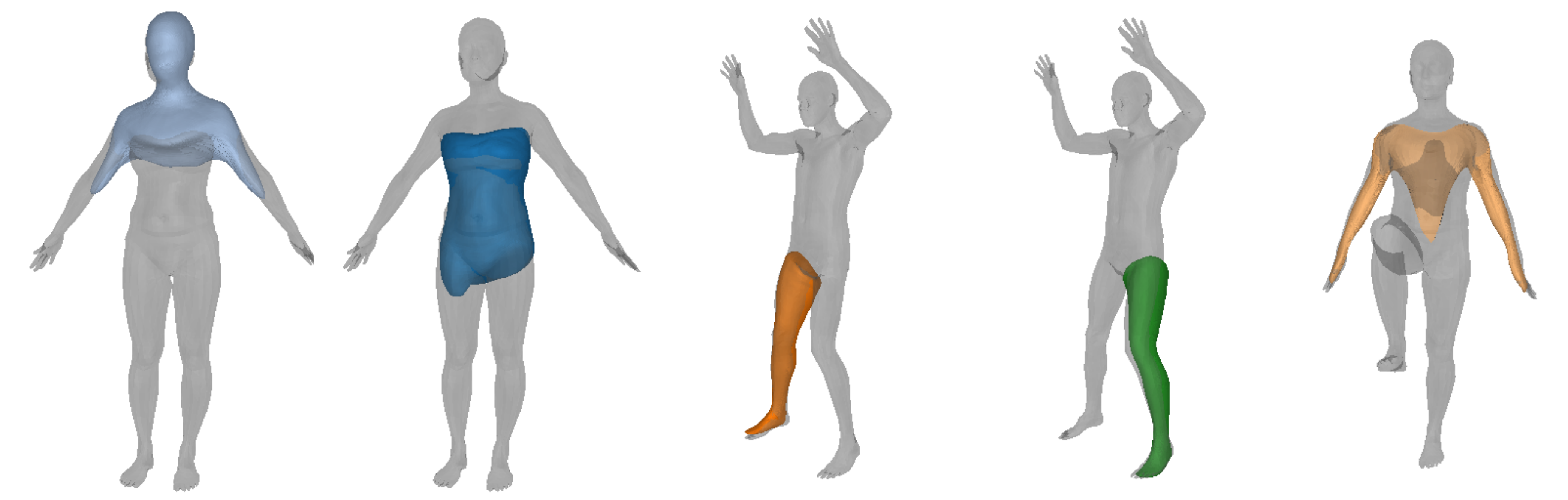}%
        \caption{Neural Parts}
    \end{subfigure}
    \centering
    \caption{{\bf{Expressive Primitives.}} We address the trade-off between reconstruction quality and sparsity (\ie number of parts) in primitive-based methods. Prior work~\cite{Tulsiani2017CVPRa, Paschalidou2019CVPR, Paschalidou2020CVPR, Deng2020CVPR} has considered convex shapes as primitives, which due to their simplicity, require a large number of parts to accurately represent complex shapes. This results in less interpretable shape abstractions (\ie primitives are not identifiable parts e.g. legs, arms \etc). In this work, we propose Neural Parts, a novel 3D primitive representation that can represent arbitrarily complex genus-zero shapes and thus yields geometrically accurate and semantically meaningful parts compared to simpler primitives.  }
    \label{fig:representation_capacity}
    \vspace{-1.2em}
\end{figure}

Recovering the geometry of a 3D shape from a single RGB image is a fundamental task in computer vision and graphics. Existing shape reconstruction models utilize a neural network to learn a parametric function that maps the input image into a mesh \cite{Liao2018CVPR, Groueix2018CVPR, Kanazawa2018ECCV, Wang2018ECCV, Yang2019ICCV, Pan2019ICCV}, a pointcloud \cite{Fan2017CVPR, Qi2017NIPS, Achlioptas2018ICML, Jiang2018ECCV, Thomas2019ICCV, Yang2019ICCV}, a voxel grid \cite{Brock2016ARXIV, Choy2016ECCV, Gadelha2017THREEDV, Rezende2016NIPS, Riegler2017CVPR, Stutz2018CVPR, Xie2019ICCV} or an implicit surface \cite{Mescheder2019CVPR, Chen2019CVPR, Park2019CVPR, Saito2019ICCV, Xu2019NIPS, Michalkiewicz2019ICCV}. An alternative line of research is focused on compact low-dimensional representations that reconstruct 3D objects by decomposing them into simpler parts, called primitives \cite{Tulsiani2017CVPRa, Paschalidou2019CVPR, Deng2020CVPR, Genova2019ICCV}. Primitive-based representations seek to infer \emph{semantically consistent part arrangements} across different object instances and provide a more interpretable alternative, compared to representations that only focus on capturing the global object geometry. Primitives are particularly useful for various applications where the notion of parts is necessary, such as shape editing, physics-based applications, graphics simulation \etc

Existing primitive-based methods rely on simple shapes for decomposing complex 3D objects into parts.
In the early days of computer vision, researchers explored various shape primitives such as 3D polyhedral shapes \cite{Roberts1963}, generalized cylinders \cite{Binford1971Visual} and geons \cite{Biederman1986Human} for representing 3D geometries.
More recently, the primitive paradigm has been revisited in the context of deep learning and \cite{Tulsiani2017CVPRa, Paschalidou2019CVPR, Hao2020CVPR, Deng2020CVPR} have demonstrated the ability of neural networks to learn part-level geometries using 3D cuboids \cite{Tulsiani2017CVPRa, Niu2018CVPR, Zou2017ICCV}, superquadrics \cite{Paschalidou2019CVPR}, spheres \cite{Hao2020CVPR} or convexes \cite{Deng2020CVPR}.
Due to their simple parametrization, these primitives have limited expressivity and cannot capture arbitrarily complex geometries.
Therefore, existing part-based methods require a large number of primitives for extracting geometrically accurate reconstructions.
However, using more primitives comes at the expense of less interpretable reconstructions.

To address this, we devise Neural Parts, a novel 3D primitive representation that is more \emph{expressive} and \emph{interpretable}, in comparison to alternatives that are limited to convex shapes.
We argue that a primitive should be a non trivial genus-zero shape with well defined implicit and explicit representations.
These characteristics allow us to efficiently combine primitives and accurately represent arbitrarily complex geometries.
To this end, we pose the task of primitive learning as the task of learning \emph{a family of homeomorphic mappings} between the 3D space of a simple genus-zero shape (\eg sphere, cube, ellipsoid) and the 3D space of the target object.
We implement this mapping using an Invertible Neural Network (INN) \cite{Dinh2015ICLR}.
Being able to map 3D points in both directions allows us to efficiently compute the explicit representation of each primitive, namely its tesselation as well as the implicit representation, \ie the relative position of a point \wrt the primitive's surface. 
In contrast to prior work~\cite{Tulsiani2017CVPRa, Paschalidou2019CVPR, Deng2020CVPR, Paschalidou2020CVPR} that directly predict the primitive parameters (\ie centroids and sizes for cuboids and superquadrics and hyperplanes for convexes), we employ the INN to fully define each primitive.
Note that while a homeomorphism preserves the genus of the shape it does not constrain it in any other way.
As a result, while existing primitives are constrained to a specific family of shapes (\eg ellipsoids), our primitives can capture arbitrarily complicated genus-zero shapes (see \figref{fig:representation_capacity}).
We demonstrate that Neural Parts can be learned in an unsupervised fashion (\ie without any primitive annotations), directly from unstructured 3D point clouds by ensuring that the assembly of predicted primitives accurately reconstructs the target.

In summary, we make the following \textbf{contributions}: We propose the first model that defines primitives as a homeomorphic mapping between two topological spaces through conditioning an INN on the input image.
Since the homeomorphism does not impose any constraints on the shape of the predicted primitive, our model effectively decouples geometric accuracy from parsimony and as a result captures complex geometries with an order of magnitude fewer primitives.
Experiments on ShapeNet objects \cite{Chang2015ARXIV}, D-FAUST humans \cite{Bogo2017CVPR} and FreiHAND hands \cite{Zimmermann2019ICCV} demonstrate the ability of our model to parse objects into more expressive and semantically meaningful shape abstractions compared to models that rely on simpler primitives \cite{Paschalidou2019CVPR, Paschalidou2020CVPR, Deng2020CVPR}. Code and data is publicly available\footnote{\url{https://paschalidoud.github.io/neural_parts}}.

\section{Related Work}
\label{sec:related}

Learning-based 3D reconstruction approaches can be categorized based on the type of their output representation to: depth-based~\cite{Kar2017NIPS, Hartmann2017ICCV, Paschalidou2018CVPR, Donne2019CVPR}, voxel-based~\cite{Choy2016ECCV, Wu2016NIPS, Gadelha2017THREEDV, Rezende2016NIPS}, point-based~\cite{Fan2017CVPR, Qi2017NIPS, Achlioptas2018ICML}, mesh-based~\cite{Liao2018CVPR, Groueix2018CVPR, Kanazawa2018ECCV}, implicit-based~\cite{Mescheder2019CVPR, Chen2019CVPR, Park2019CVPR} and primitive-based~\cite{Tulsiani2017CVPRa, Niu2018CVPR, Paschalidou2019CVPR}.
Here, we primarily focus on primitive-based methods that are more relevant to our work.
Since our formulation is independent of a specific INN implementation, a thorough discussion of INNs is beyond the scope of this paper, thus we refer the reader to \cite{Ardizzone2019ICLR} for a detailed overview. 

\boldparagraph{3D Representations}%
Voxels~\cite{Brock2016ARXIV, Choy2016ECCV, Wu2016NIPS, Gadelha2017THREEDV, Rezende2016NIPS, Stutz2018CVPR, Xie2019ICCV} naturally capture the 3D geometry by discretizing the shape into a regular grid.
While several efficient space partitioning techniques \cite{Maturana2015IROS, Riegler2017CVPR, Tatarchenko2017ICCV, Haene2017ARXIV, Riegler2017THREEDV} have been proposed to address their high memory and computation requirements, their application is still limited. A promising new direction explored learning a deformation of the grid itself to better capture geometric details~\cite{Gao2020NIPS}.
Pointclouds~\cite{Fan2017CVPR, Qi2017NIPS, Achlioptas2018ICML, Jiang2018ECCV, Thomas2019ICCV, Yang2019ICCV} are more memory efficient but lack surface connectivity, thus post-processing is necessary for generating the final mesh. 
Most mesh-based methods~\cite{Liao2018CVPR, Groueix2018CVPR, Kanazawa2018ECCV, Wang2018ECCV, Yang2019ICCV, Gkioxari2019ICCV, Pan2019ICCV, Chen2019NIPS, Deng2020CVPR, Zhang2020ARXIV} naturally yield smooth reconstructions but either require a deformable template mesh~\cite{Wang2018ECCV} or represent the geometry as an atlas of multiple mappings~\cite{Groueix2018CVPR, Deng2020THREEDV, Ma2021CVPR}.
To address these limitations, implicit models~\cite{Mescheder2019CVPR, Chen2019CVPR, Park2019CVPR, Saito2019ICCV, Xu2019NIPS, Michalkiewicz2019ICCV, Niemeyer2019ICCV, Oechsle2019ICCV, Atzmon2019NIPS, Wang2019NIPSa, Genova2019ARXIV, Chen2019ICCV, Niemeyer2020CVPR, Atzmon2020CVPR, Gropp2020ICML, Saito2020CVPR, Jiang2020CVPR, Chibane2020CVPR, Peng2020ECCV, Remelli2020ARXIV, Oechsle2020THREEDV} have recently gained popularity.
These methods represent a 3D shape as the level-set of a distance or occupancy field implemented as a neural network, that takes a context vector and a query point and predicts either a signed distance value \cite{Park2019CVPR, Michalkiewicz2019ICCV, Atzmon2020CVPR, Gropp2020ICML, takikawa2021nglod} or a binary occupancy value \cite{Mescheder2019CVPR, Chen2019CVPR} for the query point.
While these methods result in accurate reconstructions, they lack interpretability as they do not consider the part-based object structure.
Instead, we focus on part-based representations and showcase that our model simultaneously yields geometrically accurate and interpretable reconstructions.
Furthermore, in contrast to implicit models, that require expensive iso-surfacing operations (\ie marching cubes) to extract a mesh, our model directly predicts a high resolution mesh for each part, without any post-processing. 

\boldparagraph{Structured-based Representations}%
Our work falls into the category of shape abstraction techniques.
This line of research seeks to decompose 3D shapes into semantically meaningful simpler parts using either supervision in terms of the primitive parameters \cite{Zou2017ICCV, Niu2018CVPR, Li2017SIGGRAPH, Mo2019SIGGRAPH, Gao2019SIGGRAPHASIA, Li2019CVPR, Sharma2020ECCV, Li2020ECCV} or without any part-level annotations \cite{Tulsiani2017CVPRa, Paschalidou2019CVPR, Deprelle2019NIPS, Paschalidou2020CVPR, Genova2019ICCV, Deng2020CVPR, Kawana2020NIPS}.
Neural Parts perform primitive-based learning in an unsupervised manner.
Traditional primitives include cuboids \cite{Tulsiani2017CVPRa, Niu2018CVPR, Zou2017ICCV, Li2017SIGGRAPH, Mo2019SIGGRAPH, Dubrovina2019ICCV}, superquadrics \cite{Paschalidou2019CVPR, Paschalidou2020CVPR}, convexes \cite{Deng2020CVPR, Chen2020CVPR, Gadelha2020CVPR}, CSG trees \cite{Sharma2018CVPR} or shape programms \cite{Ellis2018NIPS, Tian2019ICLR, Liu2019ICLR}.
Due to the simplicity of the shapes of traditional primitives, the reconstruction quality of existing part-based methods is coupled with the number of primitives, namely a larger number of primitives results in more accurate reconstructions (see \figref{fig:human_body_modelling}).
However, these reconstructions are less parsimonious and the constituent primitives often lack a semantic interpretation (\ie are not recognizable parts).
Instead, Neural Parts are not restricted to convex shapes and can capture complex geometries with a few primitives.
In recent work, \cite{Genova2019ARXIV} propose a 3D representation that decomposes space into a structured set of implicit functions \cite{Genova2019ICCV}. However, extracting a single part from their prediction is not possible. This is not the case for our model.

\boldparagraph{Shape Deformations}%
Deforming a single genus-zero shape into more complicated shapes with graph convolutions~\cite{Wang2018ECCV}, MLPs~\cite{Groueix2018CVPR, Tulsiani2020ARXIV} and Neural ODEs~\cite{Gupta2020ARXIV} has demonstrated impressive results.
Groueix \etal \cite{Groueix2018CVPR} were among the first to employ an MLP to implement a homeomorphism between a sphere and a complicated shape.
Note that since the deformation is implemented via an MLP, computing the inverse mapping becomes infeasible.
Very recently, \cite{Gupta2020ARXIV} proposed to learn the deformation of a single ellipsoid using several Neural ODEs. Similar to our Neural Parts, their model is invertible, however it does not consider any part decomposition or latent object structure.
Instead, we use the inverse mapping of the homeomorphism to define the predicted shape as the union of primitives, by discarding points from a part's surface that are internal to any other part. Thus our model learns to combine multiple genus-zero primitives and is able to reconstruct shapes of arbitrary genus.
In addition, we formulate our losses using the forward and the inverse mapping of the homeomorphism, which in turn, allows us to utilize both volumetric as well as surface information during training.
This facilitates imposing additional constraints on the predicted primitives \eg normal consistency and parsimony and leads to more accurate results as evidenced by our experiments.

\section{Method}
\label{sec:method}

\begin{figure*}
    \centering
    \includegraphics[width=1.0\textwidth]{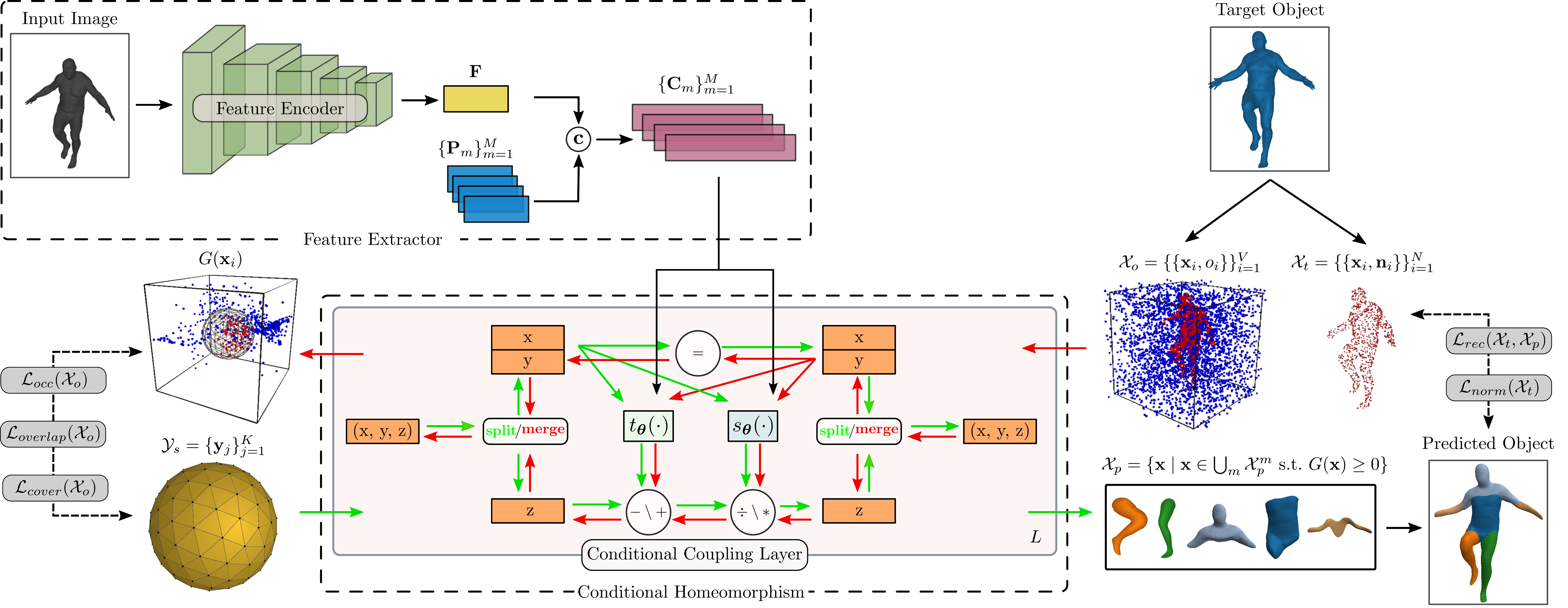}
    \caption{
        {\bf{Method Overview.}}
        Our model comprises two main components: A \emph{Feature Extractor}
        which maps an input image into a per-primitive shape embedding and
        a \emph{Conditional Homeomorphism} that deforms a sphere into $M$
        primitives and vice-versa.
        First, the \emph{feature encoder} maps
        the input to a global feature representation $\bF$.
        Then, for every primitive $m$, $\bF$ is concatenated
        with a learnable primitive embedding $\bP_m$ to generate the
        shape embedding $\bC_m$ for this primitive.
        The Conditional Homeomorphism $\phi_{\btheta}(\cdot; \bC_m)$ is
        implemented by a stack of $L$ conditional coupling layers. Applying the
        {\color{darkgreen}{\emph{forward mapping}}} on a set of points $\cY_s$,
        randomly sampled on the surface of the sphere, generates points on the
        surface of the $m$-th primitive $\cX_p^m$ \eqref{eq:primitive_explicit_full}.
        Using the {\color{red}{\emph{inverse mapping}}}
        $\phi_{\btheta}^{-1}(\cdot; \bC_m)$, allows us to compute whether any
        point in 3D space lies inside or outside a primitive
        \eqref{eq:primitive_implicit_full}.
        We train our model using both \emph{surface} ($\cL_{rec}, \cL_{norm}$) and \emph{occupancy} ($\cL_{occ}$) losses to simultaneously capture fine object details and volumetric characteristics of the target object.
        The use of the {\color{red}{\emph{inverse mapping}}} allows us to impose additional constraints (\eg discouraging inter-penetration) on the predicted primitives ($\cL_{overlap}, \cL_{cover}$).
    }
    \label{fig:method_overview}
    \vspace{-0.5em}
\end{figure*}

Given an input image we seek to learn a representation with $M$ primitives that best describes the target object.
We define our primitives via a \emph{deformation between shapes} that is parametrized as a \emph{learned homeomorphism} implemented with an Invertible Neural Network (INN).
Using an INN allows us to efficiently compute the implicit and explicit representation of the predicted shape and impose various constraints on the predicted parts.

In particular, for each primitive, we seek to learn a homeomorphism between the 3D space of a simple genus-zero shape and the 3D space of the target object, such that the deformed shape matches a part of the target object.
Due to its simple implicit surface definition and tesselation, we employ a sphere as our genus-zero shape.
We refer to the 3D space of the sphere as \emph{latent space} and to the 3D space of the target object as \emph{primitive space}.

In \secref{subsec:primitives}, we present the explicit and implicit
representation of our primitives as a homeomorphism of a sphere.
Subsequently, in \secref{subsec:network}, we present our novel
architecture for predicting multiple primitives using homeomorphisms
conditioned on the input image. Finally, in \secref{subsec:training}, we
formulate our optimization objective.

\subsection{Primitives as Homeomorphic Mappings}
\label{subsec:primitives}

A homeomorphism is a continuous map between two topological spaces $Y$ and $X$ that preserves all topological properties.
Intuitively a homeomorphism is a continuous stretching and bending of $Y$ into a new space $X$.
In our 3D topology, a homeomorphism $\phi_{\btheta}: \mathbb{R}^3 \to \mathbb{R}^3$ is
\begin{equation}
    \bx = \phi_{\btheta}(\by) \,\, \text{and} \,\, \by = \phi_{\btheta}^{-1}(\bx)
    \label{eq:homeomorphism}
\end{equation}
where $\bx$ and $\by$ are 3D points in $X$ and $Y$ and $\displaystyle \phi_{\btheta}:Y \to X$, $\displaystyle \phi_{\btheta}^{-1}: X \to Y$ are continuous bijections.
In our setting, $Y$ and $X$ correspond to the latent and the primitive space respectively.
Using the explicit and implicit representation of a sphere with radius $r$, positioned at $(0,0,0)$ and the homeomorphic mapping from \eqref{eq:homeomorphism}, we can now define the implicit and explicit representation of a single primitive.

\boldparagraph{Explicit Representation}%
The explicit representation of a primitive, parametrized as a mesh with vertices $\cV_p$ and faces $\cF_p$, can be obtained by applying the homeomorphism on the sphere vertices $\cV$ and faces $\cF$ as follows:
\begin{equation}
    \begin{aligned}
        \cV_p &= \{\phi_{\btheta}(\bv_j),\,\, \forall \,\, \bv_j \in \cV\} \\
        \cF_p &= \cF.
    \end{aligned}
    \label{eq:primitive_explicit}
\end{equation}
It is evident that applying $\phi_{\btheta}$ on the sphere vertices $\cV$ alters their location in the primitive space, while the vertex arrangements (\ie faces) remain unchanged.
Since our primitives are defined as a deformation of a sphere mesh of arbitrarily high resolution, we can also obtain primitive meshes of arbitrary resolutions without any post-processing, such as marching-cubes which is required for traditional neural implicit representations \cite{Mescheder2019CVPR}.

\boldparagraph{Implicit Representation}%
The implicit representation of a primitive can be derived by applying the inverse homeomorphic mapping on a 3D point $\bx$ as follows
\begin{equation}
    g(\bx) = {\Vert \phi_{\btheta}^{-1}(\bx) \Vert}_2 - r.
    \label{eq:primitive_implicit}
\end{equation}
To evaluate the relative position of a point $\bx$ \wrt the primitive surface it suffices to evaluate whether $\phi_{\bf\theta}^{-1}(\bx)$ lies inside, outside or on the surface of the sphere.
Namely, points that are internal to the sphere are also inside the primitive and points that are outside the sphere are also outside the primitive surface.
Note that computing $g(\bx)$ in \eqref{eq:primitive_implicit} is only possible because $\phi_{\btheta}(\bx)$ is implemented with an INN.

\boldparagraph{Multiple Primitives}%
The homeomorphism in \eqref{eq:homeomorphism} implements a single deformation. However, we seek to predict multiple primitives (\ie deformations) conditioned on the input. Hence, we define a conditional homeomorphism as:
\begin{equation}
    \bx = \phi_{\btheta}(\by; \bC_m) \,\, \text{and} \,\, \by = \phi_{\btheta}^{-1}(\bx; \bC_m)
    \label{eq:conditional_homeomorphism}
\end{equation}
where $\bC_m$ is the shape embedding for the $m$-th primitive and is predicted from the input. Note that for different shape embeddings a different homeomorphism is defined.

\subsection{Network Architecture}
\label{subsec:network}

Our architecture comprises two main components: (i) the \emph{feature extractor} that maps the input to a vector of per-primitive shape embeddings $\{\bC_m\}_{m=1}^M$ and (ii) the \emph{conditional homeomorphism} that learns a homeomorphic mapping conditioned on the shape embedding.
The overall architecture is illustrated in \figref{fig:method_overview}.

\boldparagraph{Feature Extractor}%
The first part of the feature extractor module is a ResNet-18 \cite{He2016CVPR} that extracts a feature representation $\bF \in \mathbb{R}^D$ from the input image.
Subsequently, for every primitive $m$, $\bF$ is concatenated with a learnable primitive embedding $\bP_m \in \mathbb{R}^D$ to derive a shape embedding $\bC_m \in \mathbb{R}^{2D}$ for this primitive.

\boldparagraph{Conditional Homeomorphism}%
We implement the INN using a Real NVP \cite{Dinh2017ICLR} due to its simple formulation.
A Real NVP models a bijective mapping by stacking a sequence of simple bijective transformation functions.
For each bijection, typically referred to as \emph{affine coupling layer}, given an input 3D point $(x_i, y_i, z_i)$, the output point $(x_o, y_o, z_o)$ is
\begin{equation}
    \begin{aligned}
        x_o &= x_i \\
        y_o &= y_i \\
        z_o &= z_i \exp\left(s_{\btheta}\left(x_i, y_i\right)\right) +
               t_{\btheta}\left(x_i, y_i\right)
    \end{aligned}
    \label{eq:coupling_layer}
\end{equation}
where $s_{\btheta} : \mathbb{R}^2 \to \mathbb{R}$ and $t_{\btheta}:\mathbb{R}^2 \to \mathbb{R}$ are \emph{scale} and \emph{translation} functions implemented with two arbitrarily complicated networks.
Namely, in each bijection, the input is split into two, $(x_i, y_i)$ and $z_i$. The first part remains unchanged and the second undergoes an affine transformation with $s_{\btheta}(\cdot)$ and $t_{\btheta}(\cdot)$.
We follow \cite{Dinh2017ICLR} and we enforce that consecutive affine coupling layers scale and translate different input dimensions. Namely, we alternate the splitting between the dimensions of the input randomly.

\begin{figure}
    \centering
    \includegraphics[width=\columnwidth]{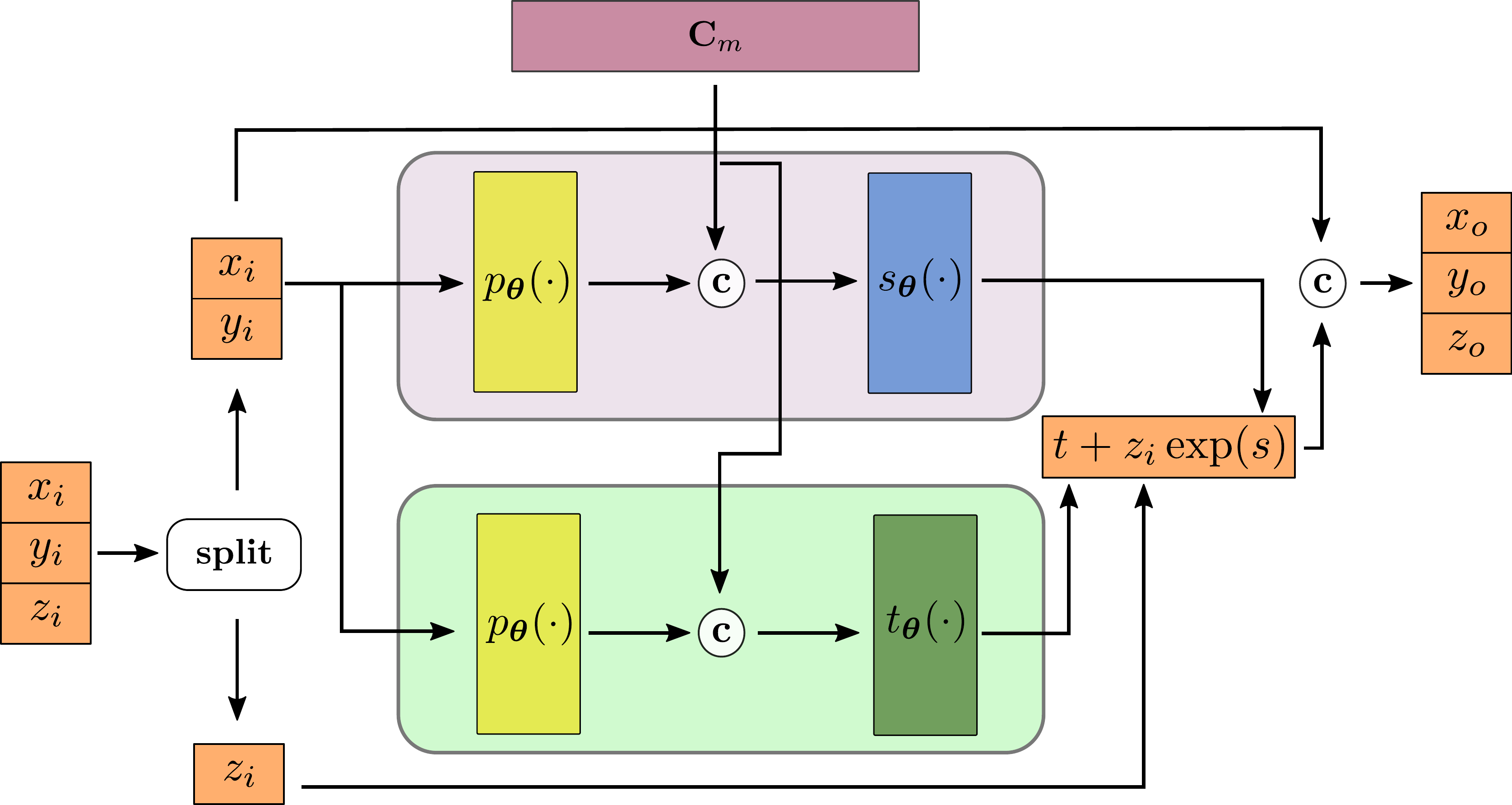}
    \caption{{\bf{Conditional Coupling Layer.}} Pictorial representation of \eqref{eq:conditional_coupling_layer}.
    The input point $(x_i, y_i, z_i)$ is passed into a \emph{coupling layer} that scales and translates one dimension of the input based on the other two and the per-primitive shape embedding $\bC_m$. The scale factor $s$ and the translation amount $t$ are predicted by two MLPs, $s_{\btheta}(\cdot)$ and $t_{\btheta}(\cdot)$. $p_{\btheta}(\cdot)$ is another MLP that increases the dimensionality of the input point before it is concatenated with $\bC_m$.
    }
    \label{fig:invertible_network}
    \vspace{-0.8em}
\end{figure}

However, the original Real NVP cannot be directly applied in our setting as it does not consider a shape embedding.
To address this, we augment the affine coupling layer as follows: we first map $(x_i, y_i)$ into a higher dimensional feature vector using a mapping, $p_{\btheta}(\cdot)$, implemented as an MLP. This is done to increase the relative importance of the input point before concatenating it with the high-dimensional shape embedding $\bC_m$. The \emph{conditional affine coupling layer} becomes
\begin{equation}
    \begin{aligned}
        x_o &= x_i \\
        y_o &= y_i \\
        z_o &= z_i \exp\left(s_{\btheta}\left(
                        \left[\bC_m ; p_{\btheta}\left(x_i, y_i\right)\right]
                    \right)\right) \\
            & \quad + t_{\btheta}\left(
                    \left[\bC_m ; p_{\btheta}\left(x_i, y_i\right)\right]
                 \right)
    \end{aligned}
    \label{eq:conditional_coupling_layer}
\end{equation}
where $[\cdot;\cdot]$ denotes concatenation.
A graphical representation of our conditional coupling layer is provided in \figref{fig:invertible_network}.

\subsection{Training}
\label{subsec:training}

Due to the lack of primitive annotations, we train our model by minimizing the geometric distance between the target and the predicted shape. Our supervision comes from a watertight mesh of the target object.
In the following, we define the implicit and explicit representation of the predicted shape as the union of $M$ primitives.

The implicit surface of the $m$-th primitive can be derived from \eqref{eq:primitive_implicit}, by applying the inverse homeomorphic mapping on points in $3D$ space:
\begin{equation}
    g^m(\bx) = {\Vert \phi_{\btheta}^{-1}(\bx; \bC_m) \Vert}_2 - r,\,\, \forall\,\, \bx \in \mathbb{R}^3
    \label{eq:primitive_implicit_full}
\end{equation}
We define the implicit surface representation of the entire object as the union of all per-primitive implicit functions
\begin{equation}
    G(\bx) = \min_{m \in 0\dots M} g^m(\bx),
    \label{eq:predicted_shape_implicit}
\end{equation}
namely we consider that a point is inside the predicted shape if it is inside at least one primitive.

Similarly, the explicit representation of the $m$-{th} primitive is a set of points on its surface, let it be $\cX_p^m$, that are generated by applying the forward homeomorphic mapping on points on the sphere surface $\cY_s$ in the latent space
\begin{equation}
    \cX_p^m = \{\phi_{\btheta}(\by_j; \bC_m),\,\, \forall \,\, \by_j \in \cY_s\}.
    \label{eq:primitive_explicit_full}
\end{equation}
To generate points on the surface of the predicted shape $\cX_p$, we first need to generate points on the surface of each primitive and then discard the ones that are inside any other primitive. From \eqref{eq:predicted_shape_implicit} this can be expressed as follows:
\begin{equation}
    \cX_p = \{\bx \mid \bx \in \bigcup_{m} \cX_p^m \,\, \text{s.t.} \,\, G(\bx) \geq 0\}.
    \label{eq:predicted_shape_explicit}
\end{equation}

\begin{figure*}
    \centering
    \begin{subfigure}[t]{\textwidth}
        \includegraphics[width=\textwidth]{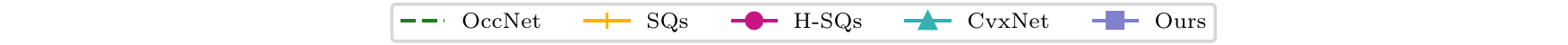}
    \end{subfigure}\\[0.7em]
    \begin{subfigure}[b]{0.5\textwidth}
        \includegraphics[width=\textwidth]{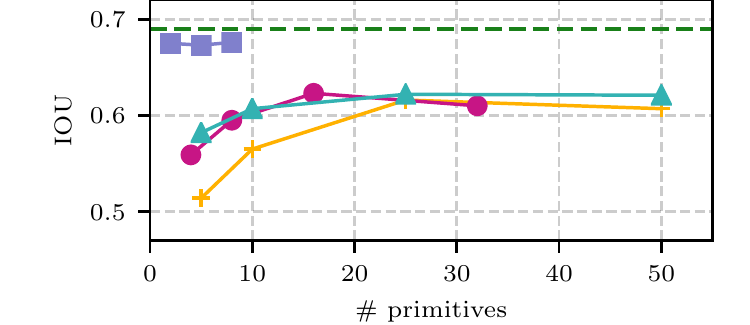}
    \end{subfigure}%
    \begin{subfigure}[b]{0.5\textwidth}
        \includegraphics[width=\textwidth]{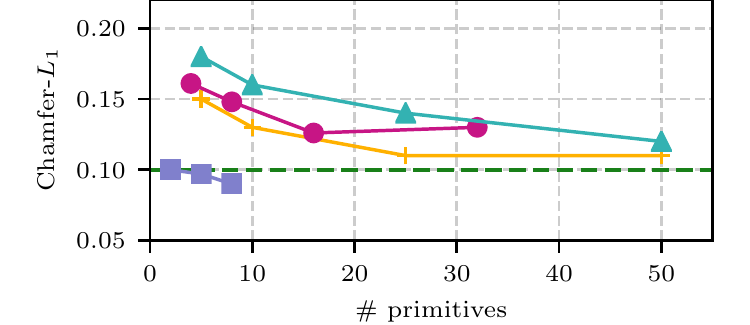}
    \end{subfigure}
    \caption{{\bf{Trade-off Reconstruction Quality and $\#$ Primitives}}. We evaluate the reconstruction quality of primitive-based methods on the D-FAUST test set for different number of primitives. Neural Parts (\textbf{{\color{ourpurple}{purple}}}) outperform CvxNet (\textbf{{\color{palgreen}{turquoise}}}), SQs (\textbf{{\color{orange}{orange}}}) and H-SQs (\textbf{{\color{magenta}{magenta}}}) in terms of both IoU ($\uparrow$) and Chamfer-L1 ($\downarrow$) for any primitive configuration, even when using as little as $2$ primitives. In addition, we show that our reconstructions are competitive to OccNet (dashed) which does not provide a primitive-based representation and requires expensive post-processing for extracting surface meshes.}
    \label{fig:rep_power}
    \vspace{-0.5em}
\end{figure*}

\boldparagraph{Loss Functions}%
Our loss $\cL$ seeks to minimize the geometric distance between the target and the predicted shape and is composed of five loss terms:
\begin{equation}
    \begin{aligned}
        \cL &= \cL_{rec}(\cX_t, \cX_p) +\, \cL_{occ}(\cX_o) +\, \cL_{norm}(\cX_t) \\
            & \quad + \cL_{overlap}(\cX_o) + \cL_{cover}(\cX_o)
    \end{aligned}
    \label{eq:total_loss}
\end{equation}
where $\cX_t = \{ \{\bx_i, \bn_i\}\}_{i=1}^N$ comprises surface samples of the target shape and the corresponding normals, and  
$\cX_o = \{\{\bx_i, o_i\}\}_{i=1}^V$ denotes a set of occupancy pairs, where $\bx_i$ corresponds to the location of the $i$-th point and $o_i$ denotes whether $\bx_i$ lies inside $(o_i=1)$ or outside $(o_i=0)$ the target. 
Note that our optimization objective comprises both occupancy \eqref{eq:occupancy_loss} and surface losses \eqref{eq:reconstruction_loss}$+$\eqref{eq:normal_loss}, since they model complementary characteristics of the target object \eg the surface loss attends to fine details that may have small volume, whereas the occupancy loss more efficiently models empty space.
We empirically observe that using both significantly improves reconstruction (see \secref{subsec:ablations}).

\boldparagraph{Reconstruction Loss}%
We measure the surface reconstruction quality using a bidirectional Chamfer loss between the points $\cX_p$ on the surface of the predicted shape and the points on the target object $\cX_t$ as follows:
\begin{equation}
    \begin{aligned}
    \cL_{rec}(\cX_t, \cX_p) = &\frac{1}{|\cX_t|} \sum_{\bx_i \in \cX_t}
    \min_{\bx_j \in \cX_p} {\| \bx_i - \bx_j\|}_2^2 \,+ \\
    &\frac{1}{|\cX_p|} \sum_{\bx_j \in \cX_p}
    \min_{\bx_i \in \cX_t} {\| \bx_i - \bx_j \|}_2^2
    \end{aligned}
    \label{eq:reconstruction_loss}
\end{equation}
The first term of \eqref{eq:reconstruction_loss} measures the completeness of the predicted shape, namely the average distance of all ground truth points to the closest predicted points. The second term measures its accuracy, namely the average distance of all predicted points to the closest ground-truth points.

\boldparagraph{Occupancy Loss}%
The occupancy loss ensures that the volume of the predicted shape matches the volume of the target.
Intuitively, we want to ensure that the free and the occupied space of the predicted and the target object coincide.
To this end, we convert the implicit surface of the predicted shape from \eqref{eq:predicted_shape_implicit} to an indicator function and compute the binary cross-entropy loss
\begin{equation}
    \cL_{occ}(\cX_o) = \sum_{(\bx, o) \in \cX_o} \cL_{ce}\left(\sigma\left(\frac{-G(\bx)}{\tau}\right), o\right),
    \label{eq:occupancy_loss}
\end{equation}
which we sum over all volume samples $\cX_o$. Here, $\cL_{ce}(\cdot, \cdot)$ is the cross-entropy classification loss, $\sigma(\cdot)$ is the sigmoid function and $\tau$ is a temperature hyperparameter that defines the sharpness of the boundary of the indicator function.
Note that $\sigma\left(\frac{-G(\bx)}{\tau}\right)$ is $1$ when $\bx$ is inside the predicted shape and $0$ otherwise.

\boldparagraph{Normal Consistency Loss}%
The normal consistency loss ensures that the orientation of the normals of the predicted shape will be aligned with the normals of the target. We penalize misalignments between the predicted and the target normals by minimizing the cosine distance as follows
\begin{equation}
    \cL_{norm}(\cX_t) = \frac{1}{|\cX_t|} \sum_{(\bx, \bn) \in \cX_t}
        \left(1 - \left\langle\frac{\nabla_{\bx} G(\bx)}{\|\nabla_{\bx} G(\bx)\|_2}, n\right\rangle\right)
    \label{eq:normal_loss}
\end{equation}
where $\langle\cdot, \cdot \rangle$ is the dot product. Note that the surface normal of the predicted shape for a point $\bx$ is simply the gradient of the implicit surface \wrt to point $\bx$ and can be efficiently computed with automatic differentiation.

\begin{figure}
    \centering
    \begin{minipage}[b]{0.2\linewidth}
        \centering
	    \includegraphics[trim=70 0 70 0,clip,width=\linewidth]{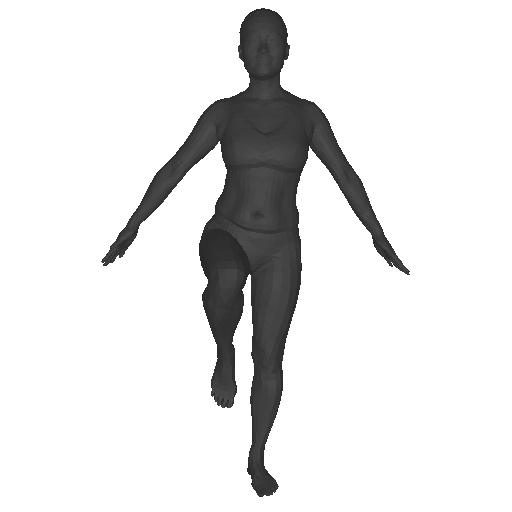}
        \vspace{0.4em}
    \end{minipage}%
    \begin{minipage}[b]{0.8\linewidth}
        \begin{subfigure}[b]{0.2\linewidth}
	    	\centering
	        \includegraphics[trim=90 0 90 0,clip,width=\linewidth]{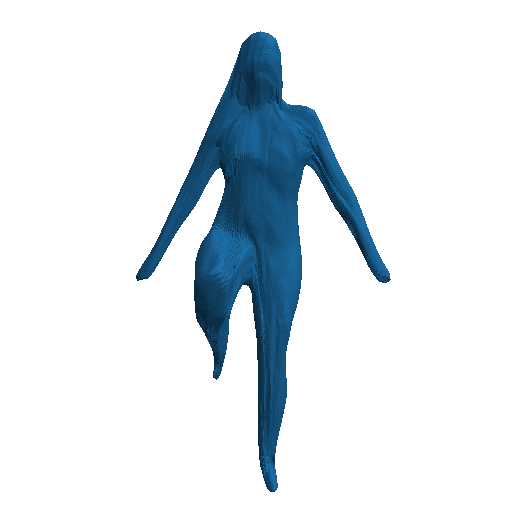}
        \end{subfigure}%
        \begin{subfigure}[b]{0.2\linewidth}
	    	\centering
	        \includegraphics[trim=90 0 90 0,clip,width=\linewidth]{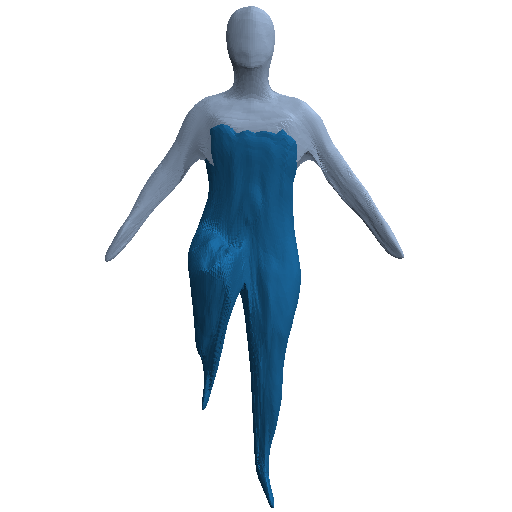}
        \end{subfigure}%
        \begin{subfigure}[b]{0.2\linewidth}
	    	\centering
	        \includegraphics[trim=90 0 90 0,clip,width=\linewidth]{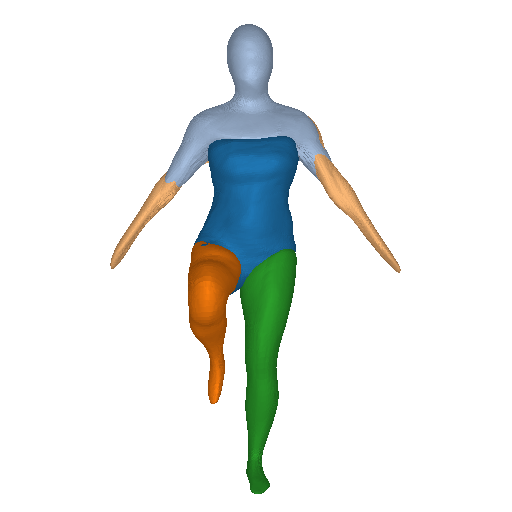}
        \end{subfigure}%
        \begin{subfigure}[b]{0.2\linewidth}
	    	\centering
	        \includegraphics[trim=90 0 90 0,clip,width=\linewidth]{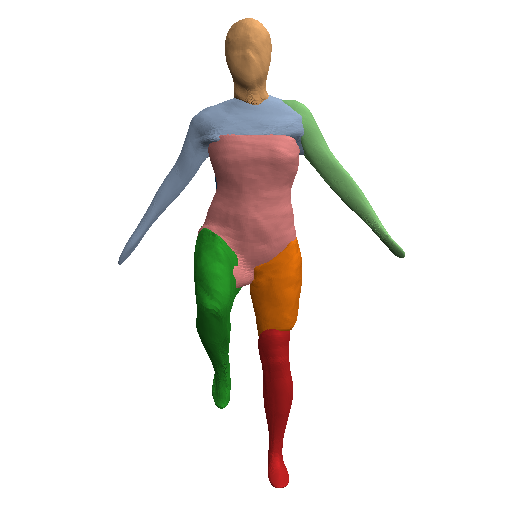}
        \end{subfigure}%
        \begin{subfigure}[b]{0.2\linewidth}
	    	\centering
	        \includegraphics[trim=90 0 90 0,clip,width=\linewidth]{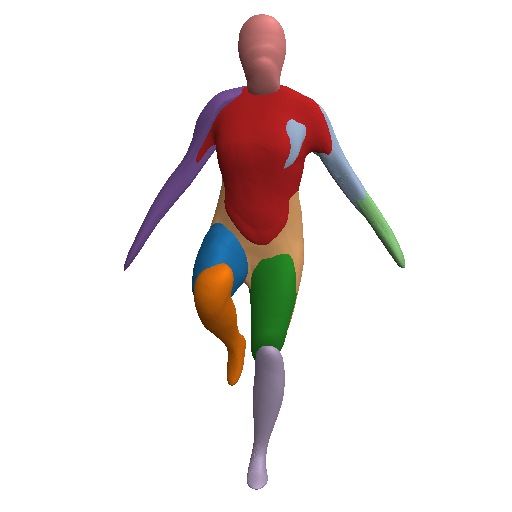}
        \end{subfigure}%
        \\
        \begin{subfigure}[b]{0.2\linewidth}
	    	\centering
	        \includegraphics[trim=70 0 70 0,clip,width=\linewidth]{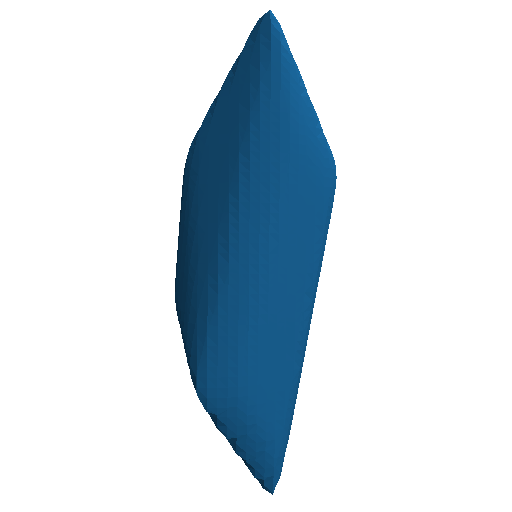}
        \end{subfigure}%
        \begin{subfigure}[b]{0.2\linewidth}
	    	\centering
	        \includegraphics[trim=80 0 80 0,clip,width=\linewidth]{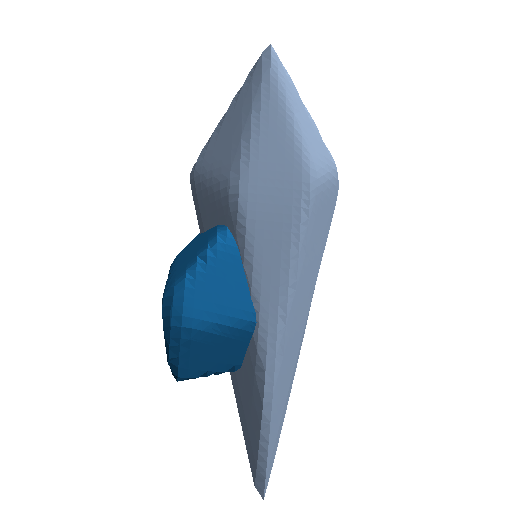}
        \end{subfigure}%
        \begin{subfigure}[b]{0.2\linewidth}
	    	\centering
	        \includegraphics[trim=80 0 80 0,clip,width=\linewidth]{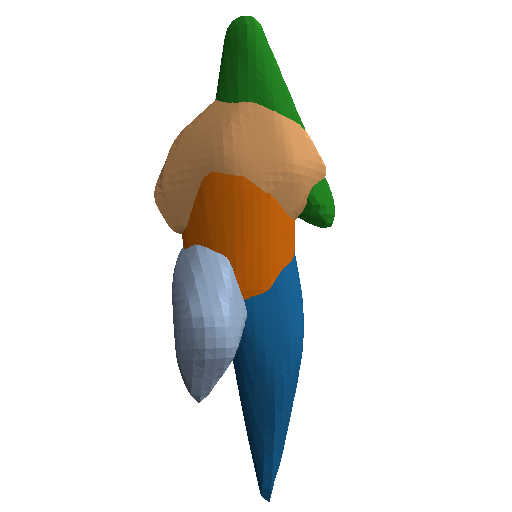}
        \end{subfigure}%
        \begin{subfigure}[b]{0.2\linewidth}
	    	\centering
	        \includegraphics[trim=80 0 80 0,clip,width=\linewidth]{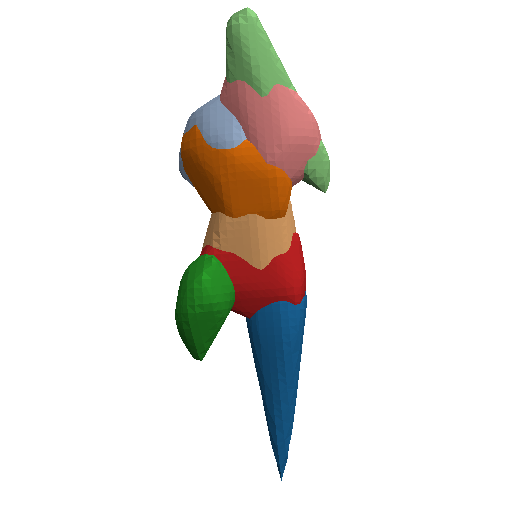}
        \end{subfigure}%
        \begin{subfigure}[b]{0.2\linewidth}
	    	\centering
	        \includegraphics[trim=80 0 80 0,clip,width=\linewidth]{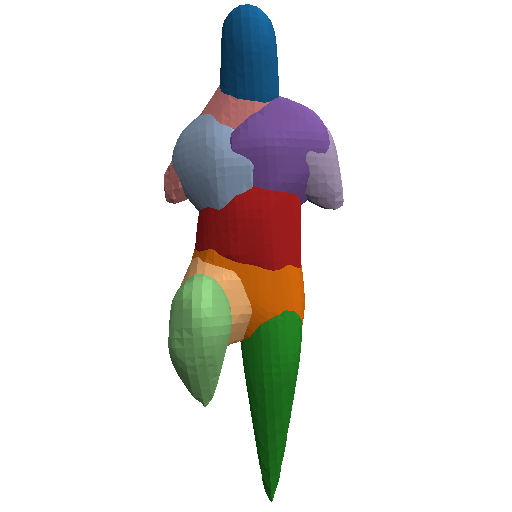}
        \end{subfigure}%
        \vspace{0.1em}
    \end{minipage}
    \vskip\baselineskip%
    \vspace{-1.2em}
    \caption{{\bf Human body modelling}. We visualize the target mesh and the predicted primitives with Neural Parts (first row) and CvxNet (second) using $1, 2, 5, 8$ and $10$ primitives.}
    \label{fig:human_body_modelling}
    \vspace{-1.2em}
\end{figure}

\boldparagraph{Overlapping Loss}%
To encourage semantically meaningful shape abstractions, where primitives represent different object parts, we introduce a non-overlapping loss that penalizes any point in space that is internal to more than $\lambda$ primitives as follows:
\begin{equation}
    \cL_{overlap}(\cX_o) = \frac{1}{|\cX_o|} \max\left(
        0,
        \sum_{m=1}^M \sigma\left(\frac{-g^m(\bx)}{\tau}\right) - \lambda
    \right)
    \label{eq:overlapping_loss}
\end{equation}

\boldparagraph{Coverage Loss}%
The coverage loss makes sure that all primitives cover parts of the predicted shape.
In practice, it prevents degenerate primitive arrangements, where some primitives are very small and do not contribute to the reconstruction. We implement this loss by encouraging that each primitive contains at least k points of the target object:
\begin{equation}
    \cL_{cover}(\cX_o) = \sum_{m=1}^M \sum_{\bx \in \cN_k^m} \max\left(0, g^m(\bx)\right).
    \label{eq:contribution_loss}
\end{equation}
Here $\cN_k^m \subset \{(\bx, o) \in \cX_o | o=1\}$ contains the $k$ points with the minimum distance from the $m$-th primitive.

\section{Experimental Evaluation}\label{sec:results}

\boldparagraph{Datasets}%
We report results on three different datasets: D-FAUST~\cite{Bogo2017CVPR}, FreiHAND~\cite{Zimmermann2019ICCV} and ShapeNet~\cite{Chang2015ARXIV}.
D-FAUST contains $38,640$ meshes of humans performing various tasks such as ``chicken wings" and ``running on spot".
For ShapeNet~\cite{Chang2015ARXIV}, we perform category specific training using the same image renderings and train/test splits as~\cite{Choy2016ECCV}.
Finally, for FreiHAND~\cite{Zimmermann2019ICCV}, we select the first $5000$ hand poses and generate meshes using the provided MANO parameters~\cite{Romero2017TOG}.
Details regarding data preprocessing and additional results are provided in the supplementary.

\boldparagraph{Baselines}%
We compare against various primitive-based methods: SQs~\cite{Paschalidou2019CVPR} that employ superquadrics, CvxNet~\cite{Deng2020CVPR} that use smooth convexes and H-SQs~\cite{Paschalidou2020CVPR} that consider a hierarchical, geometrically more accurate decomposition of parts using superquadrics.
Finally, we also report results for OccNet \cite{Mescheder2019CVPR}, a state-of-the-art implicit-based method that does not reason about object parts. For all baselines, we use the code provided by the authors.

\begin{figure}
    \centering
    \vspace{-1.2em}
    \begin{subfigure}[b]{0.15\linewidth}
		\centering
        Target
    \end{subfigure}%
    \hfill%
    \begin{subfigure}[b]{0.15\linewidth}
		\centering
        OccNet
    \end{subfigure}%
    \hfill%
    \begin{subfigure}[b]{0.15\linewidth}
		\centering
        SQs
    \end{subfigure}%
    \hfill%
    \begin{subfigure}[b]{0.15\linewidth}
		\centering
        H-SQs
    \end{subfigure}%
    \hfill%
    \begin{subfigure}[b]{0.15\linewidth}
		\centering
        CvxNet
    \end{subfigure}%
    \hfill%
    \begin{subfigure}[b]{0.15\linewidth}
        \centering
        Ours
    \end{subfigure}
    \vskip\baselineskip%
    \vspace{-0.5em}
    \begin{subfigure}[b]{0.15\linewidth}
		\centering
		\includegraphics[trim=100 0 100 0,clip,width=\linewidth]{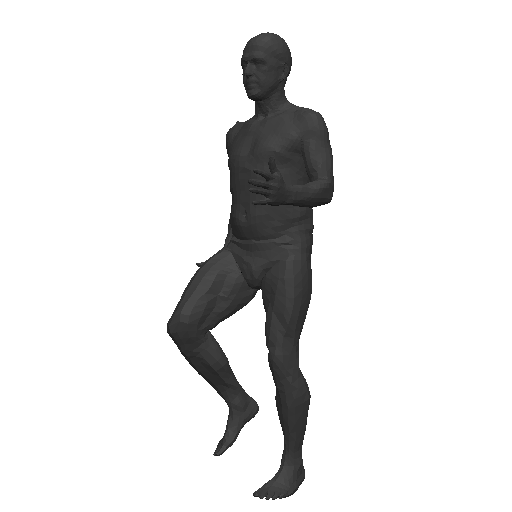}
    \end{subfigure}%
    \hfill%
    \begin{subfigure}[b]{0.15\linewidth}
		\centering
		\includegraphics[trim=100 0 100 0,clip,width=\linewidth]{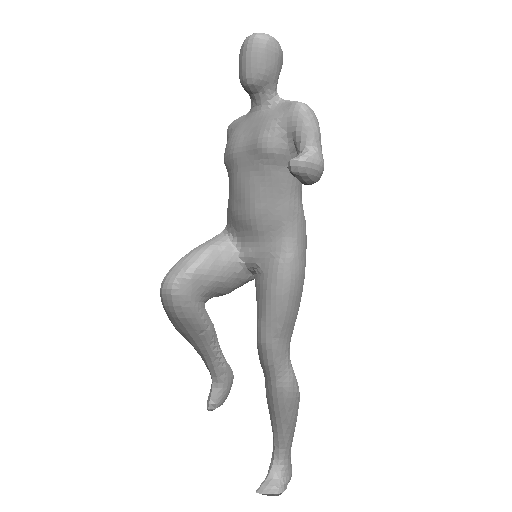}
    \end{subfigure}%
    \hfill%
    \begin{subfigure}[b]{0.15\linewidth}
		\centering
		\includegraphics[trim=100 0 100 0,clip,width=\linewidth]{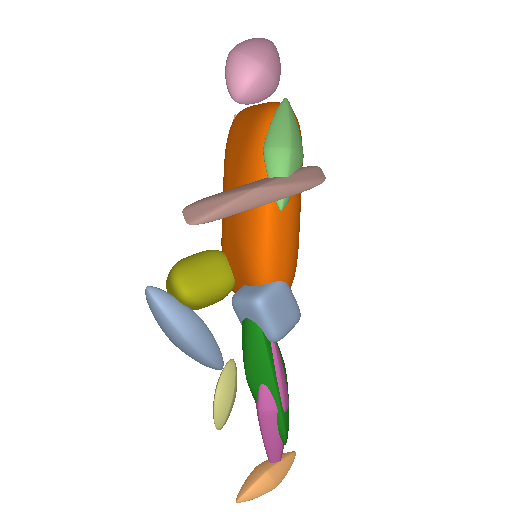}
    \end{subfigure}%
    \hfill%
    \begin{subfigure}[b]{0.15\linewidth}
		\centering
		\includegraphics[trim=100 0 100 0,clip,width=\linewidth]{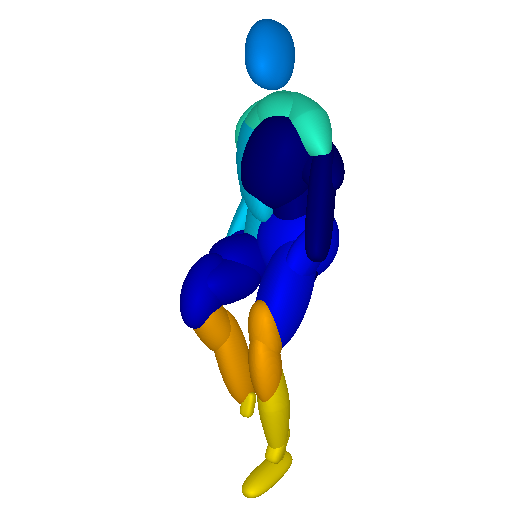}
    \end{subfigure}%
    \hfill%
    \begin{subfigure}[b]{0.15\linewidth}
		\centering
		\includegraphics[trim=100 0 100 0,clip,width=\linewidth]{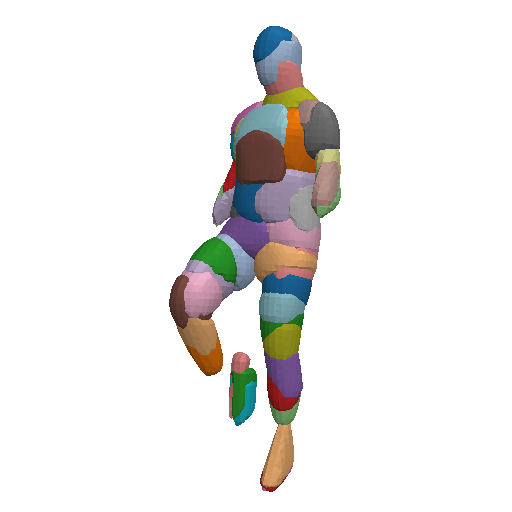}
    \end{subfigure}%
    \hfill%
    \begin{subfigure}[b]{0.15\linewidth}
		\centering
		\includegraphics[trim=100 0 100 0,clip,width=\linewidth]{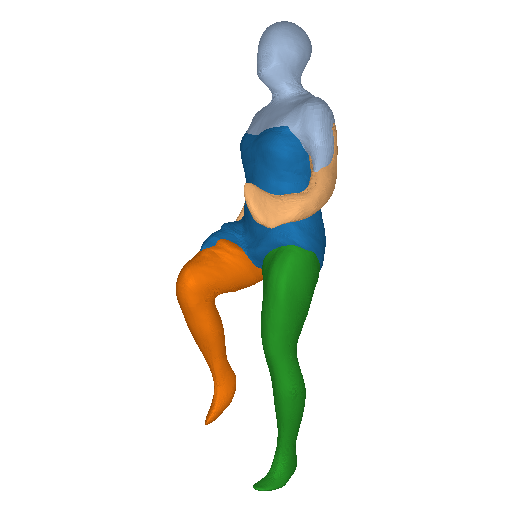}
    \end{subfigure}%
            		            		            		            		            		            		        \vspace{-1.2em}
    \vskip\baselineskip%
    \begin{subfigure}[b]{0.15\linewidth}
		\centering
		\includegraphics[trim=100 0 100 0,clip,width=\linewidth]{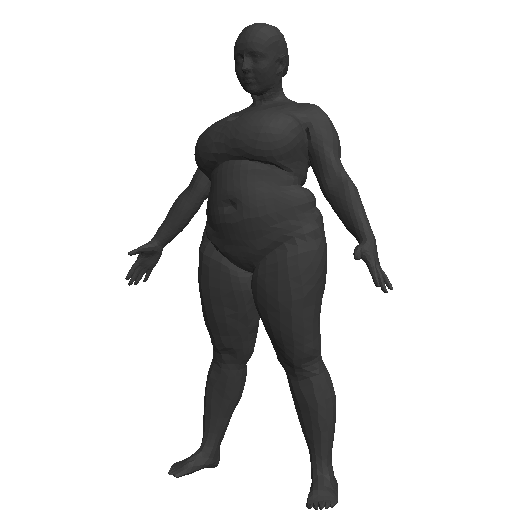}
    \end{subfigure}%
    \hfill%
    \begin{subfigure}[b]{0.15\linewidth}
		\centering
		\includegraphics[trim=100 0 100 0,clip,width=\linewidth]{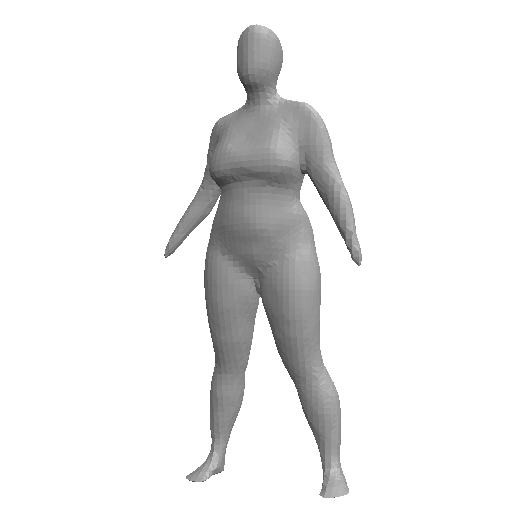}
    \end{subfigure}%
    \hfill%
    \begin{subfigure}[b]{0.15\linewidth}
		\centering
		\includegraphics[trim=100 0 100 0,clip,width=\linewidth]{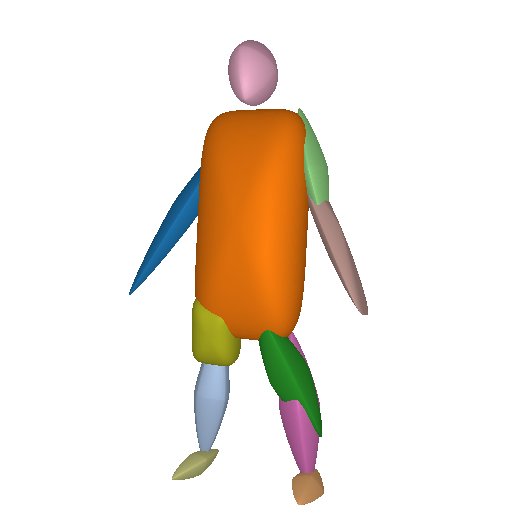}
    \end{subfigure}%
    \hfill%
    \begin{subfigure}[b]{0.15\linewidth}
		\centering
		\includegraphics[trim=100 0 100 0,clip,width=\linewidth]{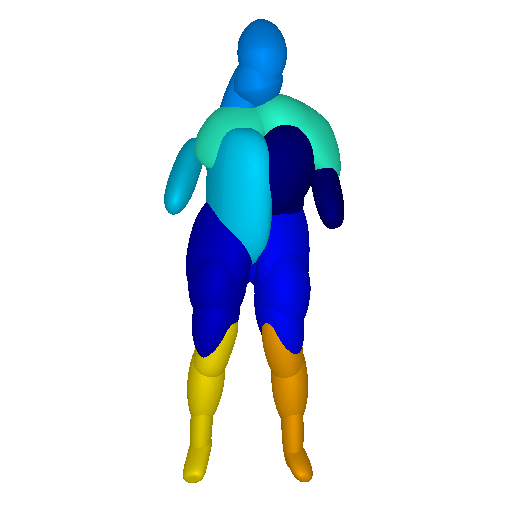}
    \end{subfigure}%
    \hfill%
    \begin{subfigure}[b]{0.15\linewidth}
		\centering
		\includegraphics[trim=100 0 100 0,clip,width=\linewidth]{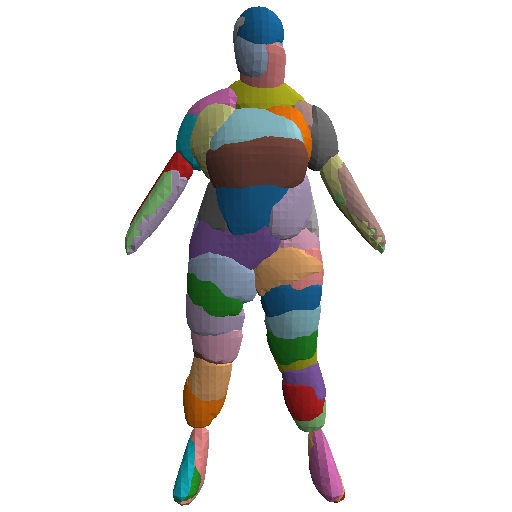}
    \end{subfigure}%
    \hfill%
    \begin{subfigure}[b]{0.15\linewidth}
		\centering
		\includegraphics[trim=100 0 100 0,clip,width=\linewidth]{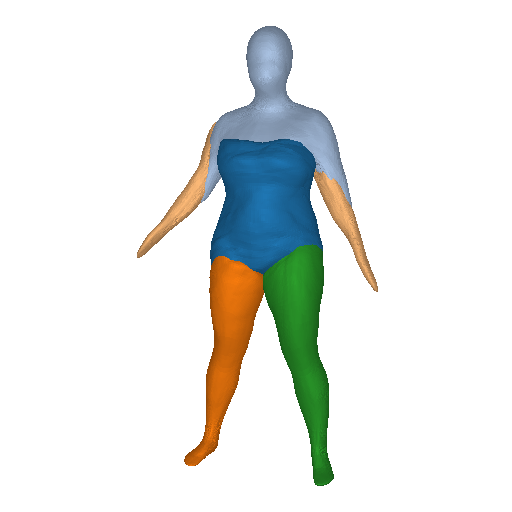}
    \end{subfigure}%
            		            		            		            		            		            		        \vskip\baselineskip%
    \vspace{-1.2em}
    \caption{{\bf Single Image 3D Reconstruction on D-FAUST}. The input image is shown on the first column and the rest contain predictions of all methods: OccNet (second), primitive-based predictions with superquadrics (third and fourth) and convexes (fifth) and ours with $5$ primitives (last).}
    \label{fig:qualitative_dfaust}
    \vspace{-1.2em}
\end{figure}

\subsection{Representation Power}
\label{subsec:representation_capacity}

To evaluate the representation power of Neural Parts, we train our model and our baselines on D-FAUST for different number of primitives and measure their reconstruction quality.
In particular, we train our model for $2, 5$ and $8$ primitives, CvxNet~\cite{Deng2020CVPR} and SQs~\cite{Paschalidou2019CVPR} for $5, 10, 25$ and $50$ primitives and H-SQs~\cite{Paschalidou2020CVPR} for $4, 8, 16$ and $32$ primitives.
In \figref{fig:rep_power}, we report the reconstruction performance \wrt IoU and Chamfer-$L_1$ distance.
Neural Parts achieve more accurate reconstructions for any given number of primitives and are competitive to OccNet that does not reason about parts. In particular, our model achieves $67.3\%$ IOU with only $5$ primitives, whereas CvxNet, with $10$ times more primitives, achieve $62\%$.
This is also validated qualitatively in \figref{fig:human_body_modelling}, where we compare the predictions of CvxNet and our model for different number of primitives.
Neural Parts accurately capture the pose and the limbs of the human with as little as one or two primitives, whereas CvxNet fail to reconstruct the arms even with $10$ primitives.

\begin{figure}
    \begin{subfigure}[t]{\columnwidth}
    \centering
    \vspace{-1.2em}
    \begin{subfigure}[b]{0.15\linewidth}
		\centering
        Input
    \end{subfigure}%
    \hfill%
    \begin{subfigure}[b]{0.15\linewidth}
		\centering
        OccNet
    \end{subfigure}%
    \hfill%
    \begin{subfigure}[b]{0.15\linewidth}
		\centering
        SQs
    \end{subfigure}%
    \hfill%
    \begin{subfigure}[b]{0.15\linewidth}
		\centering
        H-SQs
    \end{subfigure}%
    \hfill%
    \begin{subfigure}[b]{0.15\linewidth}
		\centering
        CvxNet
    \end{subfigure}%
    \hfill%
    \begin{subfigure}[b]{0.15\linewidth}
        \centering
        Ours
    \end{subfigure}
    \vspace{-1em}
    \vskip\baselineskip%
    \begin{subfigure}[b]{0.15\linewidth}
		\centering
		\includegraphics[trim=15 0 15 0,clip,width=\linewidth]{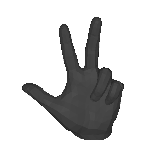}
    \end{subfigure}%
    \hfill%
    \begin{subfigure}[b]{0.15\linewidth}
		\centering
		\includegraphics[trim=50 0 50 0,clip,width=\linewidth]{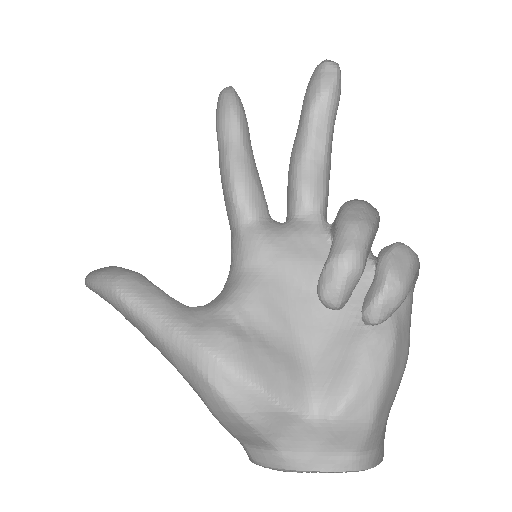}
    \end{subfigure}%
    \hfill%
    \begin{subfigure}[b]{0.15\linewidth}
		\centering
		\includegraphics[trim=20 0 20 0,clip,width=\linewidth]{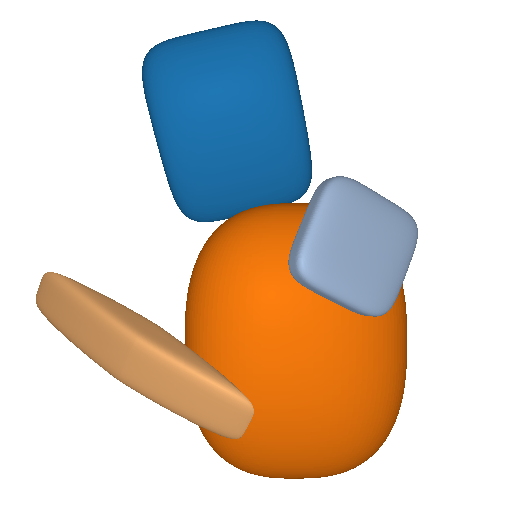}
    \end{subfigure}%
    \hfill%
    \begin{subfigure}[b]{0.15\linewidth}
		\centering
		\includegraphics[trim=25 0 25 0,clip,width=\linewidth]{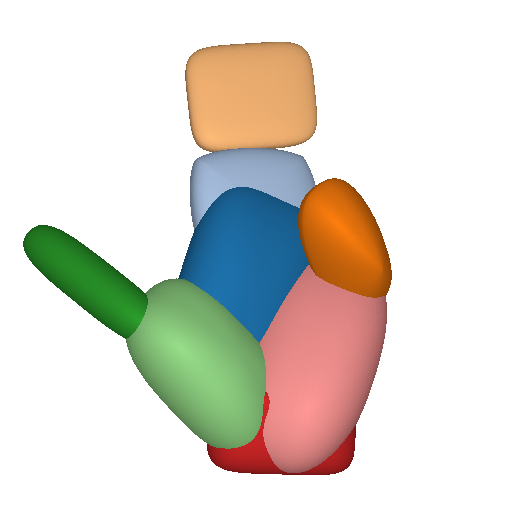}
    \end{subfigure}%
    \hfill%
    \begin{subfigure}[b]{0.15\linewidth}
		\centering
		\includegraphics[trim=50 0 50 0,clip,width=\linewidth]{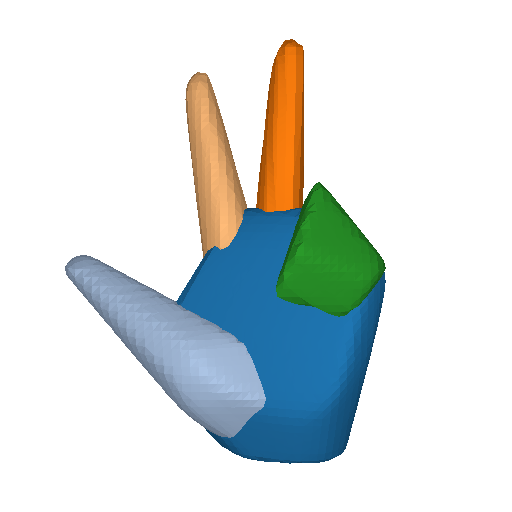}
    \end{subfigure}%
    \hfill%
    \begin{subfigure}[b]{0.15\linewidth}
		\centering
		\includegraphics[trim=50 0 50 0,clip,width=\linewidth]{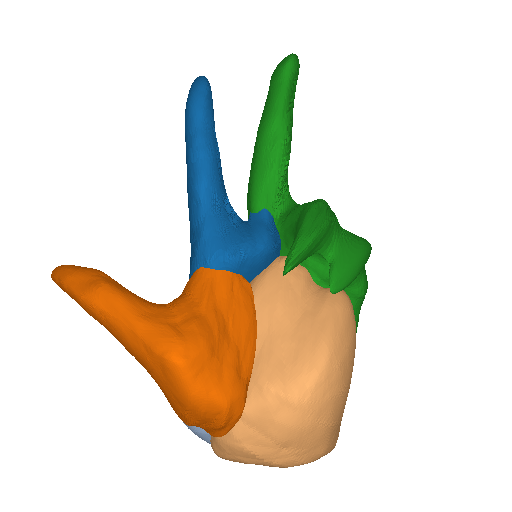}
    \end{subfigure}%
    \vspace{-1.75em}
    \vskip\baselineskip%
    \begin{subfigure}[b]{0.15\linewidth}
		\centering
		\includegraphics[trim=20 20 20 -20,clip,width=\linewidth]{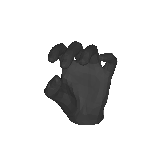}
    \end{subfigure}%
    \hfill%
    \begin{subfigure}[b]{0.15\linewidth}
		\centering
		\includegraphics[width=0.9\linewidth]{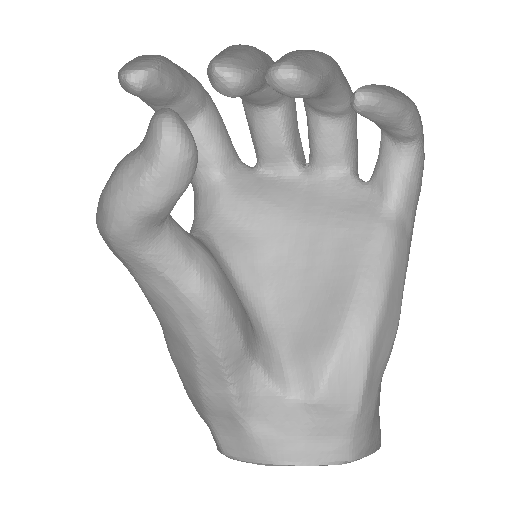}
    \end{subfigure}%
    \hfill%
    \begin{subfigure}[b]{0.15\linewidth}
		\centering
		\includegraphics[trim=50 0 50 0,clip,width=\linewidth]{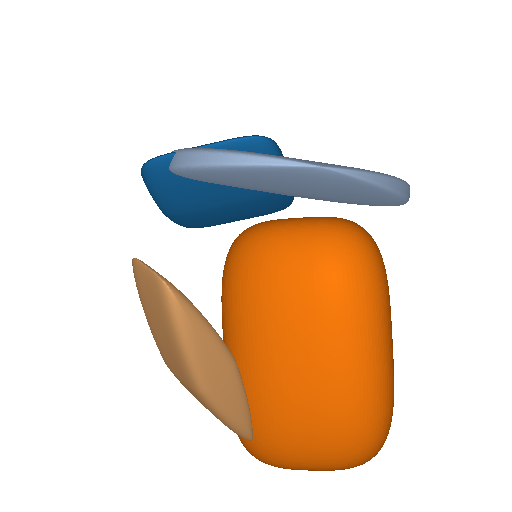}
    \end{subfigure}%
    \hfill%
    \begin{subfigure}[b]{0.15\linewidth}
		\centering
		\includegraphics[trim=40 0 40 0,clip,width=\linewidth]{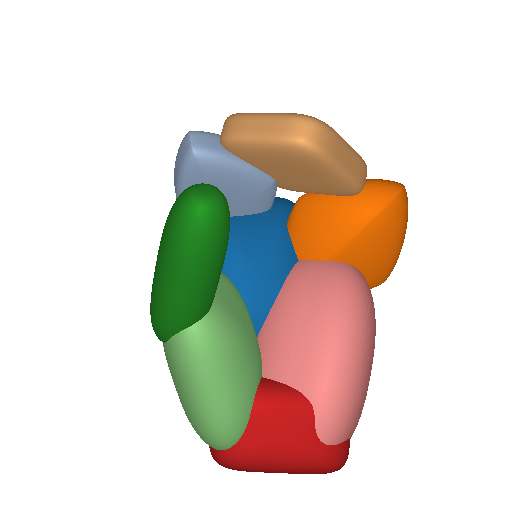}
    \end{subfigure}%
    \hfill%
    \begin{subfigure}[b]{0.15\linewidth}
		\centering
		\includegraphics[trim=50 0 50 0,clip,width=\linewidth]{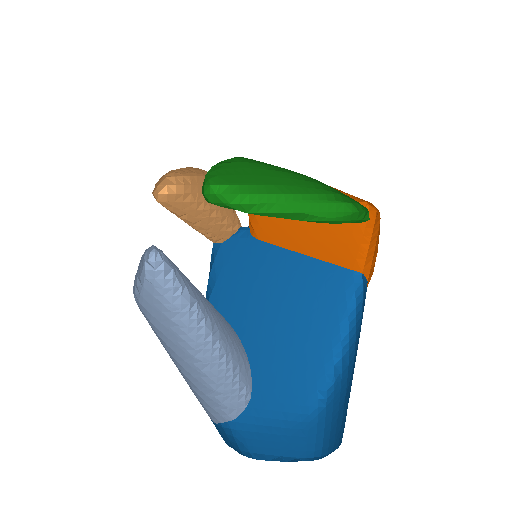}
    \end{subfigure}%
    \hfill%
    \begin{subfigure}[b]{0.15\linewidth}
		\centering
		\includegraphics[trim=50 40 50 -40,clip,width=\linewidth]{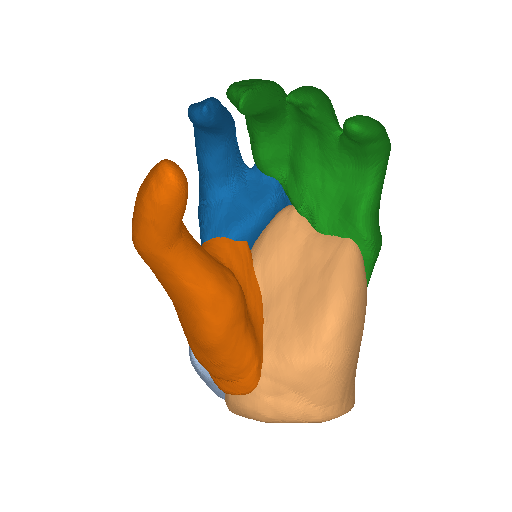}
    \end{subfigure}%
    \vspace{-1.75em}
    \vskip\baselineskip%
    \begin{subfigure}[b]{0.15\linewidth}
		\centering
		\includegraphics[trim=20 8 20 -8,clip,width=\linewidth]{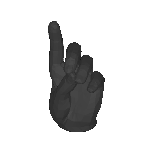}
    \end{subfigure}%
    \hfill%
    \begin{subfigure}[b]{0.15\linewidth}
		\centering
		\includegraphics[trim=50 0 50 0,clip,width=\linewidth]{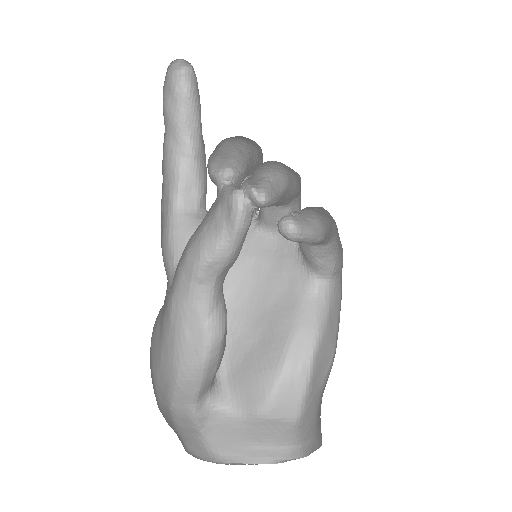}
    \end{subfigure}%
    \hfill%
    \begin{subfigure}[b]{0.15\linewidth}
		\centering
		\includegraphics[trim=50 0 50 0,clip,width=\linewidth]{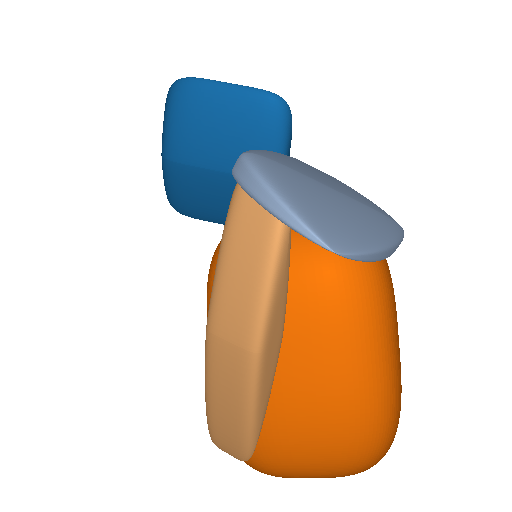}
    \end{subfigure}%
    \hfill%
    \begin{subfigure}[b]{0.15\linewidth}
		\centering
		\includegraphics[trim=50 0 50 0,clip,width=\linewidth]{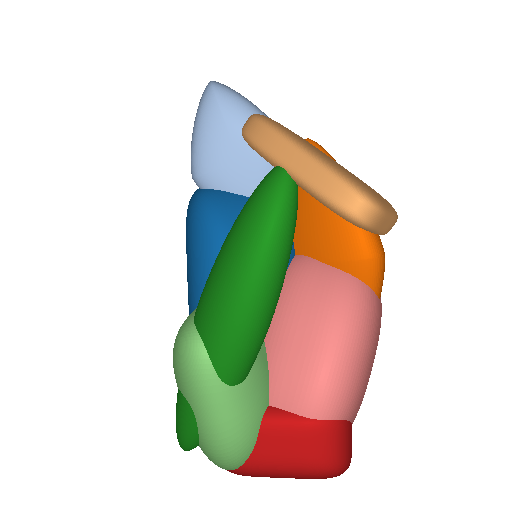}
    \end{subfigure}%
    \hfill%
    \begin{subfigure}[b]{0.15\linewidth}
		\centering
		\includegraphics[trim=50 0 50 0,clip,width=\linewidth]{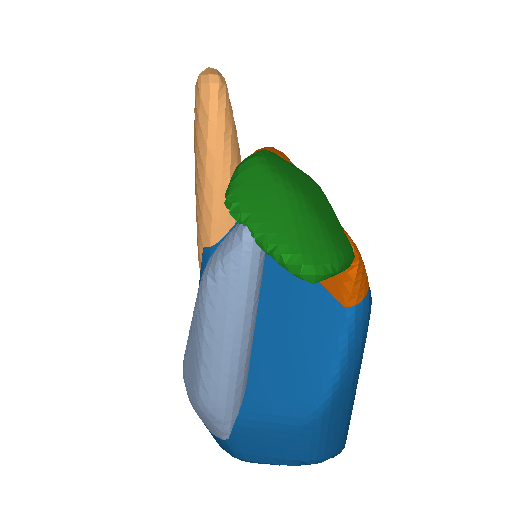}
    \end{subfigure}%
    \hfill%
    \begin{subfigure}[b]{0.15\linewidth}
		\centering
		\includegraphics[trim=50 0 50 0,clip,width=\linewidth]{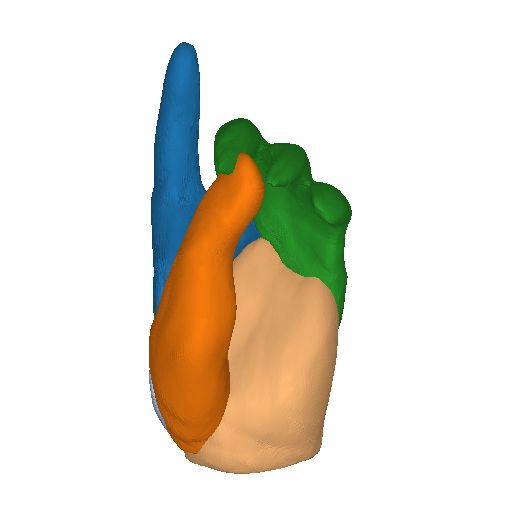}
    \end{subfigure}
    \end{subfigure}\\[1.5em]
    \begin{subfigure}[t]{\columnwidth}
        \resizebox{\textwidth}{!}{
        \begin{tabular}{l||c|cccc}
            \toprule
            & OccNet & SQs & H-SQs & CvxNet & Ours\\
            \midrule
            IoU & 0.891 & 0.693 & 0.768 & 0.832 & \bf{0.879} \\
            Chamfer-$L_1$ & 0.038 & 0.093 & 0.077 & 0.059 & \bf{0.057} \\
            \bottomrule
        \end{tabular}}
    \end{subfigure}
    \caption{{\bf{Single Image 3D Reconstruction on FreiHAND}}.
    We compare our model with OccNet, SQs and CvxNet with $5$ primitives and
    H-SQs with $8$ primitives. Our model outperforms all primitive based
    methods in terms of both IoU ($\uparrow$) and Chamfer-L1 ($\downarrow$)
    distance.}
    \label{fig:freihand_results}
    \vspace{-1.2em}
\end{figure}

\subsection{Reconstruction Accuracy}

\boldparagraph{Dynamic FAUST}%
In this experiment, we compare CvxNet and SQs with $50$ primitives and H-SQs for a maximum number of $32$ primitives to Neural Parts with $5$ primitives to showcase that our model can capture the human body's geometry using an order of magnitude less primitives.
Qualitative results from the predicted primitives using our model and the baselines are summarized in \figref{fig:qualitative_dfaust}.
Note that OccNet is not directly comparable with part-based methods, however, we include it in our analysis as a typical representative of powerful implicit shape extraction techniques.
We observe that while all methods roughly capture the human pose, Neural Parts result in a part assembly that is very close to the target object. While CvxNet with $50$ primitives yield fairly accurate reconstructions, the final representation lacks any part-level semantic interpretation.
The quantitative evaluation of this experiment is provided in \figref{fig:rep_power}.

\boldparagraph{FreiHAND}%
For this experiment, we train our model, CvxNet and SQs with $5$ and H-SQs with $8$ primitives.
\figref{fig:freihand_results} summarizes the reconstruction performance for all methods. We observe that Neural Parts yield more geometrically accurate reconstructions that faithfully capture fine details, \ie the position of the thumb, (see \figref{fig:freihand_results}) whereas CvxNet and SQs focus primarily on the structure of the predicted shape and miss out fine details.

\begin{figure}
    \begin{subfigure}[t]{\columnwidth}
    \centering
    \vspace{-1.2em}
    \begin{subfigure}[b]{0.20\linewidth}
		\centering
        Input
    \end{subfigure}%
    \hfill%
    \begin{subfigure}[b]{0.20\linewidth}
		\centering
        OccNet
    \end{subfigure}%
    \hfill%
    \begin{subfigure}[b]{0.20\linewidth}
		\centering
        CvxNet-5
    \end{subfigure}%
    \hfill%
    \begin{subfigure}[b]{0.20\linewidth}
		\centering
        CvxNet-25
    \end{subfigure}%
    \hfill%
    \begin{subfigure}[b]{0.20\linewidth}
        \centering
        Ours
    \end{subfigure}
    \vspace{-2.2em}
    \vskip\baselineskip%
    \begin{subfigure}[b]{0.20\linewidth}
		\centering
		\includegraphics[width=\linewidth]{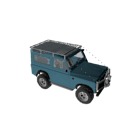}
    \end{subfigure}%
    \hfill%
    \begin{subfigure}[b]{0.20\linewidth}
		\centering
		\includegraphics[width=\linewidth]{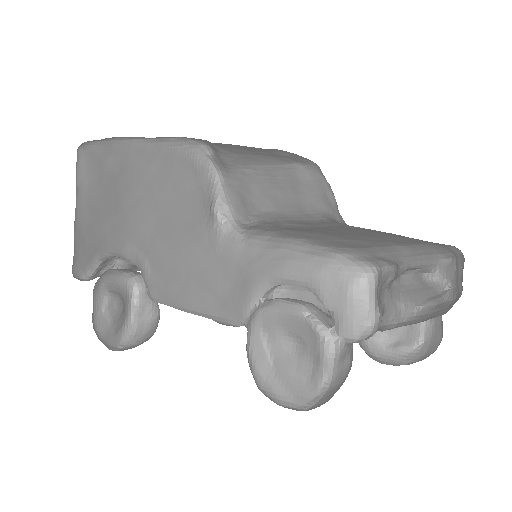}
    \end{subfigure}%
    \hfill%
    \begin{subfigure}[b]{0.20\linewidth}
		\centering
		\includegraphics[width=\linewidth]{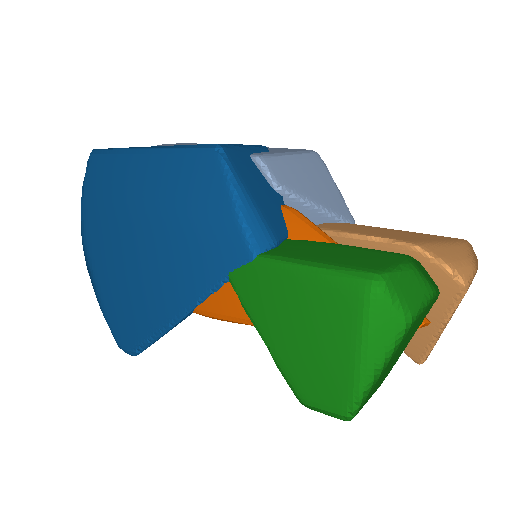}
    \end{subfigure}%
    \hfill%
    \begin{subfigure}[b]{0.20\linewidth}
		\centering
		\includegraphics[width=\linewidth]{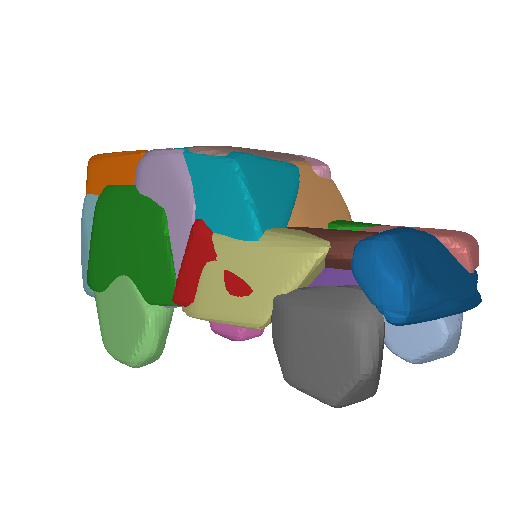}
    \end{subfigure}%
    \hfill%
    \begin{subfigure}[b]{0.20\linewidth}
		\centering
		\includegraphics[width=\linewidth]{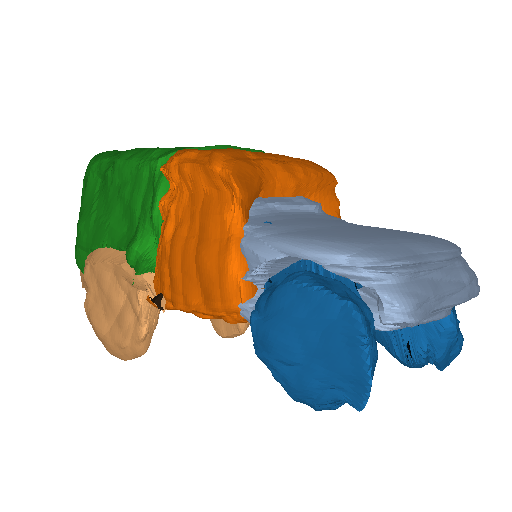}
    \end{subfigure}%
    \vspace{-2.2em}
    \vskip\baselineskip%
    \begin{subfigure}[b]{0.20\linewidth}
    	\centering
		\includegraphics[trim=30 30 30 30,clip,width=\textwidth]{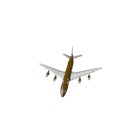}
    \end{subfigure}%
    \hfill%
    \begin{subfigure}[b]{0.20\linewidth}
    	\centering
		\includegraphics[width=0.8\linewidth]{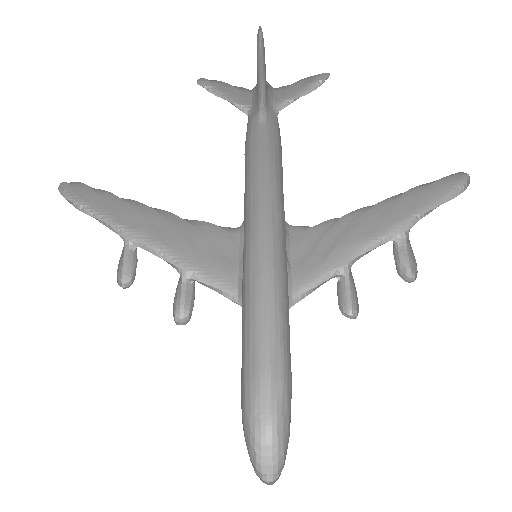}
    \end{subfigure}%
    \hfill%
    \begin{subfigure}[b]{0.20\linewidth}
    	\centering
		\includegraphics[width=0.8\linewidth]{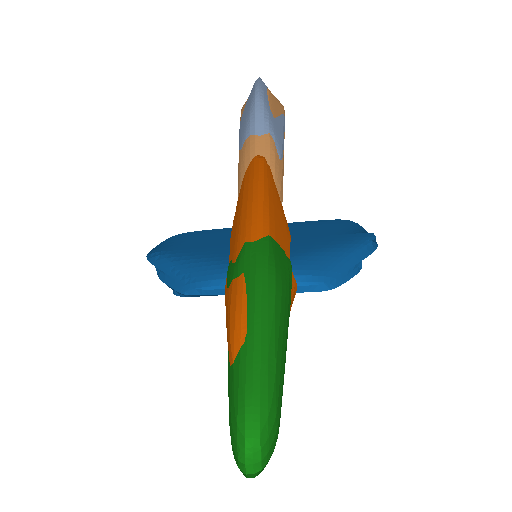}
    \end{subfigure}%
    \hfill%
    \begin{subfigure}[b]{0.20\linewidth}
    	\centering
		\includegraphics[width=0.8\linewidth]{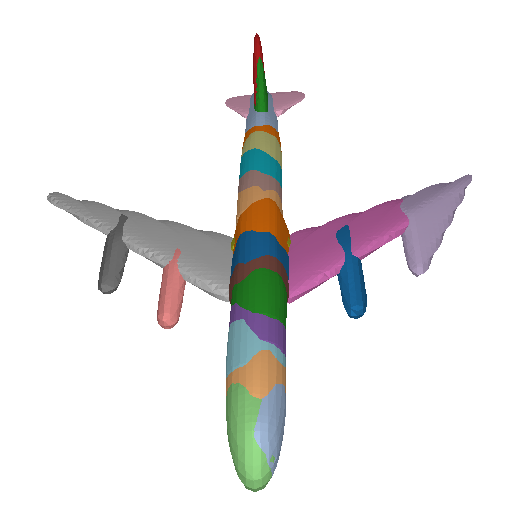}
    \end{subfigure}%
    \hfill%
    \begin{subfigure}[b]{0.20\linewidth}
    	\centering
		\includegraphics[width=0.8\linewidth]{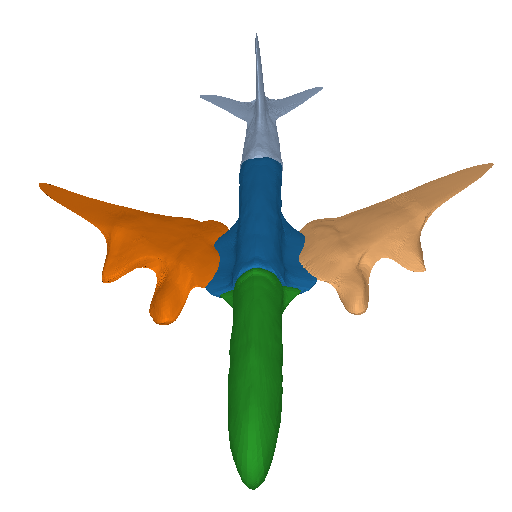}
    \end{subfigure}%
    \vspace{-1.2em}
    \vskip\baselineskip%
    \begin{subfigure}[b]{0.20\linewidth}
    	\centering
		\includegraphics[width=0.8\textwidth]{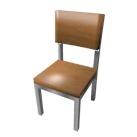}
    \end{subfigure}%
    \hfill%
    \begin{subfigure}[b]{0.20\linewidth}
    	\centering
		\includegraphics[width=0.8\textwidth]{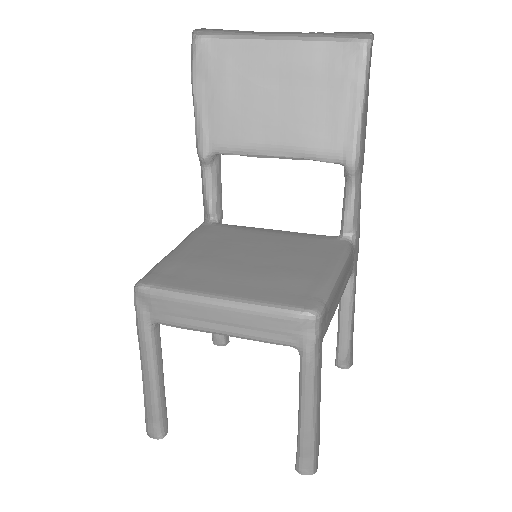}
    \end{subfigure}%
    \hfill%
    \begin{subfigure}[b]{0.20\linewidth}
    	\centering
		\includegraphics[width=0.8\textwidth]{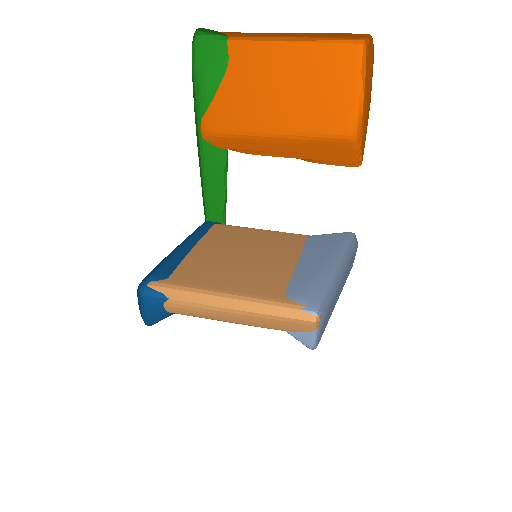}
    \end{subfigure}%
    \hfill%
    \begin{subfigure}[b]{0.20\linewidth}
    	\centering
		\includegraphics[width=0.8\textwidth]{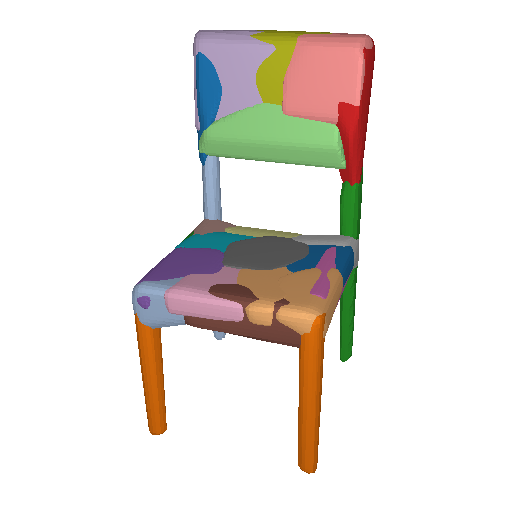}
    \end{subfigure}%
    \hfill%
    \begin{subfigure}[b]{0.20\linewidth}
    	\centering
		\includegraphics[width=0.8\textwidth]{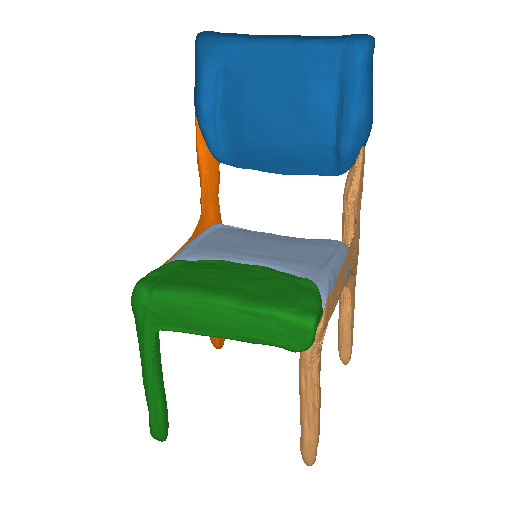}
    \end{subfigure}%
    \end{subfigure}\\[1.2em]
    \begin{subfigure}[t]{0.95\columnwidth}
        \resizebox{\columnwidth}{!}{%
        \centering
        \begin{tabular}{l||c|ccc}
            \toprule
            IoU & OccNet & CvxNet - 5 & CvxNet - 25 & Ours \\
            \midrule
            cars & 0.763 & 0.650 & 0.666 &  \bf{0.697} \\
            planes & 0.451 & 0.425 & 0.448 &  \bf{0.454} \\
            chairs & 0.432 & 0.364 & 0.392 &  \bf{0.412} \\
            \bottomrule
        \end{tabular}}
    \end{subfigure}
    \caption{{\bf Single Image 3D Reconstruction on ShapeNet}. We compare
    Neural Parts to OccNet and CvxNet with $5$ and $25$ primitives. Our model
    yields semantic and more accurate reconstructions with $5\times$ less
    primitives.}
    \label{fig:shapenet_results}
    \vspace{-1.2em}
\end{figure}

\boldparagraph{ShapeNet}%
We train Neural Parts with $5$ primitives and CvxNet with $5$ and $25$ primitives on various ShapeNet objects.
We observe that our model results in more accurate reconstructions than CvxNet with both $5$ and $25$ primitives (see \figref{fig:shapenet_results}).
When increasing the number of primitives to $25$, CvxNet improves in terms of
reconstruction quality but the predicted primitives lack semantic
interpretation. Furthermore, for the case of chairs, CvxNet with $5$ primitives
do not have enough representation power, thus entire object parts are missing.
Additional results as well as comparison \wrt Chamfer-$L_1$ distance are provided in the supplementary.

\subsection{Ablation Study}
\label{subsec:ablations}

In this section, we train our model on D-FAUST with $5$ primitives and investigate the impact of its components.
Additional ablations are provided in the supplementary.

\begin{figure}
    \vspace{-1.5em}
    \begin{subfigure}[b]{\columnwidth}
        \begin{subfigure}[t]{0.25\columnwidth}
            \centering
            \caption{w/o $\phi_{\btheta}^{-1}(\bx)$}
            \includegraphics[trim=25 0 25 0,clip,width=\textwidth]{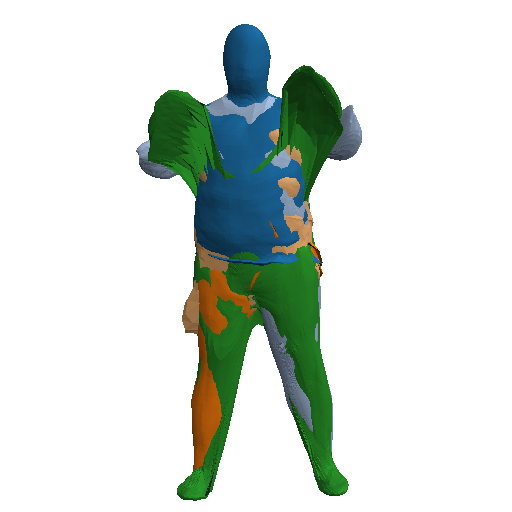}
            \label{fig:ablation_inn}
        \end{subfigure}%
        \begin{subfigure}[t]{0.25\columnwidth}
            \centering
            \caption{w/o $\cL_{occ}$}
            \includegraphics[trim=25 0 25 0,clip,width=\textwidth]{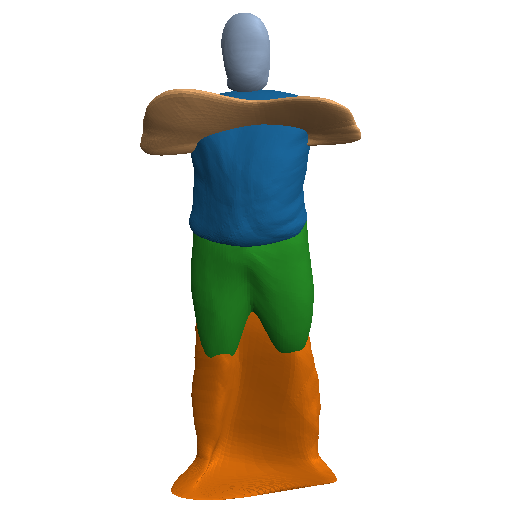}
            \label{fig:ablation_occ}
        \end{subfigure}%
        \begin{subfigure}[t]{0.25\columnwidth}
            \centering
            \caption{w/o  $\cL_{rec}$}
            \includegraphics[trim=25 0 25 0,clip,width=\textwidth]{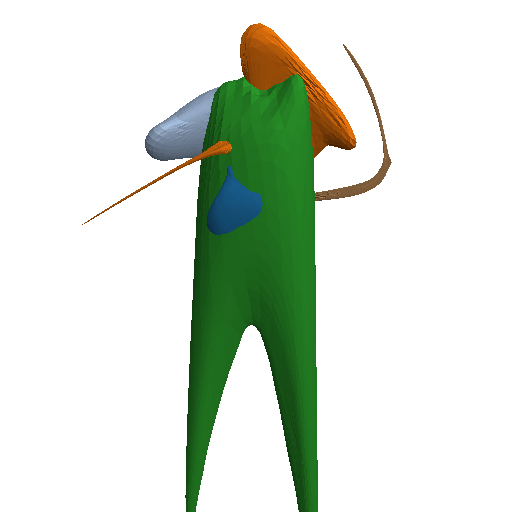}
            \label{fig:ablation_ch}
        \end{subfigure}%
        \begin{subfigure}[t]{0.25\columnwidth}
            \centering
            \caption{Ours}
            \includegraphics[trim=25 0 25 0,clip,width=\textwidth]{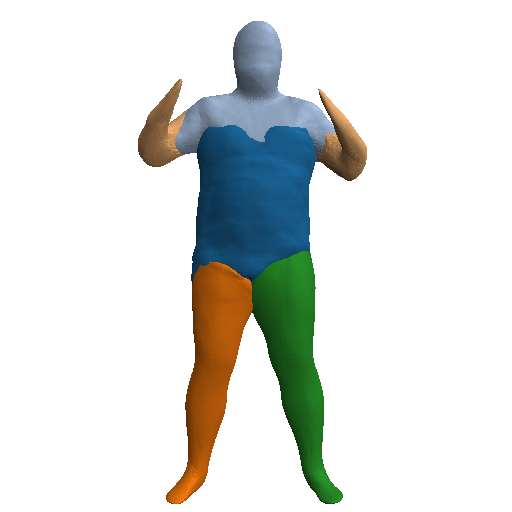}
            \label{fig:ablation_ours}
        \end{subfigure}
    \end{subfigure}\\[0em]
    \begin{subfigure}[b]{\columnwidth}
    \centering
    \resizebox{\columnwidth}{!}{
    \begin{tabular}{l||ccc|c}
        \toprule
        & w/o $\phi_{\btheta}^{-1}(\bx)$ & w/o $\cL_{occ}$ & w/o  $\cL_{rec}$ & Ours \\
        \midrule
        IoU & 0.639 & 0.642 & 0.643 &  0.673 \\
        Chamfer-$L_1$ & 0.119 & 0.125 & 0.150  & 0.09 \\
        \bottomrule
    \end{tabular}}
    \end{subfigure}
    \caption{{\bf Ablation Study.} We ablate the INN as well as the volumetric and the surface loss and show their impact both quantitatively and qualitatively.}
    \label{fig:ablation}
\end{figure}

\boldparagraph{Invertibility}%
To investigate the impact of the INN, we train our model without using the inverse mapping, $\phi_{\btheta}^{-1}(\bx)$.
As a result, it is not possible to either compute the union of parts or enforce additional constraints on the predicted primitives \ie we only train with the loss of \eqref{eq:reconstruction_loss}, without discarding internal points as in \eqref{eq:primitive_explicit_full}.
From \figref{fig:ablation_inn} it becomes evident that each primitive seeks to cover the entire shape thus resulting in redundant and non semantic primitives. In addition, the lack of an inverse mapping prevents us from using any occupancy loss, which results in poor modelling of empty space (\ie the hands are connected with the body).

\boldparagraph{Occupancy and Surface Losses}%
We compare the performance of our model using either the occupancy \eqref{eq:occupancy_loss} or the surface loss \eqref{eq:reconstruction_loss}.
The reconstruction without $\cL_{occ}$ (\figref{fig:ablation_occ}) accurately captures the geometry of the human but fails to efficiently model empty space (\ie legs are connected).
Similarly, the reconstruction without $\cL_{rec}$ (\figref{fig:ablation_ch}) results in degenerate primitives with small volume that are ``pushed'' far away from the human body.

\begin{figure}
    \begin{subfigure}[t]{\columnwidth}
        \includegraphics[width=1.0\linewidth]{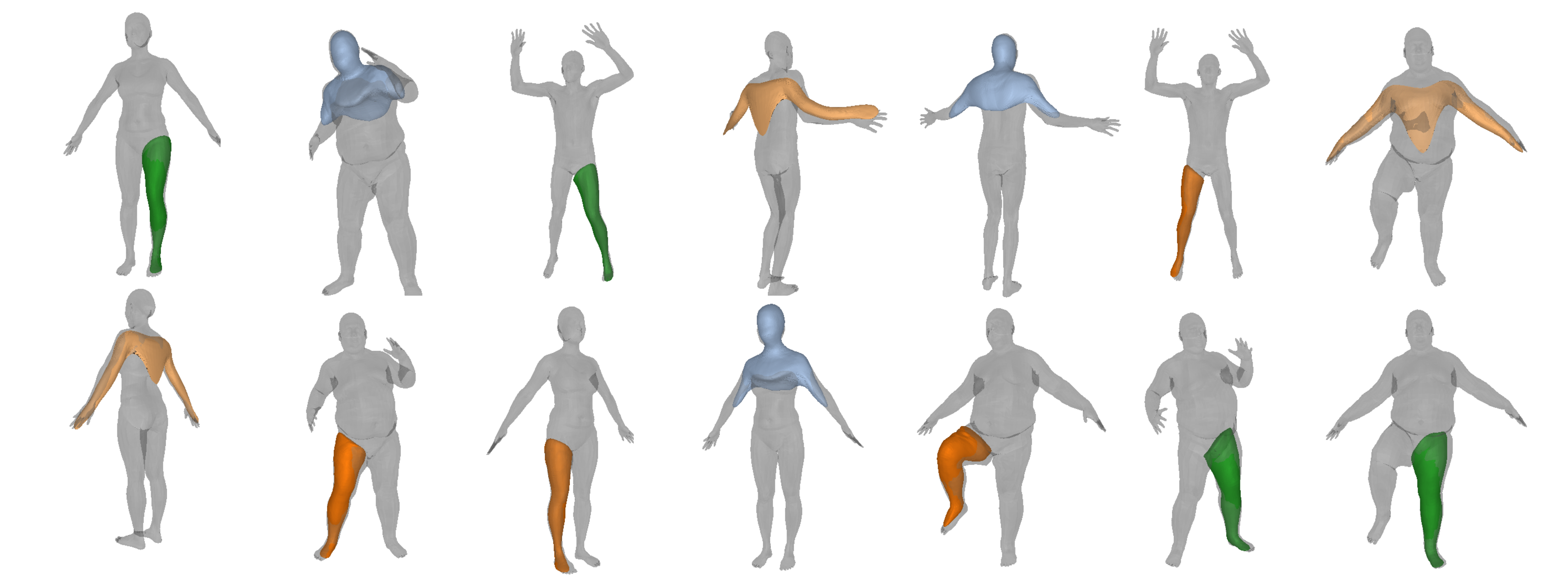}
    \end{subfigure}\\[1em]
    \begin{subfigure}[t]{\columnwidth}
        \resizebox{\columnwidth}{!}{
        \begin{tabular}{l|ccccc}
            \toprule
            & L-thumb & R-thumb & L-toe & R-toe & Nose\\
            \midrule
            CvxNet-5 & 61.1$\%$ & 67.1$\%$ & 98.2$\%$ & 91.2$\%$ & 98$\%$ \\
            CvxNet-50 & 29$\%$ & 37$\%$ & 56.3$\%$ & 58.1$\%$ & 52$\%$ \\
            Ours & 91.9$\%$ & 88.2$\%$ & 99.8$\%$ & 92.5$\%$ & 100$\%$ \\
            \bottomrule
        \end{tabular}}
    \end{subfigure}
    \caption{{\bf Semantic Consistency.}
    We report the classification accuracy of semantic vertices on the human
    body using the label of the closest primitive. Our predicted primitives are
    consistently used for representing the same human part.}
    \label{fig:semantic_consistency}
    \vspace{-1.2em}
\end{figure}

\boldparagraph{Semantic Consistency}%
We now investigate the ability of our model to decompose 3D objects into semantically consistent parts.
Similar to \cite{Deng2020CVPR}, we use $5$ representative vertex indices provided by SMPL-X~\cite{Pavlakos2019CVPR} (\ie thumbs, toes and nose) and compute the classification accuracy of those points when using the label of the closest primitive.
We compare with CvxNet with $5$ and $50$ primitives and note that Neural Parts are more semantically consistent (see \figref{fig:semantic_consistency}).

\section{Conclusion}

We consider this paper as a step towards bridging the gap between interpretable
and high fidelity primitive-based reconstructions. Our experiments demonstrate
the ability of our model to learn geometrically accurate reconstructions that
preserve the semantic interpretation of each part. In addition, we show that
our expressive primitives outperform existing methods, that rely on simple
shapes, both in terms of accuracy and semanticness. In future work, we plan to
investigate learning primitive-based decompositions without access to target
meshes using differentiable rendering techniques. In addition, we aim to
investigate replacing the sphere with more complex shapes in order to obtain
even more expressive primitives.

\section*{Acknowledgments}

This research was supported by the Max Planck ETH Center for Learning Systems
and an NVIDIA research gift. 

{\small
	\bibliographystyle{ieee_fullname}
	\bibliography{bibliography_long,bibliography,bibliography_custom}
}

\newpage
\clearpage
\appendix
\onecolumn

\begin{minipage}{\textwidth}
   \null
   \vskip .375in
   \begin{center}
      {\Large \bf Supplementary Material for\\ Neural Parts: Learning Expressive 3D Shape Abstractions \\ with Invertible Neural Networks\par}
            \vspace*{24pt}
      {
      \large
      \lineskip .5em
      \begin{tabular}[t]{c}
        Despoina Paschalidou$^{1,5,6}$ \quad Angelos Katharopoulos$^{3,4}$ \quad Andreas Geiger$^{1,2,5}$ \quad Sanja Fidler$^{6,7,8}$\\
        $^1$Max Planck Institute for Intelligent Systems T{\"u}bingen\quad
        $^2$University of T{\"u}bingen\\
        $^3$Idiap Research Institute, Switzerland\quad
        $^4${\'E}cole Polytechique F{\'e}d{\'e}rale de Lausanne (EPFL)\\
        $^5$Max Planck ETH Center for Learning Systems\quad
        $^6$NVIDIA\quad
        $^7$University of Toronto\quad
        $^8$Vector Institute \\
        {\tt\small \{firstname.lastname\}@tue.mpg.de \quad angelos.katharopoulos@idiap.ch \quad sfidler@nvidia.com}
        \vspace*{1pt}
      \end{tabular}
      \par
      }
            \vskip .5em
            \vspace*{12pt}
   \end{center}
\end{minipage}
\begin{abstract}

In this \textbf{supplementary document}, we first present a detailed overview
of our network architecture and the training procedure. We then provide
ablations on how different components of our system impact the performance of
our model on the single-view 3D reconstruction task. Subsequently, we provide
additional experimental results on the D-FAUST~\cite{Bogo2017CVPR}, the
FreiHAND~\cite{Zimmermann2019ICCV} and ShapeNet~\cite{Chang2015ARXIV} dataset.
Next, we provide additional evidence for the representation consistency of our
method. \end{abstract}

\section{Implementation Details}

In this section, we provide a detailed description of our network architecture.
We then describe our training protocol, our sampling strategy and provide
details on the computation of the metrics during training and testing.
Finally, we also provide additional details regarding our baselines.

\subsection{Network Architecture}

Here we describe the architecture of each individual component of our model
(from Fig. $2$ in main submission). Our architecture comprises two main
components: (i) a \emph{feature extractor} that maps the input into a vector of
per-primitive shape embeddings $\{\bC_m\}_{m=1}^M$ and (ii) the
\emph{conditional homeomorphism} that learns a family of homeomorphic mappings
between the sphere and the target shape, conditioned on the shape embedding.

\begin{figure}[!h]
    \centering
    \includegraphics[width=0.8\textwidth]{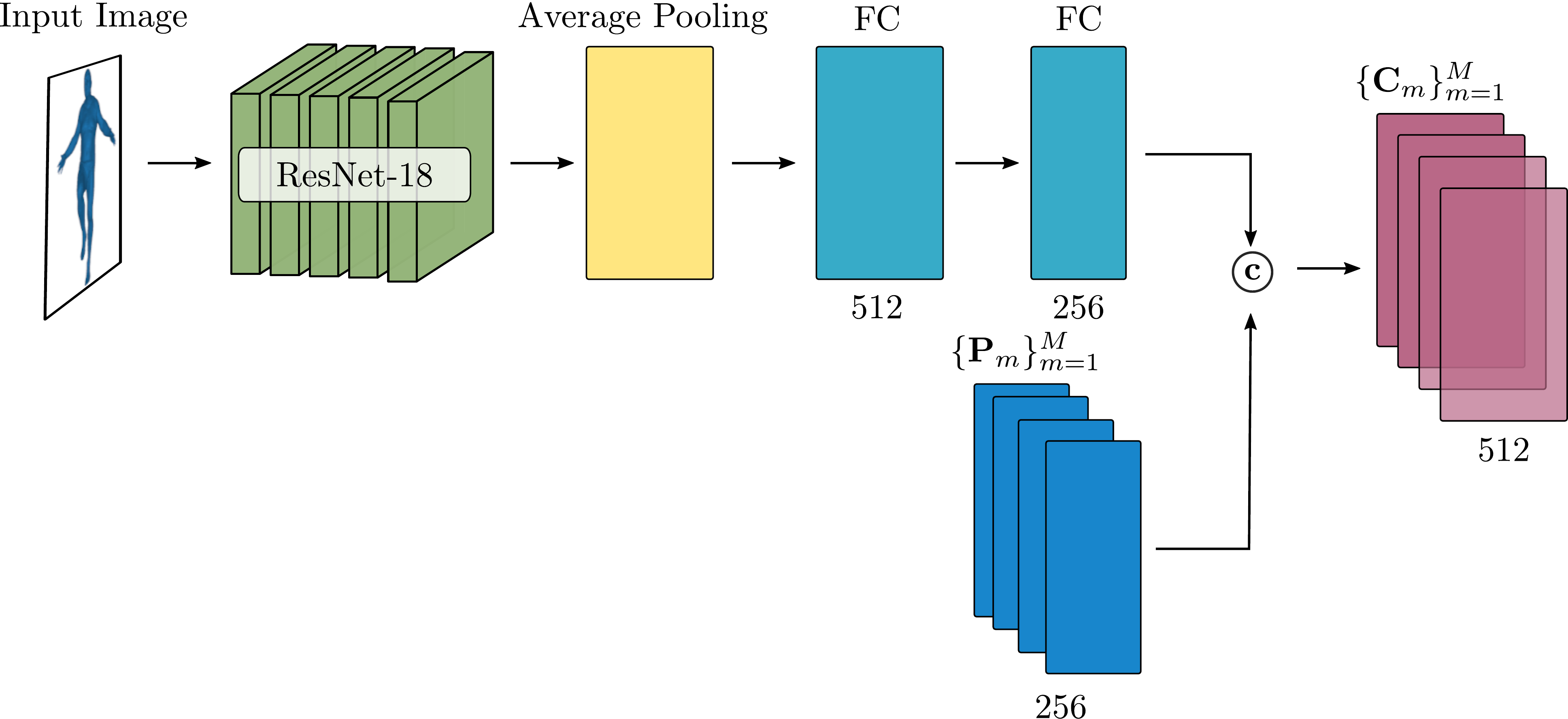}
    \caption{{\bf{Feature Extractor.}}. The feature extractor predicts a vector of per-primitive shape embeddings $\{\bC_m\}_{m=1}^M$ conditioned on the input image. Note that depending on the type of the input (\eg pointcloud, voxel grid), a different feature encoder module needs to be employed.}
    \label{fig:feature_extractor}
    \vspace{-0.8em}
\end{figure}

\boldparagraph{Feature Extractor}%
The first part of the \emph{feature extractor} is the feature encoder. In our
experiments, the feature encoder is implemented with a ResNet-18
architecture~\cite{He2016CVPR} that is pre-trained on
ImageNet~\cite{Deng2009CVPR}.  From the original architecture, we remove the
final fully connected layer and keep only the feature vector of length $512$
after average pooling.
Subsequently, we use $2$ fully connected layers to map this to the global feature vector $\bF\in \mathbb{R}^{256}$.
During training, we use the pre-trained batch statistics for normalization. The
primitive embedding $\{\bP_m\}_{m=1}^M$ is a matrix of size $M \times 256$,
that stores a vector per-primitive index that is concatenated with the global
image feature $\bF$ and outputs a vector of shape embeddings $\{\bC_m \in \mathbb{R}^{512}\}_{m=1}^M$ for every primitive.
A pictorial representation of the feature extractor is provided in \figref{fig:feature_extractor}.

\begin{figure}[!h]
    \centering
    \begin{subfigure}[t]{0.48\columnwidth}
        \centering
        \includegraphics[width=\textwidth]{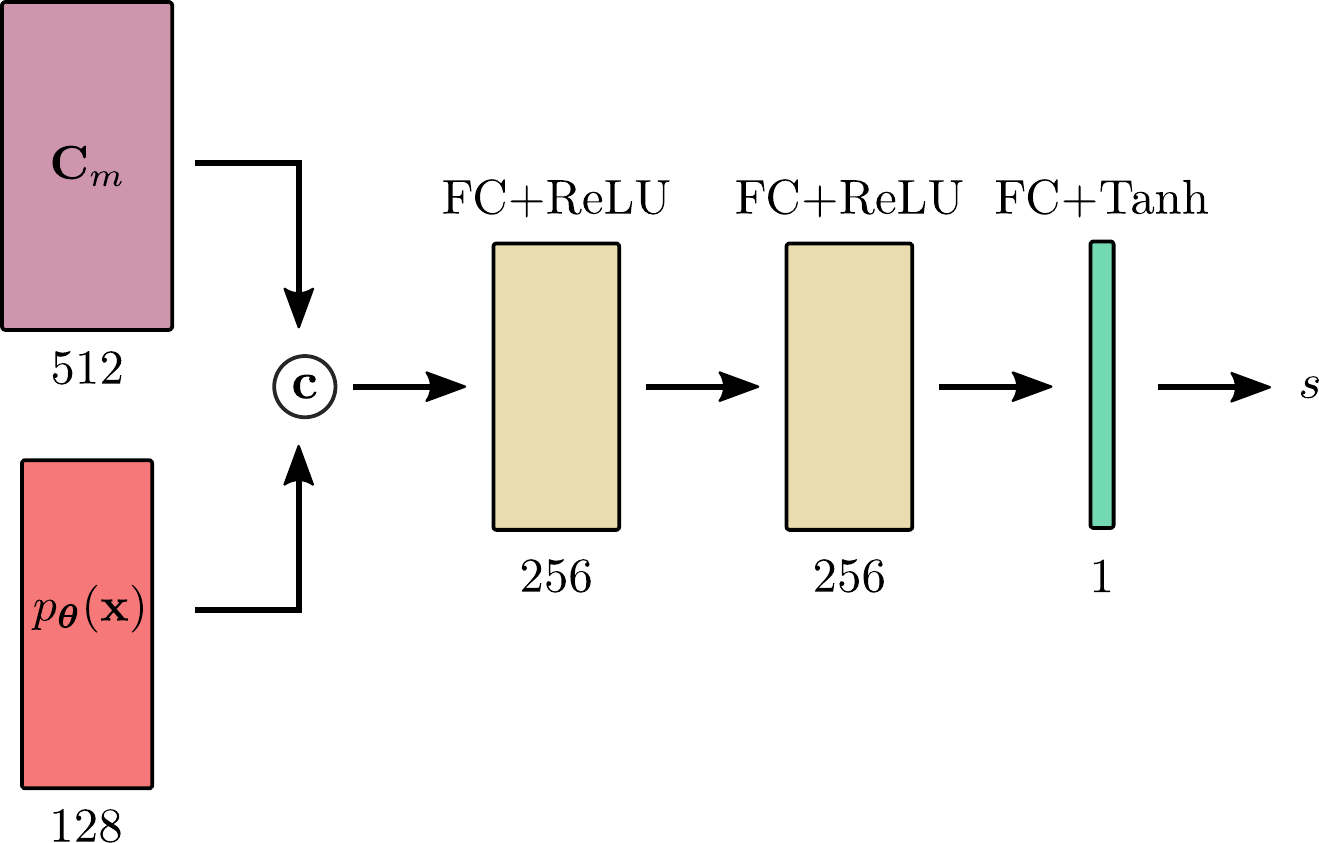}
        \vspace{-3.2em}
        \caption{$s_{\btheta}(\cdot)$ predicts the scaling factor $s$}
        \label{fig:map_s}
    \end{subfigure}%
    \hfill
    \begin{subfigure}[t]{0.48\columnwidth}
        \centering
        \includegraphics[width=\textwidth]{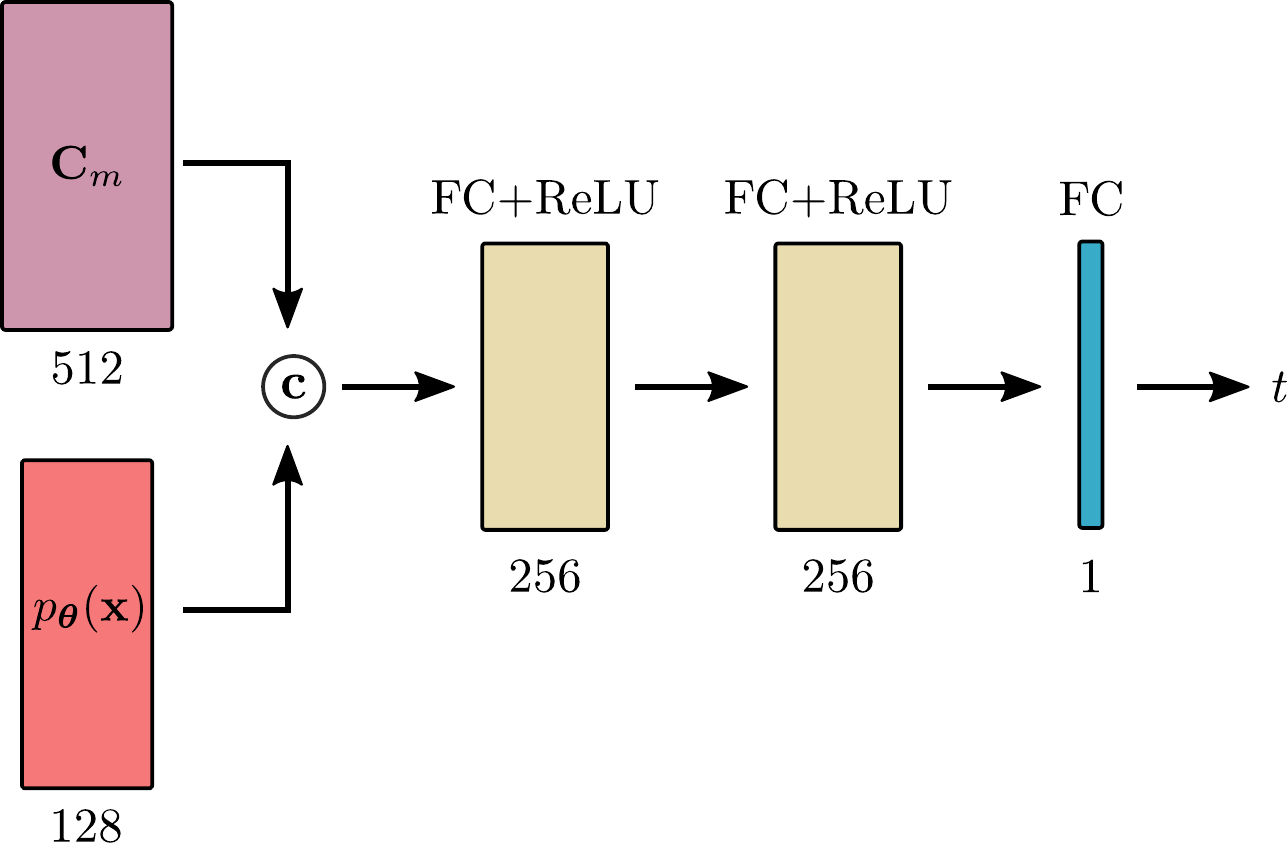}
        \vspace{-3.2em}
        \caption{$t_{\btheta}(\cdot)$ predicts the translation amount $t$}
        \label{fig:map_t}
    \end{subfigure}\\[1.5em]
    \begin{subfigure}[t]{0.80\columnwidth}
        \centering
        \includegraphics[width=\textwidth]{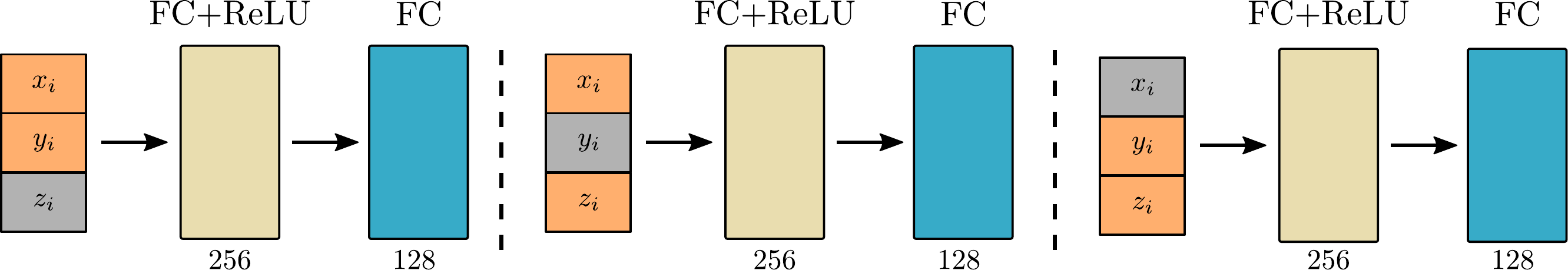}
        \caption{$p_{\btheta}(\cdot)$ maps the input point $(x_i, y_i, z_i)$ into a higher dimensional feature space}
        \label{fig:p_theta}
    \end{subfigure}%
    \caption{{\bf{Conditional Homeomorphism.}} The conditional homeomorphism is the backbone of our architecture that given a vector of per-primitive shape embeddings and a set of points on the sphere in the latent space maps them in the primitive space of the target object and vice versa. Here, we visualize $s_{\btheta}(\cdot)$, $t_{\btheta}(\cdot)$ and $p_{\btheta}(\cdot)$ that comprise our \emph{affine coupling layer}.}
    \label{fig:conditional_homeomorphism_components}
    \vspace{-0.8em}
\end{figure}

\boldparagraph{Conditional Homeomorphism}%
The conditional homeomorphism is implemented with a Real NVP \cite{Dinh2017ICLR} augmented as
described in the main submission. It comprises 4 coupling layers and each
coupling layer consists of three components: $s_{\btheta}(\cdot)$ that predicts
the \textbf{scaling factor} $s$, $t_{\btheta}(\cdot)$ that predicts the \textbf{translation amount} $t$
and $p_{\btheta}(\cdot)$ that maps the input 3D point to a higher dimensional
space before concatenating it with the per-primitive shape embedding $\bC_m$.
In particular, $s_{\btheta}(\cdot)$ is a 3-layer MLP with hidden size equal to
$256$. After each layer we add ReLU non-linearities except for the last layer,
where we use a hard tanh with thresholds $-10$ and $10$ (see
\figref{fig:map_s}). Note that we use the
tanh to avoid numerical instabilities when computing the exponential of the
scaling factor. $t_{\btheta}(\cdot)$ is implemented with the same architecture
apart from the tanh activation in the final layer (see \figref{fig:map_t}).  Finally,
$p_{\btheta}(\cdot)$ is implemented with a 2-layer MLP with ReLU
non-linearities with hidden size $256$ and output size $128$. An illustration of
the components of the conditional homeomorphism is provided in
\figref{fig:conditional_homeomorphism_components}.
As discussed in the main submission, we split the dimensions of the input point
$(x_i, y_i, z_i)$ in order to scale and translate one of them based on the
other two. In practice, this is efficiently implemented via masking one of the
dimensions, as illustrated in \figref{fig:p_theta} with gray.

\subsection{Training Protocol}

In all our experiments, we use the Adam optimizer \cite{Kingma2015ICLR} with learning rate $\eta=10^{-4}$ and no weight decay.
For the other hyperparameters of Adam we use the PyTorch defaults.
Depending on the number of primitives, we train our model with a different batch size and a different number of iterations.
In particular, for $2$ primitives, we use a batch size of
$6$ for $100$k iterations and we aggregate the gradients over $2$ batches.
For the experiments, with $5$ primitives, we use a batch size of $4$ for $150$k
iterations and we again aggregate the gradients over $2$ batches.
For $8$ and $10$ primitives, we use a batch size of $4$ for $200$k iterations
and aggregate the gradients over $4$ batches.  We do not perform any data
augmentation.

We weigh the loss terms of Eq. $11$ in our main submission with $1.0, 0.1, 0.01, 0.1, 0.01$.
Note that we use smaller weights for $\cL_{norm}$ and $\cL_{cover}$ since they
act as regularizers and we want our model to focus primarily on learning
geometrically accurate and semantically meaningful primitives.
The impact of each term is discussed in detail in \secref{subsec:losses}.
The temperature parameter $\tau$ in the occupancy $\cL_{occ}$ and the overlap
loss $\cL_{overlap}$ is set to $4 \times 10^{-3}$. In addition, the $k$ term for the
$\cL_{overlap}$ is set to $1.95$ since we want to penalize overlapping primitives
but ensure that they will be connected. Note that $k=1$ results in disconnected
primitives and $k=2$ allows primitives to overlap in pairs of two. Hence we
select $k=1.95$ as we empirically observe that it balances connectivity and
interpenetration.
Finally, we set the $k$ to be equal to $10$ for the $\cL_{cover}$, as we
empirically observe that it leads to good performance.

\subsection{Sampling Strategy}%
In this section, we discuss our sampling strategy for generating the surface
samples $\cX_t =\{\bx_i, \bn_i\}_{i=1}^N$, the occupancy pairs $\cX_o=\{\bx_i,
o_i\}_{i=1}^V$ and the points on the sphere surface $\cY_s =
\{\by_j\}_{j=1}^K$.

\boldparagraph{Surface Samples}%
We generate samples on the surface of the target mesh by sampling a face with
probability proportional to its area and then sampling a point uniformly in
that face. We use the corresponding face normals to generate point-normal pairs
from the target mesh. During training, we randomly sample $2,000$ points and
normals on the target mesh.

\boldparagraph{Occupancy Pairs}%
We generate occupancy pairs by sampling $100,000$ points uniformly in the unit
cube centered at $(0, 0, 0)$ for each mesh. Subsequently, we compute which of
these points lie inside or outside the mesh to generate the occupancy labels.
However, because the target objects have a small volume compared to the unit
cube, resampling uniformly from the $100,000$ occupancy pairs results in a small
amount of points with positive labels. This yields bad reconstructions for
parts with small volume such as hands and fingers. To address this, we sample
from an unbalanced distribution that, in expectation, results in an equal
number of points with positive and negative labels. However, we also compute
importance sampling weights in order to reweigh our loss and create an unbiased
estimator of the loss with uniform sampling similar to
\cite{Paschalidou2019CVPR}. During training, we sample $5,000$ occupancy pairs.

\boldparagraph{Sphere Points}%
In order to generate points on the surface of each primitive, we sample points
uniformly on the surface of a sphere with radius $r$ by sampling points from a
3-D isotropic Gaussian and normalizing their length to $r$. Note that this does
not generate uniform samples on the surface of the primitive due to different
amounts of stretching and contracting of the sphere by each homeomorphism.
However, we empirically observe that this sampling does not impact negatively
the training of our model, thus we leave sampling uniformly on the surface of
each primitive for future work. During training we sample $200$ points for each
primitive. In addition, to generate meshes for the predicted primitives, we
also need to compute sphere surface points and corresponding faces. We achieve
this using the \emph{UV} parametrization of the sphere~\cite{Christensen1978}.
For visualization purposes we use $2,000$ points sampled uniformly in latitude
and longitude.

\subsection{Metrics}

As mentioned in the main submission, we evaluate our model and our baselines
using the volumetric Intersection-over-Union (IoU) and the Chamfer-$L_1$
distance.  We obtain unbiased low-variance estimates for the IoU by sampling
$100,000$ points from the bounding volume and determining if the points lie
inside or outside the target/predicted mesh.  Similarly, we obtain an unbiased
low-variance estimator of the Chamfer-$L_1$ distance by sampling $10,000$
points on the surface of the target/predicted mesh.

Volumetric IoU is defined as the quotient of the volume of the intersection of the target $O_t$ and the predicted $O_p$ object and the volume of their union.
\begin{equation}
    \text{IoU}(O_t, O_p) = \frac{\mid V(O_t \cap O_p)\mid}{\mid V(O_t \cup O_p)\mid}
\end{equation}
where $V(.)$ is a function that computes the volume of a mesh.

The Chamfer-$L_1$ distance is defined between a set of points sampled on the surface of the target and the predicted mesh.
We denote $\cX=\{\bx_i\}_{i=1}^N$ the set of points sampled on the surface of the target mesh and $\cY=\{\by_i\}_{i=1}^M$ the set of points sampled on the surface of the predicted mesh.
\begin{equation}
    {\text{Chamfer-}L_1}(\cX, \cY) = \frac{1}{N} \sum_{\bx \in \cX}
    \min_{\by \in \cY}\| \bx - \by\| +
    \frac{1}{M} \sum_{\by \in \cY}
    \min_{\bx \in \cX}\| \by - \bx \|
    \label{eq:chamfer}
\end{equation}
The first term of \eqref{eq:chamfer} measures the \textit{completeness} of the predicted shape, namely how far is on average the closest predicted point from a ground-truth point. The second term measures the \textit{accuracy} of the predicted shape, namely how far on average is the closest ground-truth point from a predicted point.
Note that during the evaluation we do not sample points on the sphere to generate
points on the surface of the primitives. Instead we use the predicted
meshes to ensure that the generated points are uniformly distributed on the
surface of the predicted object.

\subsection{Baselines}
In this section, we provide additional details regarding our baselines. For a
fair comparison, all baselines use the same feature extractor network, namely
ResNet18~\cite{He2016CVPR}.

\boldparagraph{SQs}%
In SQs~\cite{Paschalidou2019CVPR}, the authors employ a convolutional neural
network (CNN) to regress the parameters of a set of superquadric surfaces that
best describe the target object, in an unsupervised manner, by minimizing the
Chamfer-distance between points on the target and the predicted shape.  In
order to have predictions, with variable number of primitives, each primitive
is associated with an existence probability.  When the existence probability of
a primitive is below a threshold this primitive is not part of the assembled
object.  We train
SQs~\cite{Paschalidou2019CVPR}\footnote{\href{https://superquadrics.com/}{https://superquadrics.com/}}
using the provide PyTorch~\cite{Paszke2016ARXIV} implementation with the
default parameters.  Similar to our model, we sample $200$ points on the
surface of each primitive and $2,000$ on the surface of the target object. We
train SQs with a batch size of $32$ until convergence.

\boldparagraph{H-SQs}%
H-SQs~\cite{Paschalidou2020CVPR} consider a hierarchical, geometrically more
accurate decomposition of parts using superquadrics.
In particular, H-SQs employ a neural network architecture that recursively
partitions objects into their constituent parts by building a latent vector
that jointly encodes both the part-level hierarchy and the part geometries.
H-SQs are trained in an unsupervised fashion, with an occupancy loss function
such that the model learns to classify whether points sampled in the bounding
box that contains the target object lie insider or outside the union of the
predicted primitives.
We train H-SQs~\cite{Paschalidou2020CVPR} using the PyTorch code provided by
the authors. Note that the maximum number of primitives for H-SQs needs to be
a power of $2$, hence we train with a different number of primitives than the
other baselines. Again, we train H-SQs with a batch size of $32$ until
convergence.

\boldparagraph{CvxNet}%
CvxNet~\cite{Deng2020CVPR} propose to reconstruct a 3D shape as a collection of convex hulls.
We train
CvxNet~\cite{Deng2020CVPR}\footnote{\href{https://cvxnet.github.io/}{https://cvxnet.github.io/}}
using the provided TensorFlow~\cite{Abadi2016ARXIV} code with the default
parameters. To ensure that the evaluation is consistent and fair, we extract
meshes using their codebase and evaluate the predicted primitives using our
evaluation code. For all datasets, we train CvxNet with a batch size of $32$
until convergence using their default parameters.

\boldparagraph{OccNet}%
In our evaluation, we also include
OccNet~\cite{Mescheder2019CVPR}\footnote{\href{https://github.com/autonomousvision/occupancy\_networks}{https://github.com/autonomousvision/occupancy\_networks}}
which we train using the provided code by the authors. In particular, OccNet is
trained with a batch size of $32$ until convergence. While OccNet does not
consider any part-based decomposition of the target object and is not directly
comparable with our model, we include it in our analysis as a typical
representative of powerful implicit shape extraction techniques

\clearpage
\section{Ablation Study}
\label{sec:ablations}

In this section, we investigate how various components of our system affect the performance of Neural Parts on the single-view 3D reconstruction task on the D-FAUST dataset.
In \secref{subsec:invertibility}, we examine the impact of the INN on the performance of our model. Next, in \secref{subsec:projection},
we investigate the impact of $p_{\btheta}(\cdot)$ and in \secref{subsec:losses}, we discuss the effect of each loss term on the overall performance of our model.
Unless stated otherwise, we train all models with $5$ primitives for $300$ epochs.

\subsection{Invertibility}
\label{subsec:invertibility}

In this section, we examine the impact of the Invertible Neural Network (INN) on the performance of Neural Parts.
To this end, we implement a variant of our model that does not consider the inverse mapping of the homeomorphism, namely without the $\phi_{\btheta}^{-1}(\bx)$.
As a result, for this variant it is not possible to compute the union of parts, as formulated  in Eq. $10$ in our main submission, namely we cannot discard points on a primitive's surface that are internal to another primitive when generating points on the surface of the predicted shape.
Moreover, we cannot impose any additional constraints on the predicted primitives.
Therefore, we train this variant using the reconstruction loss between the target $\cX_t$ and the predicted $\cX_{\hat{p}}$ shape, $\cL_{rec}(\cX_t, \cX_{\hat{p}})$ from Eq $12$ in our main submission as follows:
\begin{equation}
    \cL_{rec}(\cX_t, \cX_{\hat{p}}) = \frac{1}{|\cX_t|} \sum_{\bx_i \in \cX_t}
    \min_{\bx_j \in \cX_{\hat{p}}} {\| \bx_i - \bx_j\|}_2^2 \,+
    \frac{1}{|\cX_{\hat{p}}|} \sum_{\bx_j \in \cX_{\hat{p}}}
    \min_{\bx_i \in \cX_t} {\| \bx_i - \bx_j \|}_2^2
    \label{eq:reconstruction_loss_approx}
\end{equation}
where $\cX_{\hat{p}} = \bx \in \bigcup_{m} \cX_p^m$, namely the union of surface points on each primitive.

Furthermore, we introduce an additional baseline that replaces the INN with an MLP. Similar to AtlasNet~\cite{Groueix2018CVPR}, we use a $4$-layer MLP to learn the homeomorphic mappings between the sphere and the target object.
Since the homeomorphism is implemented via an MLP it is not possible to define its inverse mapping. Thus, also this baseline is trained using the loss of \eqref{eq:reconstruction_loss_approx}.
Note that this baseline is slightly different from AtlasNet~\cite{Groueix2018CVPR} because instead of having a single MLP for each homeomorphism, we have one MLP for the $M$ homeomorphisms.
In particular, for this baseline, we implement the decoder architecture proposed in \cite{Groueix2018CVPR}.

\begin{table}[!h]
    \centering
    \begin{tabular}{l||cc|c}
        \toprule
        & w/o $\phi_{\btheta}^{-1}(\bx)$ & AtlasNet - sphere & Ours \\
        \midrule
        IoU & 0.639 & $*$ & 0.673 \\
        Chamfer-$L_1$ & 0.119 & 0.087 & 0.097 \\
        \bottomrule
    \end{tabular}
    \caption{{\bf Ablation Study on INN.} This table shows a quantitative comparison of our approach \wrt a variant of our model that does not consider the inverse mapping of the homeomorphism and a second baseline that replaces the INN with an MLP for learning the homeomorphic mapping. We refer to the latter as AtlasNet - sphere, as it is a variant of the original AtlasNet method~\cite{Groueix2018CVPR}. We note that our model outperforms both baseline both in terms of IoU $(\uparrow)$ and Chamfer-$L_1$ distance $(\downarrow)$.}
    \label{tab:ablation_invertibility}
\end{table}

\begin{figure}[H]
    \begin{subfigure}[t]{0.16\columnwidth}
        \centering
        \caption{w/o $\phi_{\btheta}^{-1}(\bx)$}
        \includegraphics[trim=25 0 25 0,clip,width=\textwidth]{gfx_ablations_invertibility_50002_chicken_wings_00056}
        \label{fig:ablation_inn_1}
    \end{subfigure}%
    \begin{subfigure}[t]{0.16\columnwidth}
        \centering
        \caption{AtlasNet-sphere}
        \includegraphics[trim=25 0 25 0,clip,width=\textwidth]{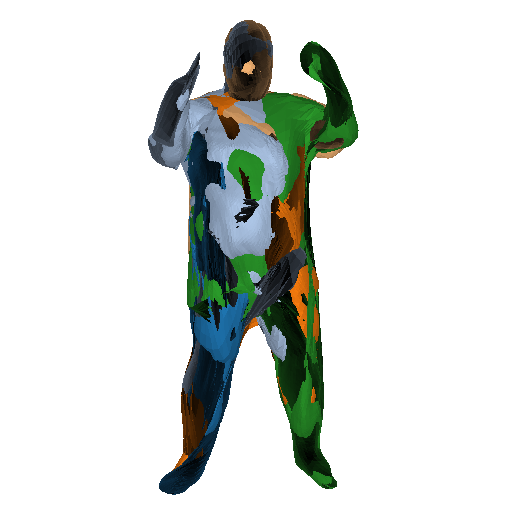}
        \label{fig:ablation_atlasnet}
    \end{subfigure}%
    \begin{subfigure}[t]{0.16\columnwidth}
        \centering
        \caption{Ours}
        \includegraphics[trim=25 0 25 0,clip,width=\textwidth]{gfx_ablations_ours_50002_chicken_wings_00056}
        \label{fig:ablation_ours}
    \end{subfigure}%
    \begin{subfigure}[t]{0.16\columnwidth}
        \centering
        \caption{w/o $\phi_{\btheta}^{-1}(\bx)$}
        \includegraphics[trim=25 0 25 0,clip,width=\textwidth]{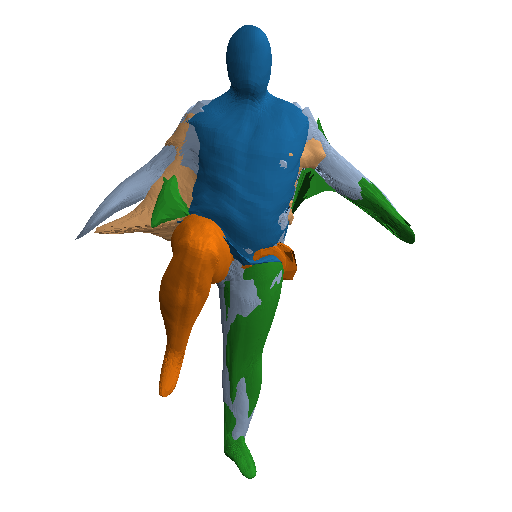}
        \label{fig:ablation_inn_2}
    \end{subfigure}%
    \begin{subfigure}[t]{0.16\columnwidth}
        \centering
        \caption{AtlasNet-sphere}
        \includegraphics[trim=25 0 25 0,clip,width=\textwidth]{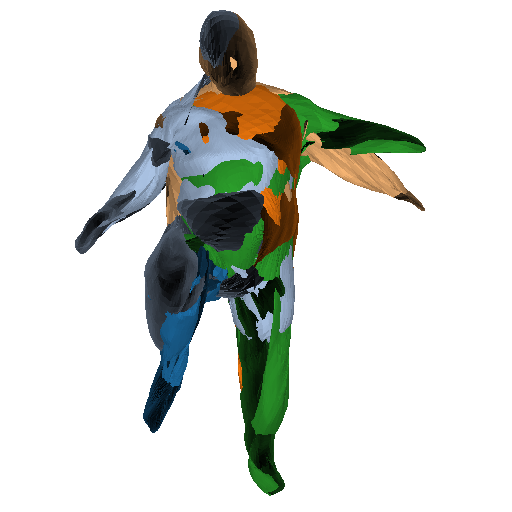}
        \label{fig:ablation_atlasnet}
    \end{subfigure}%
    \begin{subfigure}[t]{0.16\columnwidth}
        \centering
        \caption{Ours}
        \includegraphics[trim=25 0 25 0,clip,width=\textwidth]{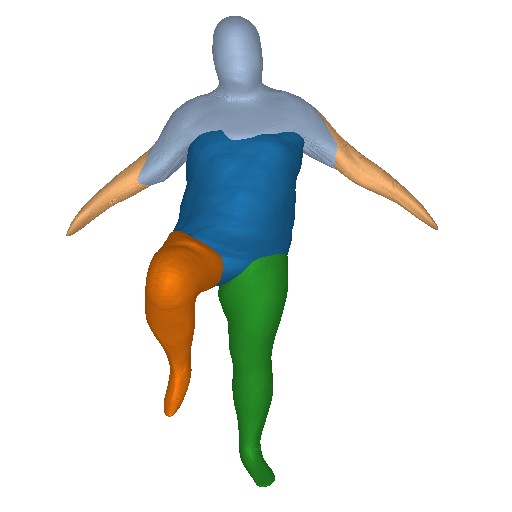}
        \label{fig:ablation_ours}
    \end{subfigure}%
    \caption{{\bf Qualitative evaluation on the impact of the INN.} We visualize the predicted primitives for our model (third and sixth column), a variant of our model that ignores the inverse mapping of the homeomorphism (first and fourth column) and the AtlasNet-sphere baseline that implements the homeomorphism with an MLP.}
    \label{fig:ablation_inn_supp}
\end{figure}
\begin{figure}[!h]
    \begin{subfigure}[b]{0.10\linewidth}
		\centering
		\includegraphics[width=\linewidth]{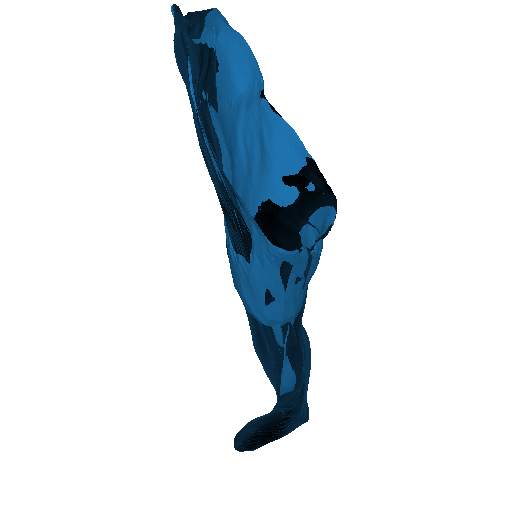}
    \end{subfigure}%
    \begin{subfigure}[b]{0.10\linewidth}
		\centering
		\includegraphics[width=\linewidth]{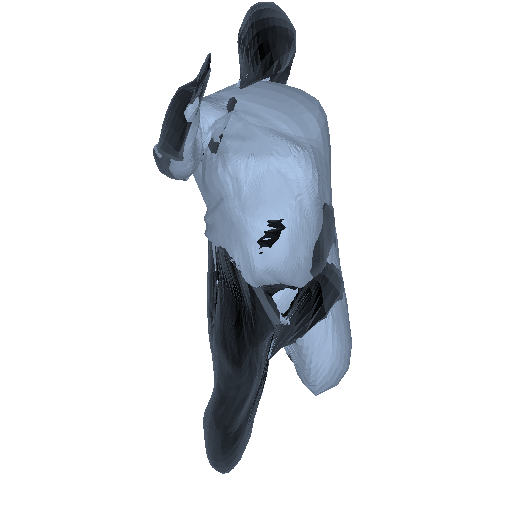}
    \end{subfigure}
    \begin{subfigure}[b]{0.10\linewidth}
		\centering
		\includegraphics[width=\linewidth]{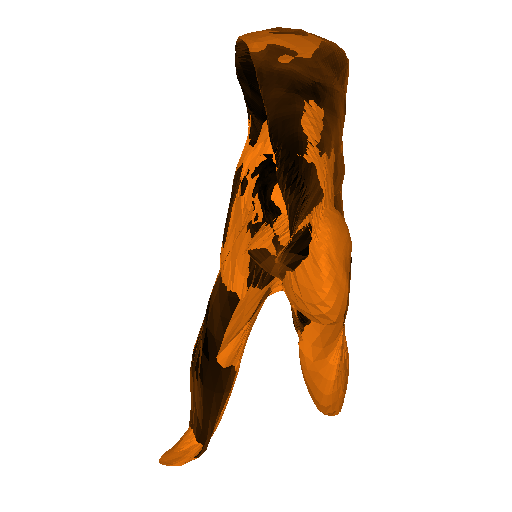}
    \end{subfigure}%
    \begin{subfigure}[b]{0.10\linewidth}
		\centering
		\includegraphics[width=\linewidth]{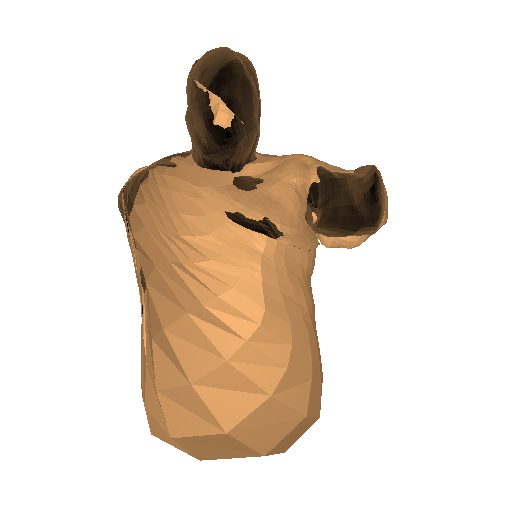}
    \end{subfigure}%
    \begin{subfigure}[b]{0.10\linewidth}
		\centering
		\includegraphics[width=\linewidth]{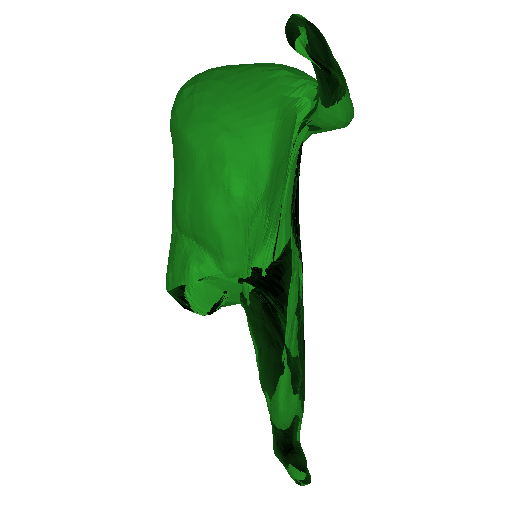}
    \end{subfigure}
    \begin{subfigure}[b]{0.10\linewidth}
		\centering
		\includegraphics[width=\linewidth]{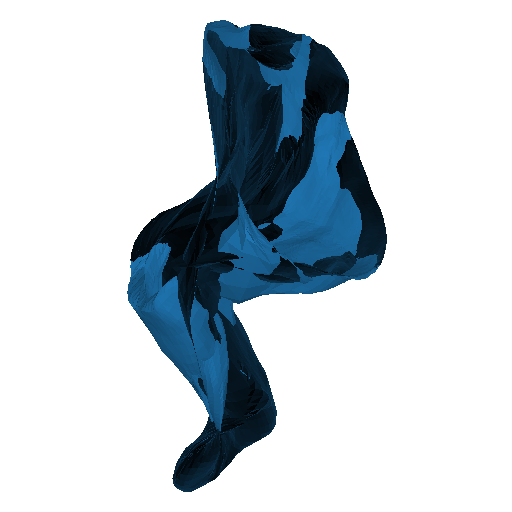}
    \end{subfigure}%
    \begin{subfigure}[b]{0.10\linewidth}
		\centering
		\includegraphics[width=\linewidth]{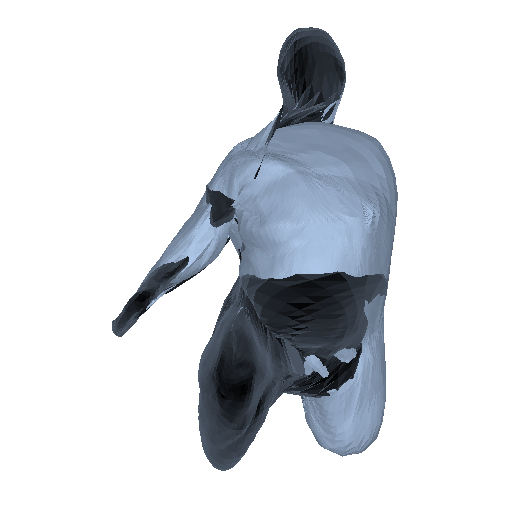}
    \end{subfigure}
    \begin{subfigure}[b]{0.10\linewidth}
		\centering
		\includegraphics[width=\linewidth]{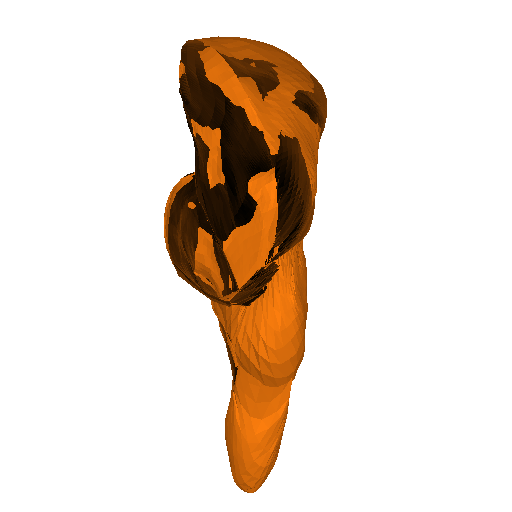}
    \end{subfigure}%
    \begin{subfigure}[b]{0.10\linewidth}
		\centering
		\includegraphics[width=\linewidth]{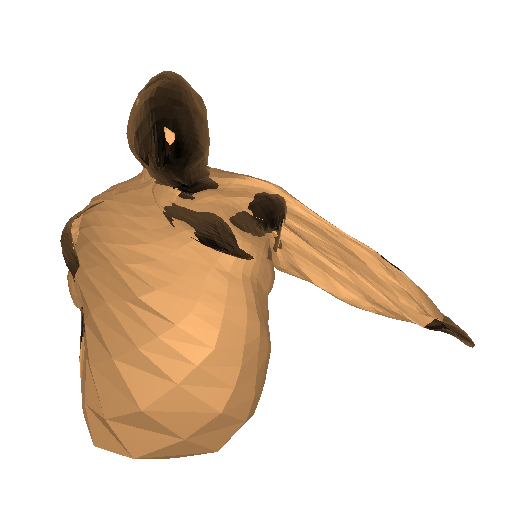}
    \end{subfigure}%
    \begin{subfigure}[b]{0.10\linewidth}
		\centering
		\includegraphics[width=\linewidth]{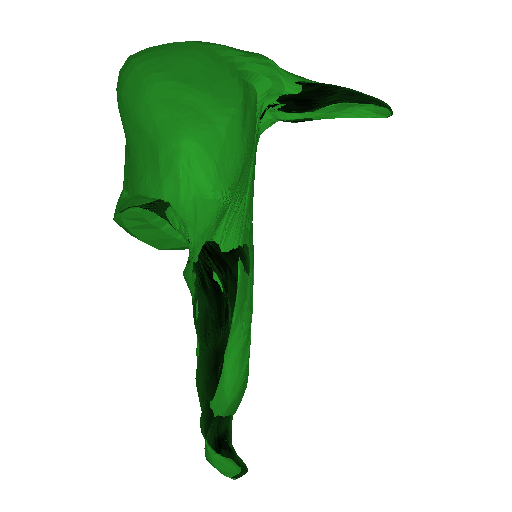}
    \end{subfigure}
    \vskip\baselineskip%
    \vspace{-1.2em}
    \caption{{\bf Predicted primitives of AtlasNet-sphere.} We visualize the individual primitives for the predictions of AtlasNet-sphere in \figref{fig:ablation_inn}. Note that some of the predicted primitives are hollow (first, third, sixth and eighth), some have holes (second, seventh) and all of them have inverted normals.}
    \label{fig:ablations_invertibility_atlasnet_prims}
\end{figure}

\tabref{tab:ablation_invertibility} summarizes the quantitative comparison between our model and the aforementioned baselines.
Note that for the AtlasNet-sphere baseline, we cannot compute the IoU since it is not possible to enforce that the predicted primitives do not degenerate to ``sheets'', which contain no points (zero volume primitives).
Indeed, from our evaluation, we observe that almost all predicted primitives
(see \figref{fig:ablations_invertibility_atlasnet_prims}) have zero volume and
also inverted normals, which means that the sphere has folded on itself
(self-intersections). This is only possible if the MLP is not implementing a
homeomorphism. Due to the aforementioned issues, the AtlasNet-sphere cannot be
used for learning primitives, however, we validate that it achieves high scores
in terms of Chamfer-$L_1$ distance, which is justified as AtlasNet-sphere is
optimized solely on this metric.

On the contrary, our method always learns valid homeomorphisms even when the
inverse mapping ($\phi_{\btheta}^{-1}(\cdot)$) is not used during training.
This becomes evident from \figref{fig:ablation_inn_1}$+$\figref{fig:ablation_inn_2},
where we observe that the predicted primitives do not have inverted normals.
However, without the inverse mapping we cannot enforce that
primitives will not interpenetrate, hence the predicted primitives lack interpretability.

\subsection{Effect of $p_{\btheta}(\cdot)$}
\label{subsec:projection}

\begin{figure}[H]
    \begin{subfigure}[t]{0.5\columnwidth}
        \centering
        \includegraphics[width=\textwidth]{gfx_coupling_layer_coupling_layer}
        \caption{Ours}
        \label{fig:coupling_layer_ours}
    \end{subfigure}%
    \hfill
    \begin{subfigure}[t]{0.45\columnwidth}
        \centering
        \includegraphics[width=\textwidth]{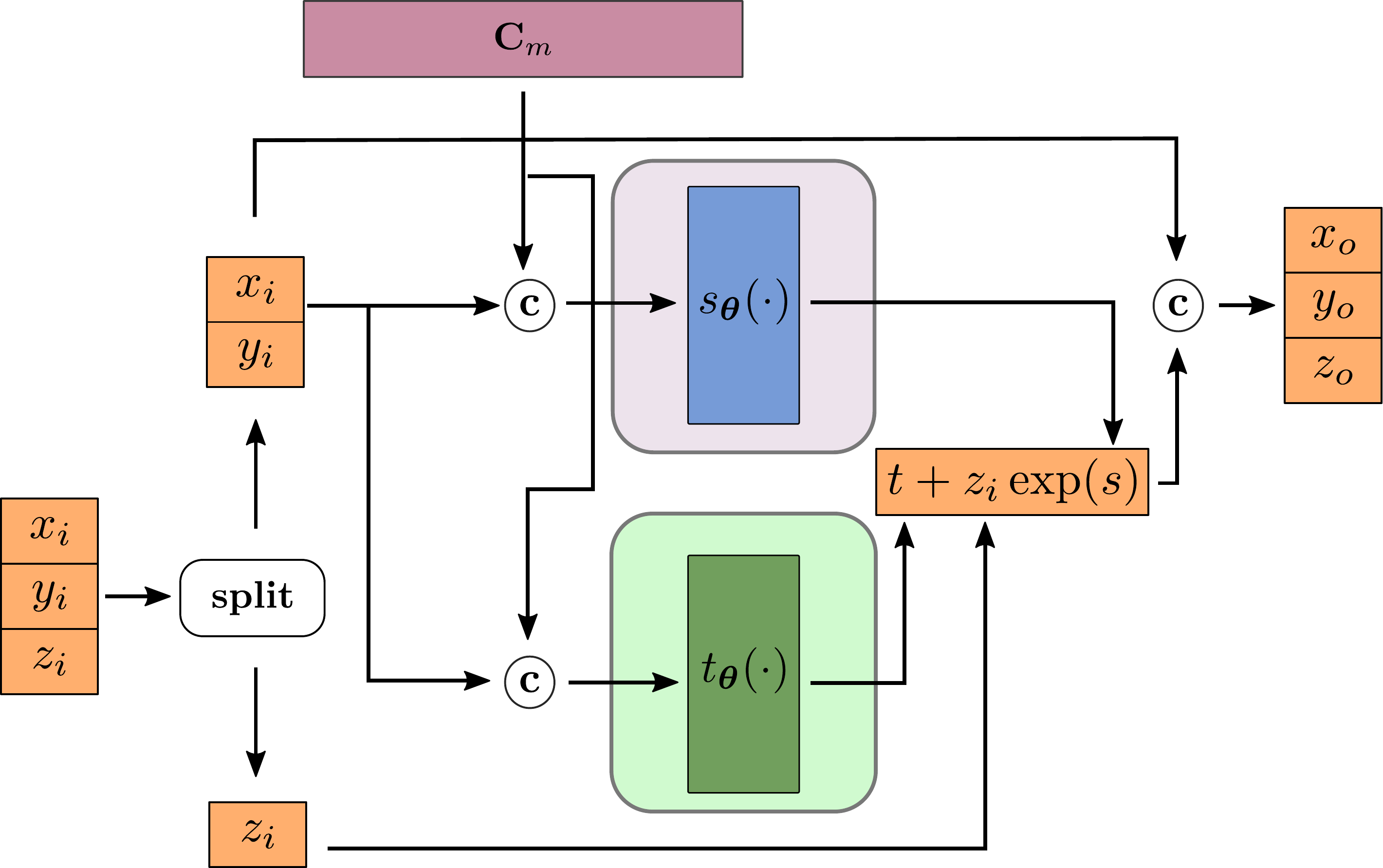}
        \caption{Conditional Coupling Layer w/o $p_{\btheta}(\cdot)$}
        \label{fig:coupling_layer_no_projection}
    \end{subfigure}%
    \caption{{\bf{Conditional Coupling Layer.}} Pictorial representation of our conditional affine coupling layer.
    The input point $(x_i, y_i, z_i)$ is passed into a \emph{coupling layer} that scales and translates one dimension of the input based on the other two and the per-primitive shape embedding $\bC_m$. The scale factor $s$ and the translation amount $t$ are predicted by two MLPs, $s_{\btheta}(\cdot)$ and $t_{\btheta}(\cdot)$. $p_{\btheta}(\cdot)$ is another MLP that increases the dimensionality of the input point before it is concatenated with $\bC_m$.
    }
    \vspace{-0.8em}
\end{figure}

The original Real NVP \cite{Dinh2017ICLR} cannot be directly applied in our setting as it does not consider a shape embedding.
To address this, we augment the affine coupling layer as follows: we first map $(x_i, y_i)$ into a higher dimensional feature vector using a mapping, $p_{\btheta}(\cdot)$, implemented as an MLP. This is done to increase the relative importance of the input point before concatenating it with the high-dimensional shape embedding $\bC_m$. The \emph{conditional affine coupling layer} becomes
\begin{equation}
    \begin{aligned}
        x_o &= x_i \\
        y_o &= y_i \\
        z_o &= z_i \exp\left(s_{\btheta}\left(
                        \left[\bC_m ; p_{\btheta}\left(x_i, y_i\right)\right]
                    \right)\right) + t_{\btheta}\left(
                    \left[\bC_m ; p_{\btheta}\left(x_i, y_i\right)\right]
                 \right)
    \end{aligned}
    \label{eq:conditional_coupling_layer}
\end{equation}
where $[\cdot;\cdot]$ denotes concatenation.
A graphical representation of our conditional coupling layer is provided in \figref{fig:coupling_layer_ours}. In this section, we examine the impact of the mapping $p_{\btheta}(\cdot)$ on the performance of our model.
In particular, instead of first mapping the input point $(x_i, y_i, z_i)$ into a higher dimensional space and then concatenating it with the per-primitive shape embedding $\bC_m$, we concatenate it as is (see \figref{fig:coupling_layer_no_projection}).

\begin{figure*}
    \begin{subfigure}[t]{\linewidth}
        \includegraphics[width=\textwidth]{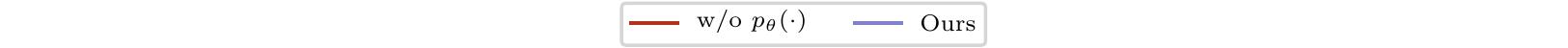}
    \end{subfigure}
    \begin{subfigure}[t]{0.5\linewidth}
        \includegraphics[width=\textwidth]{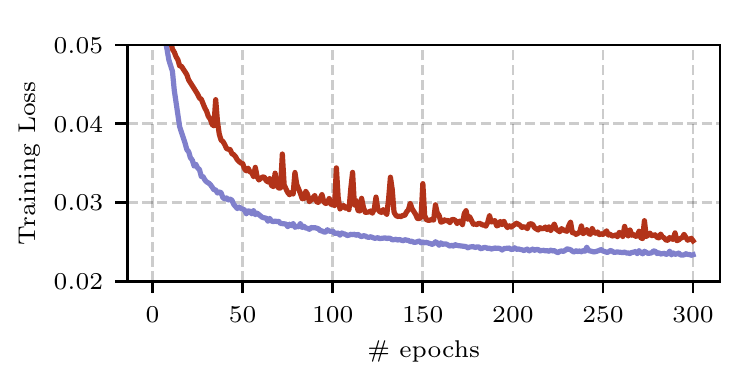}
        \caption{Evolution of Training Loss} \label{fig:convergence_no_proj_loss}
    \end{subfigure}
    \begin{subfigure}[t]{0.5\linewidth}
        \includegraphics[width=\textwidth]{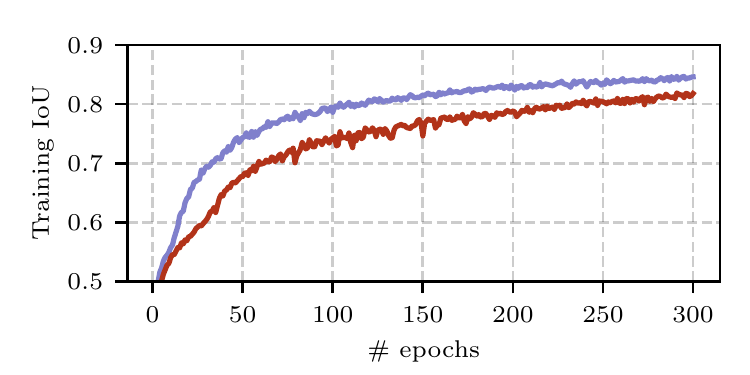}
        \caption{Evolution of Training IoU} \label{fig:convergence_no_proj_iou}
    \end{subfigure}
    \caption{
        {\bf Training Convergence w/o $p_{\btheta}(\cdot)$.}
        We illustrate the training evolution over $300$ epochs in terms of
        training loss (Eq. $11$ in the main submission) and IoU for our method
        with ({\bf \color{ourpurple} purple}) and without $p_{\btheta}(\cdot)$
        ({\bf \color{darkred} red}). We observe that our method converges
        smoother and to a lower loss and higher IoU when using
        $p_{\btheta}(\cdot)$.
    }
    \vspace{-0.8em}
    \label{fig:convergence_no_proj}
\end{figure*}

We observe that removing $p_{\btheta}(\cdot)$ makes it harder for the network
to produce a scaling and translation dependent on the input point due to the
large difference in the number of dimensions compared to the shape embedding
$\bC_m$.  This results in slower convergence in comparison to our full model
(see \figref{fig:convergence_no_proj}).  Moreover, we also notice that the
reconstruction quality becomes worse both in terms of IoU and Chamfer-$L_1$
distance (see \tabref{tab:ablation_projection}).
\begin{table}[!h]
    \centering
    \begin{tabular}{l||c|c}
        \toprule
        & w/o $p_{\btheta}(\cdot)$ & Ours \\
        \midrule
        IoU & 0.638 & 0.673 \\
        Chamfer-$L_1$ & 0.106 & 0.097 \\
        \bottomrule
    \end{tabular}
    \caption{{\bf Ablation Study on $p_{\btheta}(\cdot)$.} This table shows a numeric comparison of our approach \wrt a variant of our model that does not utilize $p_{\btheta}(.)$, namely instead of first mapping the input point $(x_i, y_i, z_i)$ into a higher dimensional space, we directly concatenate the input point with the per-primitive shape embedding $\bC_m$.}
    \label{tab:ablation_projection}
\end{table}

\subsection{Loss Functions}
\label{subsec:losses}

Neural Parts are trained in an unsupervised fashion, without any primitive annotations. To address, the lack of part-level supervision, we learn this task by minimizing the geometric distance between the target and the predicted shape. In particular, our optimization objective comprises of five loss terms that enforce that the predicted primitives are geometrically accurate and semantically meaningful.
\begin{equation}
    \cL = \cL_{rec}(\cX_t, \cX_p) +\, \cL_{occ}(\cX_o) +\, \cL_{norm}(\cX_t) + \cL_{overlap}(\cX_o) + \cL_{cover}(\cX_o).
    \label{eq:total_loss}
\end{equation}
In this section, we discuss how each loss term affects the performance of our model on the single-view 3D reconstruction task. In particular, we train $5$ variants of our model and for each one we omit one of each loss terms. We provide both quantitative (see \tabref{tab:ablation}) and qualitative comparison (see \figref{fig:ablation_supp}) for the $5$ different scenarios.
For a fair comparison all models are trained for the same number of epochs.
\begin{table}[h!]
    \centering
    \begin{tabular}{l||ccccc|c}
        \toprule
        & w/o $\cL_{occ}$ & w/o  $\cL_{rec}$ & w/o $\cL_{norm}$ & w/o $\cL_{overlap}$ & w/o $\cL_{cover}$ & Ours \\
        \midrule
        IoU & 0.642 & 0.643 & 0.669 & 0.670 & 0.668 & 0.673 \\
        Chamfer-$L_1$ & 0.125 & 0.150 & 0.096 & 0.098 & 0.096 & 0.097 \\
        \bottomrule
    \end{tabular}
    \caption{{\bf Ablation Study on Loss Terms.} We investigate the impact of each loss term by training our model without each one of them. We report the IoU $(\uparrow)$ and the Chamfer-$L_1$ distance $(\downarrow)$ on the single-view 3D reconstruction task on D-FAUST dataset.}
    \label{tab:ablation}
\end{table}

{
\renewcommand{\thesubfigure}{\arabic{subfigure}}

\begin{figure}[H]
    \begin{subfigure}[b]{\columnwidth}
        \begin{subfigure}[t]{0.16\columnwidth}
            \centering
            \caption{w/o $\cL_{occ}$}
            \includegraphics[trim=25 0 25 0,clip,width=\textwidth]{gfx_ablations_no_occupancy_50002_chicken_wings_00056}
        \end{subfigure}%
        \begin{subfigure}[t]{0.16\columnwidth}
            \centering
            \caption{w/o  $\cL_{rec}$}
            \includegraphics[trim=25 0 25 0,clip,width=\textwidth]{gfx_ablations_no_chamfer_50002_chicken_wings_00056}
        \end{subfigure}%
        \begin{subfigure}[t]{0.16\columnwidth}
            \centering
            \caption{w/o $\cL_{norm}$}
            \includegraphics[trim=25 0 25 0,clip,width=\textwidth]{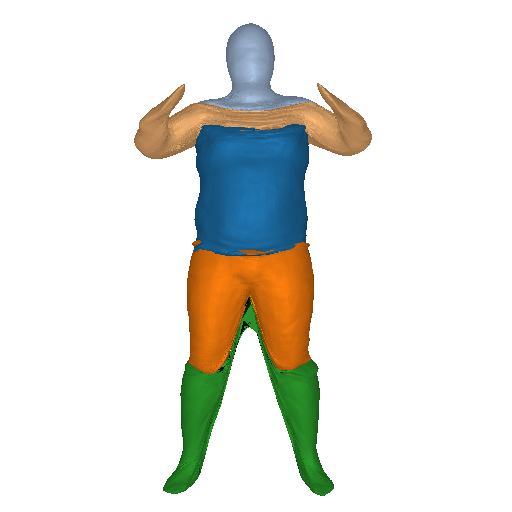}
        \end{subfigure}%
        \begin{subfigure}[t]{0.16\columnwidth}
            \centering
            \caption{w/o $\cL_{overlap}$}
            \includegraphics[trim=25 0 25 0,clip,width=\textwidth]{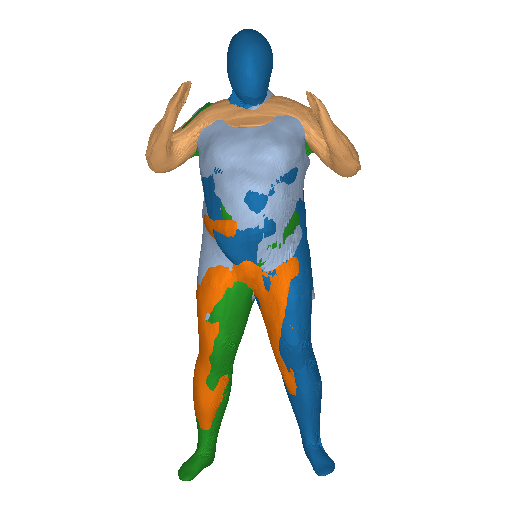}
        \end{subfigure}%
        \begin{subfigure}[t]{0.16\columnwidth}
            \centering
            \caption{w/o $\cL_{cover}$}
            \includegraphics[trim=25 0 25 0,clip,width=\textwidth]{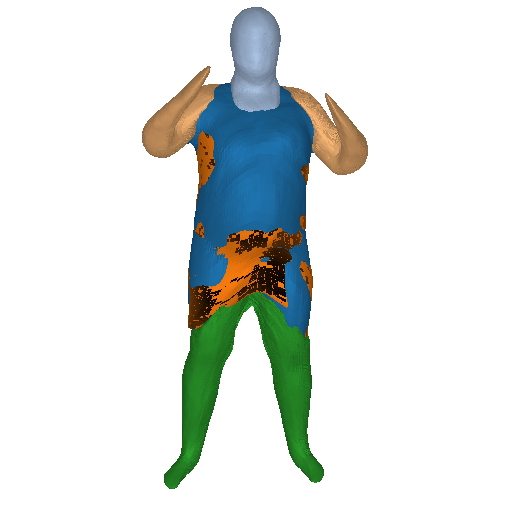}
        \end{subfigure}%
        \begin{subfigure}[t]{0.16\columnwidth}
            \centering
            \caption{Ours}
            \includegraphics[trim=25 0 25 0,clip,width=\textwidth]{gfx_ablations_ours_50002_chicken_wings_00056}
        \end{subfigure}
    \end{subfigure}\\[0em]
    \begin{subfigure}[b]{\columnwidth}
        \begin{subfigure}[t]{0.16\columnwidth}
            \centering
            \caption{w/o $\cL_{occ}$}
            \includegraphics[trim=25 0 25 0,clip,width=\textwidth]{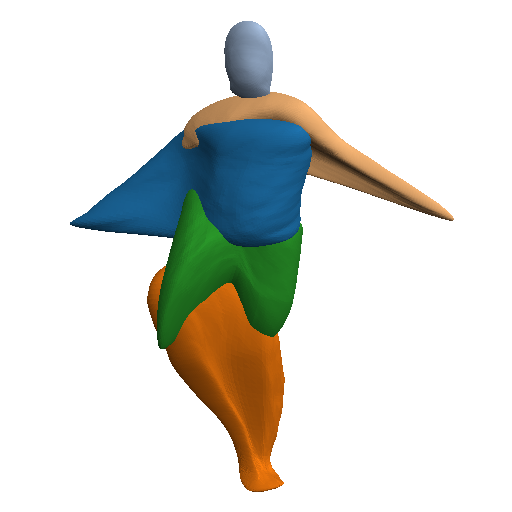}
        \end{subfigure}%
        \begin{subfigure}[t]{0.16\columnwidth}
            \centering
            \caption{w/o  $\cL_{rec}$}
            \includegraphics[trim=25 0 25 0,clip,width=\textwidth]{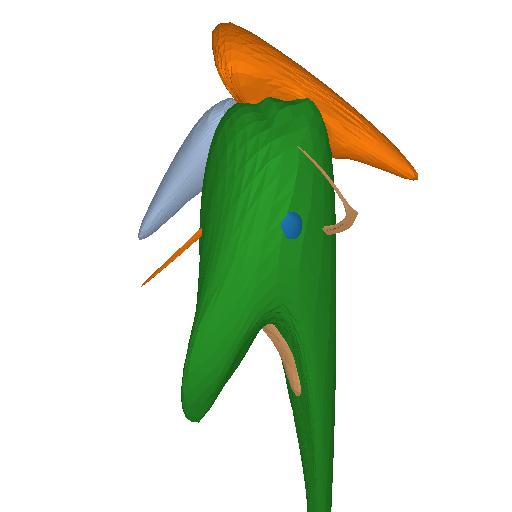}
        \end{subfigure}%
        \begin{subfigure}[t]{0.16\columnwidth}
            \centering
            \caption{w/o $\cL_{norm}$}
            \includegraphics[trim=25 0 25 0,clip,width=\textwidth]{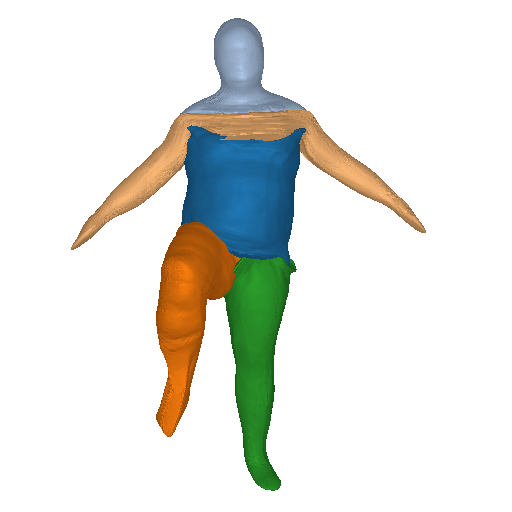}
        \end{subfigure}%
        \begin{subfigure}[t]{0.16\columnwidth}
            \centering
            \caption{w/o $\cL_{overlap}$}
            \includegraphics[trim=25 0 25 0,clip,width=\textwidth]{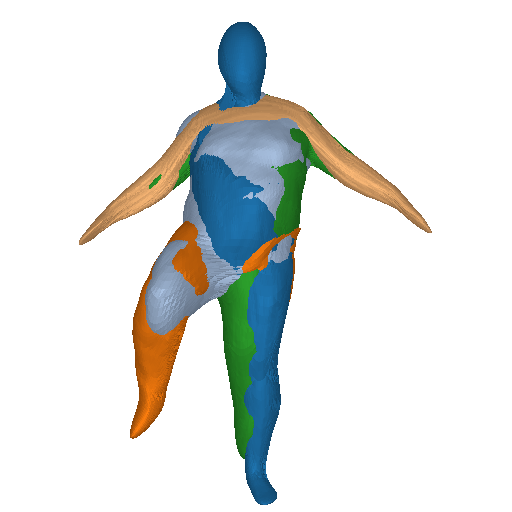}
        \end{subfigure}%
        \begin{subfigure}[t]{0.16\columnwidth}
            \centering
            \caption{w/o $\cL_{cover}$}
            \includegraphics[trim=25 0 25 0,clip,width=\textwidth]{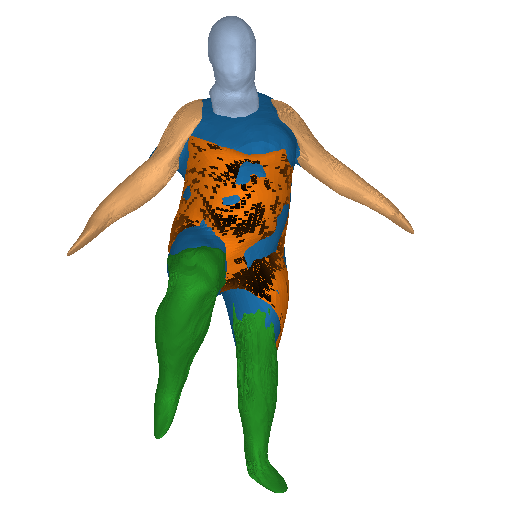}
        \end{subfigure}%
        \begin{subfigure}[t]{0.16\columnwidth}
            \centering
            \caption{Ours}
            \includegraphics[trim=25 0 25 0,clip,width=\textwidth]{gfx_ablations_50002_knees_00046}
        \end{subfigure}
    \end{subfigure}\\[0em]
    \begin{subfigure}[b]{\columnwidth}
        \begin{subfigure}[t]{0.16\columnwidth}
            \centering
            \caption{w/o $\cL_{occ}$}
            \includegraphics[trim=25 0 25 0,clip,width=\textwidth]{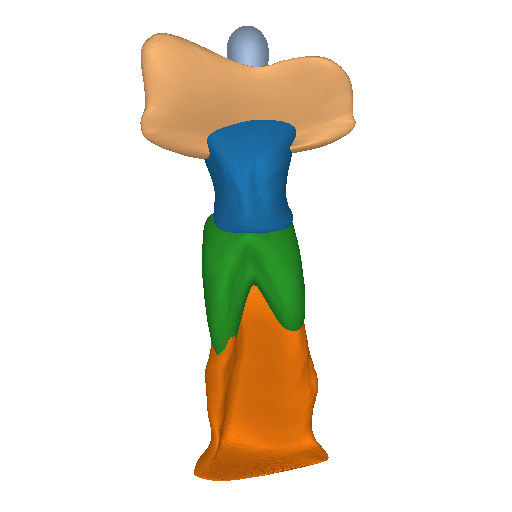}
        \end{subfigure}%
        \begin{subfigure}[t]{0.16\columnwidth}
            \centering
            \caption{w/o  $\cL_{rec}$}
            \includegraphics[trim=25 0 25 0,clip,width=\textwidth]{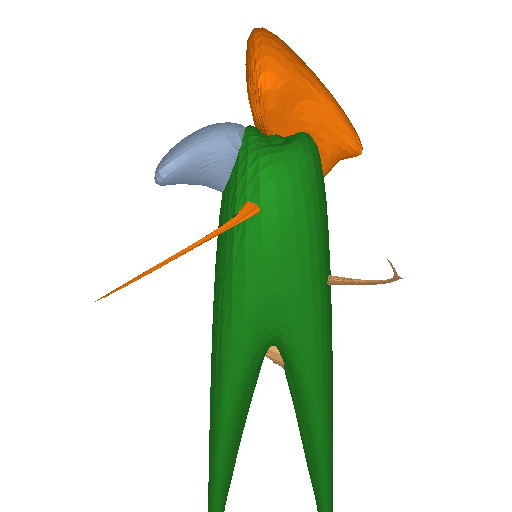}
        \end{subfigure}%
        \begin{subfigure}[t]{0.16\columnwidth}
            \centering
            \caption{w/o $\cL_{norm}$}
            \includegraphics[trim=25 0 25 0,clip,width=\textwidth]{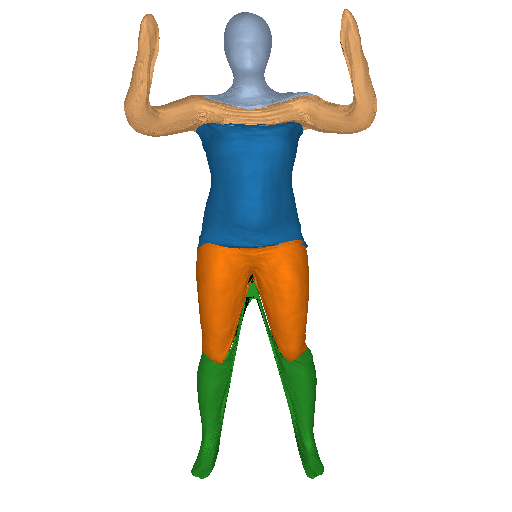}
        \end{subfigure}%
        \begin{subfigure}[t]{0.16\columnwidth}
            \centering
            \caption{w/o $\cL_{overlap}$}
            \includegraphics[trim=25 0 25 0,clip,width=\textwidth]{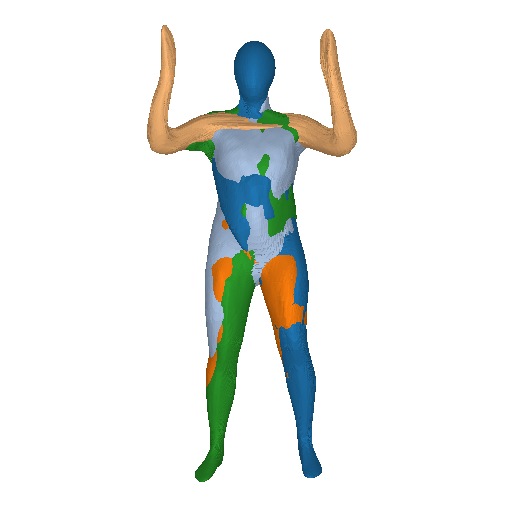}
        \end{subfigure}%
        \begin{subfigure}[t]{0.16\columnwidth}
            \centering
            \caption{w/o $\cL_{cover}$}
            \includegraphics[trim=25 0 25 0,clip,width=\textwidth]{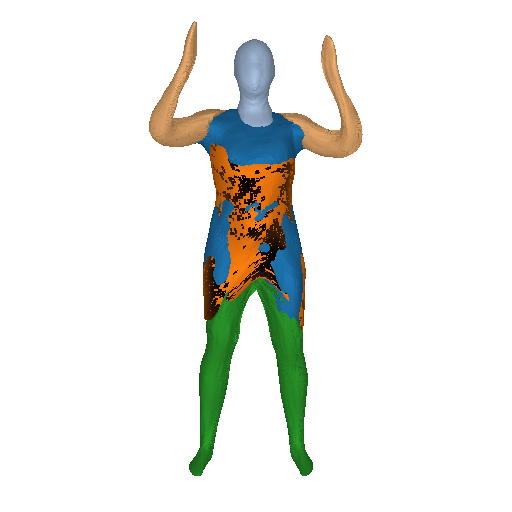}
        \end{subfigure}%
        \begin{subfigure}[t]{0.16\columnwidth}
            \centering
            \caption{Ours}
            \includegraphics[trim=25 0 25 0,clip,width=\textwidth]{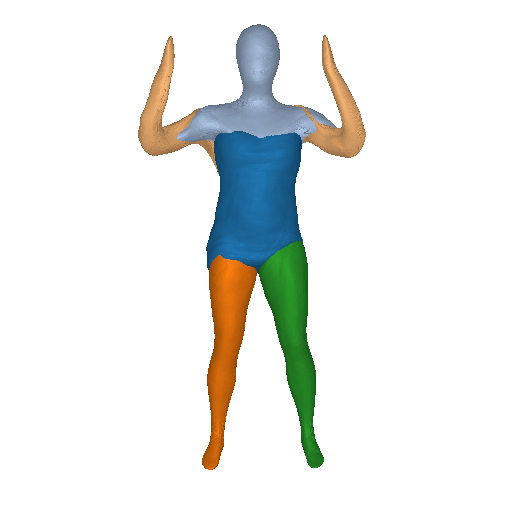}
        \end{subfigure}
    \end{subfigure}\\[0em]
    \begin{subfigure}[b]{\columnwidth}
        \begin{subfigure}[t]{0.16\columnwidth}
            \centering
            \caption{w/o $\cL_{occ}$}
            \includegraphics[trim=25 0 25 0,clip,width=\textwidth]{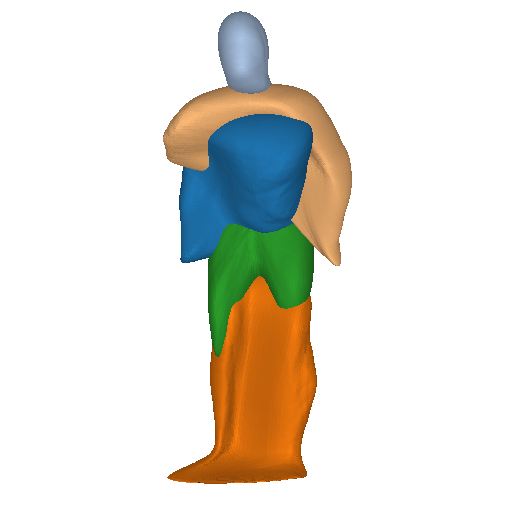}
        \end{subfigure}%
        \begin{subfigure}[t]{0.16\columnwidth}
            \centering
            \caption{w/o  $\cL_{rec}$}
            \includegraphics[trim=25 0 25 0,clip,width=\textwidth]{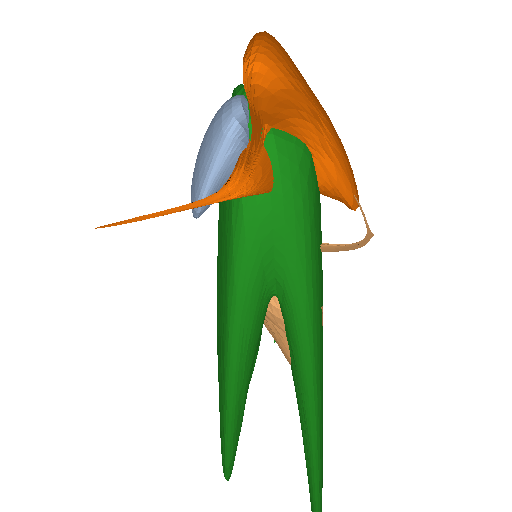}
        \end{subfigure}%
        \begin{subfigure}[t]{0.16\columnwidth}
            \centering
            \caption{w/o $\cL_{norm}$}
            \includegraphics[trim=25 0 25 0,clip,width=\textwidth]{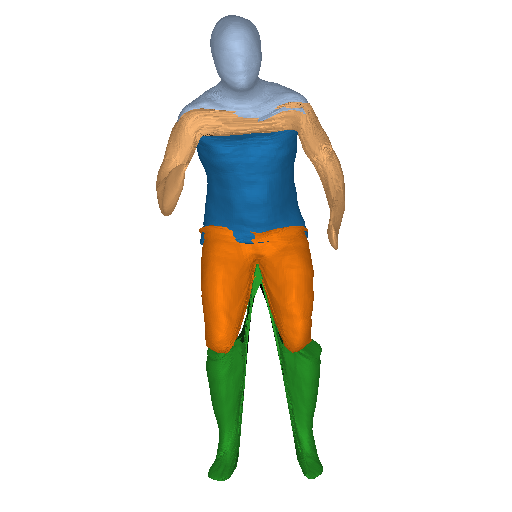}
        \end{subfigure}%
        \begin{subfigure}[t]{0.16\columnwidth}
            \centering
            \caption{w/o $\cL_{overlap}$}
            \includegraphics[trim=25 0 25 0,clip,width=\textwidth]{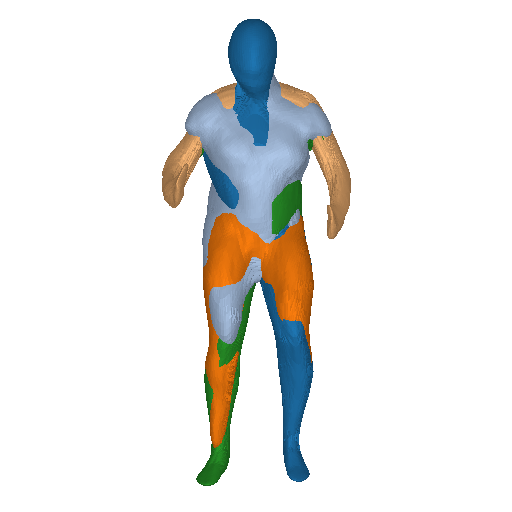}
        \end{subfigure}%
        \begin{subfigure}[t]{0.16\columnwidth}
            \centering
            \caption{w/o $\cL_{cover}$}
            \includegraphics[trim=25 0 25 0,clip,width=\textwidth]{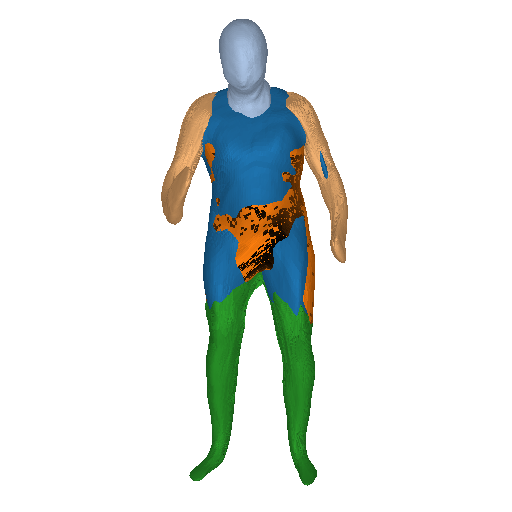}
        \end{subfigure}%
        \begin{subfigure}[t]{0.16\columnwidth}
            \centering
            \caption{Ours}
            \includegraphics[trim=25 0 25 0,clip,width=\textwidth]{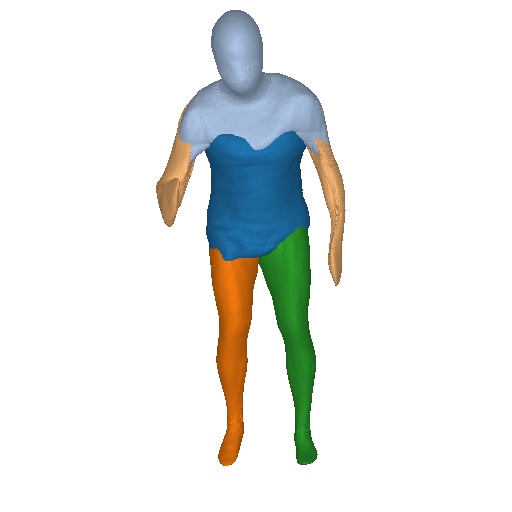}
        \end{subfigure}
    \end{subfigure}\\[0em]
    \begin{subfigure}[b]{\columnwidth}
        \begin{subfigure}[t]{0.16\columnwidth}
            \centering
            \caption{w/o $\cL_{occ}$}
            \includegraphics[trim=25 0 25 0,clip,width=\textwidth]{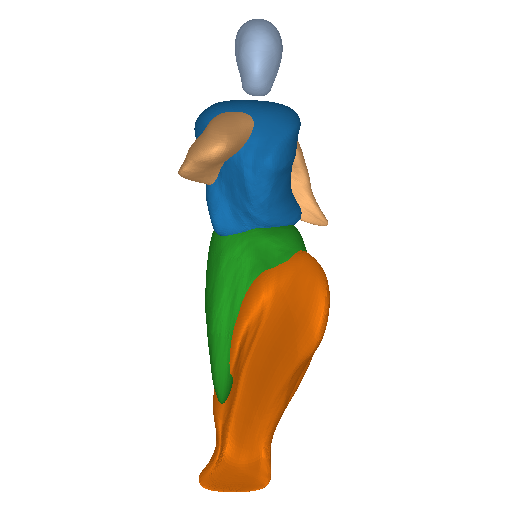}
        \end{subfigure}%
        \begin{subfigure}[t]{0.16\columnwidth}
            \centering
            \caption{w/o  $\cL_{rec}$}
            \includegraphics[trim=25 0 25 0,clip,width=\textwidth]{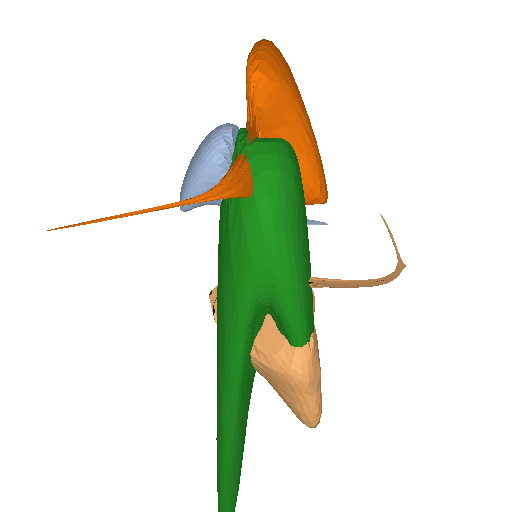}
        \end{subfigure}%
        \begin{subfigure}[t]{0.16\columnwidth}
            \centering
            \caption{w/o $\cL_{norm}$}
            \includegraphics[trim=25 0 25 0,clip,width=\textwidth]{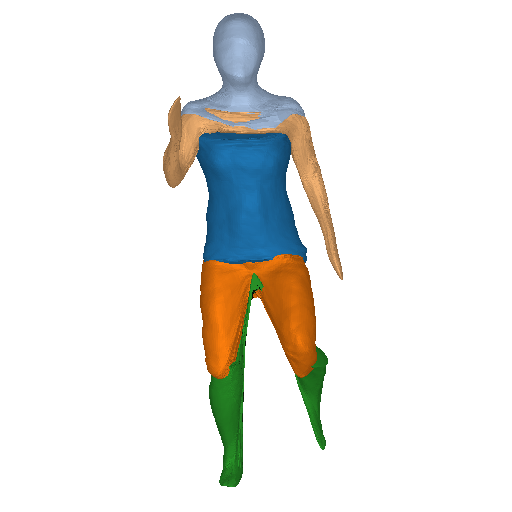}
        \end{subfigure}%
        \begin{subfigure}[t]{0.16\columnwidth}
            \centering
            \caption{w/o $\cL_{overlap}$}
            \includegraphics[trim=25 0 25 0,clip,width=\textwidth]{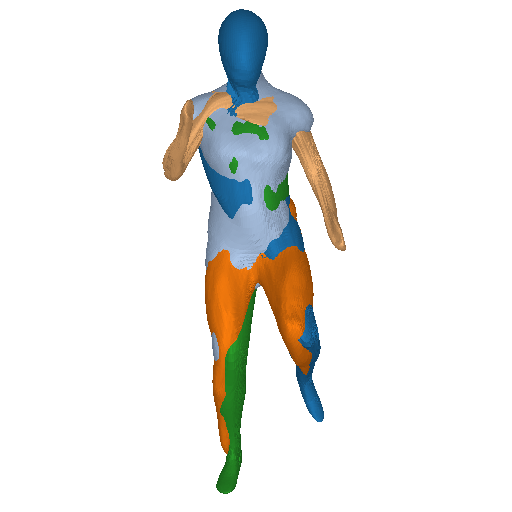}
        \end{subfigure}%
        \begin{subfigure}[t]{0.16\columnwidth}
            \centering
            \caption{w/o $\cL_{cover}$}
            \includegraphics[trim=25 0 25 0,clip,width=\textwidth]{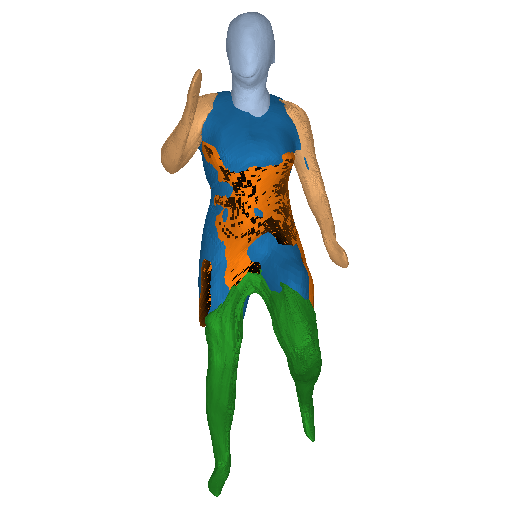}
        \end{subfigure}%
        \begin{subfigure}[t]{0.16\columnwidth}
            \centering
            \caption{Ours}
            \includegraphics[trim=25 0 25 0,clip,width=\textwidth]{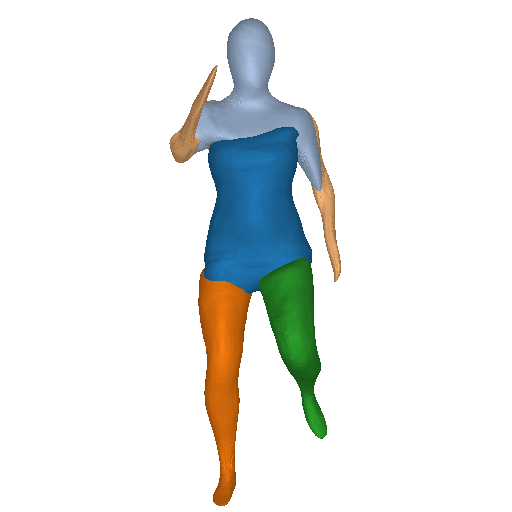}
        \end{subfigure}
    \end{subfigure}\\[0em]
    \caption{{\bf Ablation Study on Loss Terms.} Qualitative evaluation of the impact of the 5 loss terms on the performance of our model. We train Neural Parts, without each loss term and visualize the predicted primitives.}
    \label{fig:ablation_supp}
\end{figure}
}

\boldparagraph{w/o Reconstruction Loss}%
The reconstruction loss is a bidirectional Chamfer loss between the surface points on the target and the predicted shape.
We note that removing $\cL_{rec}(\cX_t, \cX_p)$ results in degenerate primitive arrangements that fail to capture the object geometry.

\boldparagraph{w/o Occupancy Loss}%
The occupancy loss enforces that the free and the occupied space of the predicted and the target object will coincide. Therefore, when we remove $\cL_{occ}(\cX_o)$ from our optimization objective, we observe that the predicted primitives accurately represent geometric parts of the human body but fail to capture the empty space (see hands and legs in first column of \figref{fig:ablation_supp}).
Note that the points on the primitive surface that capture empty space have a small distance from points on the target object and as a result, $\cL_{rec}(\cX_t, \cX_p)$ has small values for these points, which makes it difficult to penalize them for covering unoccupied space.

\begin{figure}[!h]
    \begin{subfigure}[b]{0.48\linewidth}
		\centering
        \caption{w/o $\cL_{normal}$}
		\includegraphics[width=\linewidth]{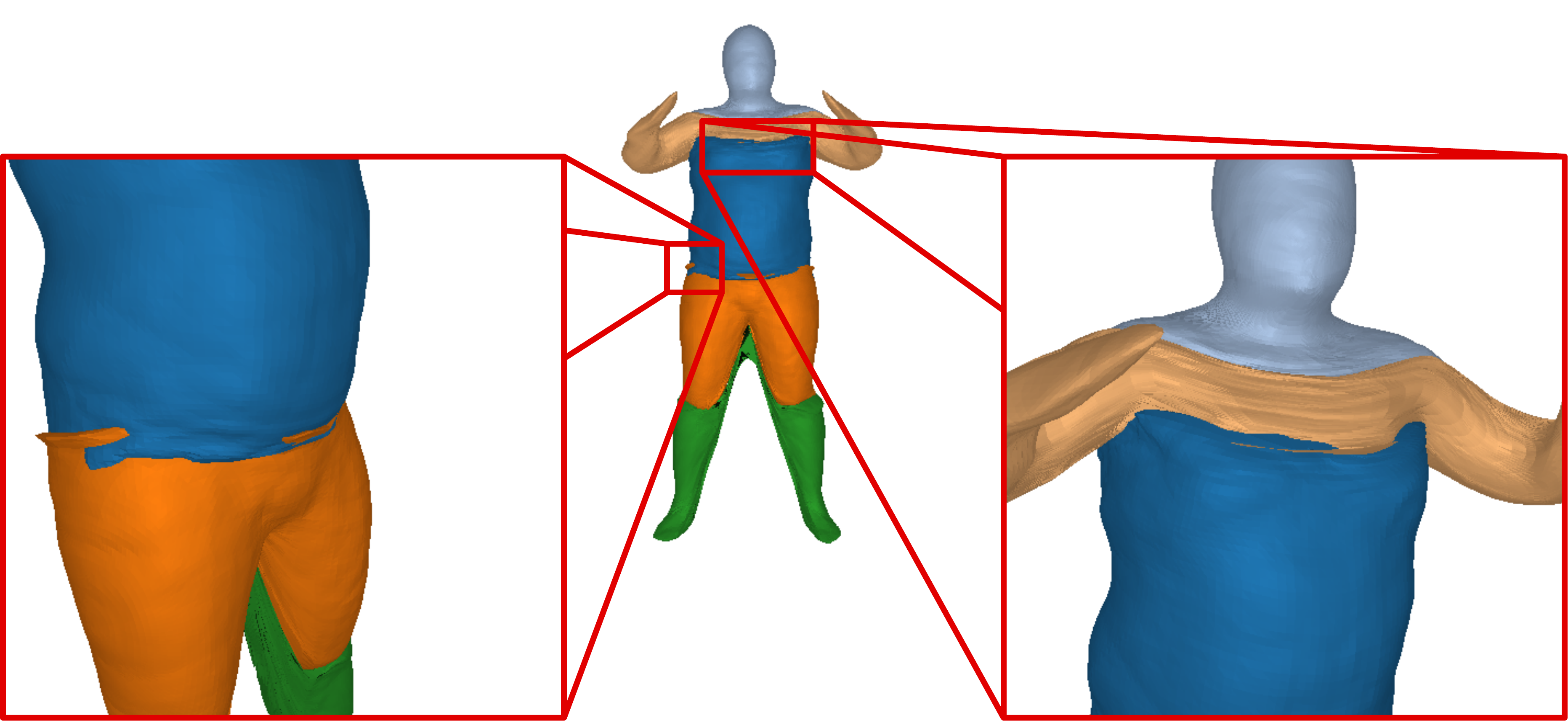}
    \end{subfigure}%
    \hfill%
    \begin{subfigure}[b]{0.48\linewidth}
		\centering
        \caption{Ours}
		\includegraphics[width=\linewidth]{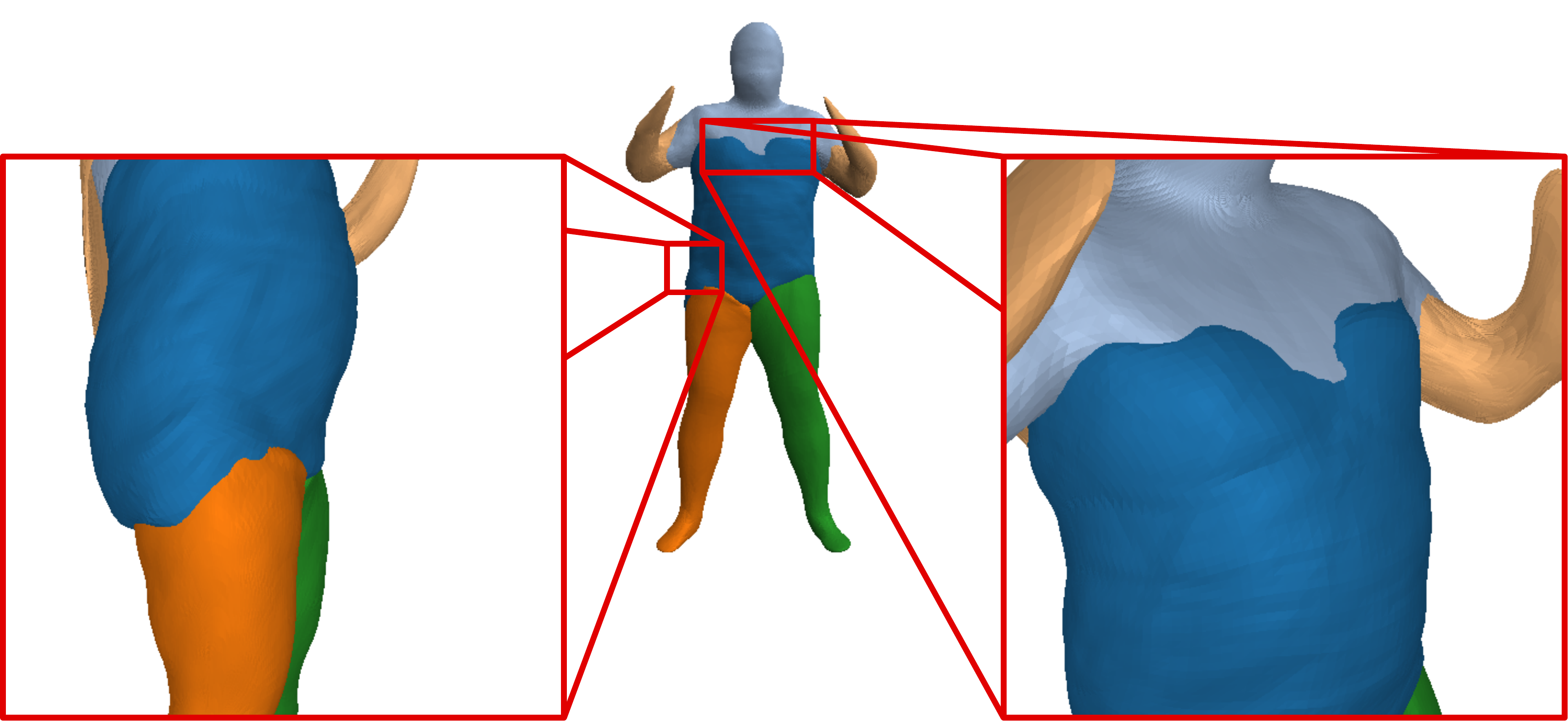}
    \end{subfigure}
    \begin{subfigure}[b]{0.48\linewidth}
		\centering
        \caption{w/o $\cL_{normal}$}
		\includegraphics[width=\linewidth]{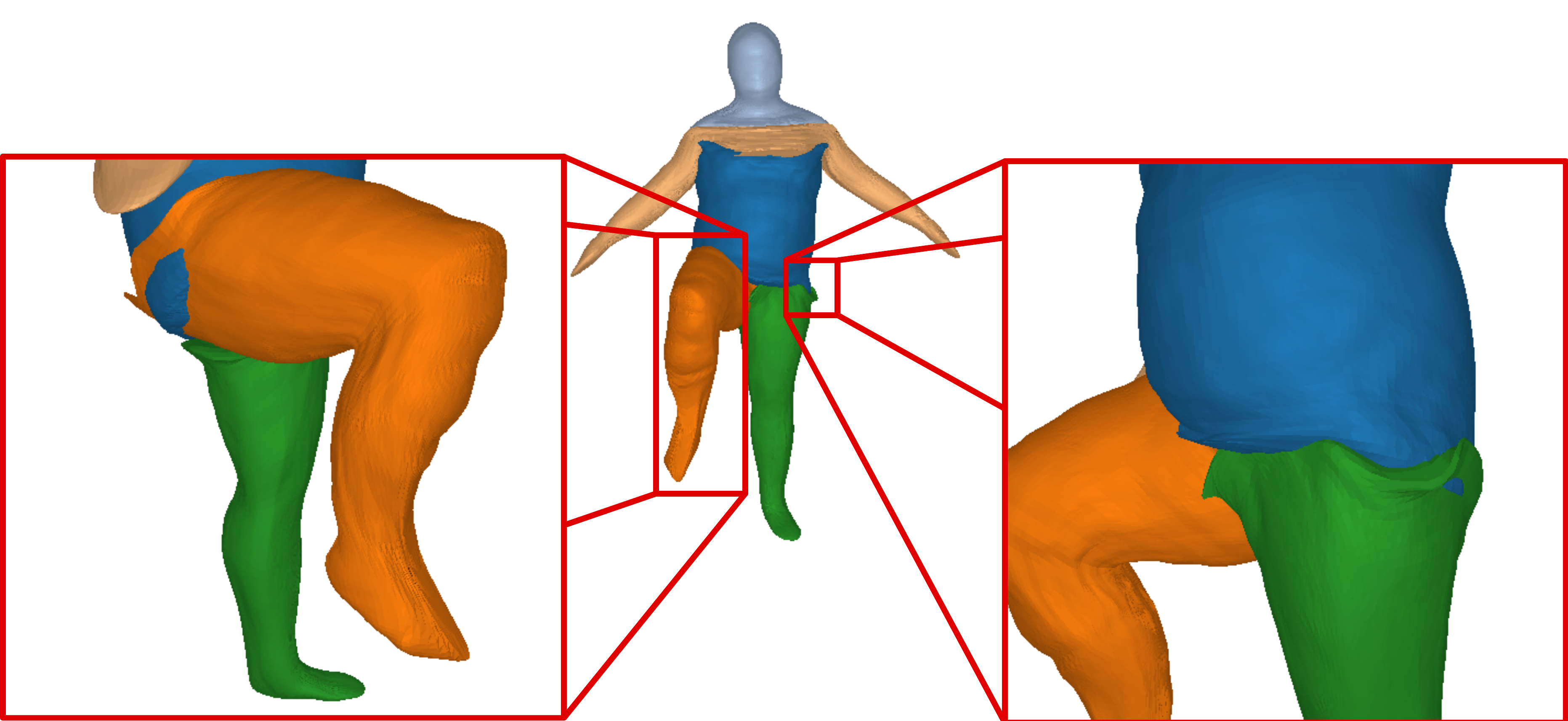}
    \end{subfigure}%
    \hfill%
    \begin{subfigure}[b]{0.48\linewidth}
		\centering
        \caption{Ours}
		\includegraphics[width=\linewidth]{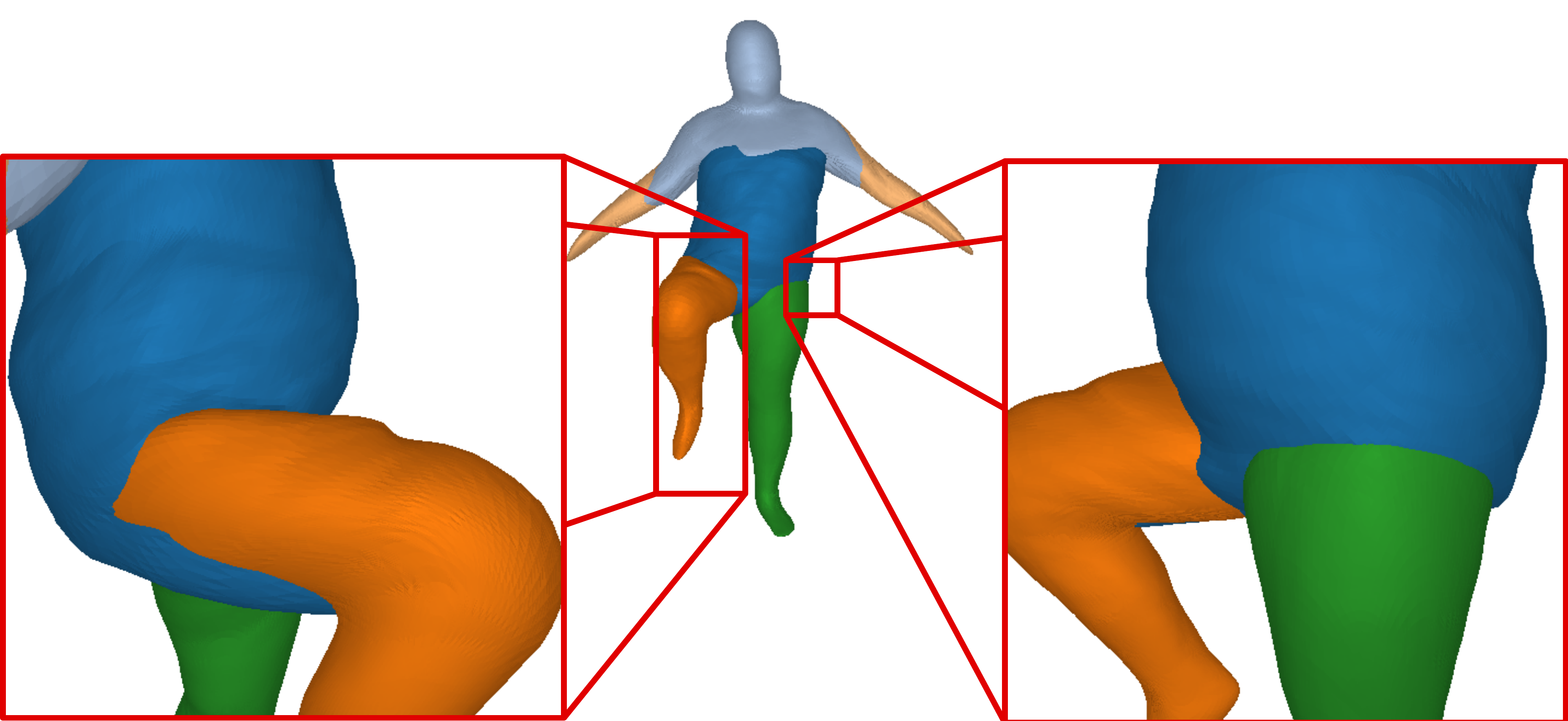}
    \end{subfigure}
    \vskip\baselineskip%
    \vspace{-1.2em}
    \caption{{\bf Impact of $\cL_{norm}$.} When removing the $\cL_{norm}$ from the optimization objective, the predicted primitives are semantically meaningful but the transitions between one primitive to another are not smooth \ie do not have consistent normals. This becomes more evident when looking at the locations where primitives meet (marked with a red rectangle). In contrast, our model has smooth transitions and looks more ``organic''.}
    \label{fig:normal_loss_comparison}
\end{figure}

\begin{figure}[!h]
    \begin{subfigure}[b]{0.166\linewidth}
		\centering
		\includegraphics[width=\linewidth]{gfx_ablations_no_normal_50002_chicken_wings_00056}
    \end{subfigure}%
    \hfill%
    \begin{subfigure}[b]{0.166\linewidth}
		\centering
		\includegraphics[width=\linewidth]{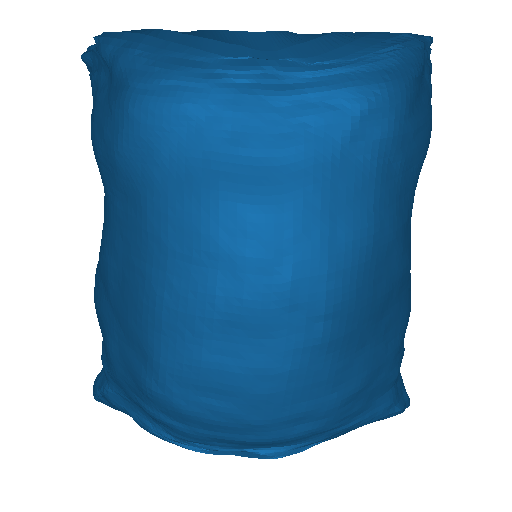}
    \end{subfigure}%
    \hfill%
    \begin{subfigure}[b]{0.166\linewidth}
		\centering
		\includegraphics[width=\linewidth]{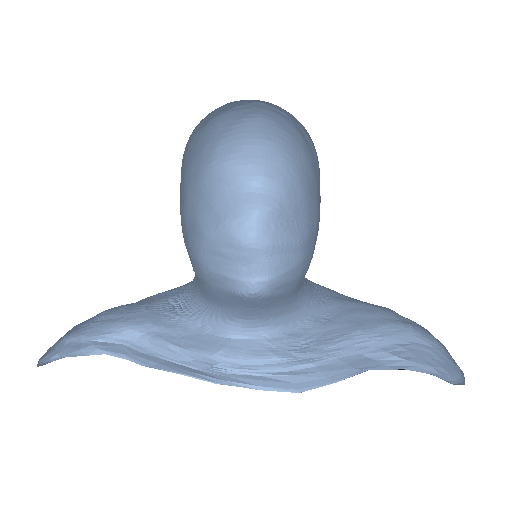}
    \end{subfigure}%
    \hfill%
    \begin{subfigure}[b]{0.166\linewidth}
		\centering
		\includegraphics[width=\linewidth]{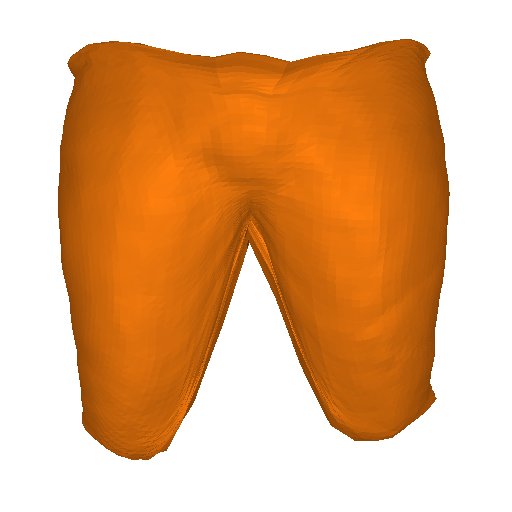}
    \end{subfigure}%
    \hfill%
    \begin{subfigure}[b]{0.166\linewidth}
		\centering
		\includegraphics[width=\linewidth]{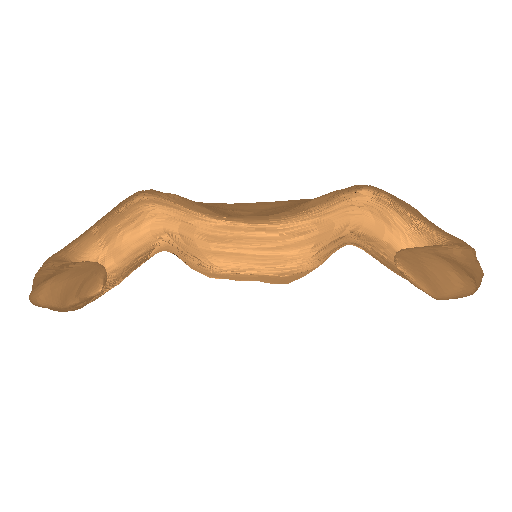}
    \end{subfigure}%
    \hfill%
    \begin{subfigure}[b]{0.166\linewidth}
		\centering
		\includegraphics[width=\linewidth]{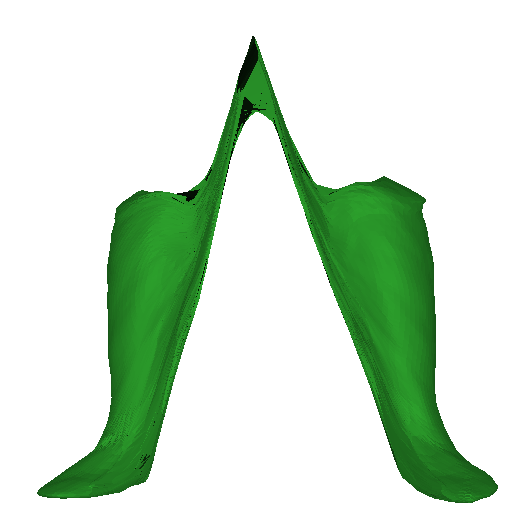}
    \end{subfigure}\\[0.2em]
    \begin{subfigure}[b]{0.166\linewidth}
		\centering
		\includegraphics[width=\linewidth]{gfx_ablations_no_normal_50002_knees_00046}
    \end{subfigure}%
    \hfill%
    \begin{subfigure}[b]{0.166\linewidth}
		\centering
		\includegraphics[width=\linewidth]{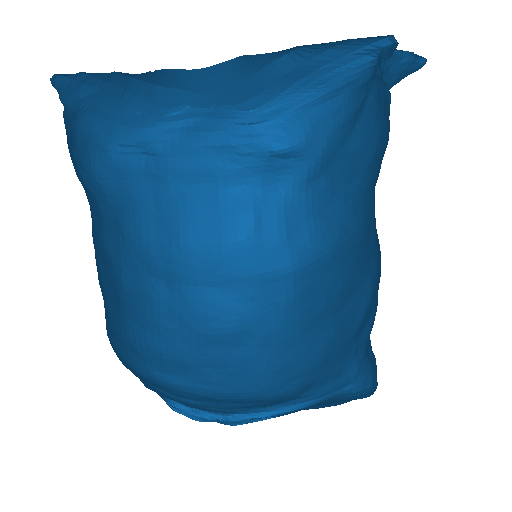}
    \end{subfigure}%
    \hfill%
    \begin{subfigure}[b]{0.166\linewidth}
		\centering
		\includegraphics[width=\linewidth]{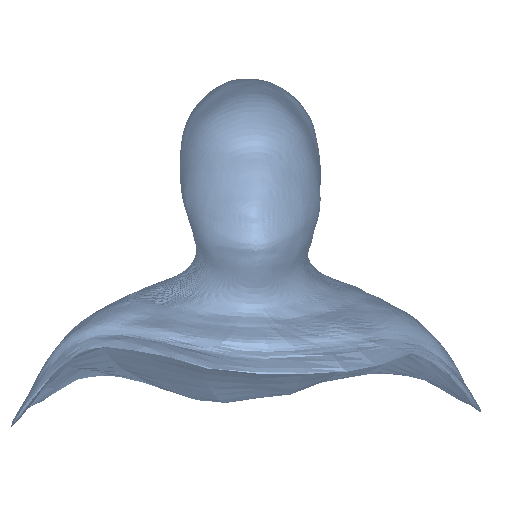}
    \end{subfigure}%
    \hfill%
    \begin{subfigure}[b]{0.166\linewidth}
		\centering
		\includegraphics[width=\linewidth]{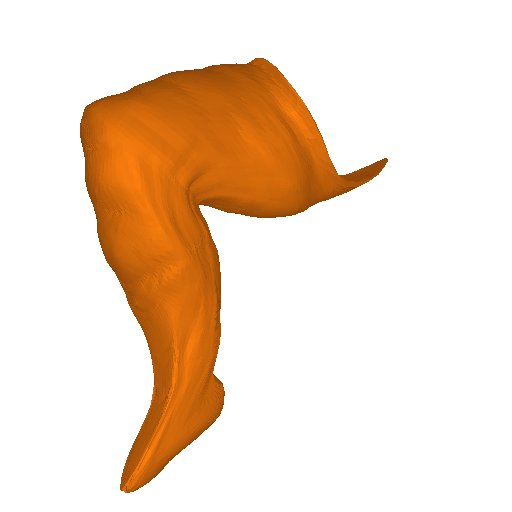}
    \end{subfigure}%
    \hfill%
    \begin{subfigure}[b]{0.166\linewidth}
		\centering
		\includegraphics[width=\linewidth]{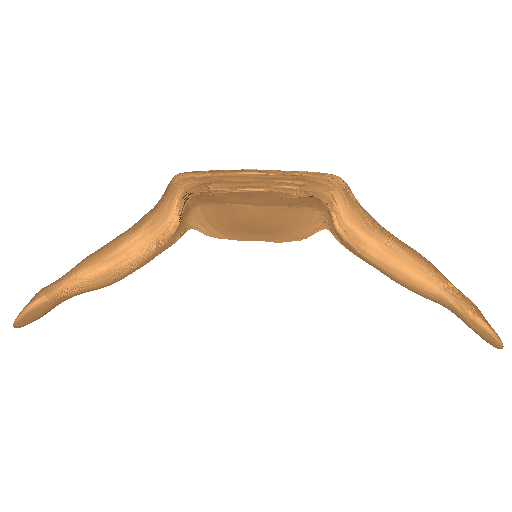}
    \end{subfigure}%
    \hfill%
    \begin{subfigure}[b]{0.166\linewidth}
		\centering
		\includegraphics[width=\linewidth]{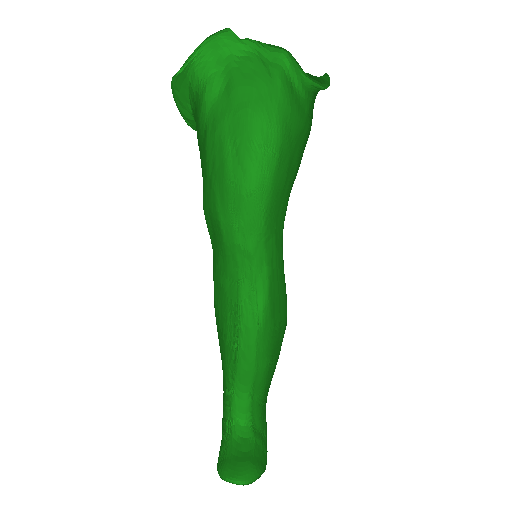}
    \end{subfigure}\\[0.2em]
    \begin{subfigure}[b]{0.166\linewidth}
		\centering
		\includegraphics[width=\linewidth]{gfx_ablations_no_normal_50020_running_on_spot_00053}
    \end{subfigure}%
    \hfill%
    \begin{subfigure}[b]{0.166\linewidth}
		\centering
		\includegraphics[width=\linewidth]{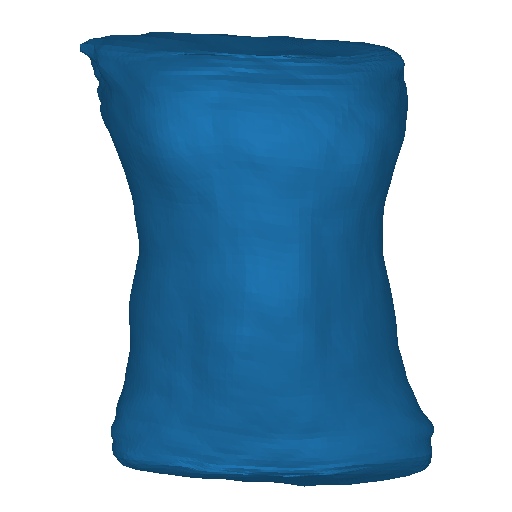}
    \end{subfigure}%
    \hfill%
    \begin{subfigure}[b]{0.166\linewidth}
		\centering
		\includegraphics[width=\linewidth]{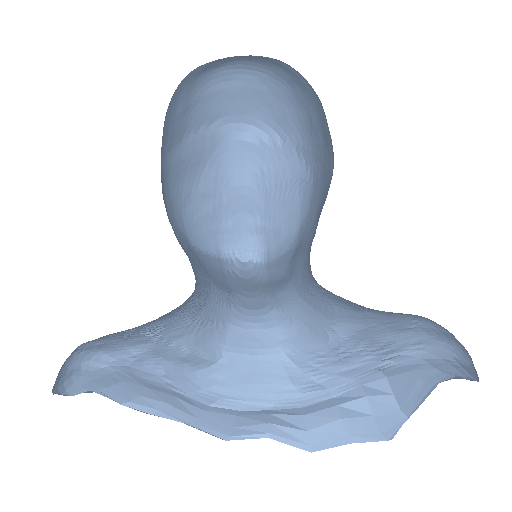}
    \end{subfigure}%
    \hfill%
    \begin{subfigure}[b]{0.166\linewidth}
		\centering
		\includegraphics[width=\linewidth]{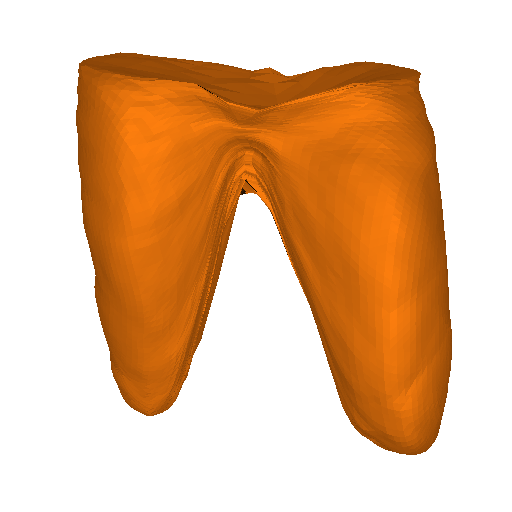}
    \end{subfigure}%
    \hfill%
    \begin{subfigure}[b]{0.166\linewidth}
		\centering
		\includegraphics[width=\linewidth]{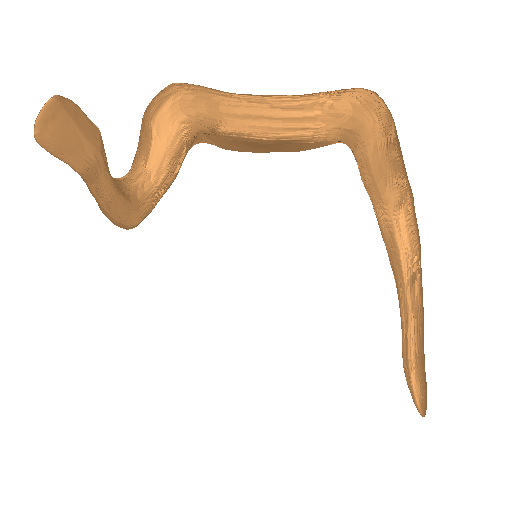}
    \end{subfigure}%
    \hfill%
    \begin{subfigure}[b]{0.166\linewidth}
		\centering
		\includegraphics[width=\linewidth]{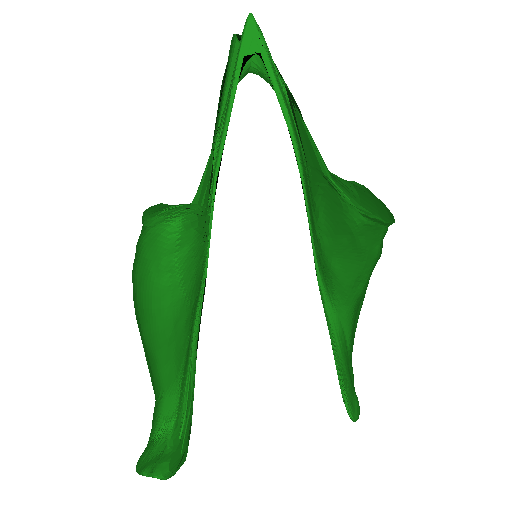}
    \end{subfigure}\\[0.2em]
    \vskip\baselineskip%
    \vspace{-1.2em}
    \caption{{\bf w/o Normal Consistency Loss}. We visualize the predicted primitives when training without the $\cL_{normal}$ and we observe that even though the predicted primitives are expressive and have a semantic interpretation, the locations where the primitives meet do not have consistent normals.}
    \label{fig:normal_loss_primitives}
    \vspace{-1.2em}
\end{figure}

\boldparagraph{w/o Normal Consistency Loss}%
The normal consistency loss enforces that the normals of the predicted primitives will be aligned with the normals of the target object.
We notice that when we remove $\cL_{norm}(\cX_t)$, our model yields non-overlapping primitives that accurately capture the geometry of different human body parts (see \figref{fig:normal_loss_primitives}).
However, at the locations where one primitive meets its adjacent, the predicted primitives have inconsistent normals, hence the connections seem unnatural. This can be better seen in \figref{fig:normal_loss_comparison}, where we provide close-ups on the locations where two primitives meet and compare with our model. Our model, yields primitives with smooth transitions that look more natural.

\begin{figure}[!h]
    \begin{subfigure}[b]{0.166\linewidth}
		\centering
		\includegraphics[width=\linewidth]{gfx_ablations_no_overlap_50002_chicken_wings_00056}
    \end{subfigure}%
    \hfill%
    \begin{subfigure}[b]{0.166\linewidth}
		\centering
		\includegraphics[width=\linewidth]{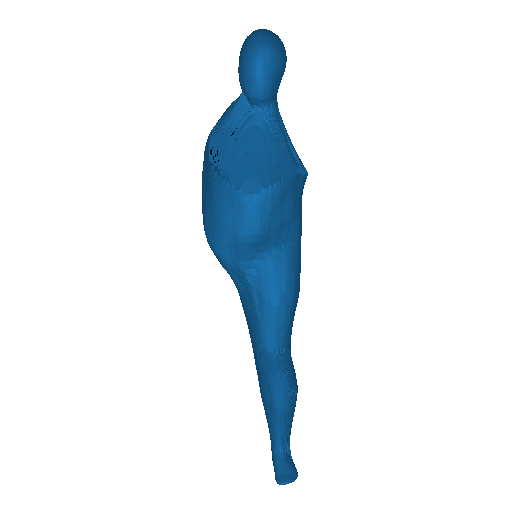}
    \end{subfigure}%
    \hfill%
    \begin{subfigure}[b]{0.166\linewidth}
		\centering
		\includegraphics[width=\linewidth]{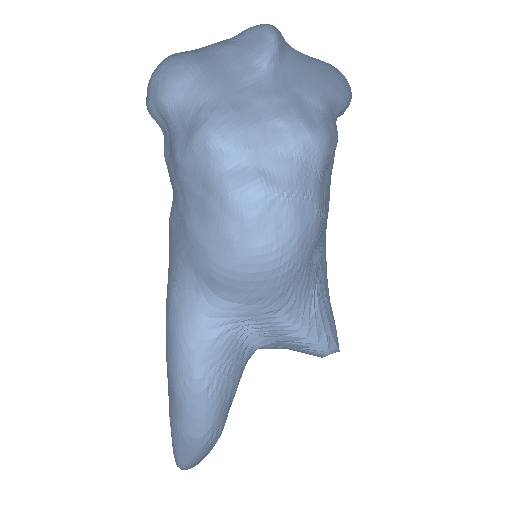}
    \end{subfigure}%
    \hfill%
    \begin{subfigure}[b]{0.166\linewidth}
		\centering
		\includegraphics[width=\linewidth]{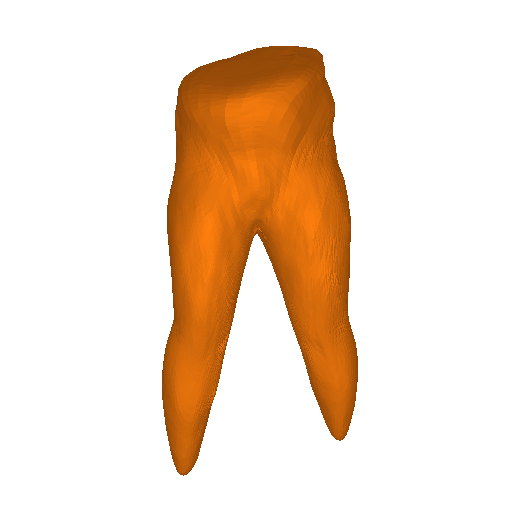}
    \end{subfigure}%
    \hfill%
    \begin{subfigure}[b]{0.166\linewidth}
		\centering
		\includegraphics[width=\linewidth]{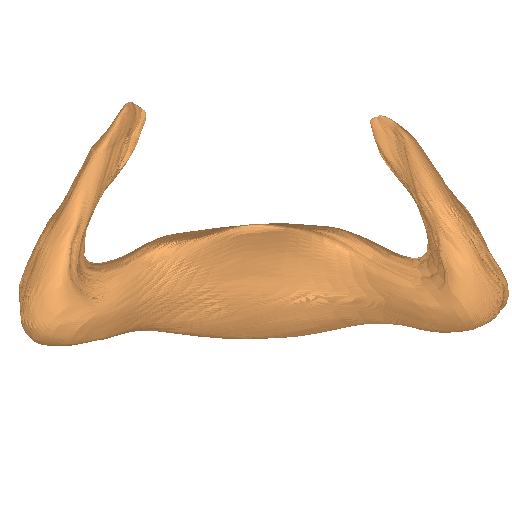}
    \end{subfigure}%
    \hfill%
    \begin{subfigure}[b]{0.166\linewidth}
		\centering
		\includegraphics[width=\linewidth]{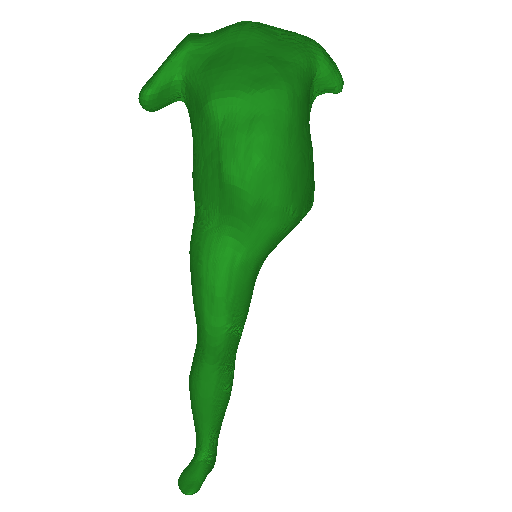}
    \end{subfigure}\\[0.2em]
    \begin{subfigure}[b]{0.166\linewidth}
		\centering
		\includegraphics[width=\linewidth]{gfx_ablations_no_overlap_50002_knees_00046}
    \end{subfigure}%
    \hfill%
    \begin{subfigure}[b]{0.166\linewidth}
		\centering
		\includegraphics[width=\linewidth]{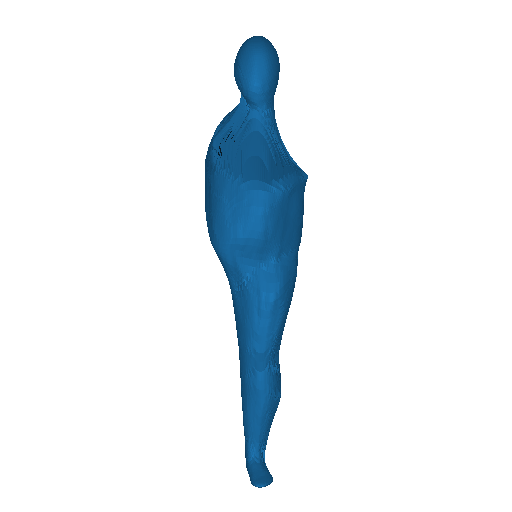}
    \end{subfigure}%
    \hfill%
    \begin{subfigure}[b]{0.166\linewidth}
		\centering
		\includegraphics[width=\linewidth]{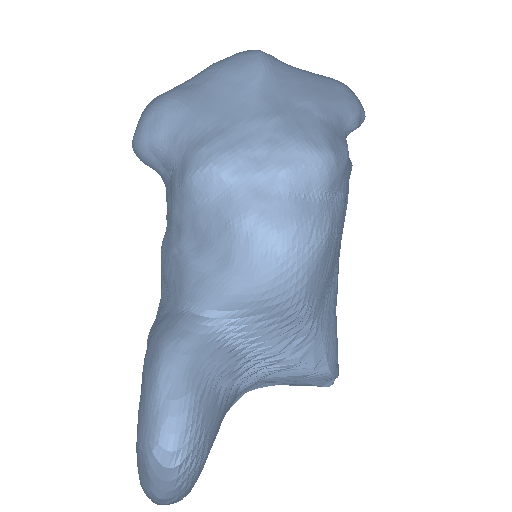}
    \end{subfigure}%
    \hfill%
    \begin{subfigure}[b]{0.166\linewidth}
		\centering
		\includegraphics[width=\linewidth]{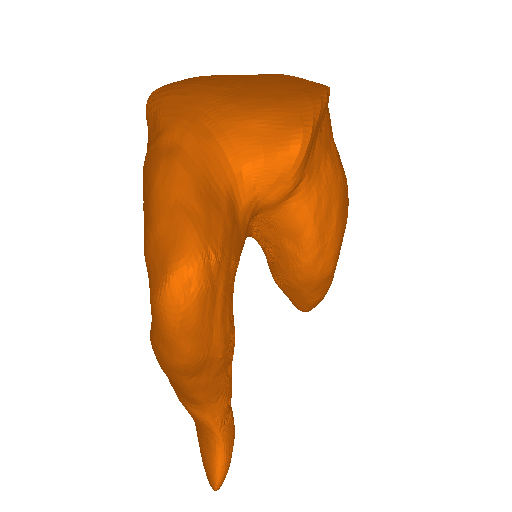}
    \end{subfigure}%
    \hfill%
    \begin{subfigure}[b]{0.166\linewidth}
		\centering
		\includegraphics[width=\linewidth]{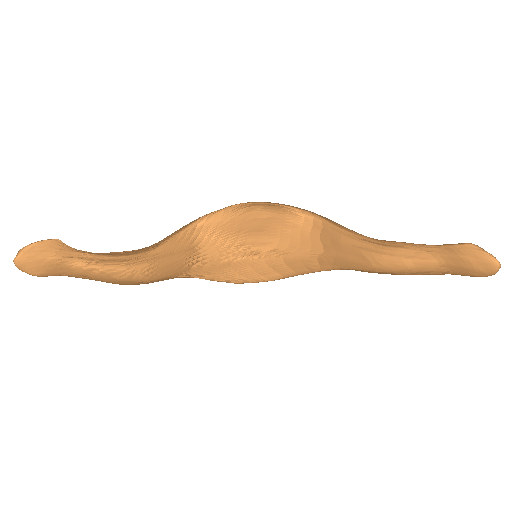}
    \end{subfigure}%
    \hfill%
    \begin{subfigure}[b]{0.166\linewidth}
		\centering
		\includegraphics[width=\linewidth]{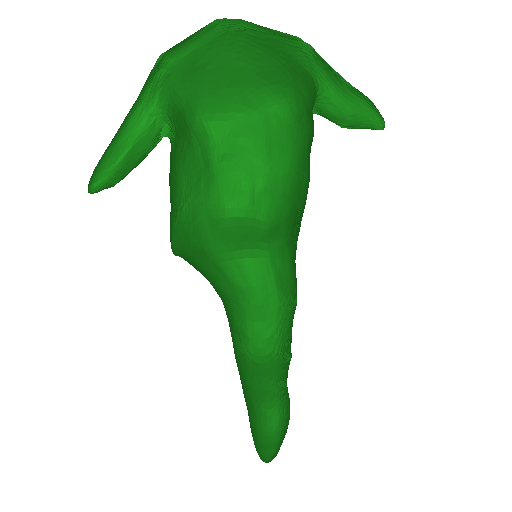}
    \end{subfigure}\\[0.2em]
    \begin{subfigure}[b]{0.166\linewidth}
		\centering
		\includegraphics[width=\linewidth]{gfx_ablations_no_overlap_50020_running_on_spot_00053}
    \end{subfigure}%
    \hfill%
    \begin{subfigure}[b]{0.166\linewidth}
		\centering
		\includegraphics[width=\linewidth]{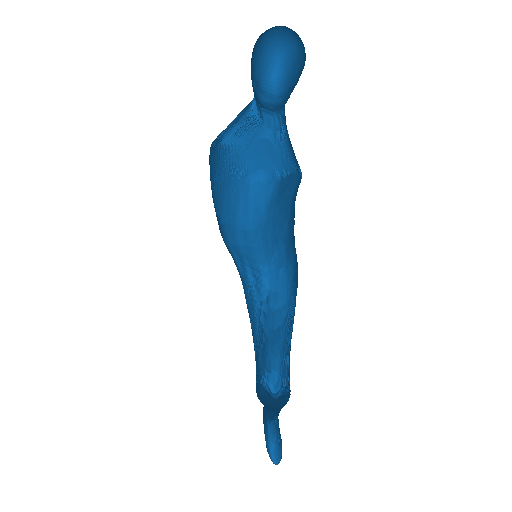}
    \end{subfigure}%
    \hfill%
    \begin{subfigure}[b]{0.166\linewidth}
		\centering
		\includegraphics[width=\linewidth]{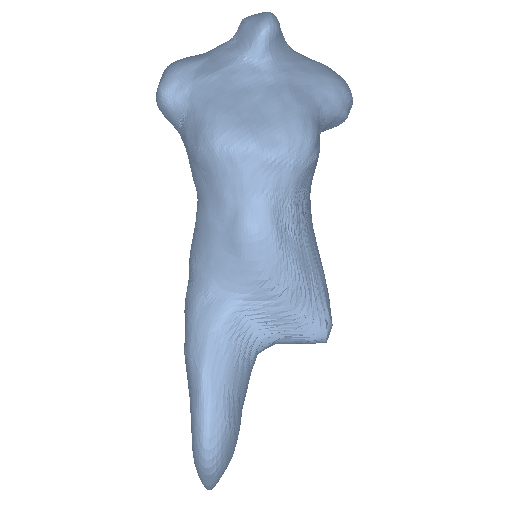}
    \end{subfigure}%
    \hfill%
    \begin{subfigure}[b]{0.166\linewidth}
		\centering
		\includegraphics[width=\linewidth]{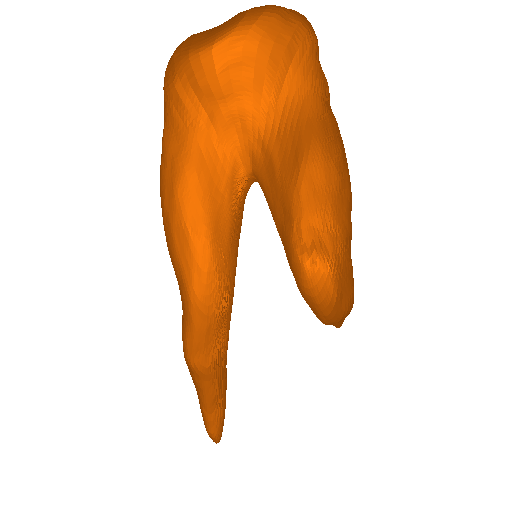}
    \end{subfigure}%
    \hfill%
    \begin{subfigure}[b]{0.166\linewidth}
		\centering
		\includegraphics[width=\linewidth]{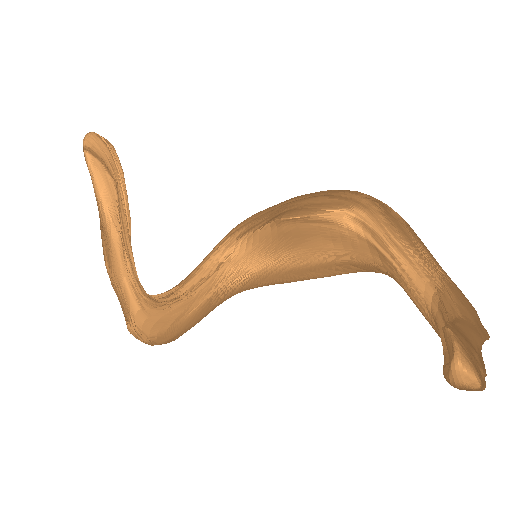}
    \end{subfigure}%
    \hfill%
    \begin{subfigure}[b]{0.166\linewidth}
		\centering
		\includegraphics[width=\linewidth]{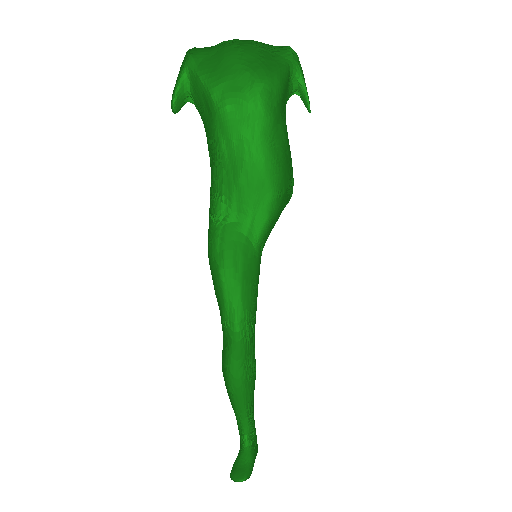}
    \end{subfigure}\\[0.2em]
    \vskip\baselineskip%
    \vspace{-1.2em}
    \caption{{\bf w/o Overlapping Loss}. We visualize the predicted primitives when training without the $\cL_{overlap}$ and we observe that even though the predicted primitives are expressive and accurately capture the 3D geometry of the human body, they do not have a semantic interpretation, namely multiple primitives describe the same object part.}
    \label{fig:overlapping_loss_primitives}
    \vspace{-1.2em}
\end{figure}

\boldparagraph{w/o Overlapping Loss}%
The overlapping loss encourages semantically meaningful shape abstractions, where primitives represent distinct geometric parts.
To this end, we introduce the $\cL_{overlap}(\cX_o)$ loss (Eq. $15$ in the main submission) to discourage having multiple primitives ``containing'' the same points from the target object.
The reconstructions without the overlapping loss accurately capture the geometry of the human body but fail to yield semantically meaningful shape abstractions, since primitives completely penetrate one another (see \figref{fig:ablation_supp}).
In \figref{fig:overlapping_loss_primitives}, we visualize the predicted
primitives for various humans and we notice that while they accurately capture
the geometry of the human body, they do not correspond to meaningful geometric
parts. In particular, there are $2$ primitives that describe the right leg
(primitive illustrated in blue and orange), $2$ for the left leg (primitives
illustrated in orange and green) and $3$ for the main body (blue, light blue
and green).
Since, we want our primitives to correspond to semantic parts, this behaviour is not desirable.

\begin{figure}[!h]
    \begin{subfigure}[b]{0.25\linewidth}
		\centering
		\includegraphics[width=\linewidth]{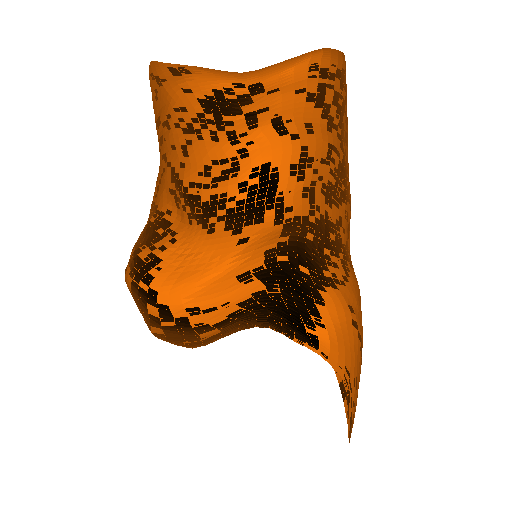}
    \end{subfigure}%
    \hfill%
    \begin{subfigure}[b]{0.25\linewidth}
		\centering
		\includegraphics[width=\linewidth]{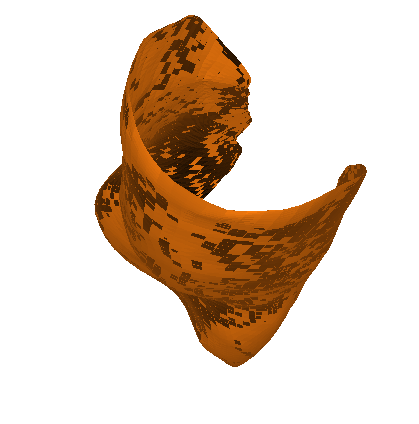}
    \end{subfigure}
    \begin{subfigure}[b]{0.25\linewidth}
		\centering
		\includegraphics[width=\linewidth]{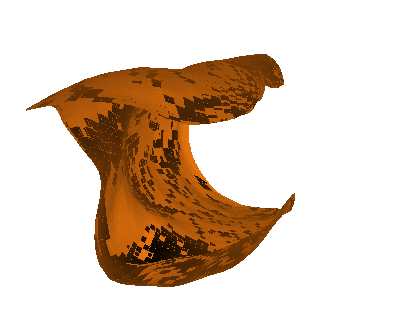}
    \end{subfigure}%
    \hfill%
    \begin{subfigure}[b]{0.24\linewidth}
		\centering
		\includegraphics[width=\linewidth]{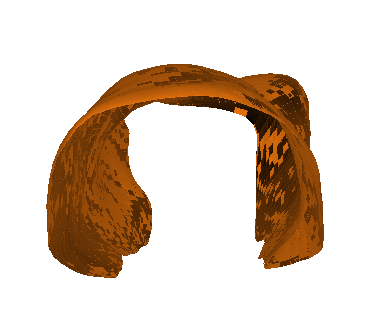}
    \end{subfigure}
    \vskip\baselineskip%
    \vspace{-1.2em}
    \caption{{\bf Impact of $\cL_{cover}$.} When removing the $\cL_{cover}$ some primitives become very thin (with minimum volume) and do not contribute to the reconstruction loss.}
    \label{fig:min_points_loss_details}
\end{figure}

\begin{figure}[!h]
    \begin{subfigure}[b]{0.166\linewidth}
		\centering
		\includegraphics[width=\linewidth]{gfx_ablations_no_min_points_50002_chicken_wings_00056}
    \end{subfigure}%
    \hfill%
    \begin{subfigure}[b]{0.166\linewidth}
		\centering
		\includegraphics[width=\linewidth]{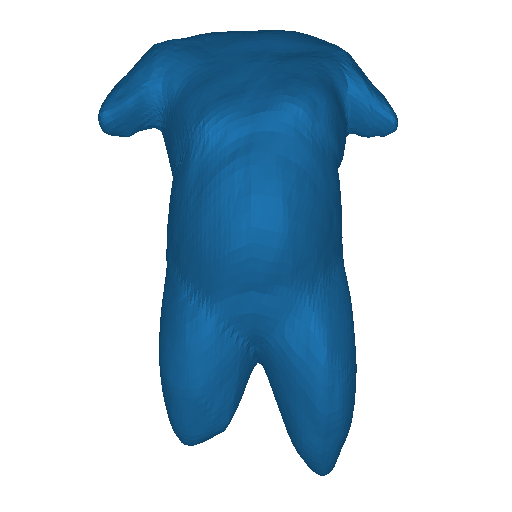}
    \end{subfigure}%
    \hfill%
    \begin{subfigure}[b]{0.166\linewidth}
		\centering
		\includegraphics[width=\linewidth]{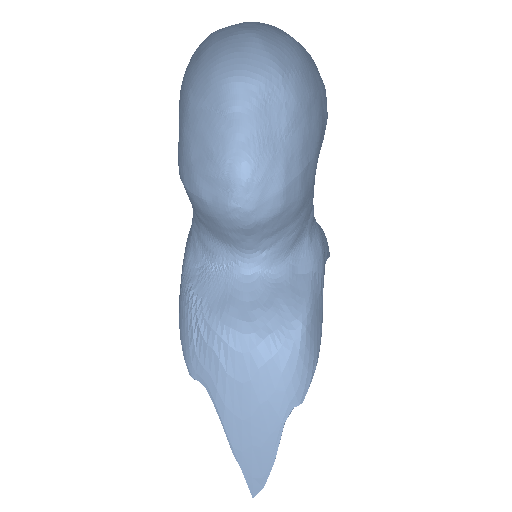}
    \end{subfigure}%
    \hfill%
    \begin{subfigure}[b]{0.166\linewidth}
		\centering
		\includegraphics[width=\linewidth]{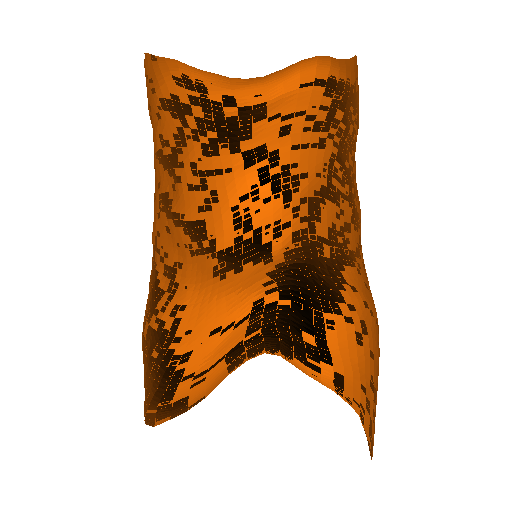}
    \end{subfigure}%
    \hfill%
    \begin{subfigure}[b]{0.166\linewidth}
		\centering
		\includegraphics[width=\linewidth]{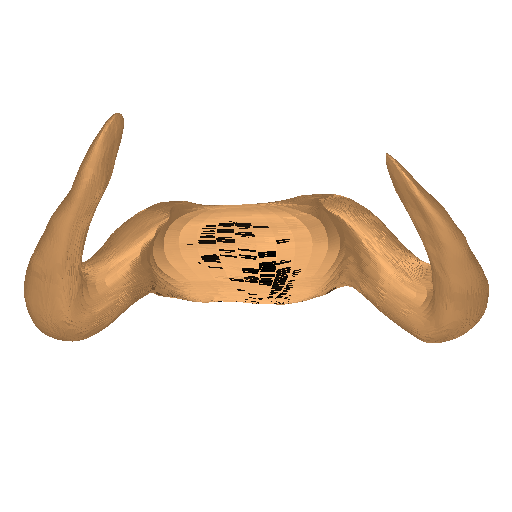}
    \end{subfigure}%
    \hfill%
    \begin{subfigure}[b]{0.166\linewidth}
		\centering
		\includegraphics[width=\linewidth]{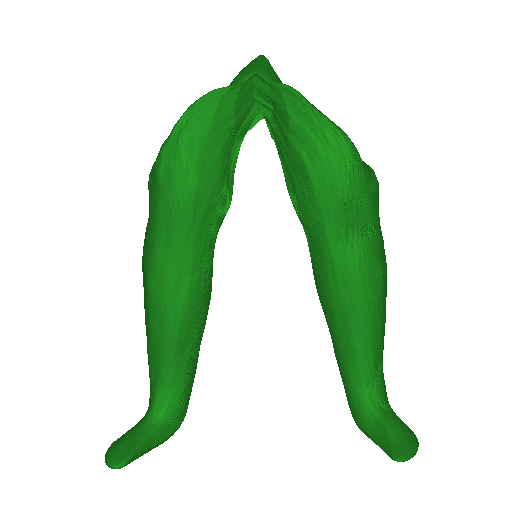}
    \end{subfigure}\\[0.2em]
    \begin{subfigure}[b]{0.166\linewidth}
		\centering
		\includegraphics[width=\linewidth]{gfx_ablations_no_min_points_50002_knees_00046}
    \end{subfigure}%
    \hfill%
    \begin{subfigure}[b]{0.166\linewidth}
		\centering
		\includegraphics[width=\linewidth]{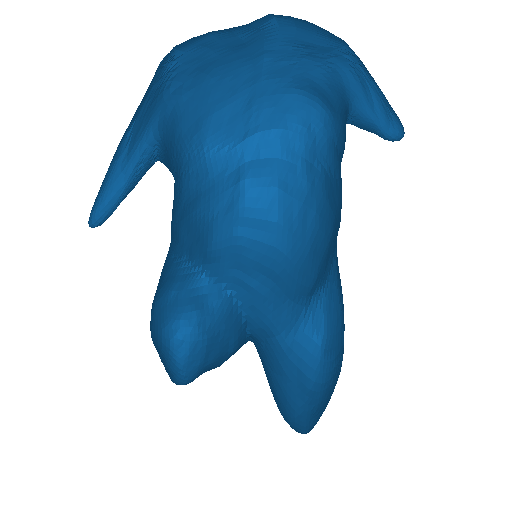}
    \end{subfigure}%
    \hfill%
    \begin{subfigure}[b]{0.166\linewidth}
		\centering
		\includegraphics[width=\linewidth]{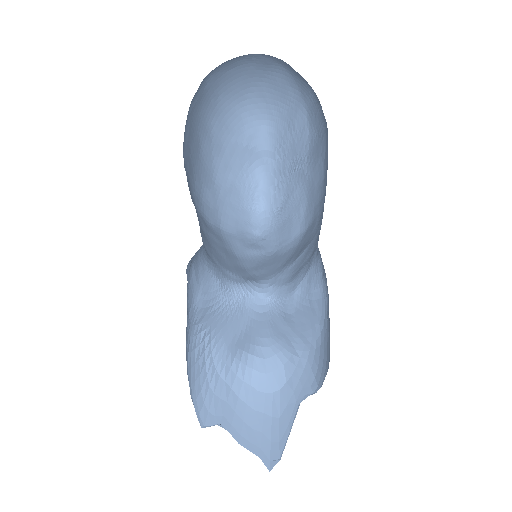}
    \end{subfigure}%
    \hfill%
    \begin{subfigure}[b]{0.166\linewidth}
		\centering
		\includegraphics[width=\linewidth]{gfx_ablations_no_min_points_50002_knees_00046_002}
    \end{subfigure}%
    \hfill%
    \begin{subfigure}[b]{0.166\linewidth}
		\centering
		\includegraphics[width=\linewidth]{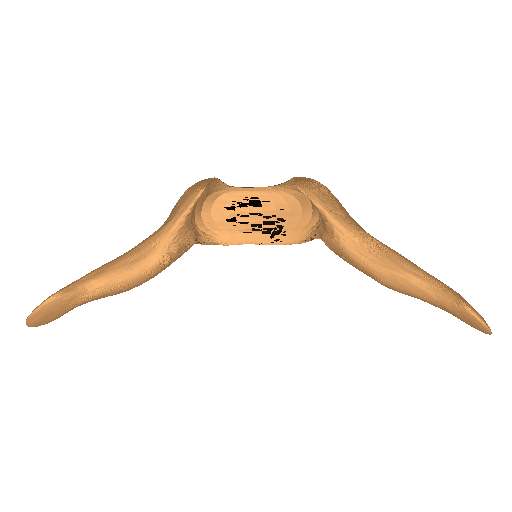}
    \end{subfigure}%
    \hfill%
    \begin{subfigure}[b]{0.166\linewidth}
		\centering
		\includegraphics[width=\linewidth]{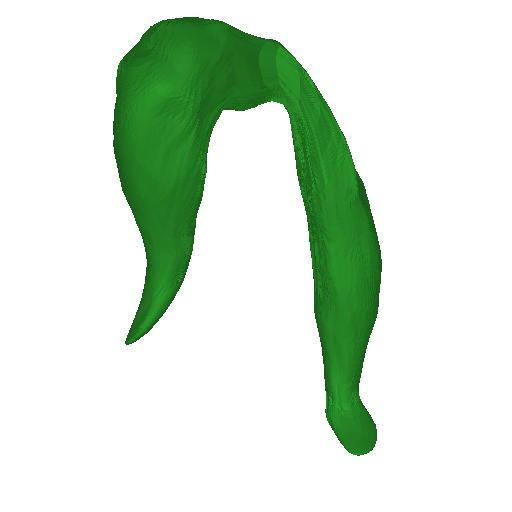}
    \end{subfigure}\\[0.2em]
    \begin{subfigure}[b]{0.166\linewidth}
		\centering
		\includegraphics[width=\linewidth]{gfx_ablations_no_min_points_50020_running_on_spot_00053}
    \end{subfigure}%
    \hfill%
    \begin{subfigure}[b]{0.166\linewidth}
		\centering
		\includegraphics[width=\linewidth]{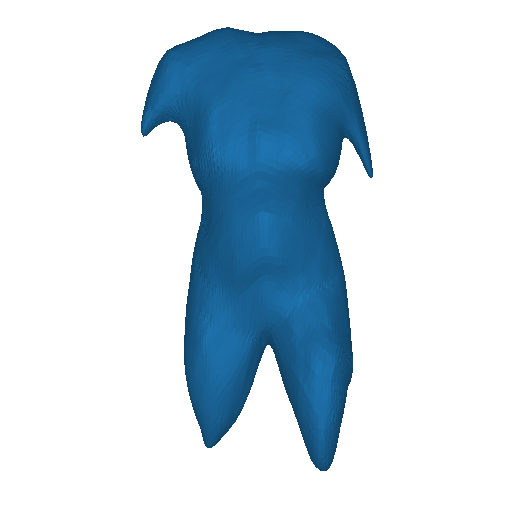}
    \end{subfigure}%
    \hfill%
    \begin{subfigure}[b]{0.166\linewidth}
		\centering
		\includegraphics[width=\linewidth]{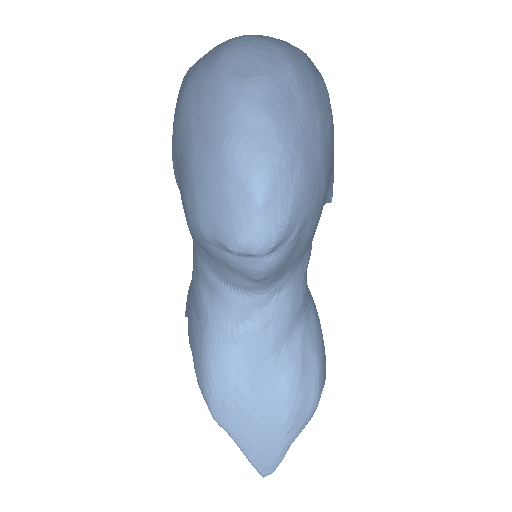}
    \end{subfigure}%
    \hfill%
    \begin{subfigure}[b]{0.166\linewidth}
		\centering
		\includegraphics[width=\linewidth]{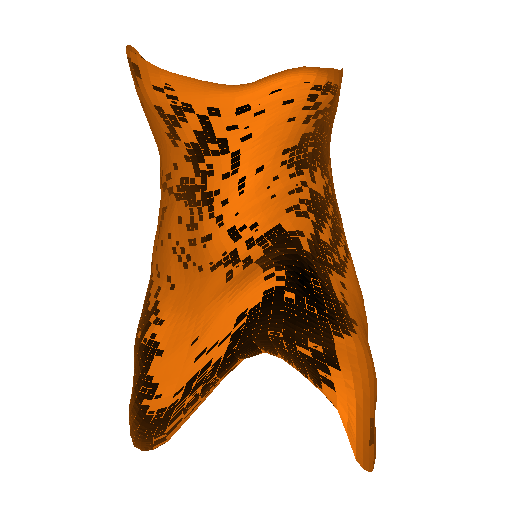}
    \end{subfigure}%
    \hfill%
    \begin{subfigure}[b]{0.166\linewidth}
		\centering
		\includegraphics[width=\linewidth]{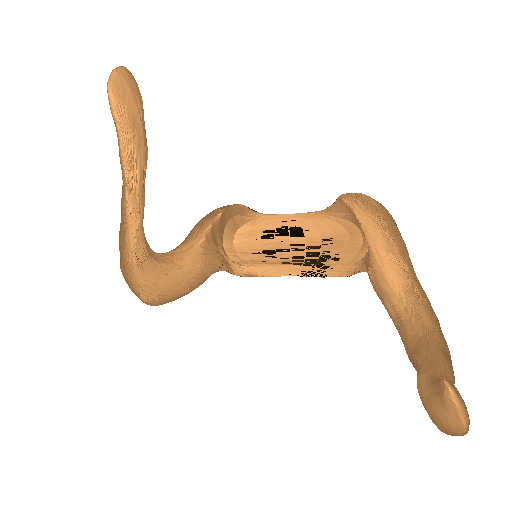}
    \end{subfigure}%
    \hfill%
    \begin{subfigure}[b]{0.166\linewidth}
		\centering
		\includegraphics[width=\linewidth]{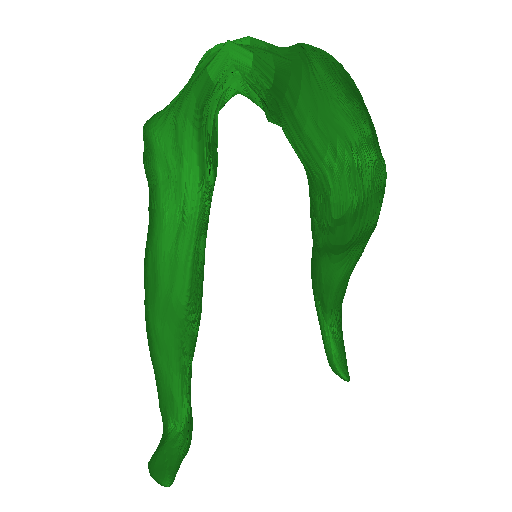}
    \end{subfigure}\\[0.2em]
    \vskip\baselineskip%
    \vspace{-1.2em}
    \caption{{\bf w/o Coverage Loss}. We visualize the predicted primitives when training without the $\cL_{cover}$ and observe that although some of them are expressive and efficiently capture the 3D geometry, some primitives have minimum volume hence, do not cover parts of the object and in turn, do not contribute to the reconstruction loss.}
    \label{fig:min_points_loss_primitives}
    \vspace{-1.2em}
\end{figure}

\boldparagraph{w/o Coverage Loss}%
The coverage loss makes sure that all primitives will ``cover'' parts of the
target object. In practice, it prevents primitives with minimal
volume (see \figref{fig:min_points_loss_primitives}, fourth column) that do not
contribute to the reconstruction.
In \figref{fig:min_points_loss_primitives}, we note that when we remove
$\cL_{cover}(\cX_o)$ from our optimization objective, a primitive degenerates
to a ``sheet'' that contains no points but only has surface area
(\figref{fig:min_points_loss_details}).

\clearpage
\section{Experiment on D-FAUST}

In this section, we provide additional information regarding our experiments on D-FAUST dataset~\cite{Bogo2017CVPR}.
D-FAUST contains $38,640$ meshes of humans performing various tasks such as ``chicken wings", ``running on spot", ``shake arms" \etc.
We follow~\cite{Paschalidou2020CVPR} and use $70\%$, $20\%$ and $10\%$ for the train, test and validation splits.
Furthermore, we filter out the first 20 frames for each sequence that contain the unnatural ``neutral pose" necessary for calibration purposes.
Note that in contrast to \cite{Paschalidou2020CVPR}, we do not normalize the meshes to the unit cube in order to retain the variety of the human body \ie tall and short humans.
This makes the single view 3D reconstruction task harder, as all models need to efficiently capture both the pose and the size of the human.

\begin{figure}[H]
    \centering
    \begin{subfigure}[b]{0.15\linewidth}
		\centering
        Target
    \end{subfigure}%
    \hfill%
    \begin{subfigure}[b]{0.15\linewidth}
		\centering
        Ours $\# 1$
    \end{subfigure}%
    \hfill%
    \begin{subfigure}[b]{0.15\linewidth}
		\centering
        Ours $\# 2$
    \end{subfigure}%
    \hfill%
    \begin{subfigure}[b]{0.15\linewidth}
		\centering
        Ours $\# 5$
    \end{subfigure}%
    \hfill%
    \begin{subfigure}[b]{0.15\linewidth}
		\centering
        Ours $\# 8$
    \end{subfigure}%
    \hfill%
    \begin{subfigure}[b]{0.15\linewidth}
        \centering
        Ours $\# 10$
    \end{subfigure}
    \vskip\baselineskip%
    \vspace{-1.0em}
    \begin{subfigure}[b]{0.15\linewidth}
		\centering
		\includegraphics[trim=80 0 80 0,clip,width=\linewidth]{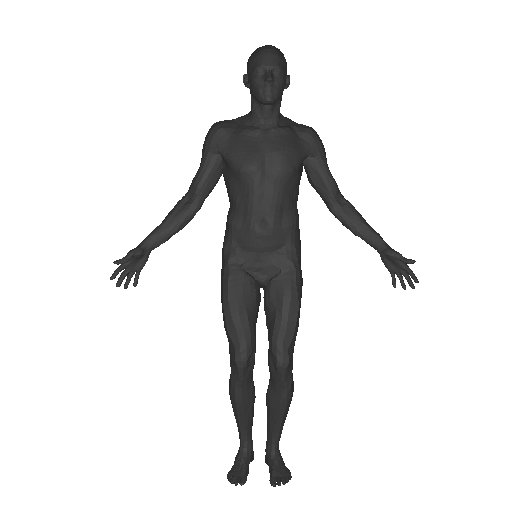}
    \end{subfigure}%
    \hfill%
    \begin{subfigure}[b]{0.15\linewidth}
		\centering
		\includegraphics[trim=80 0 80 0,clip,width=\linewidth]{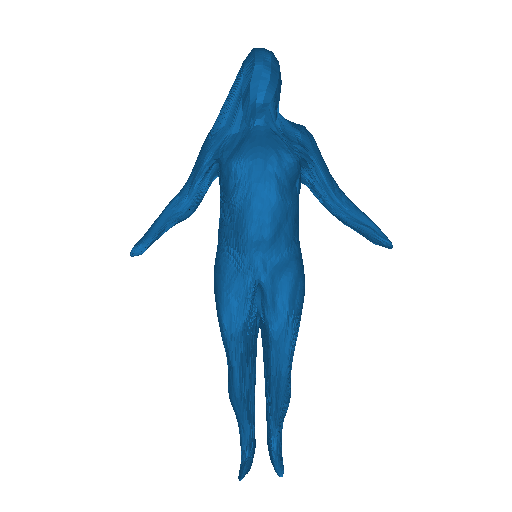}
    \end{subfigure}%
    \hfill%
    \begin{subfigure}[b]{0.15\linewidth}
		\centering
		\includegraphics[trim=80 0 80 0,clip,width=\linewidth]{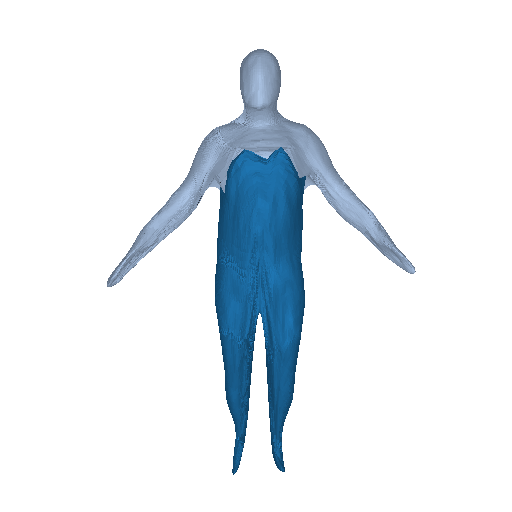}
    \end{subfigure}%
    \hfill%
    \begin{subfigure}[b]{0.15\linewidth}
		\centering
		\includegraphics[trim=75 0 75 0,clip,width=\linewidth]{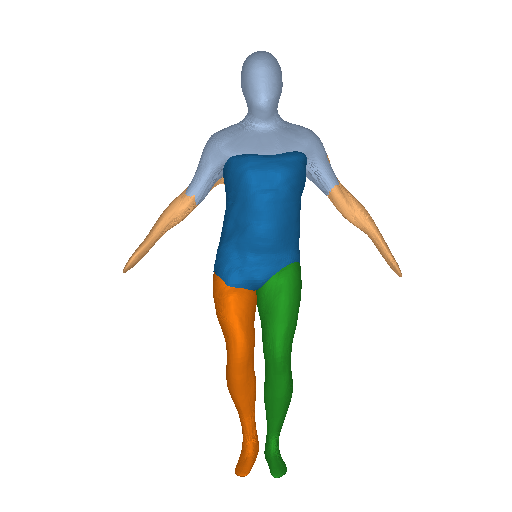}
    \end{subfigure}%
    \hfill%
    \begin{subfigure}[b]{0.15\linewidth}
		\centering
		\includegraphics[trim=80 0 80 0,clip,width=\linewidth]{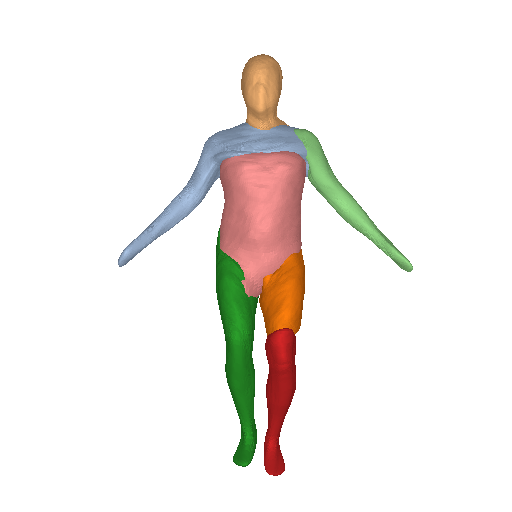}
    \end{subfigure}%
    \hfill%
    \begin{subfigure}[b]{0.15\linewidth}
		\centering
		\includegraphics[trim=75 0 75 0,clip,width=\linewidth]{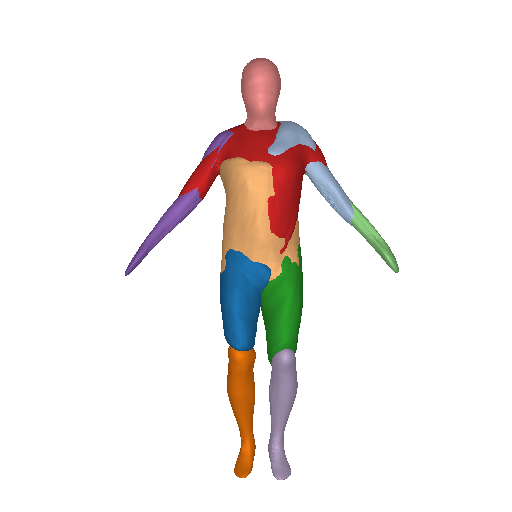}
    \end{subfigure}%
    \vspace{-1.2em}
    \vskip\baselineskip%
    \begin{subfigure}[b]{0.15\linewidth}
		\centering
		\includegraphics[trim=80 0 80 0,clip,width=\linewidth]{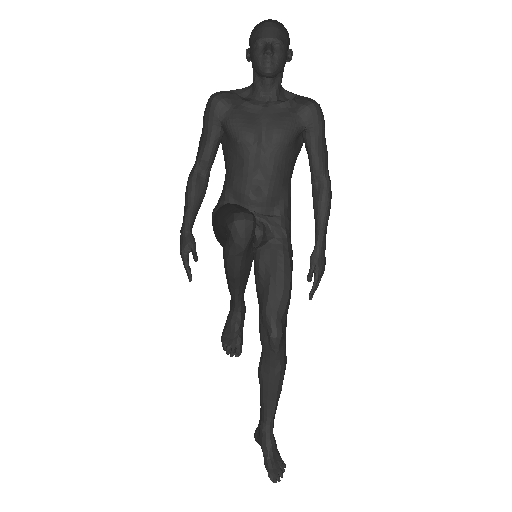}
    \end{subfigure}%
    \hfill%
    \begin{subfigure}[b]{0.15\linewidth}
		\centering
		\includegraphics[trim=80 0 80 0,clip,width=\linewidth]{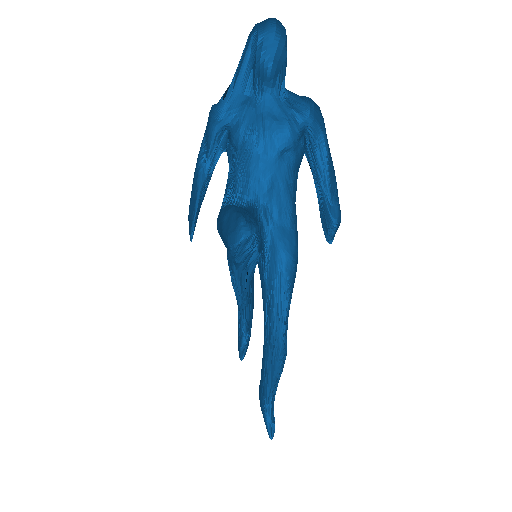}
    \end{subfigure}%
    \hfill%
    \begin{subfigure}[b]{0.15\linewidth}
		\centering
		\includegraphics[trim=80 0 80 0,clip,width=\linewidth]{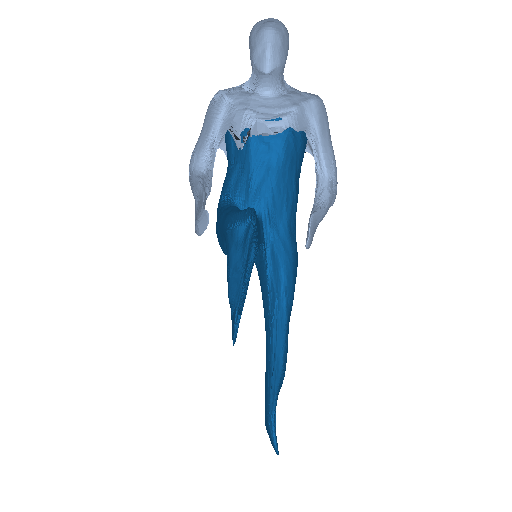}
    \end{subfigure}%
    \hfill%
    \begin{subfigure}[b]{0.15\linewidth}
		\centering
		\includegraphics[trim=75 0 75 0,clip,width=\linewidth]{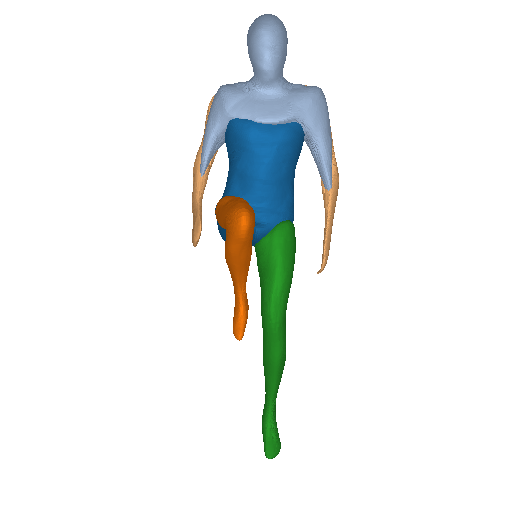}
    \end{subfigure}%
    \hfill%
    \begin{subfigure}[b]{0.15\linewidth}
		\centering
		\includegraphics[trim=80 0 80 0,clip,width=\linewidth]{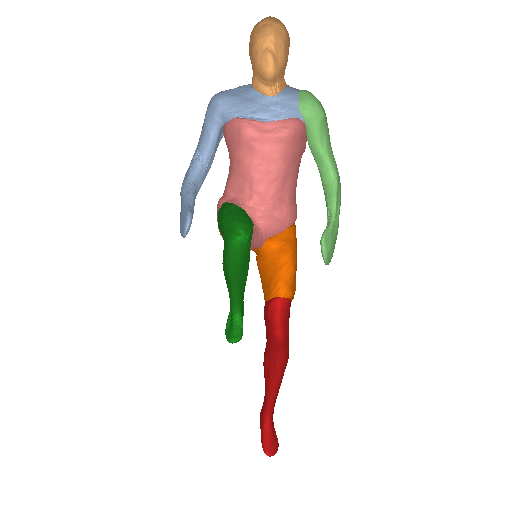}
    \end{subfigure}%
    \hfill%
    \begin{subfigure}[b]{0.15\linewidth}
		\centering
		\includegraphics[trim=75 0 75 0,clip,width=\linewidth]{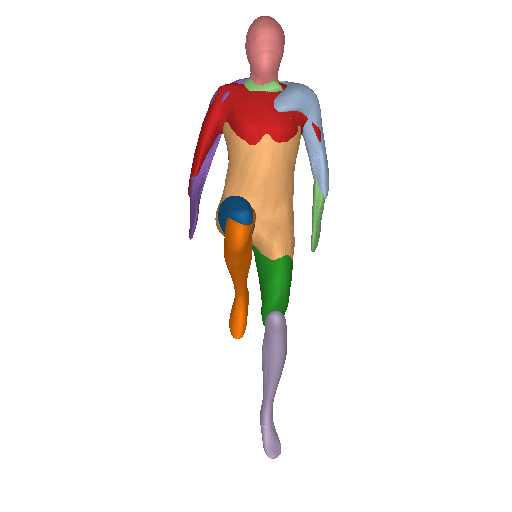}
    \end{subfigure}%
        \vspace{-1.2em}
    \caption{{\bf Single Image 3D Reconstruction on D-FAUST}. The input image is shown on the first column and the rest contain predictions of our method with different number of primitives.}
    \label{fig:qualitative_dfaust_different_num_of_prims}
    \vspace{-1.2em}
\end{figure}

\tabref{tab:dfaust_quantitative} summarizes the quantitative results from Fig. $4$ in our main submission.
We notice that for all primitive-based baselines, increasing the number of parts results in improved reconstruction quality in terms of volumetric $IoU$ and Chamfer-$L_1$ distance.
Instead, Neural Parts, decouple the number of primitives from the reconstruction accuracy as the performance of our model is independent of the number of the predicted primitives.
This is expected, since our primitives are highly expressive and our model can capture the 3D shape geometry using even a single primitive (see \figref{fig:qualitative_dfaust_different_num_of_prims}).
We argue that this is a desirable property, as Neural Parts enable us to select the number of parts based on the expected number of semantic parts and not the desired reconstruction quality.
\begin{table}[H]
   \centering
    \resizebox{\columnwidth}{!}{
    \begin{tabular}{l||c||cccc|cccc|cccc|cccc}
        \toprule
        \multicolumn{2}{c}{\,} & \multicolumn{4}{c}{SQs} & \multicolumn{4}{c}{H-SQs} & \multicolumn{4}{c}{CvxNet} & \multicolumn{4}{c}{Ours}\\
        & OccNet & 5 & 10 & 25 & 50 & 4 & 8 & 16 & 32 & 5 & 10 & 25 & 50 & 2 & 5 & 8 & 10 \\
        \midrule
        IoU & 0.691 & 0.514 & 0.565 & 0.616 & 0.607 & 0.559 & 0.595 & 0.623 & 0.610 & 0.582 & 0.607 & 0.622 & 0.621 & 0.675 & 0.673 & 0.676 & 0.678 \\
        Chamfer-$L_1$ & 0.102 & 0.147 & 0.129 & 0.114 & 0.116 & 0.161 & 0.148 & 0.126 & 0.129 & 0.183 & 0.161 & 0.139 & 0.121 & 0.101 & 0.097 & 0.090 & 0.095 \\
        \bottomrule
    \end{tabular}}
    \caption{{\bf Single Image Reconstruction on D-FAUST.} We report the volumetric IoU $(\uparrow)$ and the Chamfer-$L_1$ $(\downarrow)$ \wrt the ground-truth mesh for our model compared to primitive-based methods SQs~\cite{Paschalidou2019CVPR} H-SQs~\cite{Paschalidou2020CVPR}, CvxNet~\cite{Deng2019ARXIV} for various number of primitives and the non primitive-based OccNet~\cite{Mescheder2019CVPR}. We observe that our model outperforms all primitive-based baselines with only $2$ primitives.}
    \label{tab:dfaust_quantitative}
\end{table}

\subsection{Additional Experimental Results}

\begin{figure}
    \centering
    \begin{subfigure}[b]{0.15\linewidth}
		\centering
        Target
    \end{subfigure}%
    \hfill%
    \begin{subfigure}[b]{0.15\linewidth}
		\centering
        OccNet
    \end{subfigure}%
    \hfill%
    \begin{subfigure}[b]{0.15\linewidth}
		\centering
        SQs
    \end{subfigure}%
    \hfill%
    \begin{subfigure}[b]{0.15\linewidth}
		\centering
        H-SQs
    \end{subfigure}%
    \hfill%
    \begin{subfigure}[b]{0.15\linewidth}
		\centering
        CvxNet
    \end{subfigure}%
    \hfill%
    \begin{subfigure}[b]{0.15\linewidth}
        \centering
        Ours
    \end{subfigure}
    \vskip\baselineskip%
    \vspace{-0.5em}
    \begin{subfigure}[b]{0.15\linewidth}
		\centering
		\includegraphics[trim=80 0 80 0,clip,width=\linewidth]{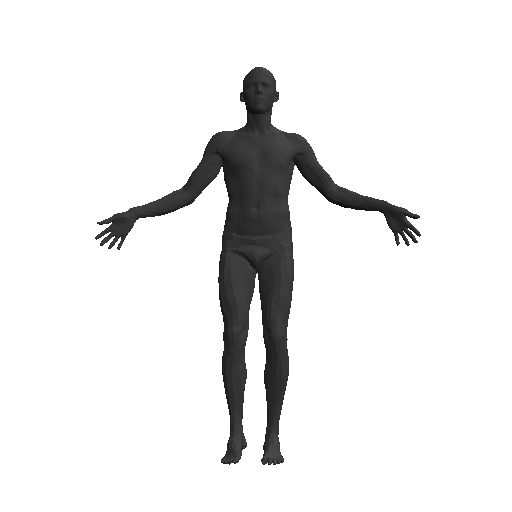}
    \end{subfigure}%
    \hfill%
    \begin{subfigure}[b]{0.15\linewidth}
		\centering
		\includegraphics[trim=80 0 80 0,clip,width=\linewidth]{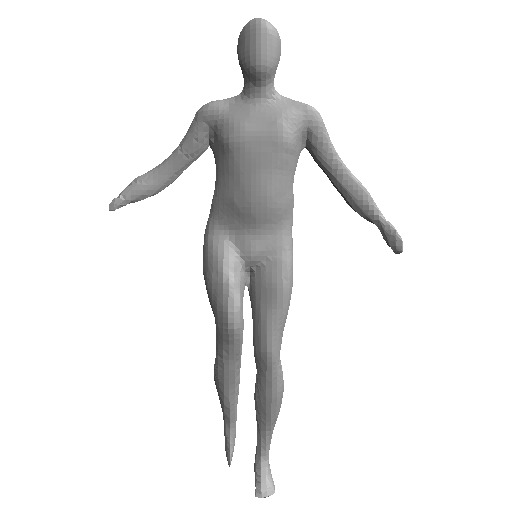}
    \end{subfigure}%
    \hfill%
    \begin{subfigure}[b]{0.15\linewidth}
		\centering
		\includegraphics[trim=80 0 80 0,clip,width=\linewidth]{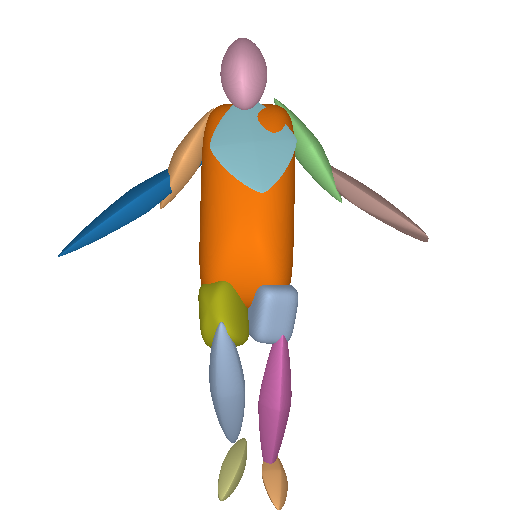}
    \end{subfigure}%
    \hfill%
    \begin{subfigure}[b]{0.15\linewidth}
		\centering
		\includegraphics[trim=80 0 80 0,clip,width=\linewidth]{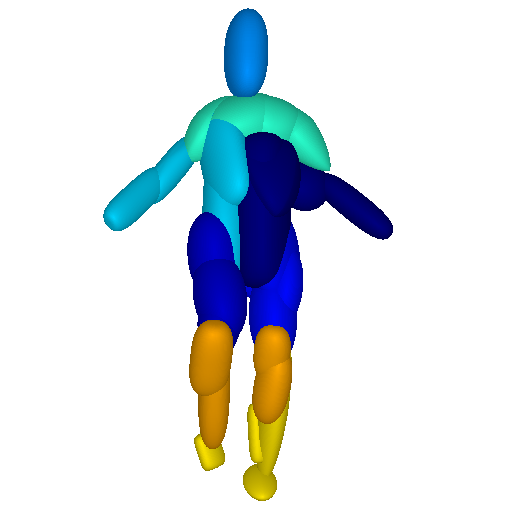}
    \end{subfigure}%
    \hfill%
    \begin{subfigure}[b]{0.15\linewidth}
		\centering
		\includegraphics[trim=80 0 80 0,clip,width=\linewidth]{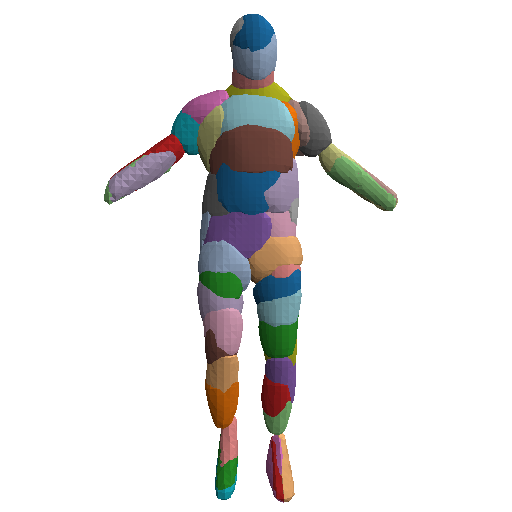}
    \end{subfigure}%
    \hfill%
    \begin{subfigure}[b]{0.15\linewidth}
		\centering
		\includegraphics[trim=80 0 80 0,clip,width=\linewidth]{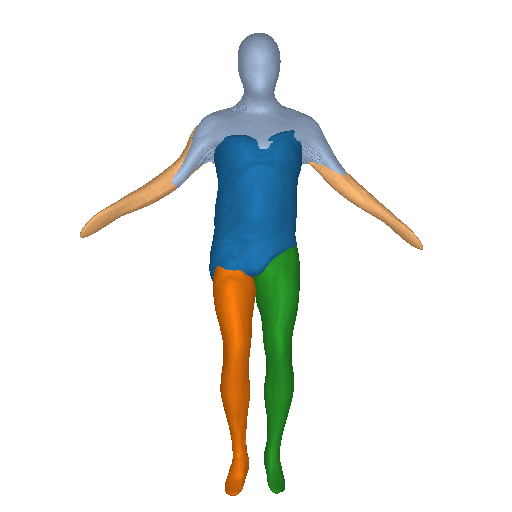}
    \end{subfigure}%
    \vspace{-1.2em}
    \vskip\baselineskip%
    \begin{subfigure}[b]{0.15\linewidth}
		\centering
		\includegraphics[trim=80 0 80 0,clip,width=\linewidth]{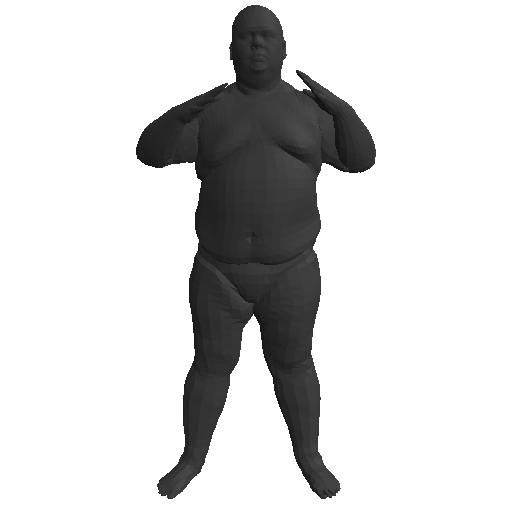}
    \end{subfigure}%
    \hfill%
    \begin{subfigure}[b]{0.15\linewidth}
		\centering
		\includegraphics[trim=80 0 80 0,clip,width=\linewidth]{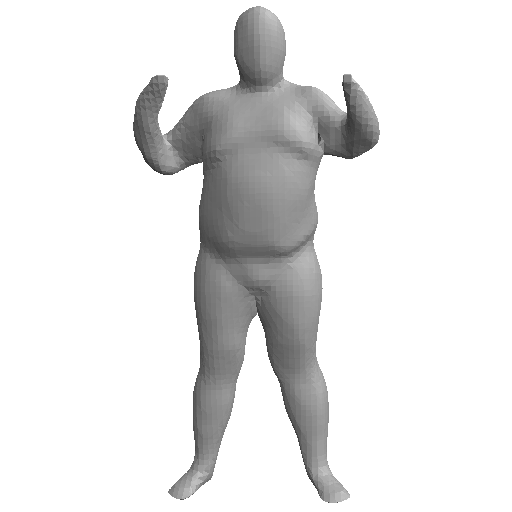}
    \end{subfigure}%
    \hfill%
    \begin{subfigure}[b]{0.15\linewidth}
		\centering
		\includegraphics[trim=80 0 80 0,clip,width=\linewidth]{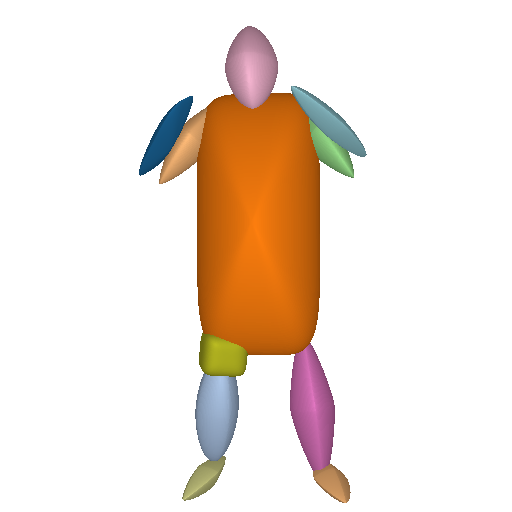}
    \end{subfigure}%
    \hfill%
    \begin{subfigure}[b]{0.15\linewidth}
		\centering
		\includegraphics[trim=80 0 80 0,clip,width=\linewidth]{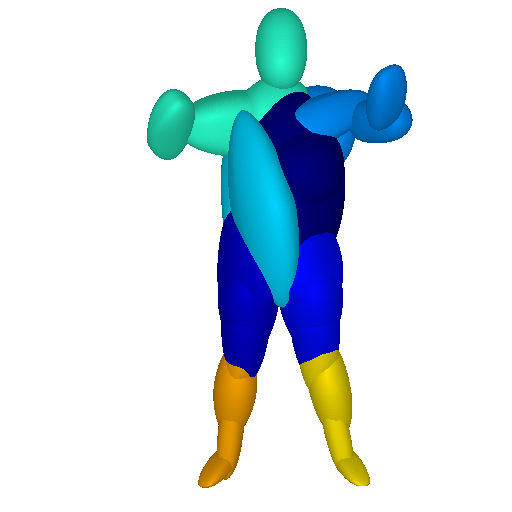}
    \end{subfigure}%
    \hfill%
    \begin{subfigure}[b]{0.15\linewidth}
		\centering
		\includegraphics[trim=80 0 80 0,clip,width=\linewidth]{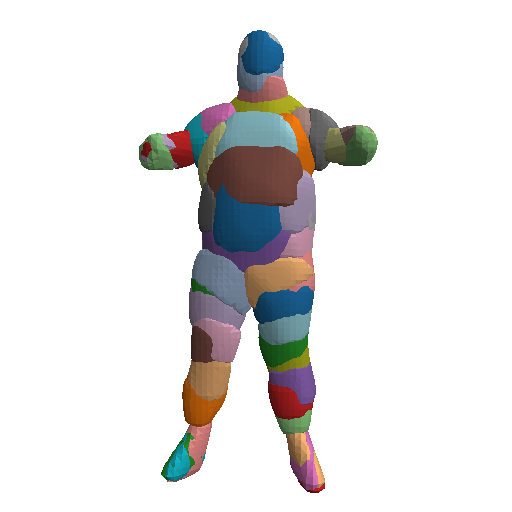}
    \end{subfigure}%
    \hfill%
    \begin{subfigure}[b]{0.15\linewidth}
		\centering
		\includegraphics[trim=80 0 80 0,clip,width=\linewidth]{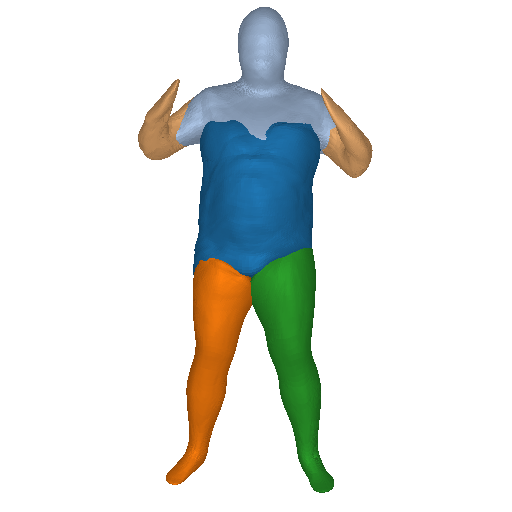}
    \end{subfigure}%
    \vspace{-1.2em}
    \vskip\baselineskip%
    \begin{subfigure}[b]{0.15\linewidth}
		\centering
		\includegraphics[trim=80 0 80 0,clip,width=\linewidth]{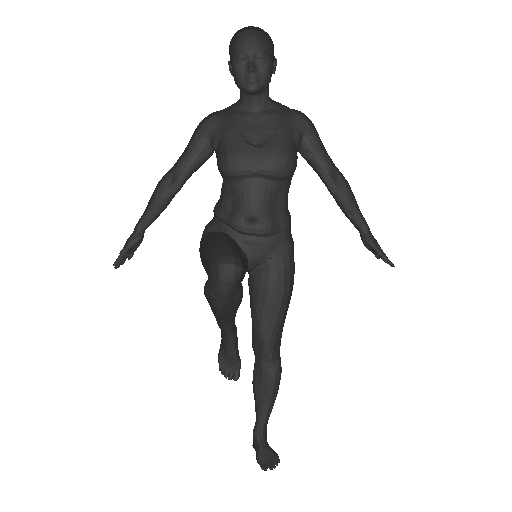}
    \end{subfigure}%
    \hfill%
    \begin{subfigure}[b]{0.15\linewidth}
		\centering
		\includegraphics[trim=80 0 80 0,clip,width=\linewidth]{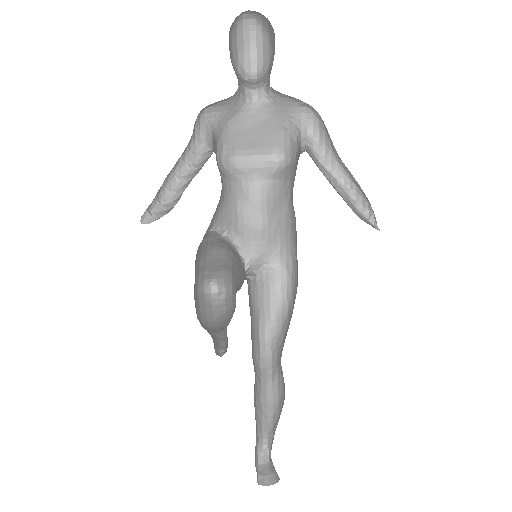}
    \end{subfigure}%
    \hfill%
    \begin{subfigure}[b]{0.15\linewidth}
		\centering
		\includegraphics[trim=80 0 80 0,clip,width=\linewidth]{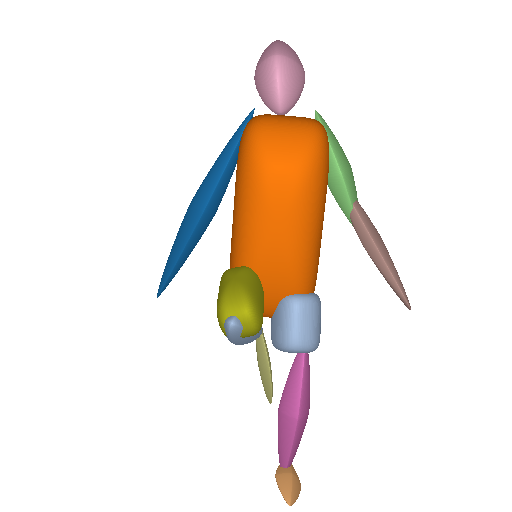}
    \end{subfigure}%
    \hfill%
    \begin{subfigure}[b]{0.15\linewidth}
		\centering
		\includegraphics[trim=80 0 80 0,clip,width=\linewidth]{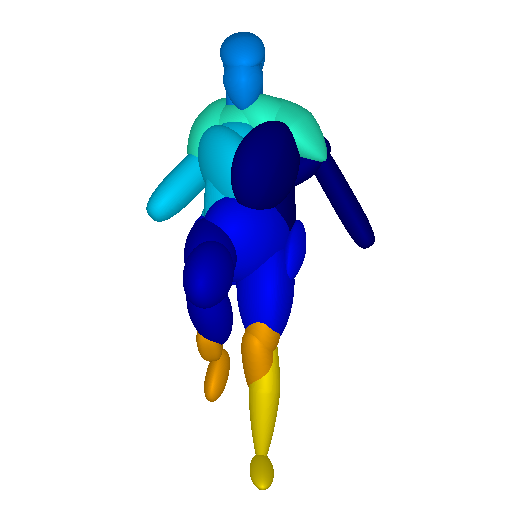}
    \end{subfigure}%
    \hfill%
    \begin{subfigure}[b]{0.15\linewidth}
		\centering
		\includegraphics[trim=80 0 80 0,clip,width=\linewidth]{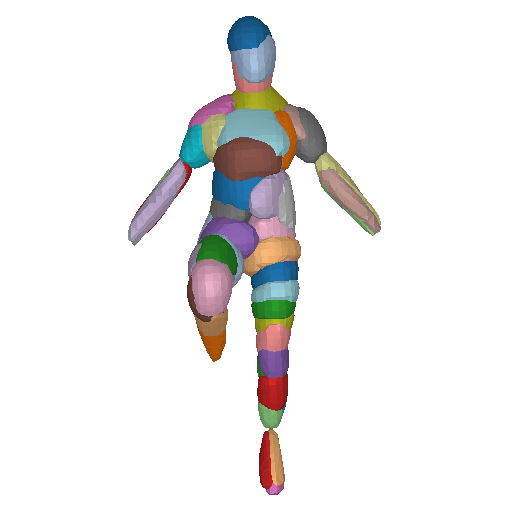}
    \end{subfigure}%
    \hfill%
    \begin{subfigure}[b]{0.15\linewidth}
		\centering
		\includegraphics[trim=80 0 80 0,clip,width=\linewidth]{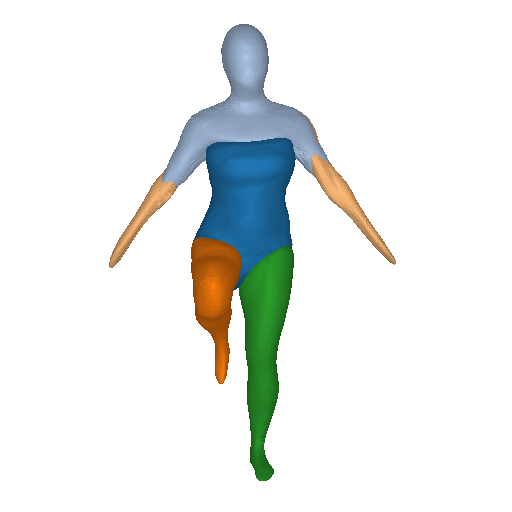}
    \end{subfigure}%
    \vspace{-1.2em}
    \vskip\baselineskip%
    \begin{subfigure}[b]{0.15\linewidth}
		\centering
		\includegraphics[trim=80 0 80 0,clip,width=\linewidth]{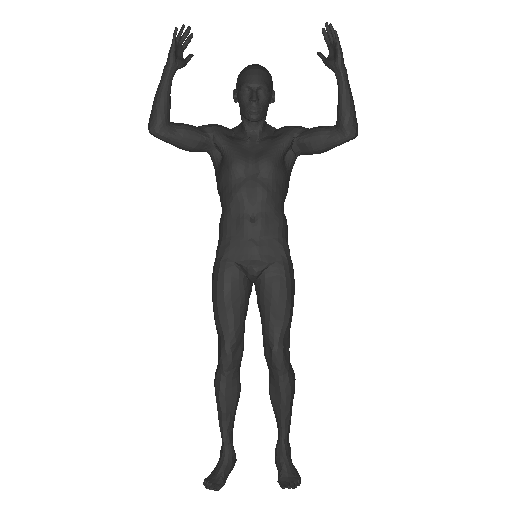}
    \end{subfigure}%
    \hfill%
    \begin{subfigure}[b]{0.15\linewidth}
		\centering
		\includegraphics[trim=80 0 80 0,clip,width=\linewidth]{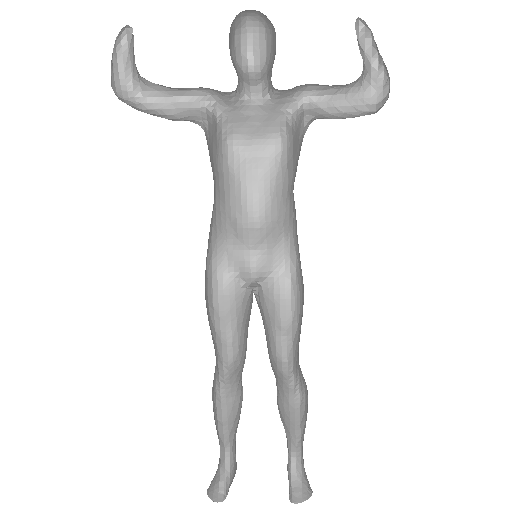}
    \end{subfigure}%
    \hfill%
    \begin{subfigure}[b]{0.15\linewidth}
		\centering
		\includegraphics[trim=80 0 80 0,clip,width=\linewidth]{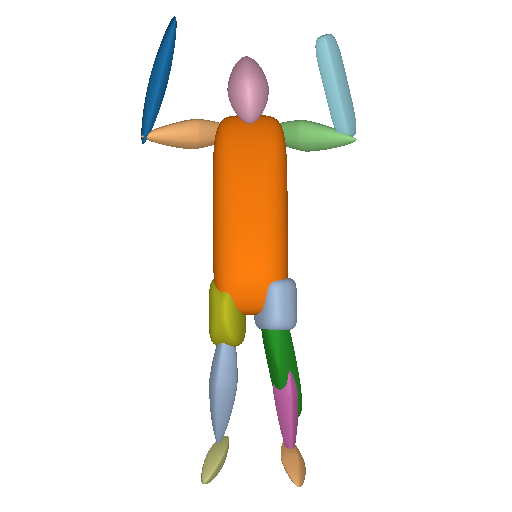}
    \end{subfigure}%
    \hfill%
    \begin{subfigure}[b]{0.15\linewidth}
		\centering
		\includegraphics[trim=80 0 80 0,clip,width=\linewidth]{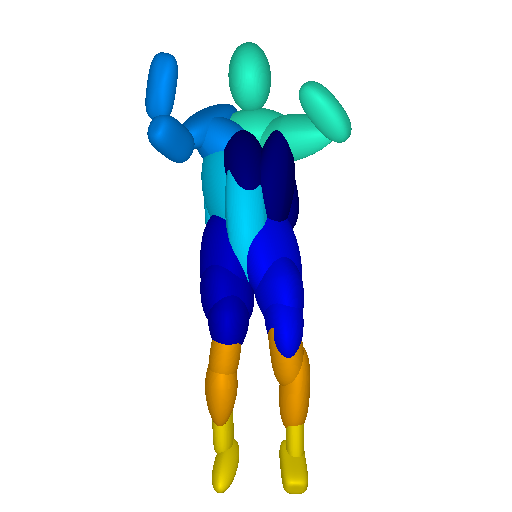}
    \end{subfigure}%
    \hfill%
    \begin{subfigure}[b]{0.15\linewidth}
		\centering
		\includegraphics[trim=80 0 80 0,clip,width=\linewidth]{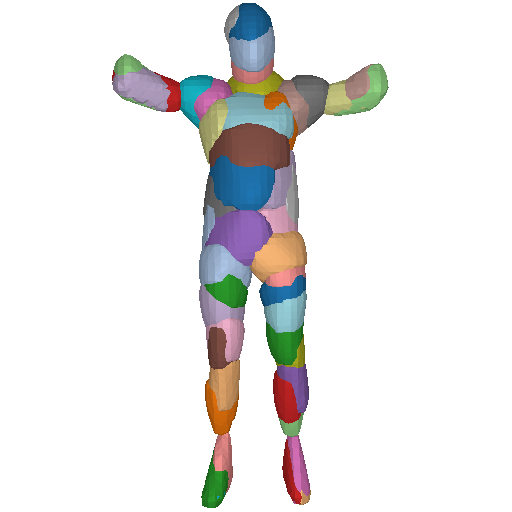}
    \end{subfigure}%
    \hfill%
    \begin{subfigure}[b]{0.15\linewidth}
		\centering
		\includegraphics[trim=80 0 80 0,clip,width=\linewidth]{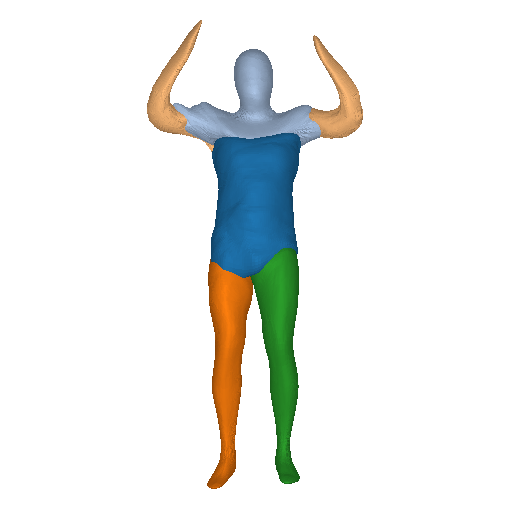}
    \end{subfigure}%
    \vspace{-1.2em}
    \vskip\baselineskip%
    \begin{subfigure}[b]{0.15\linewidth}
		\centering
		\includegraphics[trim=80 0 80 0,clip,width=\linewidth]{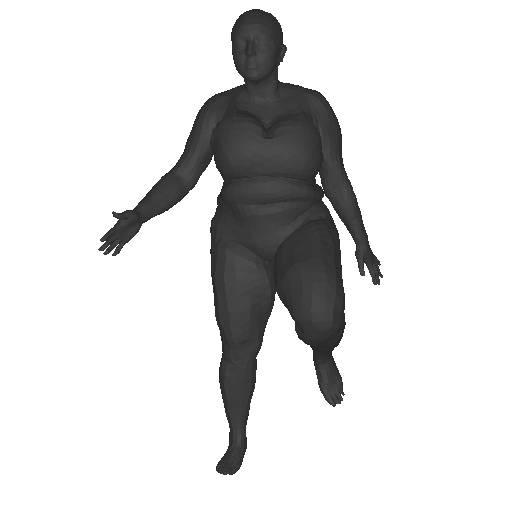}
    \end{subfigure}%
    \hfill%
    \begin{subfigure}[b]{0.15\linewidth}
		\centering
		\includegraphics[trim=80 0 80 0,clip,width=\linewidth]{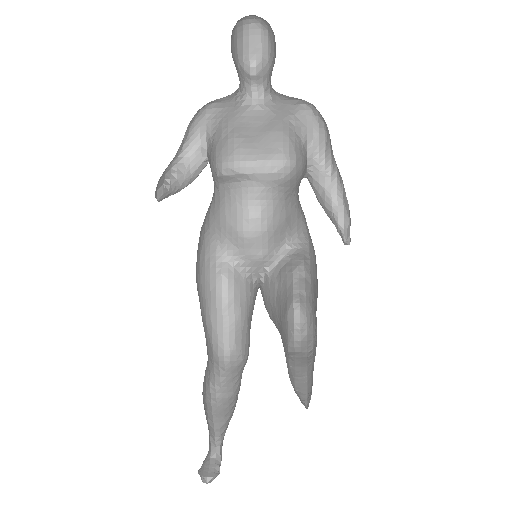}
    \end{subfigure}%
    \hfill%
    \begin{subfigure}[b]{0.15\linewidth}
		\centering
		\includegraphics[trim=80 0 80 0,clip,width=\linewidth]{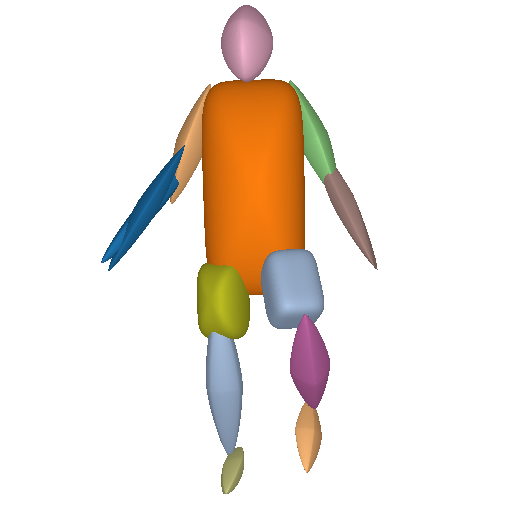}
    \end{subfigure}%
    \hfill%
    \begin{subfigure}[b]{0.15\linewidth}
		\centering
		\includegraphics[trim=80 0 80 0,clip,width=\linewidth]{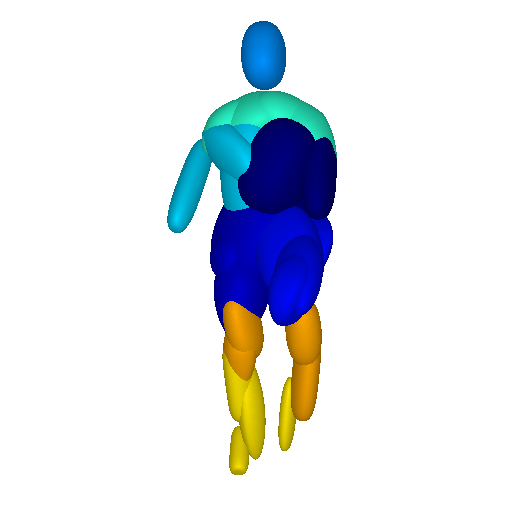}
    \end{subfigure}%
    \hfill%
    \begin{subfigure}[b]{0.15\linewidth}
		\centering
		\includegraphics[trim=80 0 80 0,clip,width=\linewidth]{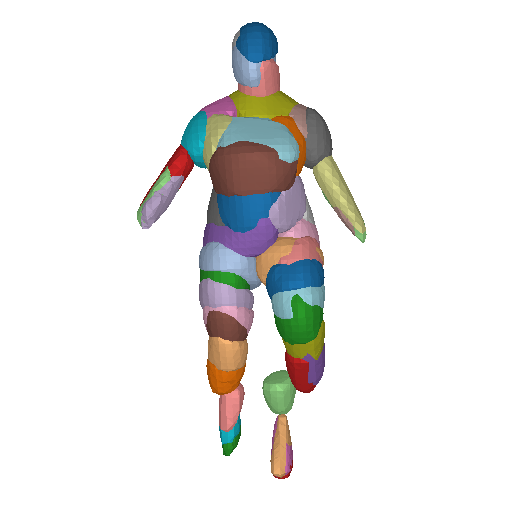}
    \end{subfigure}%
    \hfill%
    \begin{subfigure}[b]{0.15\linewidth}
		\centering
		\includegraphics[trim=80 0 80 0,clip,width=\linewidth]{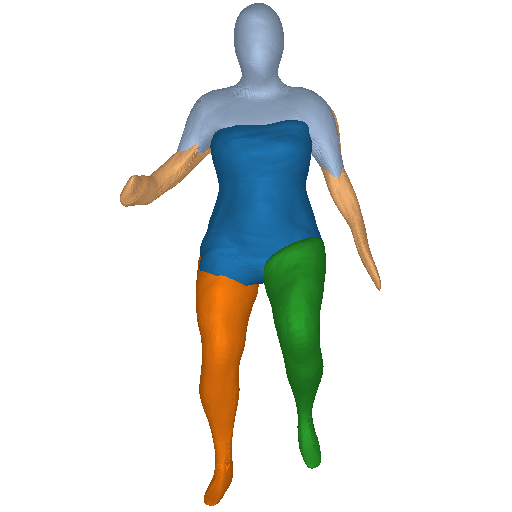}
    \end{subfigure}%
    \vskip\baselineskip%
    \vspace{-1.2em}
    \caption{{\bf Single Image 3D Reconstruction on D-FAUST}. The input image is shown on the first column and the rest contain predictions of all methods: OccNet (second), primitive-based predictions with superquadrics (third and fourth) and convexes (fifth) and ours with $5$ primitives (last).}
    \label{fig:qualitative_dfaust_supp}
    \vspace{-1.2em}
\end{figure}

In \figref{fig:qualitative_dfaust_supp}, we provide additional experimental results on the single-view 3D reconstruction task on D-FAUST. Similar, to the results in section 4.2 of the main submission, we compare our model with $5$ primitives, with CvxNet and SQs with $50$ primitives and H-SQs with $32$ primitives.
We observe that while CvxNet and H-SQs accurately capture the geometry of the human body, their predicted primitives lack any semantic interpretation, as primitives do not correspond to actual geometric parts of the human body.
Instead, SQs lead to semantic and parsimonious shape abstractions that are not accurate.
On the other hand, Neural Parts achieve both semantic interpretability and high reconstruction quality.

\clearpage
\section{Experiment on FreiHAND}

In this section, we provide additional information regarding our experiments on FreiHAND dataset~\cite{Zimmermann2019ICCV}.
For FreiHAND, we select the first $5000$ hand poses and generate meshes using the provided MANO parameters~\cite{Romero2017TOG}.
We render them from a fixed orientation and use $70\%$, $20\%$, $10\%$ for the train, test and validation splits.
Note that MANO does not generate watertight meshes, thus we need to post-process the meshes in order to create occupancy pairs.
This is achieved by adding triangular faces using the following vertex indices $38, 92, 234, 239, 279, 215, 214, 121, 78, 79, 108, 120, 119, 117, 118, 122$.

\subsection{Additional Experimental Results}

\begin{figure}[H]
    \centering
    \begin{subfigure}[b]{0.15\linewidth}
		\centering
        Input
    \end{subfigure}%
    \hfill%
    \begin{subfigure}[b]{0.15\linewidth}
		\centering
        OccNet
    \end{subfigure}%
    \hfill%
    \begin{subfigure}[b]{0.15\linewidth}
		\centering
        SQs
    \end{subfigure}%
    \hfill%
    \begin{subfigure}[b]{0.15\linewidth}
		\centering
        H-SQs
    \end{subfigure}%
    \hfill%
    \begin{subfigure}[b]{0.15\linewidth}
		\centering
        CvxNet
    \end{subfigure}%
    \hfill%
    \begin{subfigure}[b]{0.15\linewidth}
        \centering
        Ours
    \end{subfigure}
    \vspace{-1em}
    \vskip\baselineskip%
    \begin{subfigure}[b]{0.15\linewidth}
		\centering
		\includegraphics[trim=0 10 0 -10,clip,width=\linewidth]{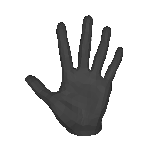}
    \end{subfigure}%
    \hfill%
    \begin{subfigure}[b]{0.15\linewidth}
		\centering
		\includegraphics[trim=10 0 10 0,clip,width=\linewidth]{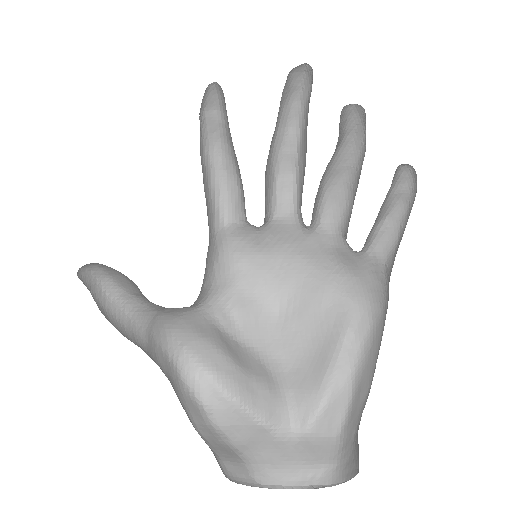}
    \end{subfigure}%
    \hfill%
    \begin{subfigure}[b]{0.15\linewidth}
		\centering
		\includegraphics[trim=-10 0 -10 0,clip,width=\linewidth]{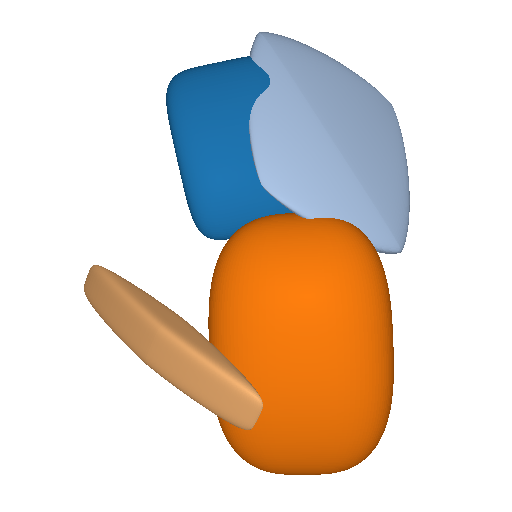}
    \end{subfigure}%
    \hfill%
    \begin{subfigure}[b]{0.15\linewidth}
		\centering
		\includegraphics[trim=5 0 5 0,clip,width=\linewidth]{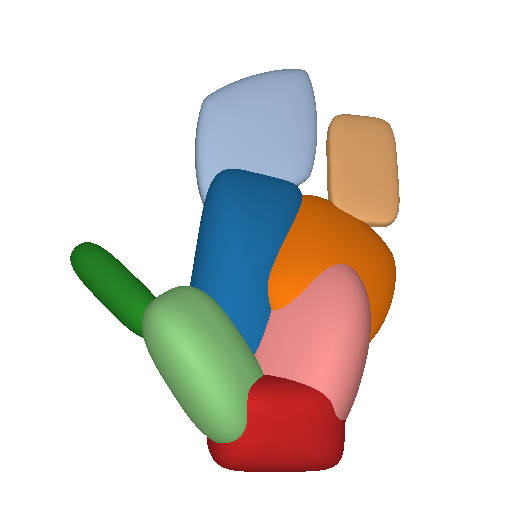}
    \end{subfigure}%
    \hfill%
    \begin{subfigure}[b]{0.15\linewidth}
		\centering
		\includegraphics[trim=20 0 20 0,clip,width=\linewidth]{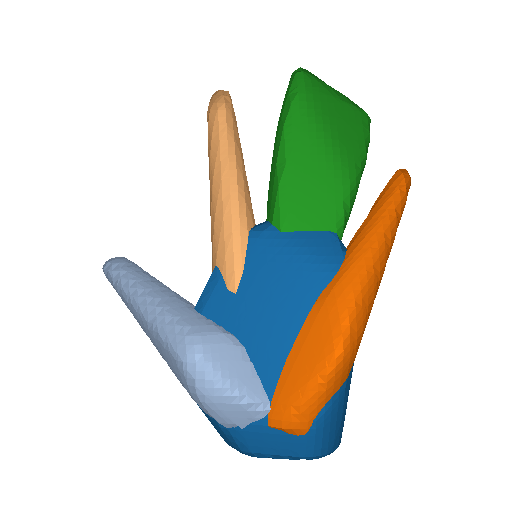}
    \end{subfigure}%
    \hfill%
    \begin{subfigure}[b]{0.15\linewidth}
		\centering
		\includegraphics[trim=20 0 20 0,clip,width=\linewidth]{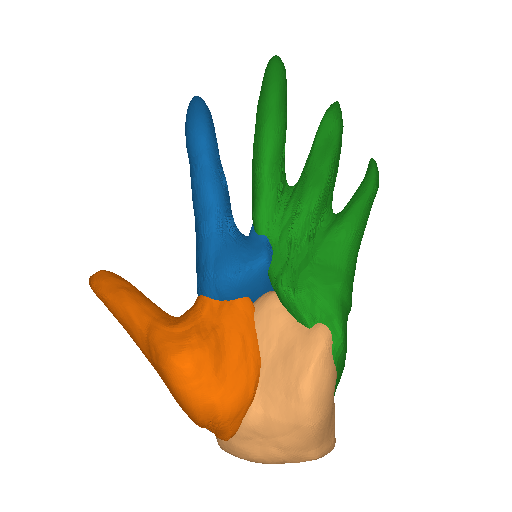}
    \end{subfigure}%
    \vspace{-1.75em}
    \vskip\baselineskip%
    \begin{subfigure}[b]{0.15\linewidth}
		\centering
		\includegraphics[trim=-4 10 -4 -10,clip,width=\linewidth]{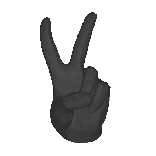}
    \end{subfigure}%
    \hfill%
    \begin{subfigure}[b]{0.15\linewidth}
		\centering
		\includegraphics[trim=10 0 10 0,clip,width=\linewidth]{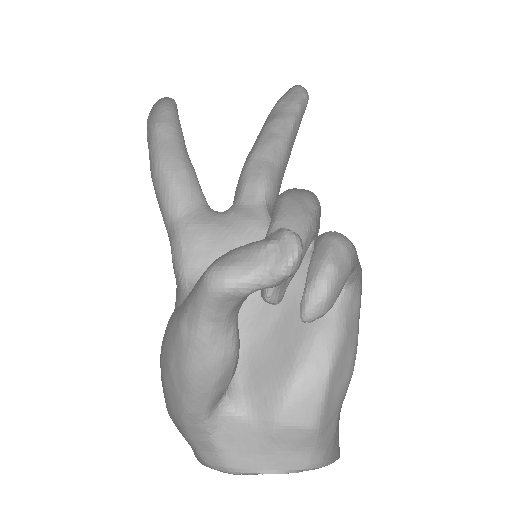}
    \end{subfigure}%
    \hfill%
    \begin{subfigure}[b]{0.15\linewidth}
		\centering
		\includegraphics[trim=-10 0 -10 0,clip,width=\linewidth]{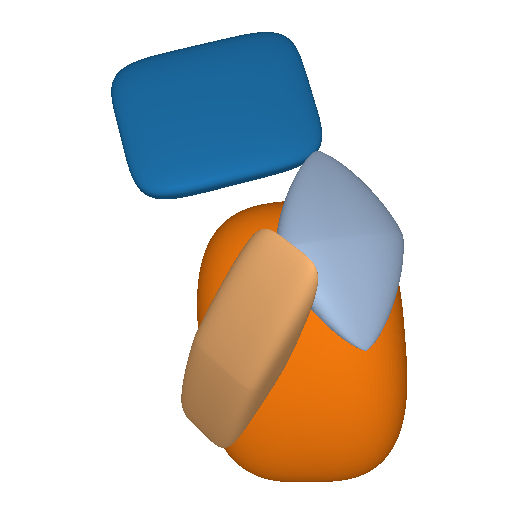}
    \end{subfigure}%
    \hfill%
    \begin{subfigure}[b]{0.15\linewidth}
		\centering
		\includegraphics[trim=0 0 0 0,clip,width=\linewidth]{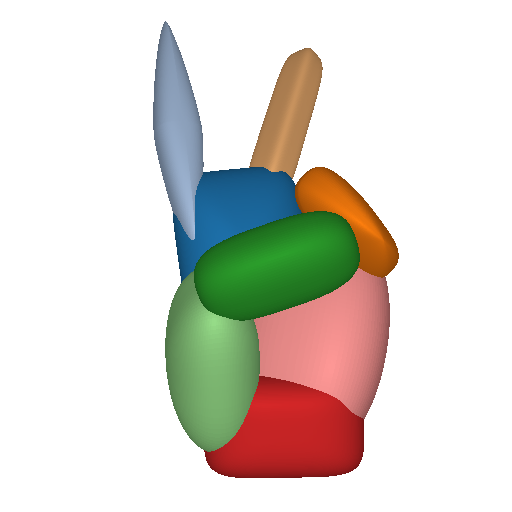}
    \end{subfigure}%
    \hfill%
    \begin{subfigure}[b]{0.15\linewidth}
		\centering
		\includegraphics[trim=10 0 10 0,clip,width=\linewidth]{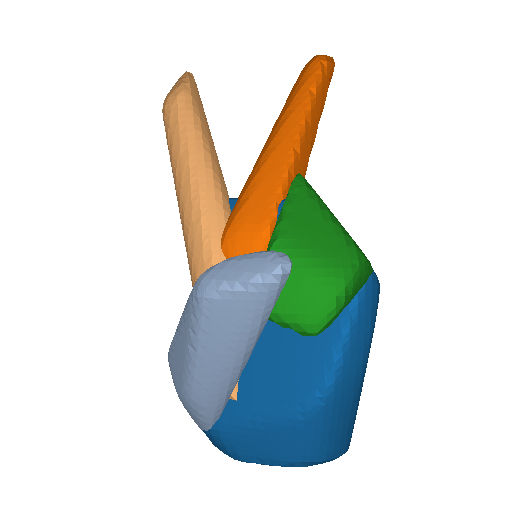}
    \end{subfigure}%
    \hfill%
    \begin{subfigure}[b]{0.15\linewidth}
		\centering
		\includegraphics[trim=30 40 30 -40,clip,width=\linewidth]{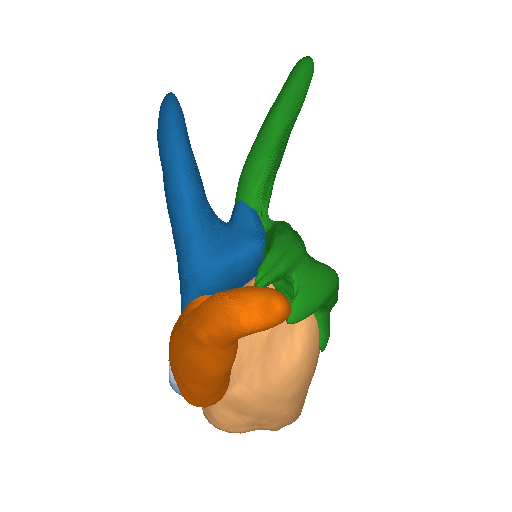}
    \end{subfigure}%
    \vspace{-1.75em}
    \vskip\baselineskip%
    \begin{subfigure}[b]{0.15\linewidth}
		\centering
		\includegraphics[trim=0 8 0 -8,clip,width=\linewidth]{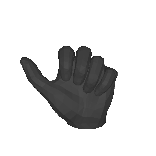}
    \end{subfigure}%
    \hfill%
    \begin{subfigure}[b]{0.15\linewidth}
		\centering
		\includegraphics[trim=-60 0 -60 0,clip,width=\linewidth]{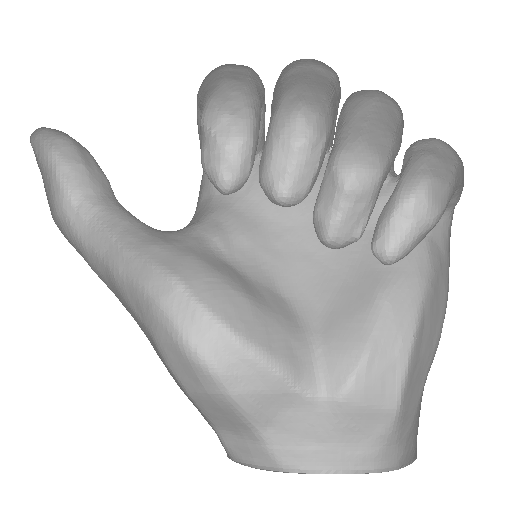}
    \end{subfigure}%
    \hfill%
    \begin{subfigure}[b]{0.15\linewidth}
		\centering
		\includegraphics[trim=0 0 0 0,clip,width=\linewidth]{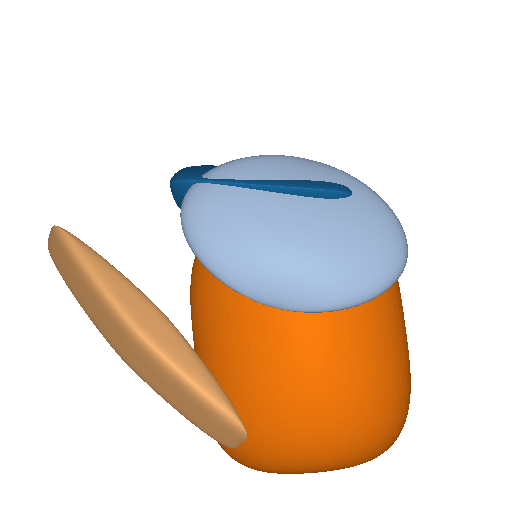}
    \end{subfigure}%
    \hfill%
    \begin{subfigure}[b]{0.15\linewidth}
		\centering
		\includegraphics[trim=0 0 0 0,clip,width=\linewidth]{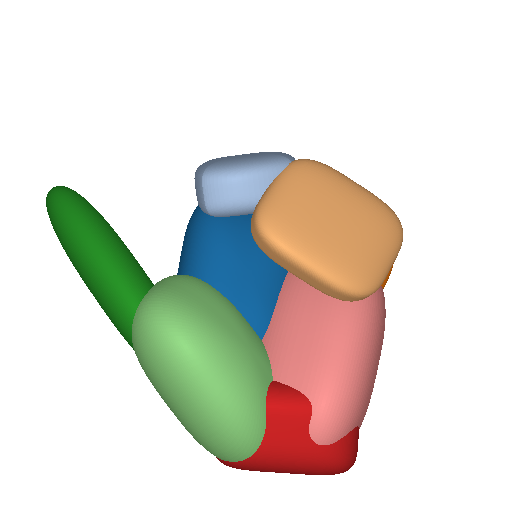}
    \end{subfigure}%
    \hfill%
    \begin{subfigure}[b]{0.15\linewidth}
		\centering
		\includegraphics[trim=0 0 0 0,clip,width=\linewidth]{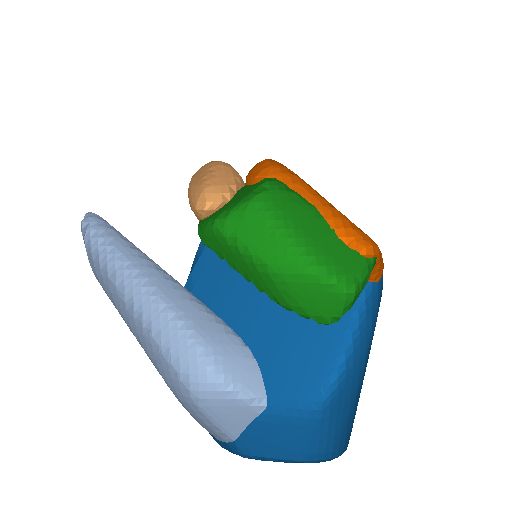}
    \end{subfigure}%
    \hfill%
    \begin{subfigure}[b]{0.15\linewidth}
		\centering
		\includegraphics[trim=-30 0 -30 0,clip,width=\linewidth]{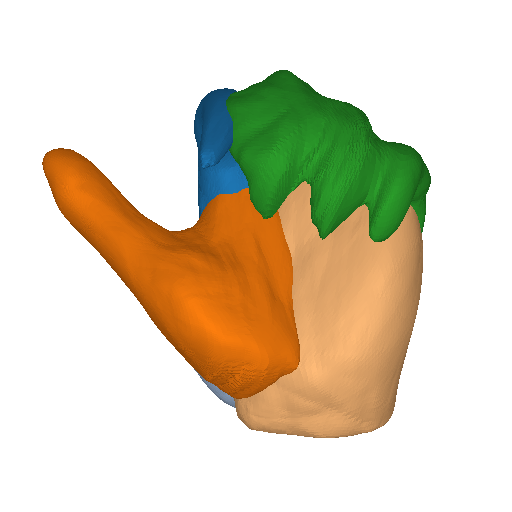}
    \end{subfigure}%
    \vspace{-4.75em}
    \vskip\baselineskip%
    \begin{subfigure}[b]{0.15\linewidth}
		\centering
		\includegraphics[trim=40 45 40 -45,clip,width=\linewidth]{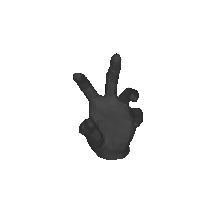}
    \end{subfigure}%
    \hfill%
    \begin{subfigure}[b]{0.15\linewidth}
		\centering
		\includegraphics[trim=0 0 0 0,clip,width=\linewidth]{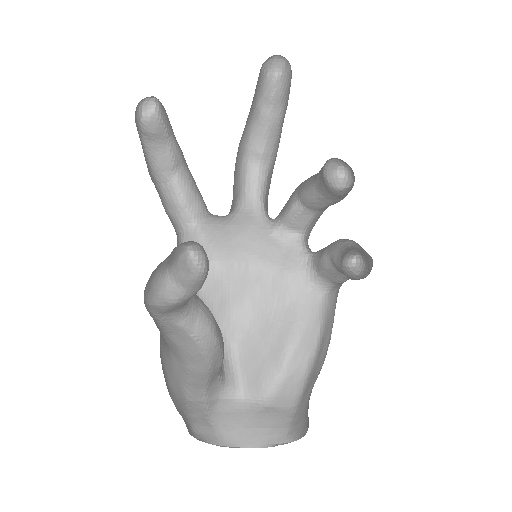}
    \end{subfigure}%
    \hfill%
    \begin{subfigure}[b]{0.15\linewidth}
		\centering
		\includegraphics[trim=0 0 0 0,clip,width=\linewidth]{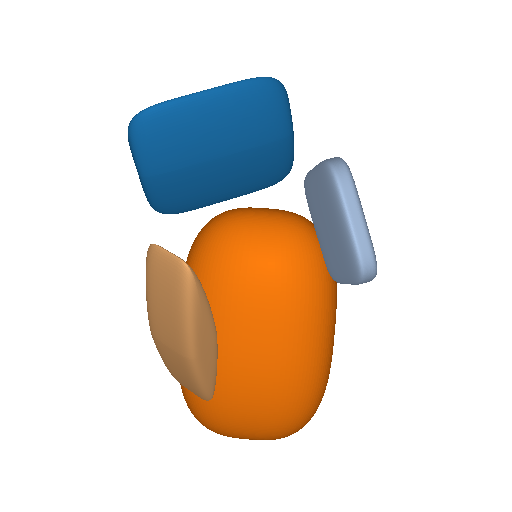}
    \end{subfigure}%
    \hfill%
    \begin{subfigure}[b]{0.15\linewidth}
		\centering
		\includegraphics[trim=0 0 0 0,clip,width=\linewidth]{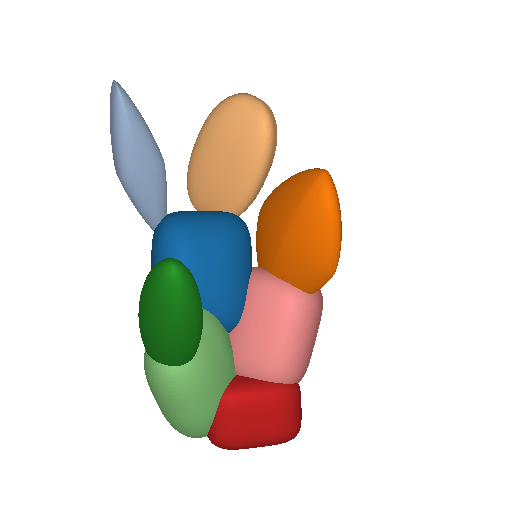}
    \end{subfigure}%
    \hfill%
    \begin{subfigure}[b]{0.15\linewidth}
		\centering
		\includegraphics[trim=0 0 0 0,clip,width=\linewidth]{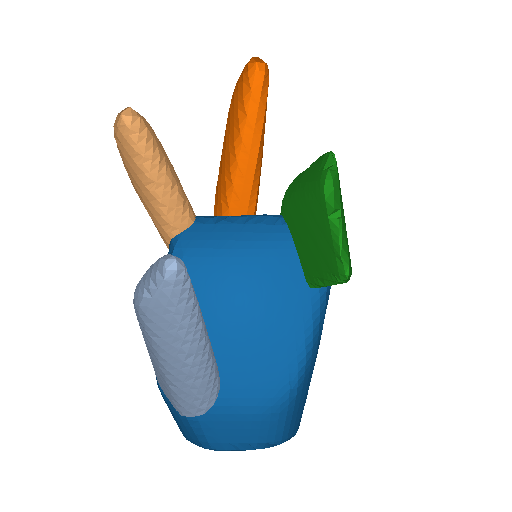}
    \end{subfigure}%
    \hfill%
    \begin{subfigure}[b]{0.15\linewidth}
		\centering
		\includegraphics[trim=0 0 0 0,clip,width=\linewidth]{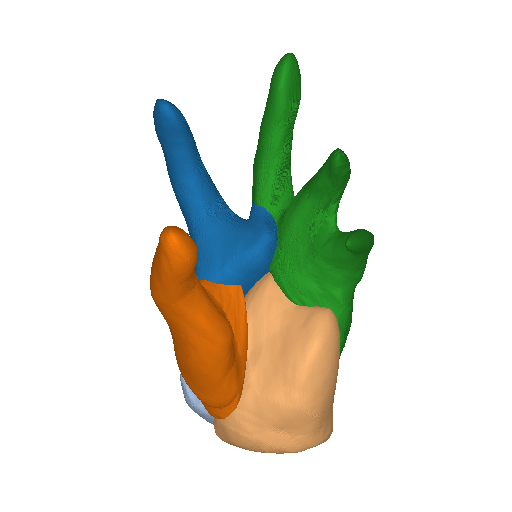}
    \end{subfigure}%
    \vspace{-4em}
    \vskip\baselineskip%
    \begin{subfigure}[b]{0.15\linewidth}
		\centering
		\includegraphics[trim=30 45 30 -45,clip,width=\linewidth]{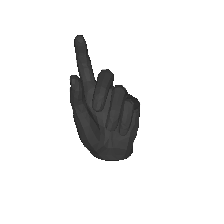}
    \end{subfigure}%
    \hfill%
    \begin{subfigure}[b]{0.15\linewidth}
		\centering
		\includegraphics[trim=5 0 5 0,clip,width=\linewidth]{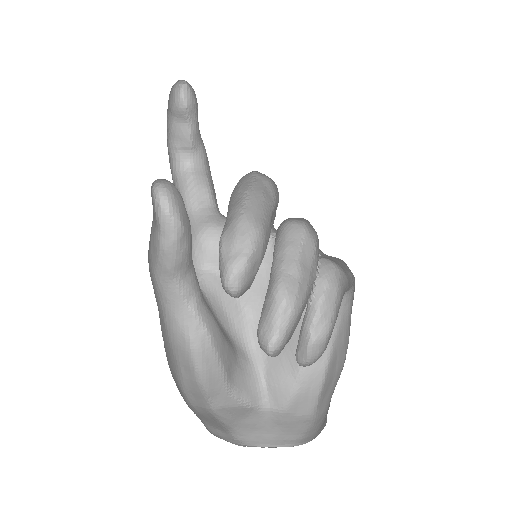}
    \end{subfigure}%
    \hfill%
    \begin{subfigure}[b]{0.15\linewidth}
		\centering
		\includegraphics[trim=5 0 5 0,clip,width=\linewidth]{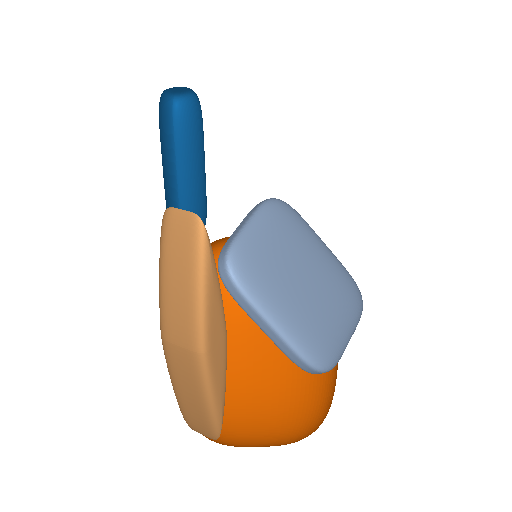}
    \end{subfigure}%
    \hfill%
    \begin{subfigure}[b]{0.15\linewidth}
		\centering
		\includegraphics[trim=5 0 5 0,clip,width=\linewidth]{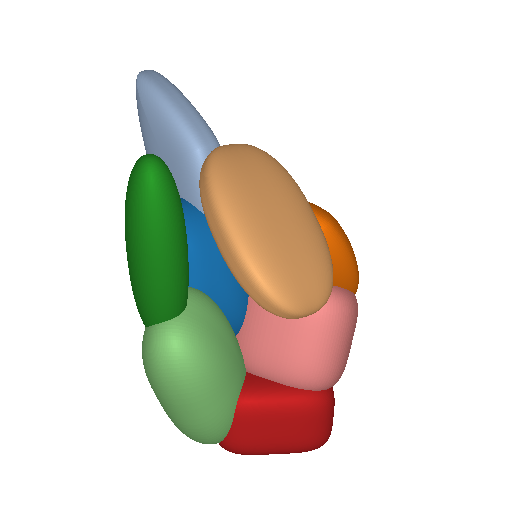}
    \end{subfigure}%
    \hfill%
    \begin{subfigure}[b]{0.15\linewidth}
		\centering
		\includegraphics[trim=5 0 5 0,clip,width=\linewidth]{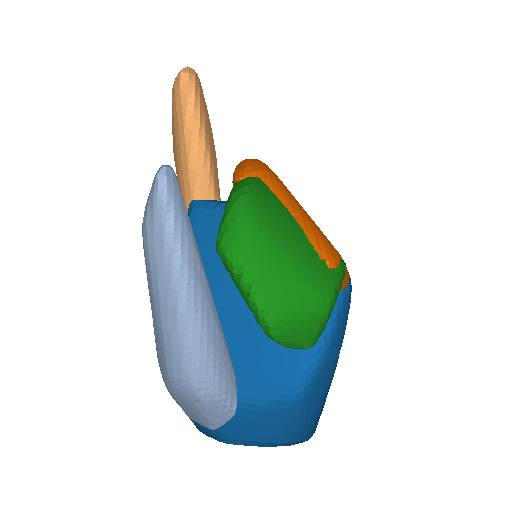}
    \end{subfigure}%
    \hfill%
    \begin{subfigure}[b]{0.15\linewidth}
		\centering
		\includegraphics[trim=5 0 5 0,clip,width=\linewidth]{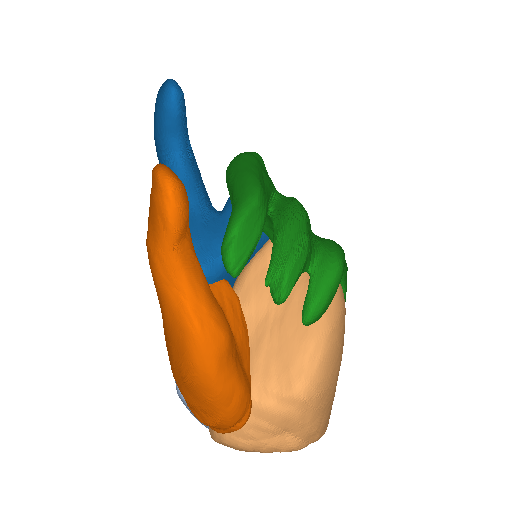}
    \end{subfigure}
    \caption{{\bf{Single Image 3D Reconstruction on FreiHAND}}.
    We compare our model with OccNet, SQs and CvxNet with $5$ primitives and
    H-SQs with $8$ primitives.}
    \label{fig:freihand_results_supp}
    \vspace{-1.2em}
\end{figure}

In \figref{fig:freihand_results_supp}, we provide additional qualitative
results on the single-view 3D reconstruction task on FreiHAND. Similar to the
results in section $4.2$ of the main submission, we compare our model with $5$
primitives to SQs and CvxNet with $5$ primitives and H-SQs with $8$. Note that
for H-SQs we can only use a maximum number of primitives that is a power of
$2$. H-SQs with $4$ primitives performed poorly; thus we increased the number
of parts. We observe that Neural Parts yield semantically meaningful parts that
accurately capture the hand geometry. Note that accurately modeling objects with
non-rigid parts, such as the flexible human fingers, requires primitives that
can bend arbitrarily (see thumb in second row
\figref{fig:freihand_results_supp}). This is only possible with Neural Parts,
as we do not impose any kind of constraint on the primitive shape.

\section{Experiment on ShapeNet}

\begin{figure}[H]
    \begin{subfigure}[b]{0.20\linewidth}
		\centering
        Input
    \end{subfigure}%
    \hfill%
    \begin{subfigure}[b]{0.20\linewidth}
		\centering
        OccNet
    \end{subfigure}%
    \hfill%
    \begin{subfigure}[b]{0.20\linewidth}
		\centering
        CvxNet-5
    \end{subfigure}%
    \hfill%
    \begin{subfigure}[b]{0.20\linewidth}
		\centering
        CvxNet-25
    \end{subfigure}%
    \hfill%
    \begin{subfigure}[b]{0.20\linewidth}
        \centering
        Ours
    \end{subfigure}
    \vspace{-2.2em}
    \vskip\baselineskip%
    \begin{subfigure}[b]{0.20\linewidth}
		\centering
		\includegraphics[width=0.8\linewidth]{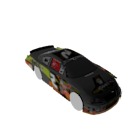}
    \end{subfigure}%
    \hfill%
    \begin{subfigure}[b]{0.20\linewidth}
		\centering
		\includegraphics[width=0.8\linewidth]{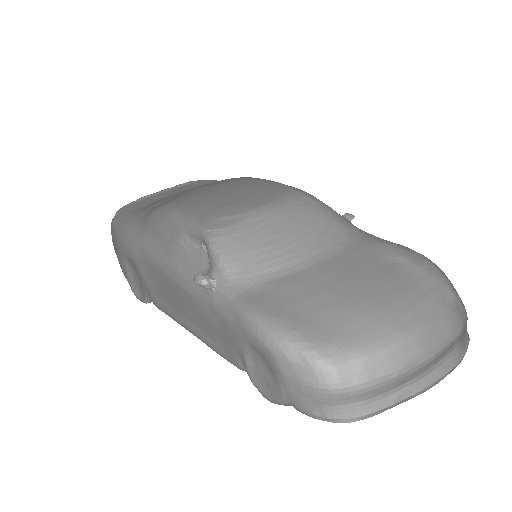}
    \end{subfigure}%
    \hfill%
    \begin{subfigure}[b]{0.20\linewidth}
		\centering
		\includegraphics[width=0.8\linewidth]{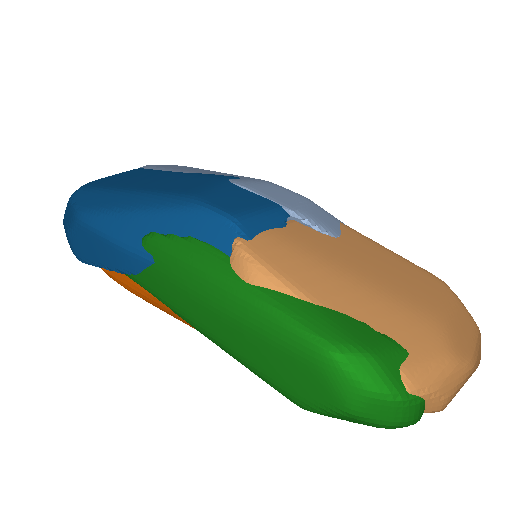}
    \end{subfigure}%
    \hfill%
    \begin{subfigure}[b]{0.20\linewidth}
		\centering
		\includegraphics[width=0.8\linewidth]{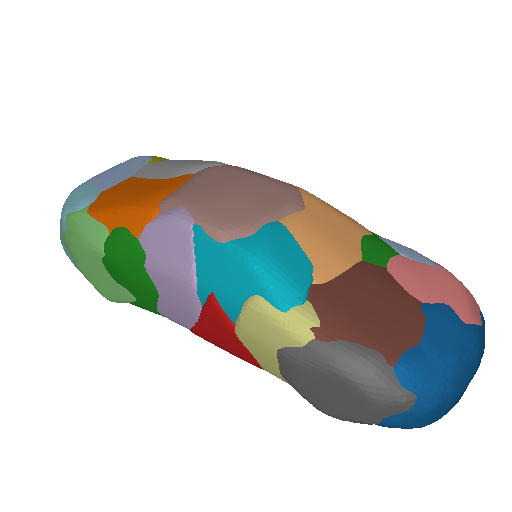}
    \end{subfigure}%
    \hfill%
    \begin{subfigure}[b]{0.20\linewidth}
		\centering
		\includegraphics[width=0.8\linewidth]{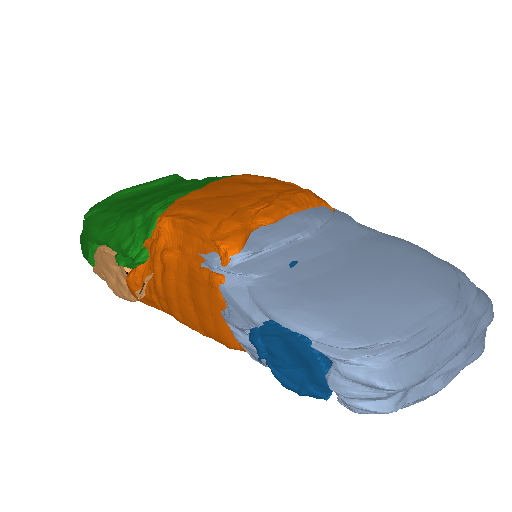}
    \end{subfigure}%
    \vspace{-2.2em}
    \vskip\baselineskip%
    \begin{subfigure}[b]{0.20\linewidth}
    	\centering
		\includegraphics[width=0.9\linewidth]{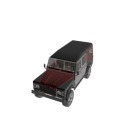}
    \end{subfigure}%
    \hfill%
    \begin{subfigure}[b]{0.20\linewidth}
    	\centering
		\includegraphics[width=0.8\linewidth]{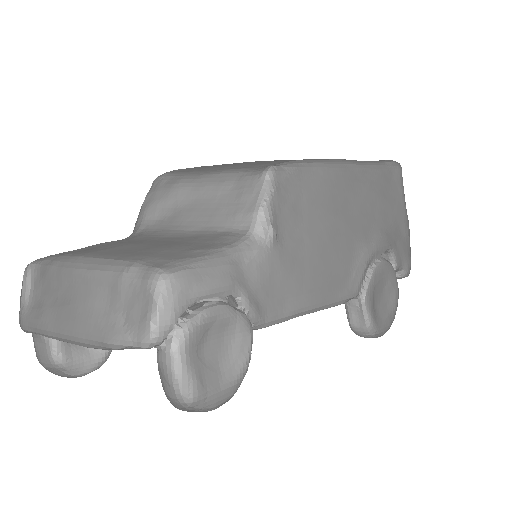}
    \end{subfigure}%
    \hfill%
    \begin{subfigure}[b]{0.20\linewidth}
    	\centering
		\includegraphics[width=0.8\linewidth]{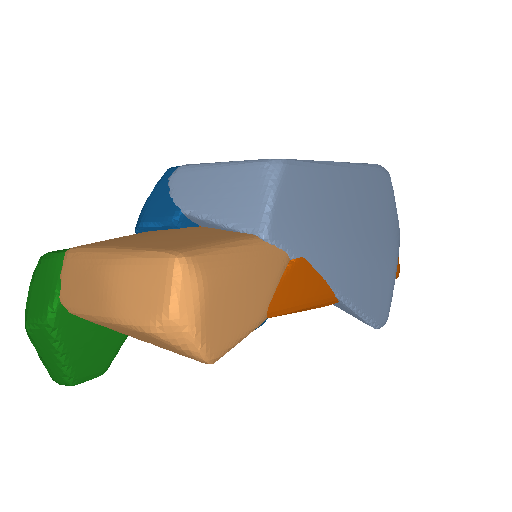}
    \end{subfigure}%
    \hfill%
    \begin{subfigure}[b]{0.20\linewidth}
    	\centering
		\includegraphics[width=0.8\linewidth]{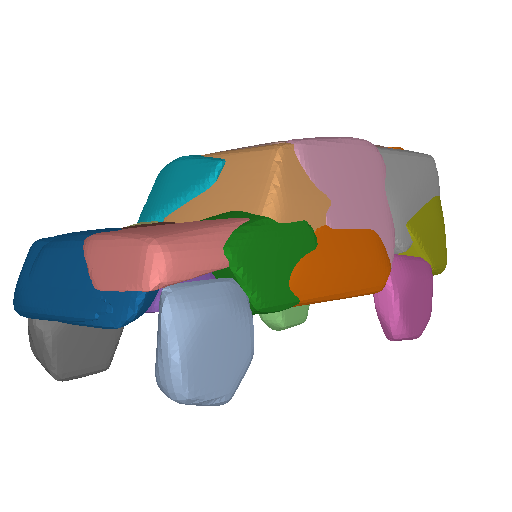}
    \end{subfigure}%
    \hfill%
    \begin{subfigure}[b]{0.20\linewidth}
    	\centering
		\includegraphics[width=0.8\linewidth]{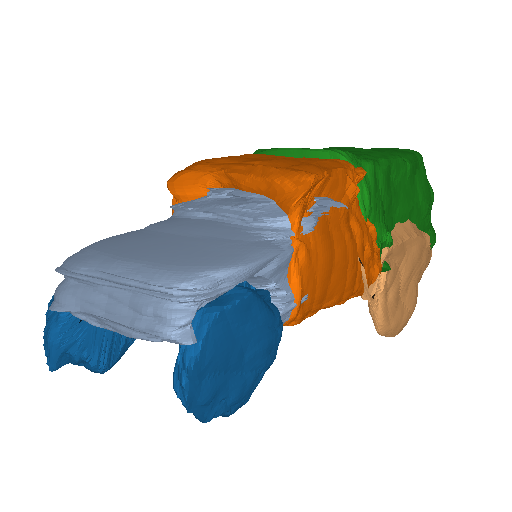}
    \end{subfigure}%
    \vspace{-3.2em}
    \vskip\baselineskip%
    \begin{subfigure}[b]{0.20\linewidth}
    	\centering
		\includegraphics[width=\linewidth]{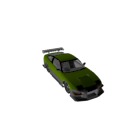}
    \end{subfigure}%
    \hfill%
    \begin{subfigure}[b]{0.20\linewidth}
    	\centering
		\includegraphics[width=0.8\linewidth]{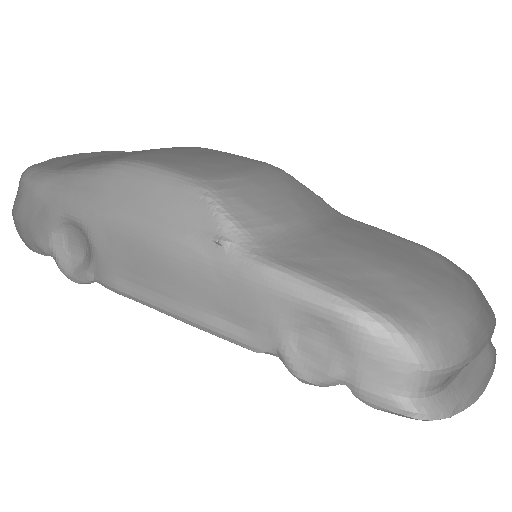}
    \end{subfigure}%
    \hfill%
    \begin{subfigure}[b]{0.20\linewidth}
    	\centering
		\includegraphics[width=0.8\linewidth]{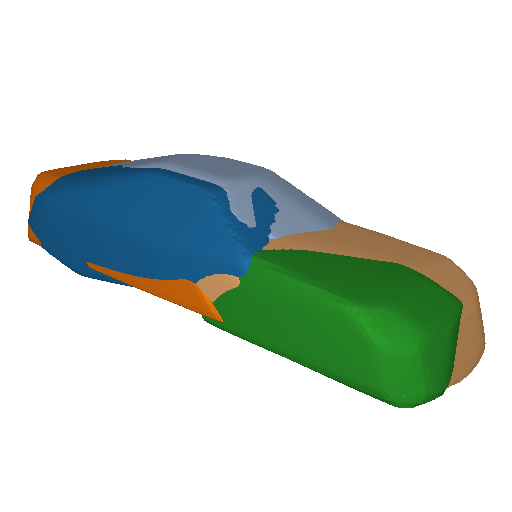}
    \end{subfigure}%
    \hfill%
    \begin{subfigure}[b]{0.20\linewidth}
    	\centering
		\includegraphics[width=0.8\linewidth]{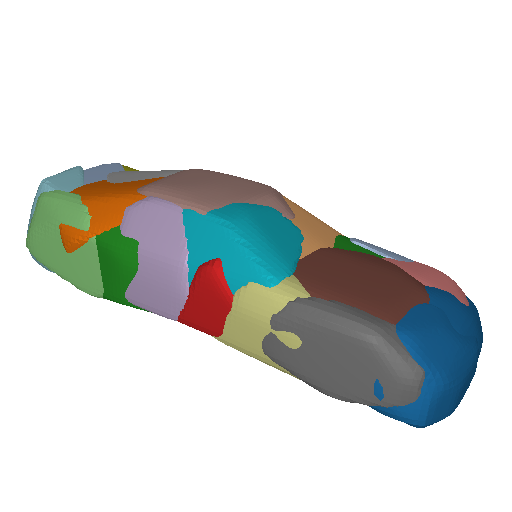}
    \end{subfigure}%
    \hfill%
    \begin{subfigure}[b]{0.20\linewidth}
    	\centering
		\includegraphics[width=0.8\linewidth]{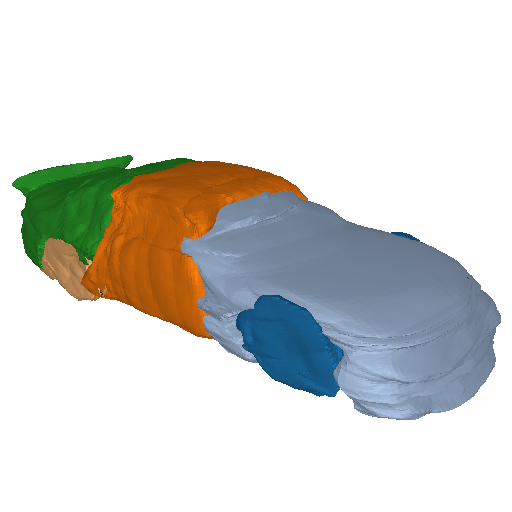}
    \end{subfigure}%
    \vspace{-2.2em}
    \vskip\baselineskip%
    \begin{subfigure}[b]{0.20\linewidth}
    	\centering
		\includegraphics[width=0.8\linewidth]{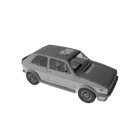}
    \end{subfigure}%
    \hfill%
    \begin{subfigure}[b]{0.20\linewidth}
    	\centering
		\includegraphics[width=0.8\linewidth]{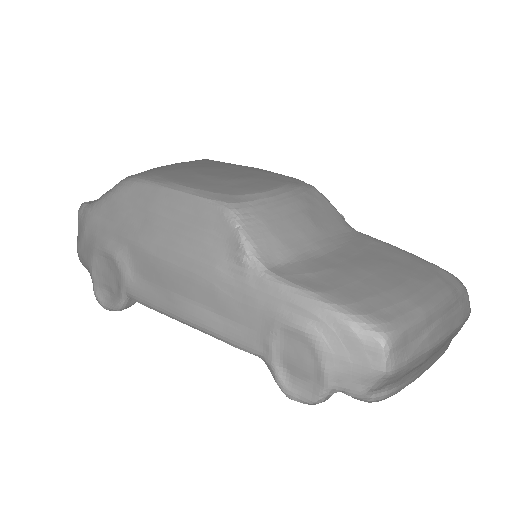}
    \end{subfigure}%
    \hfill%
    \begin{subfigure}[b]{0.20\linewidth}
    	\centering
		\includegraphics[width=0.8\linewidth]{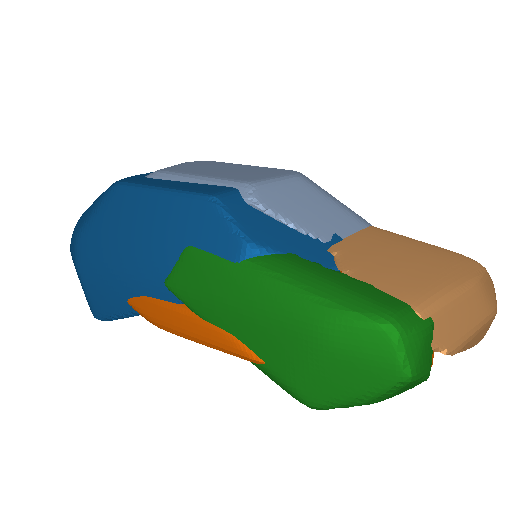}
    \end{subfigure}%
    \hfill%
    \begin{subfigure}[b]{0.20\linewidth}
    	\centering
		\includegraphics[width=0.8\linewidth]{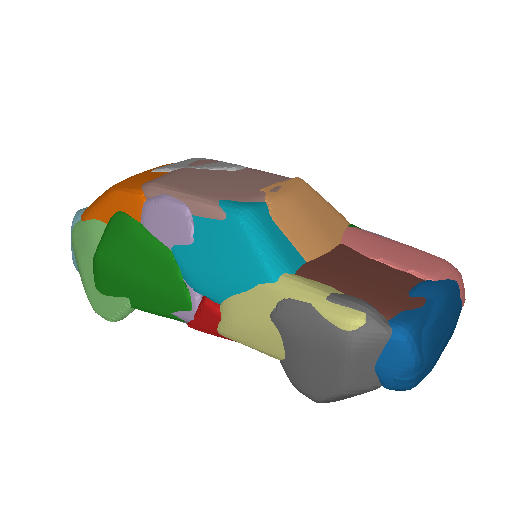}
    \end{subfigure}%
    \hfill%
    \begin{subfigure}[b]{0.20\linewidth}
    	\centering
		\includegraphics[width=0.8\linewidth]{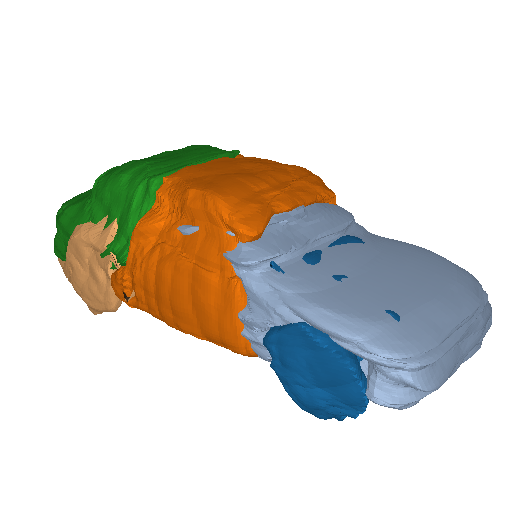}
    \end{subfigure}%
    \vspace{-3.2em}
    \vskip\baselineskip%
    \begin{subfigure}[b]{0.20\linewidth}
    	\centering
		\includegraphics[width=\linewidth]{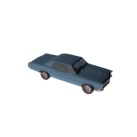}
    \end{subfigure}%
    \hfill%
    \begin{subfigure}[b]{0.20\linewidth}
    	\centering
		\includegraphics[width=0.8\linewidth]{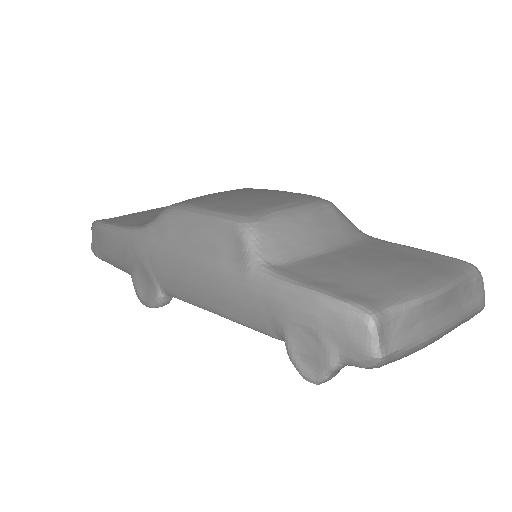}
    \end{subfigure}%
    \hfill%
    \begin{subfigure}[b]{0.20\linewidth}
    	\centering
		\includegraphics[width=0.8\linewidth]{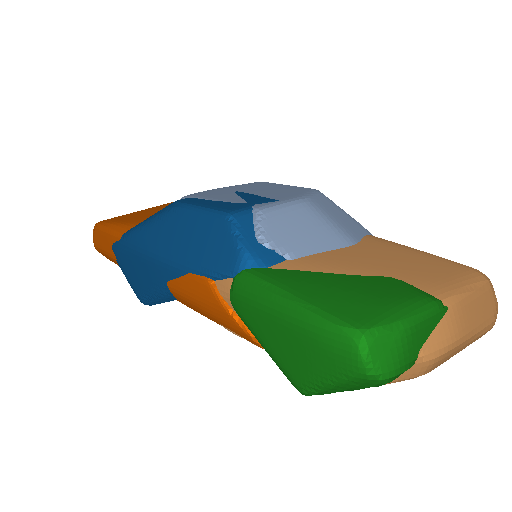}
    \end{subfigure}%
    \hfill%
    \begin{subfigure}[b]{0.20\linewidth}
    	\centering
		\includegraphics[width=0.8\linewidth]{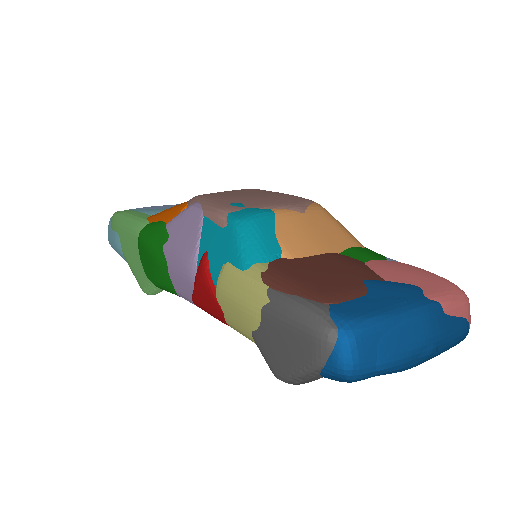}
    \end{subfigure}%
    \hfill%
    \begin{subfigure}[b]{0.20\linewidth}
    	\centering
		\includegraphics[width=0.8\linewidth]{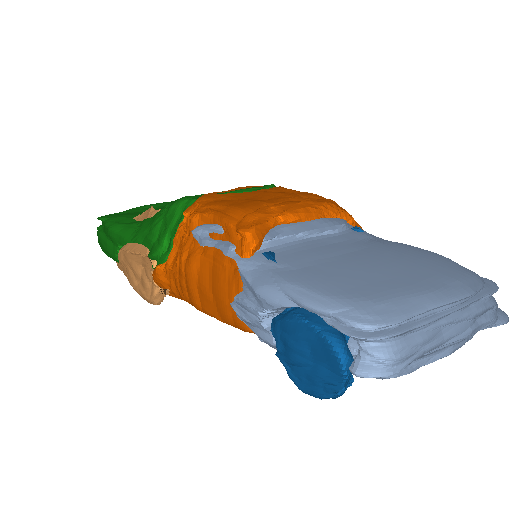}
    \end{subfigure}%
    \vspace{-4.2em}
    \vskip\baselineskip%
    \begin{subfigure}[b]{0.20\linewidth}
    	\centering
		\includegraphics[width=\linewidth]{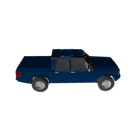}
    \end{subfigure}%
    \hfill%
    \begin{subfigure}[b]{0.20\linewidth}
    	\centering
		\includegraphics[width=0.8\linewidth]{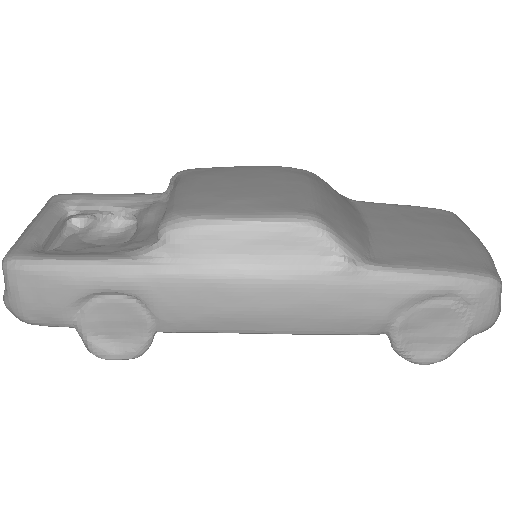}
    \end{subfigure}%
    \hfill%
    \begin{subfigure}[b]{0.20\linewidth}
    	\centering
		\includegraphics[width=0.8\linewidth]{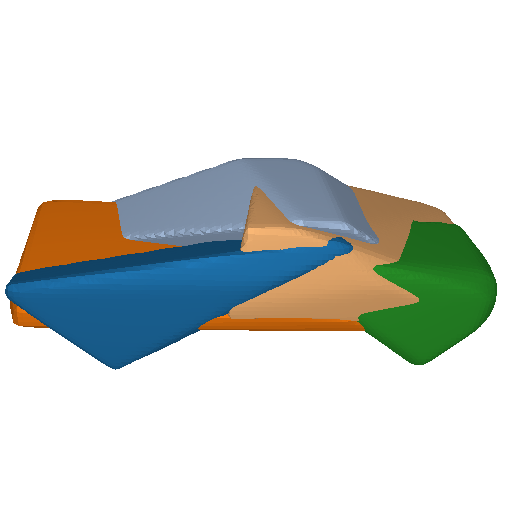}
    \end{subfigure}%
    \hfill%
    \begin{subfigure}[b]{0.20\linewidth}
    	\centering
		\includegraphics[width=0.8\linewidth]{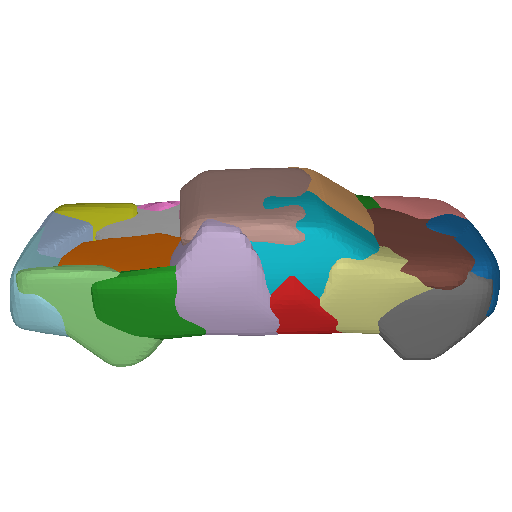}
    \end{subfigure}%
    \hfill%
    \begin{subfigure}[b]{0.20\linewidth}
    	\centering
		\includegraphics[width=0.8\linewidth]{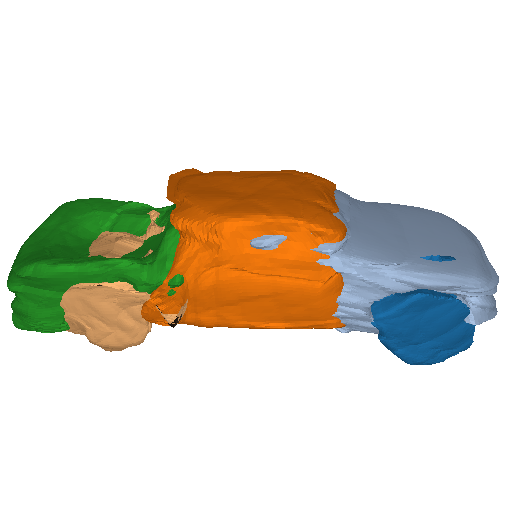}
    \end{subfigure}%
    \vspace{-1.2em}
    \vskip\baselineskip%
    \caption{{\bf Single Image 3D Reconstruction on ShapeNet}. We compare
    Neural Parts to OccNet and CvxNet with $5$ and $25$ primitives. Our model
    yields semantic and more accurate reconstructions with $5$ times less
    primitives.}
    \label{fig:shapenet_results_supp}
\end{figure}

In this section, we provide additional information regarding our experiments on
ShapeNet dataset~\cite{Chang2015ARXIV}. In particular, we use the same image
renderings and train/test splits of Choy \etal~\cite{Choy2016ECCV}.  In order
to determine whether points lie inside or outside the target mesh (\ie for
generating the occupancy pairs $\cX_o$) we need the meshes to be watertight.
For this, we follow \cite{Mescheder2019CVPR} and use the code provided by Stutz
\etal
\cite{Stutz2018CVPR}\footnote{\href{https://github.com/davidstutz/mesh-fusion}{https://github.com/davidstutz/mesh-fusion}}
which performs TSDF-fusion on random depth renderings of the object, to create
watertight meshes of the object.  In \figref{fig:shapenet_results_supp}, we
provide additional qualitative results on various ShapeNet cars and compare our
model with $5$ primitives to CvxNet with $5$ and $25$ and with the non
primitive-based OccNet.

\begin{table}
    \centering
    \begin{tabular}{l||c|ccc}
        \toprule
        & OccNet & CvxNet - 5 & CvxNet - 25 & Ours \\
        \midrule
        IoU & 0.763 & 0.650 & 0.666 &  \bf{0.697} \\
        Chamfer-$L_1$ & 0.186 & 0.218 & 0.210  & \bf{0.185} \\
        \bottomrule
    \end{tabular}
    \caption{{\bf Single Image Reconstruction on ShapeNet cars.} Quantitative evaluation of our method against OccNet
    \cite{Mescheder2019CVPR} and CvxNet \cite{Deng2020CVPR} with 5 and 25 primitives.}
    \label{tab:shapenet_results_supp_chairs}
    \vspace{-0.4em}
\end{table}

\begin{figure}[H]
    \begin{subfigure}[b]{0.20\linewidth}
		\centering
        Input
    \end{subfigure}%
    \hfill%
    \begin{subfigure}[b]{0.20\linewidth}
		\centering
        OccNet
    \end{subfigure}%
    \hfill%
    \begin{subfigure}[b]{0.20\linewidth}
		\centering
        CvxNet-5
    \end{subfigure}%
    \hfill%
    \begin{subfigure}[b]{0.20\linewidth}
		\centering
        CvxNet-25
    \end{subfigure}%
    \hfill%
    \begin{subfigure}[b]{0.20\linewidth}
        \centering
        Ours
    \end{subfigure}
    \vspace{-2.5em}
    \vskip\baselineskip%
    \begin{subfigure}[b]{0.20\linewidth}
		\centering
		\includegraphics[trim=30 30 30 30,clip,width=\textwidth]{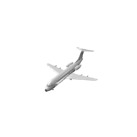}
    \end{subfigure}%
    \hfill%
    \begin{subfigure}[b]{0.20\linewidth}
		\centering
		\includegraphics[width=0.7\linewidth]{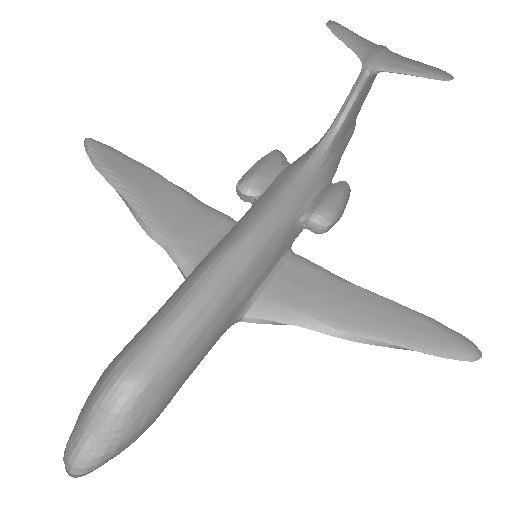}
    \end{subfigure}%
    \hfill%
    \begin{subfigure}[b]{0.20\linewidth}
		\centering
		\includegraphics[width=0.8\linewidth]{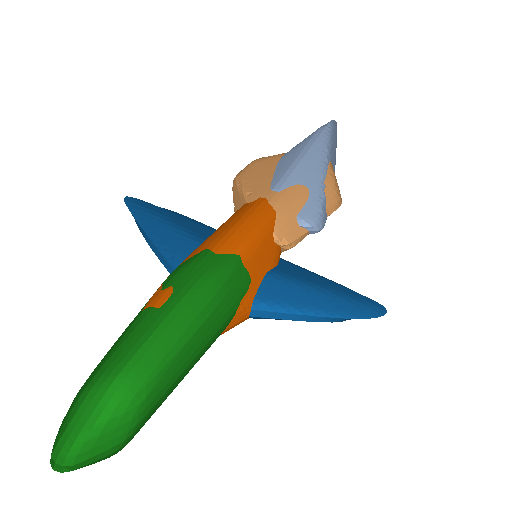}
    \end{subfigure}%
    \hfill%
    \begin{subfigure}[b]{0.20\linewidth}
		\centering
		\includegraphics[width=0.8\linewidth]{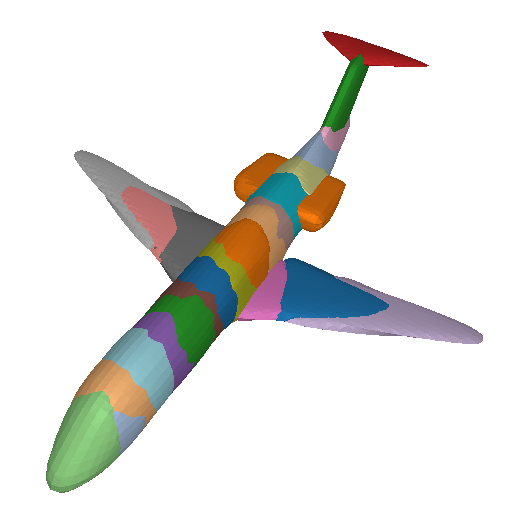}
    \end{subfigure}%
    \hfill%
    \begin{subfigure}[b]{0.20\linewidth}
		\centering
		\includegraphics[width=0.8\linewidth]{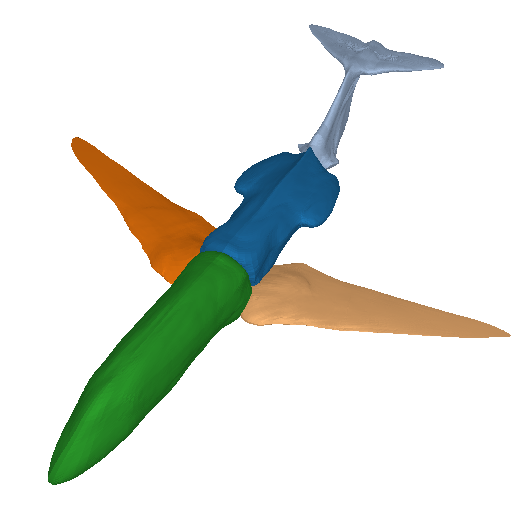}
    \end{subfigure}%
    \vspace{-2.0em}
    \vskip\baselineskip%
    \begin{subfigure}[b]{0.20\linewidth}
    	\centering
		\includegraphics[trim=20 20 20 20,clip,width=\textwidth]{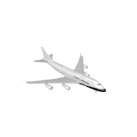}
    \end{subfigure}%
    \hfill%
    \begin{subfigure}[b]{0.20\linewidth}
    	\centering
		\includegraphics[width=0.85\linewidth]{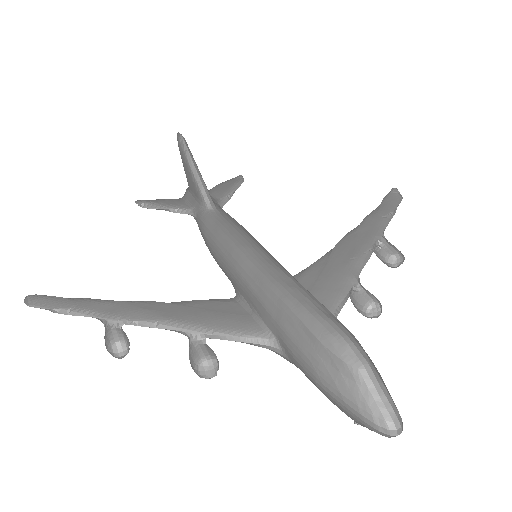}
    \end{subfigure}%
    \hfill%
    \begin{subfigure}[b]{0.20\linewidth}
    	\centering
		\includegraphics[width=0.9\linewidth]{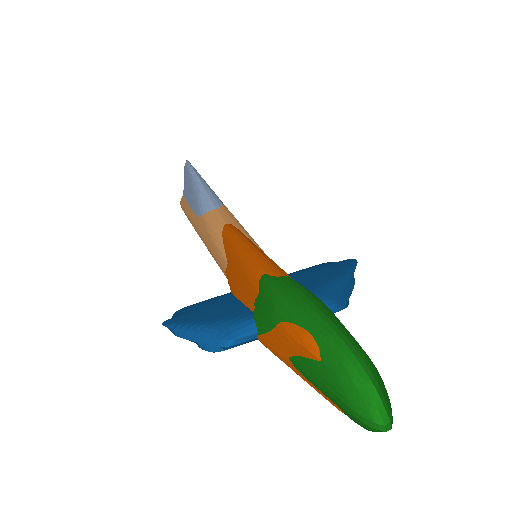}
    \end{subfigure}%
    \hfill%
    \begin{subfigure}[b]{0.20\linewidth}
    	\centering
		\includegraphics[width=0.9\linewidth]{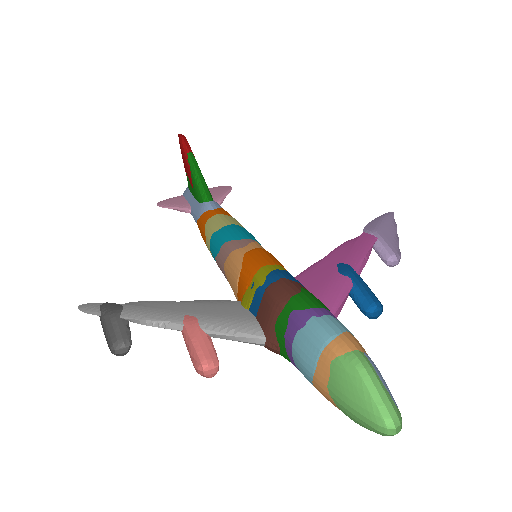}
    \end{subfigure}%
    \hfill%
    \begin{subfigure}[b]{0.20\linewidth}
    	\centering
		\includegraphics[width=0.9\linewidth]{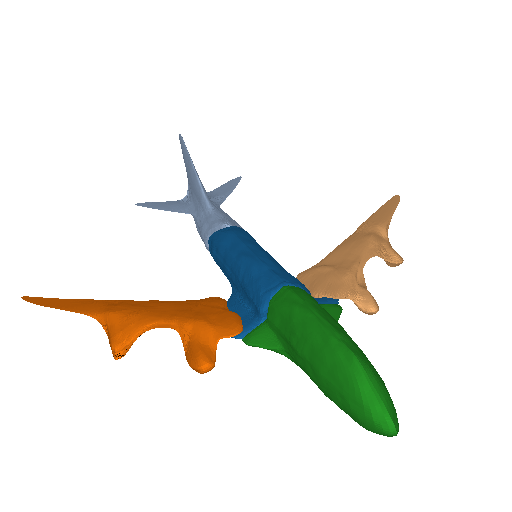}
    \end{subfigure}%
    \vspace{-2.5em}
    \vskip\baselineskip%
    \begin{subfigure}[b]{0.20\linewidth}
    	\centering
		\includegraphics[trim=20 20 20 20,clip,width=\textwidth]{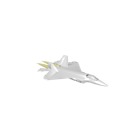}
    \end{subfigure}%
    \hfill%
    \begin{subfigure}[b]{0.20\linewidth}
    	\centering
		\includegraphics[width=0.8\linewidth]{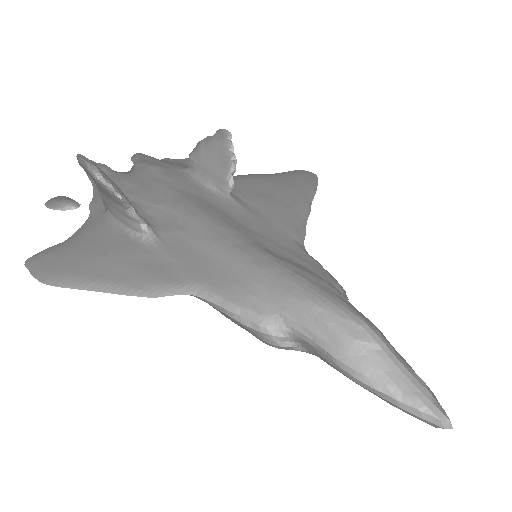}
    \end{subfigure}%
    \hfill%
    \begin{subfigure}[b]{0.20\linewidth}
    	\centering
		\includegraphics[width=0.9\linewidth]{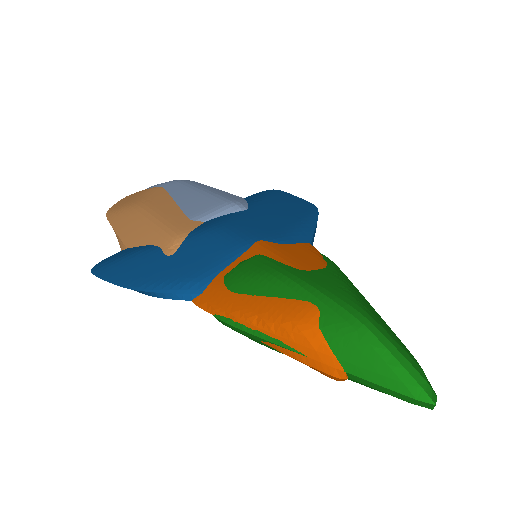}
    \end{subfigure}%
    \hfill%
    \begin{subfigure}[b]{0.20\linewidth}
    	\centering
		\includegraphics[width=0.9\linewidth]{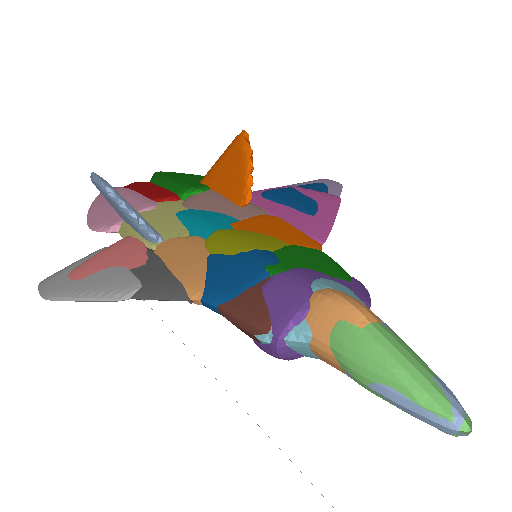}
    \end{subfigure}%
    \hfill%
    \begin{subfigure}[b]{0.20\linewidth}
    	\centering
		\includegraphics[width=0.9\linewidth]{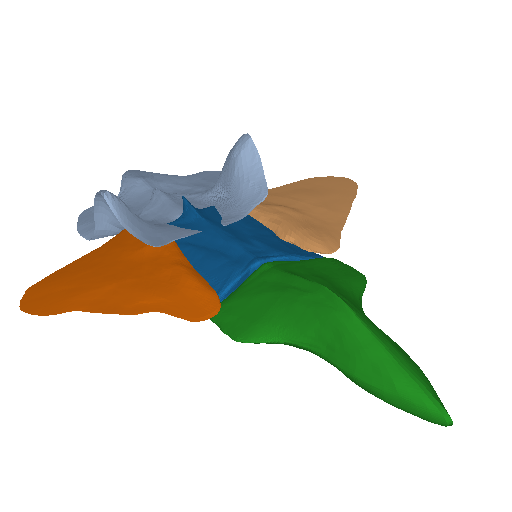}
    \end{subfigure}%
    \vspace{-3.5em}
    \vskip\baselineskip%
    \begin{subfigure}[b]{0.20\linewidth}
    	\centering
		\includegraphics[trim=20 20 20 20,clip,width=\textwidth]{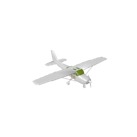}
    \end{subfigure}%
    \hfill%
    \begin{subfigure}[b]{0.20\linewidth}
    	\centering
		\includegraphics[width=0.8\linewidth]{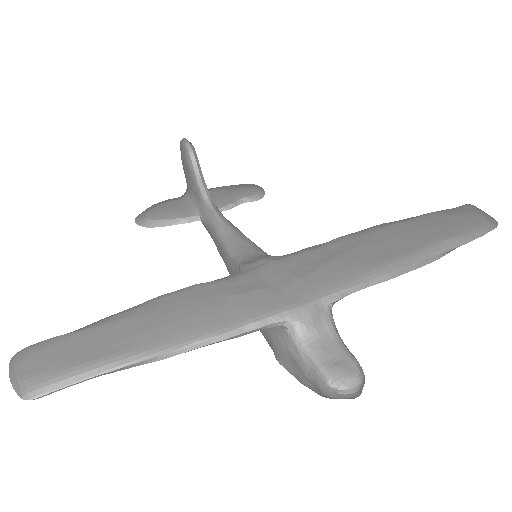}
    \end{subfigure}%
    \hfill%
    \begin{subfigure}[b]{0.20\linewidth}
    	\centering
		\includegraphics[width=0.8\linewidth]{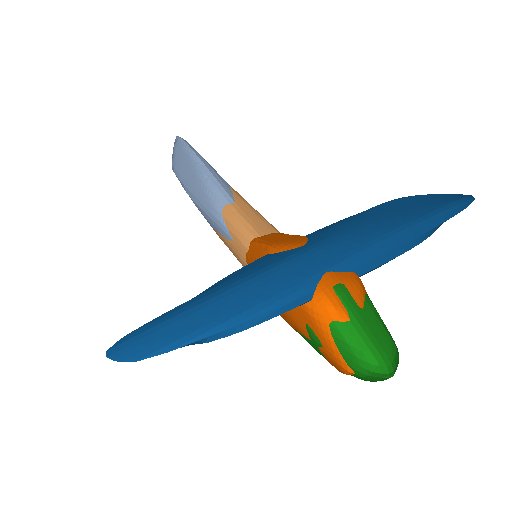}
    \end{subfigure}%
    \hfill%
    \begin{subfigure}[b]{0.20\linewidth}
    	\centering
		\includegraphics[width=0.8\linewidth]{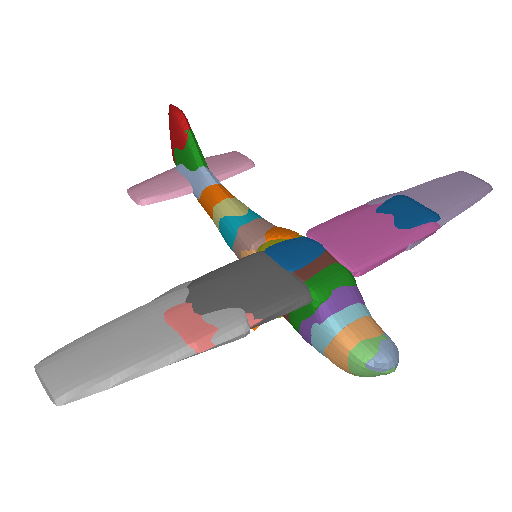}
    \end{subfigure}%
    \hfill%
    \begin{subfigure}[b]{0.20\linewidth}
    	\centering
		\includegraphics[width=0.8\linewidth]{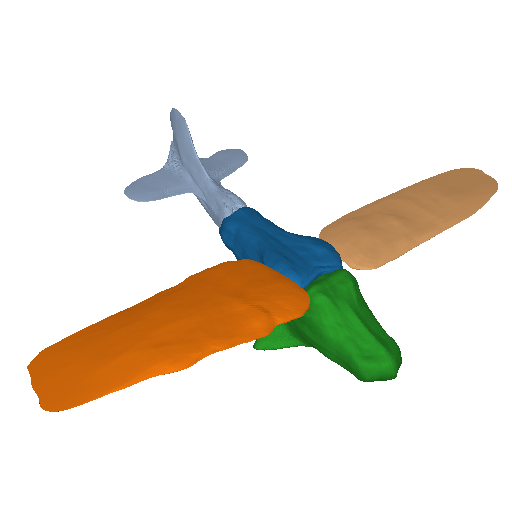}
    \end{subfigure}%
    \vspace{-3.5em}
    \vskip\baselineskip%
    \begin{subfigure}[b]{0.20\linewidth}
    	\centering
		\includegraphics[trim=20 20 20 20,clip,width=\textwidth]{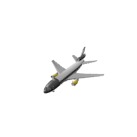}
    \end{subfigure}%
    \hfill%
    \begin{subfigure}[b]{0.20\linewidth}
    	\centering
		\includegraphics[width=0.6\linewidth]{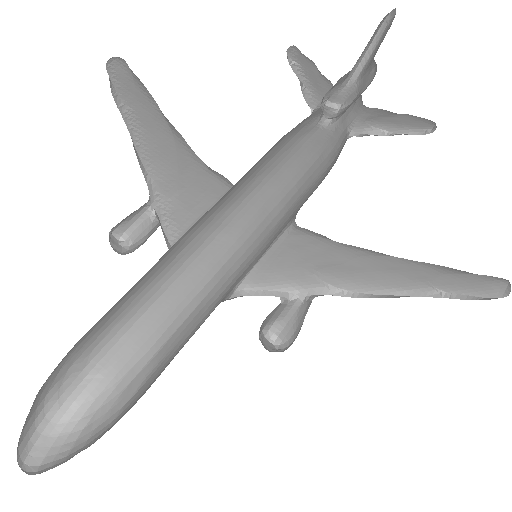}
    \end{subfigure}%
    \hfill%
    \begin{subfigure}[b]{0.20\linewidth}
    	\centering
		\includegraphics[width=0.7\linewidth]{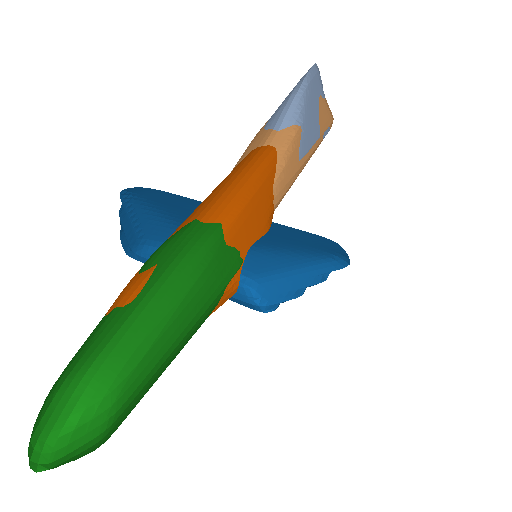}
    \end{subfigure}%
    \hfill%
    \begin{subfigure}[b]{0.20\linewidth}
    	\centering
		\includegraphics[width=0.7\linewidth]{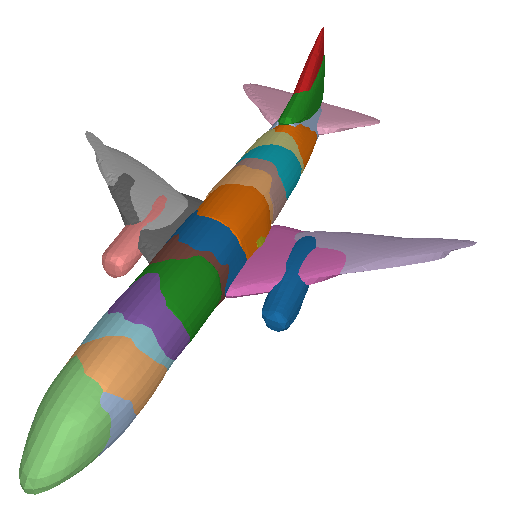}
    \end{subfigure}%
    \hfill%
    \begin{subfigure}[b]{0.20\linewidth}
    	\centering
		\includegraphics[width=0.7\linewidth]{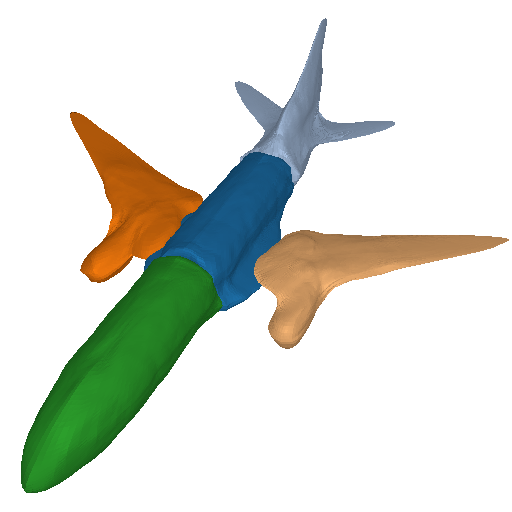}
    \end{subfigure}%
    \vspace{-2.5em}
    \vskip\baselineskip%
    \begin{subfigure}[b]{0.20\linewidth}
    	\centering
		\includegraphics[trim=20 20 20 20,clip,width=\textwidth]{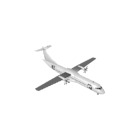}
    \end{subfigure}%
    \hfill%
    \begin{subfigure}[b]{0.20\linewidth}
    	\centering
		\includegraphics[width=0.9\linewidth]{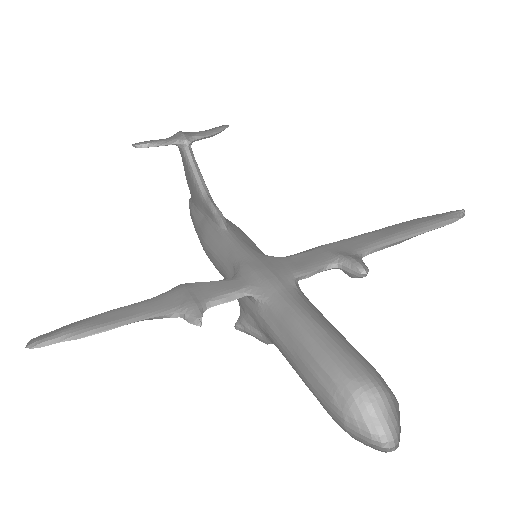}
    \end{subfigure}%
    \hfill%
    \begin{subfigure}[b]{0.20\linewidth}
    	\centering
		\includegraphics[width=0.9\linewidth]{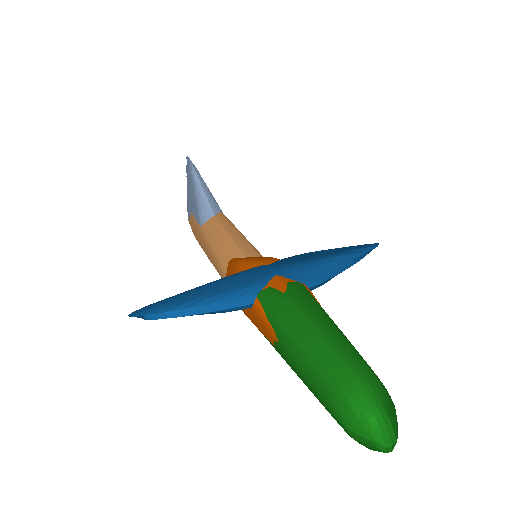}
    \end{subfigure}%
    \hfill%
    \begin{subfigure}[b]{0.20\linewidth}
    	\centering
		\includegraphics[width=0.9\linewidth]{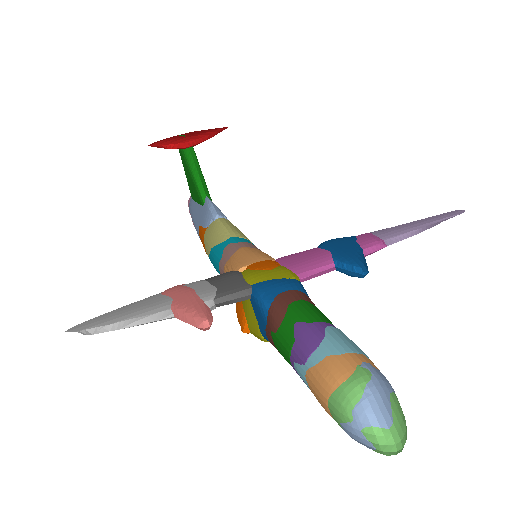}
    \end{subfigure}%
    \hfill%
    \begin{subfigure}[b]{0.20\linewidth}
    	\centering
		\includegraphics[width=0.9\linewidth]{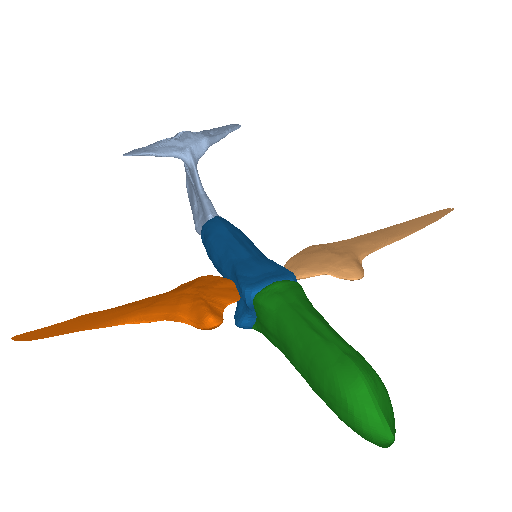}
    \end{subfigure}%
    \vspace{-1.2em}
    \vskip\baselineskip%
    \caption{{\bf Single Image 3D Reconstruction on ShapeNet}. We compare
    Neural Parts to OccNet and CvxNet with $5$ and $25$ primitives. Our model
    yields semantic and more accurate reconstructions with $5$ times less
    primitives.}
    \label{fig:shapenet_results_supp_planes}
\end{figure}

\begin{figure}[H]
    \begin{subfigure}[b]{0.20\linewidth}
		\centering
        Input
    \end{subfigure}%
    \hfill%
    \begin{subfigure}[b]{0.20\linewidth}
		\centering
        OccNet
    \end{subfigure}%
    \hfill%
    \begin{subfigure}[b]{0.20\linewidth}
		\centering
        CvxNet-5
    \end{subfigure}%
    \hfill%
    \begin{subfigure}[b]{0.20\linewidth}
		\centering
        CvxNet-25
    \end{subfigure}%
    \hfill%
    \begin{subfigure}[b]{0.20\linewidth}
        \centering
        Ours
    \end{subfigure}
    \vspace{-2.2em}
    \vskip\baselineskip%
    \begin{subfigure}[b]{0.20\linewidth}
		\centering
		\includegraphics[width=0.8\textwidth]{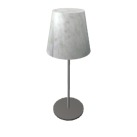}
    \end{subfigure}%
    \hfill%
    \begin{subfigure}[b]{0.20\linewidth}
		\centering
		\includegraphics[width=0.8\textwidth]{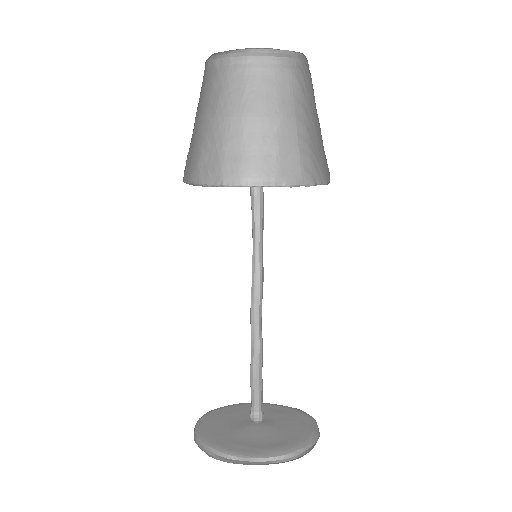}
    \end{subfigure}%
    \hfill%
    \begin{subfigure}[b]{0.20\linewidth}
		\centering
		\includegraphics[width=0.8\textwidth]{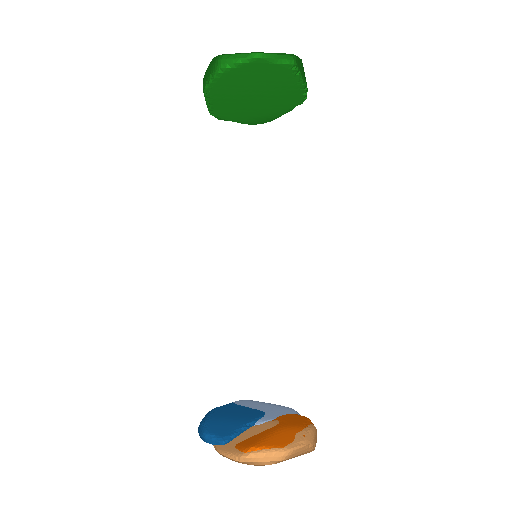}
    \end{subfigure}%
    \hfill%
    \begin{subfigure}[b]{0.20\linewidth}
		\centering
		\includegraphics[width=0.8\textwidth]{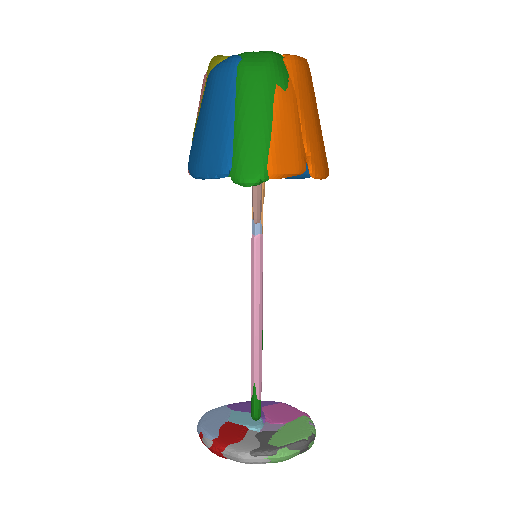}
    \end{subfigure}%
    \hfill%
    \begin{subfigure}[b]{0.20\linewidth}
		\centering
		\includegraphics[width=0.8\textwidth]{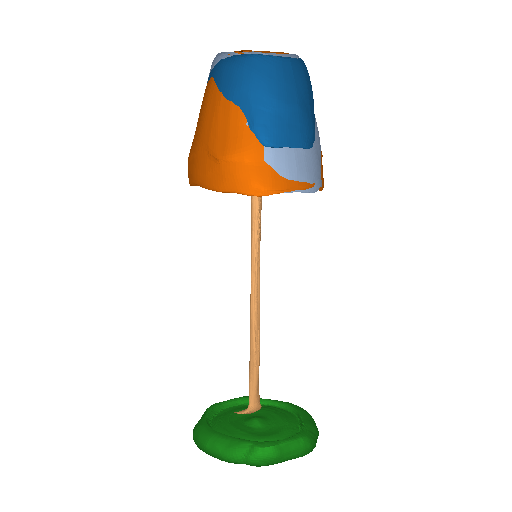}
    \end{subfigure}%
    \vspace{-1.5em}
    \vskip\baselineskip%
    \begin{subfigure}[b]{0.20\linewidth}
    	\centering
		\includegraphics[width=\textwidth]{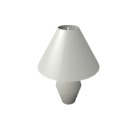}
    \end{subfigure}%
    \hfill%
    \begin{subfigure}[b]{0.20\linewidth}
    	\centering
		\includegraphics[width=0.7\textwidth]{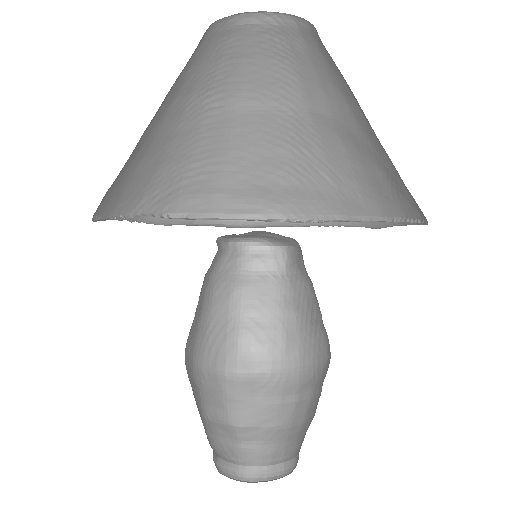}
    \end{subfigure}%
    \hfill%
    \begin{subfigure}[b]{0.20\linewidth}
    	\centering
		\includegraphics[width=0.7\textwidth]{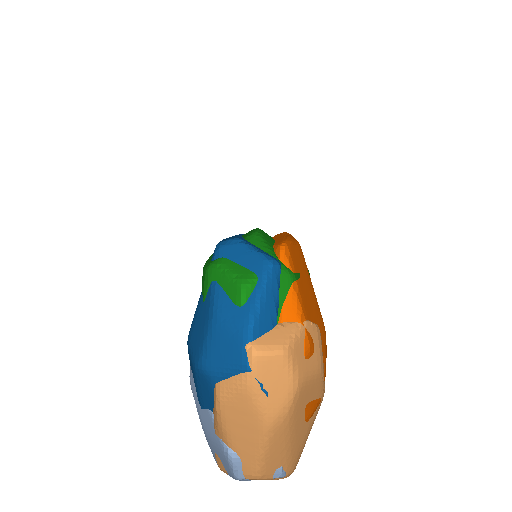}
    \end{subfigure}%
    \hfill%
    \begin{subfigure}[b]{0.20\linewidth}
    	\centering
		\includegraphics[width=0.7\textwidth]{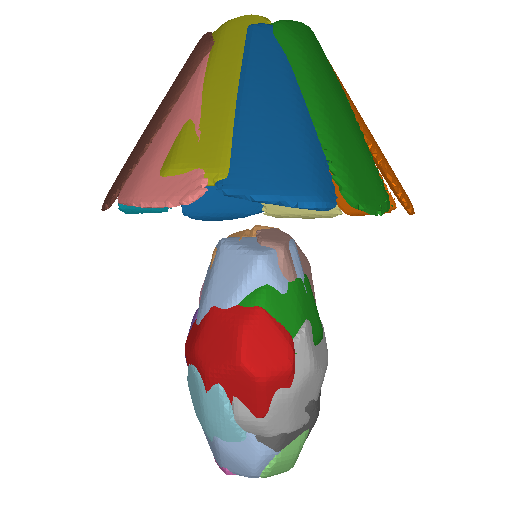}
    \end{subfigure}%
    \hfill%
    \begin{subfigure}[b]{0.20\linewidth}
    	\centering
		\includegraphics[width=0.7\textwidth]{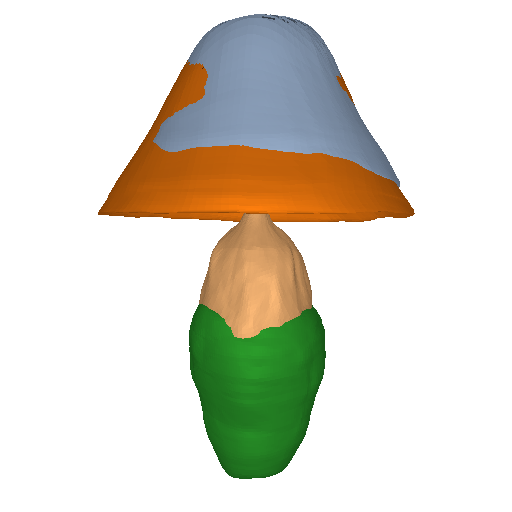}
    \end{subfigure}%
    \vspace{-1.5em}
    \vskip\baselineskip%
    \begin{subfigure}[b]{0.20\linewidth}
    	\centering
		\includegraphics[width=0.8\textwidth]{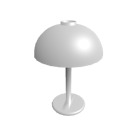}
    \end{subfigure}%
    \hfill%
    \begin{subfigure}[b]{0.20\linewidth}
    	\centering
		\includegraphics[width=0.7\textwidth]{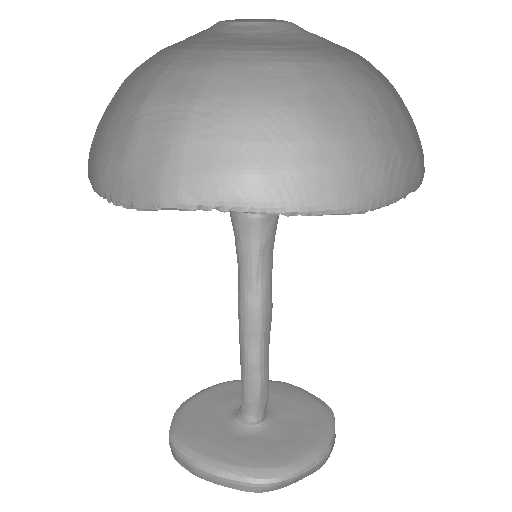}
    \end{subfigure}%
    \hfill%
    \begin{subfigure}[b]{0.20\linewidth}
    	\centering
		\includegraphics[width=0.7\textwidth]{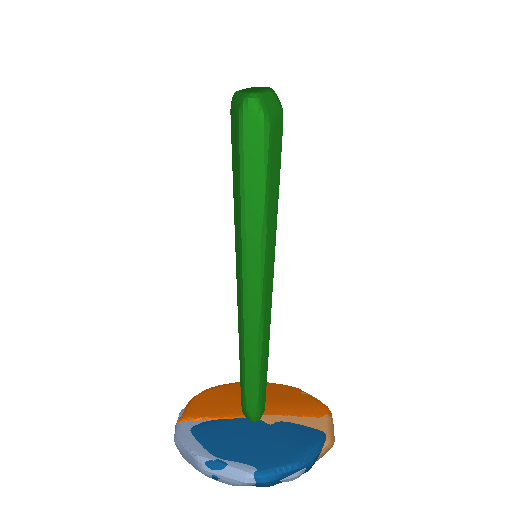}
    \end{subfigure}%
    \hfill%
    \begin{subfigure}[b]{0.20\linewidth}
    	\centering
		\includegraphics[width=0.7\textwidth]{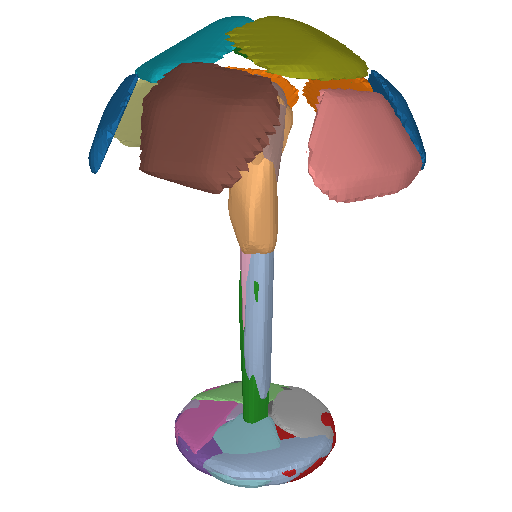}
    \end{subfigure}%
    \hfill%
    \begin{subfigure}[b]{0.20\linewidth}
    	\centering
		\includegraphics[width=0.7\textwidth]{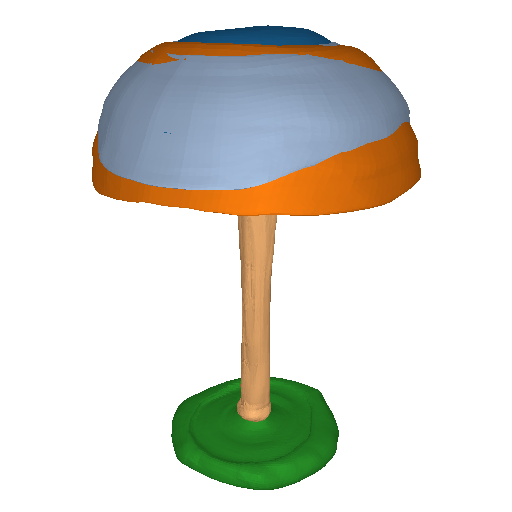}
    \end{subfigure}%
    \vspace{-1.5em}
    \vskip\baselineskip%
    \begin{subfigure}[b]{0.20\linewidth}
    	\centering
		\includegraphics[width=0.8\textwidth]{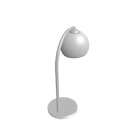}
    \end{subfigure}%
    \hfill%
    \begin{subfigure}[b]{0.20\linewidth}
    	\centering
		\includegraphics[width=0.7\textwidth]{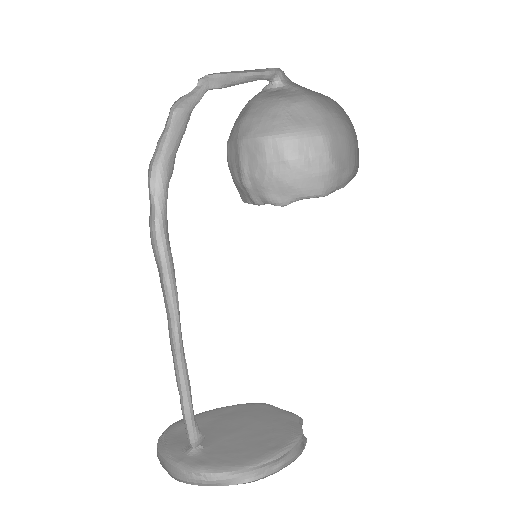}
    \end{subfigure}%
    \hfill%
    \begin{subfigure}[b]{0.20\linewidth}
    	\centering
		\includegraphics[width=0.7\textwidth]{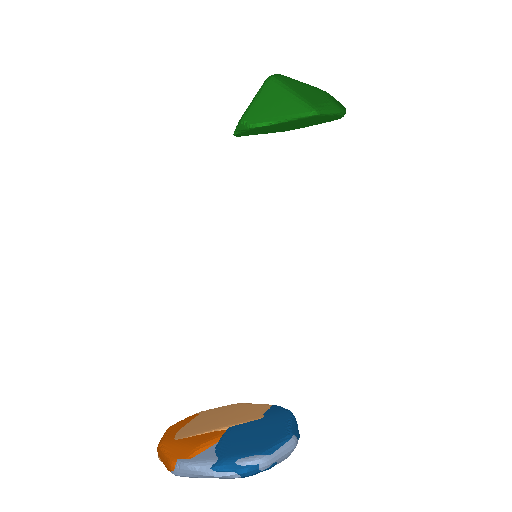}
    \end{subfigure}%
    \hfill%
    \begin{subfigure}[b]{0.20\linewidth}
    	\centering
		\includegraphics[width=0.7\textwidth]{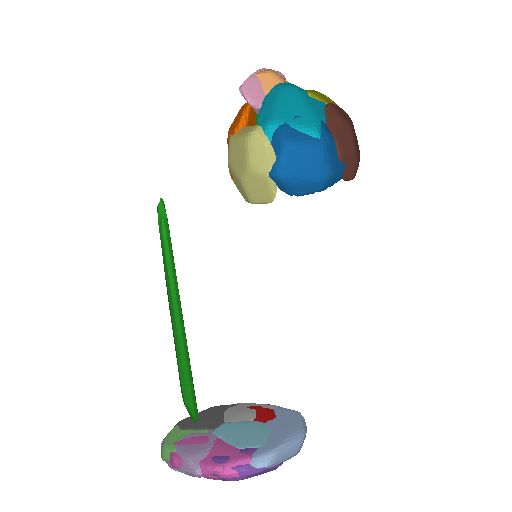}
    \end{subfigure}%
    \hfill%
    \begin{subfigure}[b]{0.20\linewidth}
    	\centering
		\includegraphics[width=0.7\textwidth]{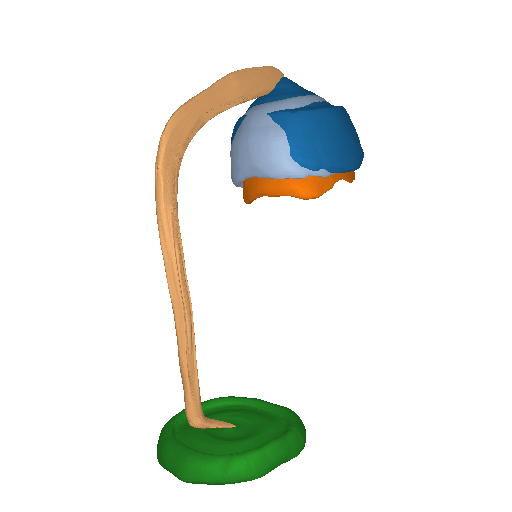}
    \end{subfigure}%

    \vspace{-1.2em}
    \vskip\baselineskip%
    \caption{{\bf Single Image 3D Reconstruction on ShapeNet}. We compare
    Neural Parts to OccNet and CvxNet with $5$ and $25$ primitives. Our model
    yields semantic and more accurate reconstructions with $5$ times less
    primitives.}
    \label{fig:shapenet_results_supp_lamps}
\end{figure}

\begin{figure}[H]
    \begin{subfigure}[b]{0.20\linewidth}
		\centering
        Input
    \end{subfigure}%
    \hfill%
    \begin{subfigure}[b]{0.20\linewidth}
		\centering
        OccNet
    \end{subfigure}%
    \hfill%
    \begin{subfigure}[b]{0.20\linewidth}
		\centering
        CvxNet-5
    \end{subfigure}%
    \hfill%
    \begin{subfigure}[b]{0.20\linewidth}
		\centering
        CvxNet-25
    \end{subfigure}%
    \hfill%
    \begin{subfigure}[b]{0.20\linewidth}
        \centering
        Ours
    \end{subfigure}
    \vspace{-2.2em}
    \vskip\baselineskip%
    \begin{subfigure}[b]{0.20\linewidth}
		\centering
		\includegraphics[width=0.8\textwidth]{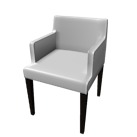}
    \end{subfigure}%
    \hfill%
    \begin{subfigure}[b]{0.20\linewidth}
		\centering
		\includegraphics[width=0.8\textwidth]{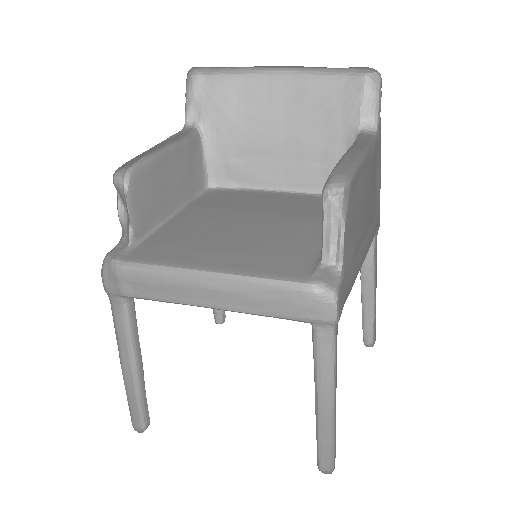}
    \end{subfigure}%
    \hfill%
    \begin{subfigure}[b]{0.20\linewidth}
		\centering
		\includegraphics[width=0.8\textwidth]{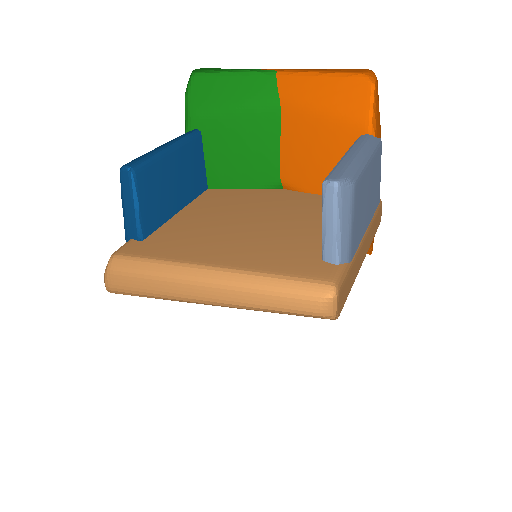}
    \end{subfigure}%
    \hfill%
    \begin{subfigure}[b]{0.20\linewidth}
		\centering
		\includegraphics[width=0.8\textwidth]{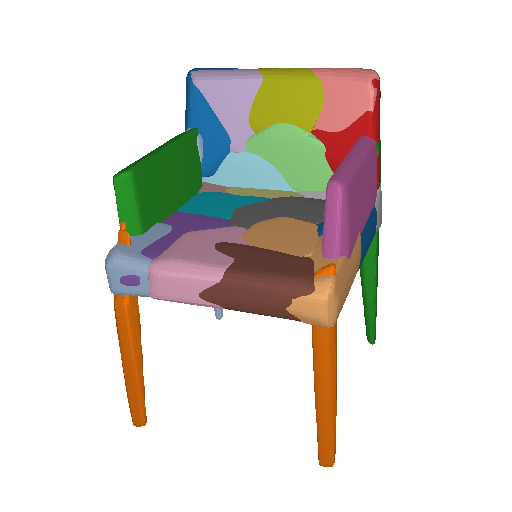}
    \end{subfigure}%
    \hfill%
    \begin{subfigure}[b]{0.20\linewidth}
		\centering
		\includegraphics[width=0.8\textwidth]{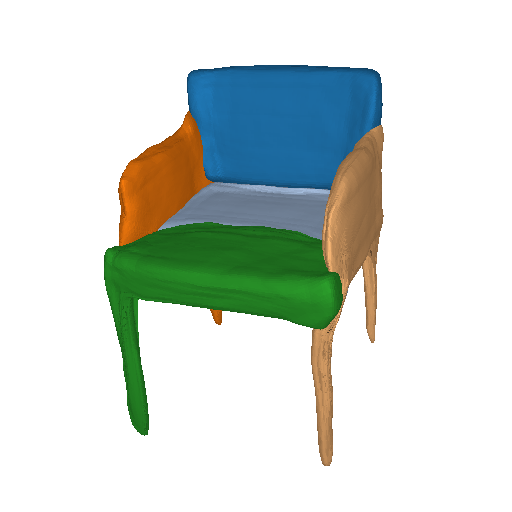}
    \end{subfigure}%
    \vspace{-1.5em}
    \vskip\baselineskip%
    \begin{subfigure}[b]{0.20\linewidth}
		\centering
		\includegraphics[width=0.8\textwidth]{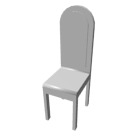}
    \end{subfigure}%
    \hfill%
    \begin{subfigure}[b]{0.20\linewidth}
		\centering
		\includegraphics[width=0.8\textwidth]{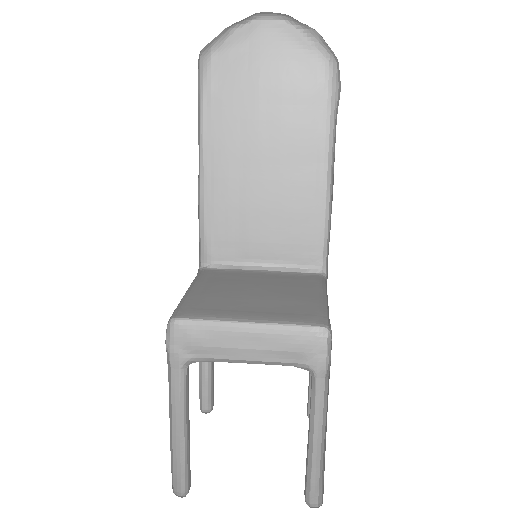}
    \end{subfigure}%
    \hfill%
    \begin{subfigure}[b]{0.20\linewidth}
		\centering
		\includegraphics[width=0.8\textwidth]{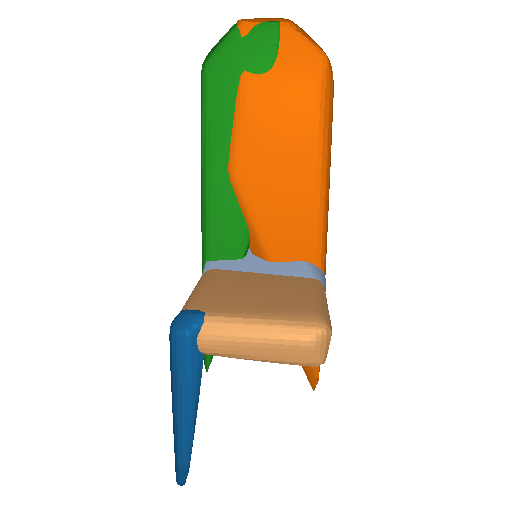}
    \end{subfigure}%
    \hfill%
    \begin{subfigure}[b]{0.20\linewidth}
		\centering
		\includegraphics[width=0.8\textwidth]{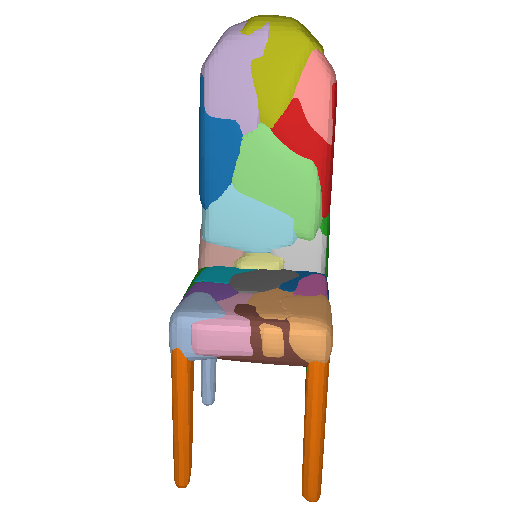}
    \end{subfigure}%
    \hfill%
    \begin{subfigure}[b]{0.20\linewidth}
		\centering
		\includegraphics[width=0.8\textwidth]{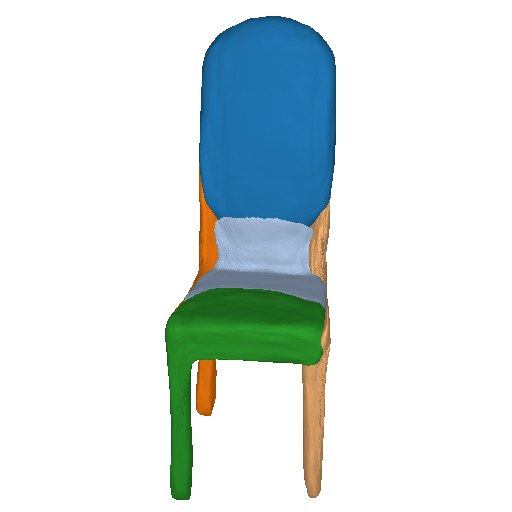}
    \end{subfigure}%
    \vspace{-1.5em}
    \vskip\baselineskip%
    \begin{subfigure}[b]{0.20\linewidth}
		\centering
		\includegraphics[width=0.8\textwidth]{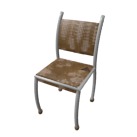}
    \end{subfigure}%
    \hfill%
    \begin{subfigure}[b]{0.20\linewidth}
		\centering
		\includegraphics[width=0.8\textwidth]{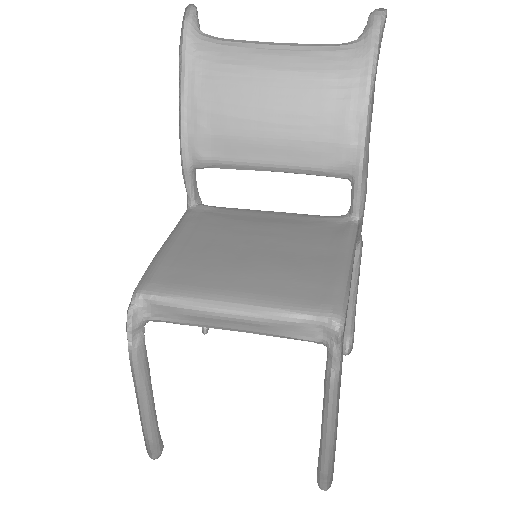}
    \end{subfigure}%
    \hfill%
    \begin{subfigure}[b]{0.20\linewidth}
		\centering
		\includegraphics[width=0.8\textwidth]{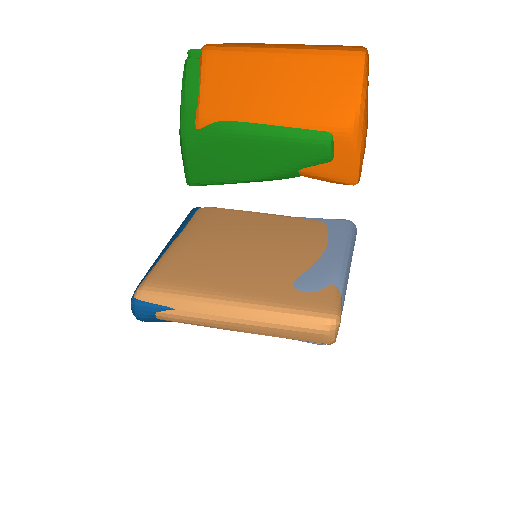}
    \end{subfigure}%
    \hfill%
    \begin{subfigure}[b]{0.20\linewidth}
		\centering
		\includegraphics[width=0.8\textwidth]{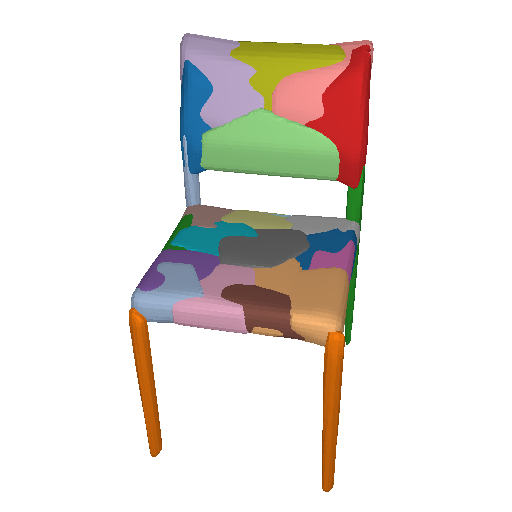}
    \end{subfigure}%
    \hfill%
    \begin{subfigure}[b]{0.20\linewidth}
		\centering
		\includegraphics[width=0.8\textwidth]{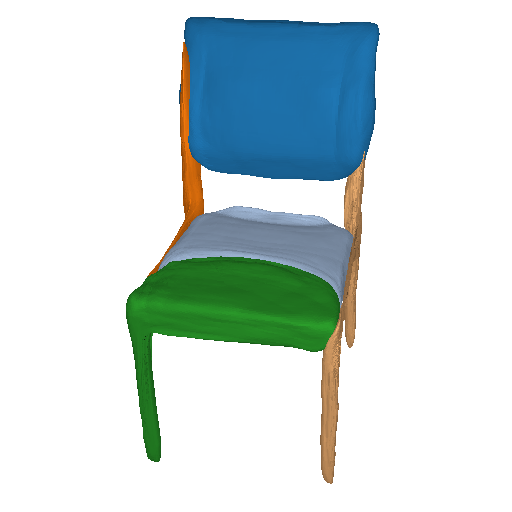}
    \end{subfigure}%
    \vspace{-1.5em}
    \vskip\baselineskip%
    \begin{subfigure}[b]{0.20\linewidth}
		\centering
		\includegraphics[width=0.8\textwidth]{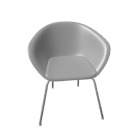}
    \end{subfigure}%
    \hfill%
    \begin{subfigure}[b]{0.20\linewidth}
		\centering
		\includegraphics[width=0.8\textwidth]{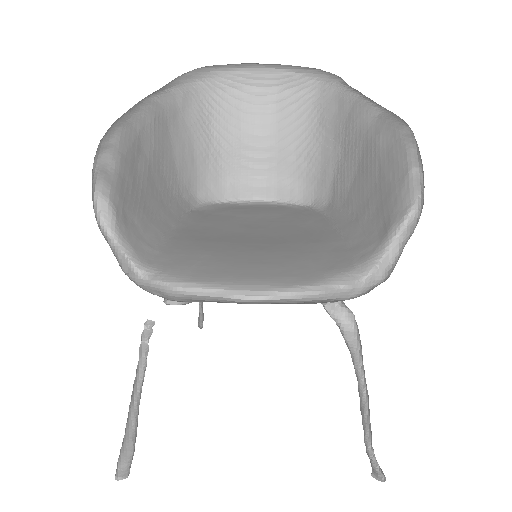}
    \end{subfigure}%
    \hfill%
    \begin{subfigure}[b]{0.20\linewidth}
		\centering
		\includegraphics[width=0.8\textwidth]{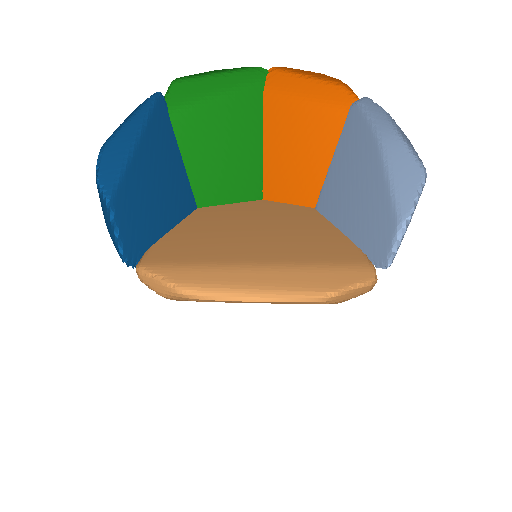}
    \end{subfigure}%
    \hfill%
    \begin{subfigure}[b]{0.20\linewidth}
		\centering
		\includegraphics[width=0.8\textwidth]{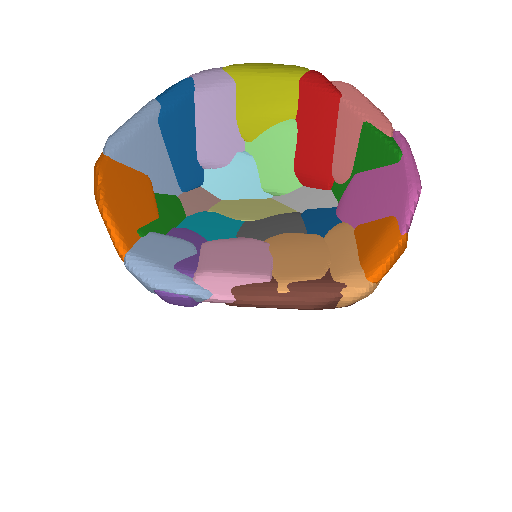}
    \end{subfigure}%
    \hfill%
    \begin{subfigure}[b]{0.20\linewidth}
		\centering
		\includegraphics[width=0.8\textwidth]{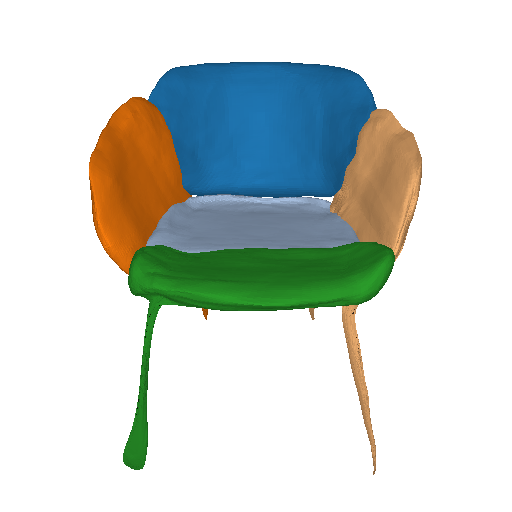}
    \end{subfigure}%
    \vspace{-1.5em}
    \vskip\baselineskip%
    \begin{subfigure}[b]{0.20\linewidth}
		\centering
		\includegraphics[width=0.8\textwidth]{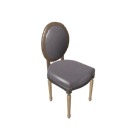}
    \end{subfigure}%
    \hfill%
    \begin{subfigure}[b]{0.20\linewidth}
		\centering
		\includegraphics[width=0.8\textwidth]{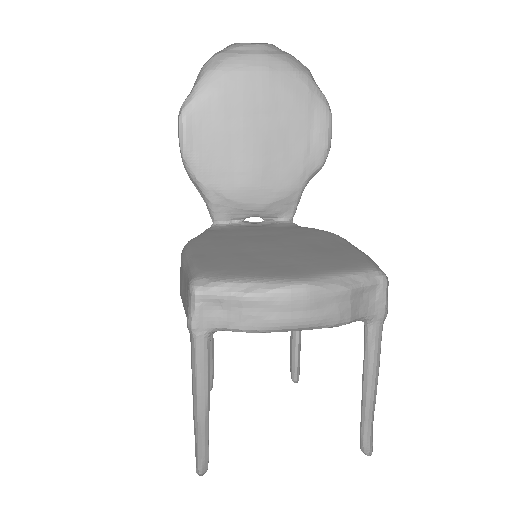}
    \end{subfigure}%
    \hfill%
    \begin{subfigure}[b]{0.20\linewidth}
		\centering
		\includegraphics[width=0.8\textwidth]{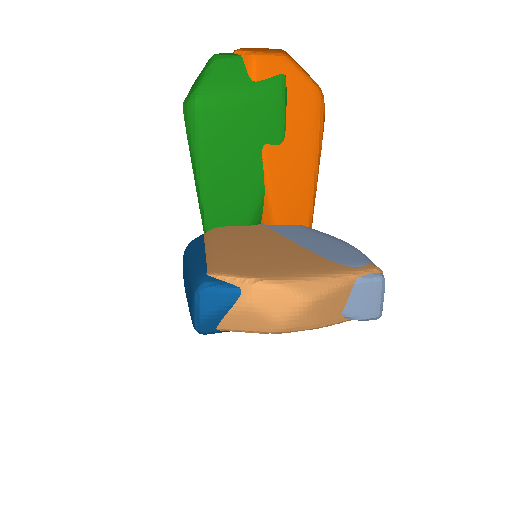}
    \end{subfigure}%
    \hfill%
    \begin{subfigure}[b]{0.20\linewidth}
		\centering
		\includegraphics[width=0.8\textwidth]{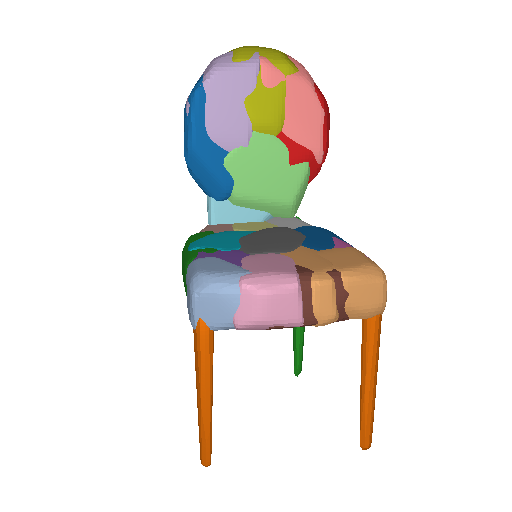}
    \end{subfigure}%
    \hfill%
    \begin{subfigure}[b]{0.20\linewidth}
		\centering
		\includegraphics[width=0.8\textwidth]{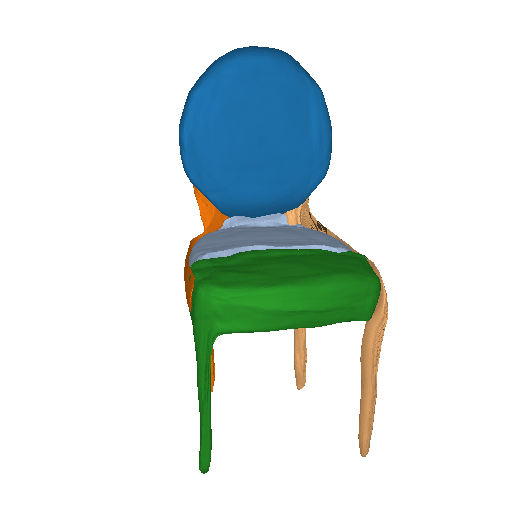}
    \end{subfigure}%
    \vspace{-1.2em}
    \vskip\baselineskip%
    \caption{{\bf Single Image 3D Reconstruction on ShapeNet}. We compare
    Neural Parts to OccNet and CvxNet with $5$ and $25$ primitives. Our model
    yields semantic and more accurate reconstructions with $5$ times less
    primitives.}
    \label{fig:shapenet_results_supp_chairs}
\end{figure}

Subsequently, we train Neural Parts with $5$ primitives and compare with CvxNet with $5$ and
$25$ primitives on ShapeNet planes
(\figref{fig:shapenet_results_supp_planes} + \tabref{tab:shapenet_results_supp_planes}), ShapeNet lamps
(\figref{fig:shapenet_results_supp_lamps} + \tabref{tab:shapenet_results_supp_lamps}) and ShapeNet chairs
(\figref{fig:shapenet_results_supp_chairs} + \tabref{tab:shapenet_results_supp_chairs}).
In \figref{fig:shapenet_results_supp_planes}, we provide a qualitative
evaluation of our model with CvxNet and OccNet on various planes. We observe
that Neural Parts yield more geometrically accurate and semantically meaningful
shape abstractions than CvxNet with multiple primitives, \ie the same primitive
is consistently used for representing the tail and the wings of the airplanes.
This is also validated quantitatively in
\tabref{tab:shapenet_results_supp_planes}, where we see that our model
outperforms CvxNet in terms of IoU and Chamfer-$L_1$ both with $5$ and $25$
primitives. We observe that CvxNet with $5$ primitives cannot represent fine
details and as a result parts of the target object are not not captured, \eg
the turbines of airplanes.
\begin{table}
    \centering
    \begin{tabular}{l||c|ccc}
        \toprule
        & OccNet & CvxNet - 5 & CvxNet - 25 & Ours \\
        \midrule
        IoU & 0.451 & 0.425 & 0.448 &  \bf{0.454} \\
        Chamfer-$L_1$ & 0.218 & 0.267 & 0.245  & \bf{0.220} \\
        \bottomrule
    \end{tabular}
    \caption{{\bf Single Image Reconstruction on ShapeNet airplanes.} Quantitative evaluation of our method against OccNet
    \cite{Mescheder2019CVPR} and CvxNet \cite{Deng2020CVPR} with 5 and 25 primitives.}
    \label{tab:shapenet_results_supp_planes}
    \vspace{-0.4em}
\end{table}

This becomes more evident for the case of lamps, where $5$ primitives do not
have enough representation power for modelling complex shapes such as the lamp
shade or the lamp body (see \figref{fig:shapenet_results_supp_lamps} third
column), thus entire object parts are missing. Similarly, also for the case of
chairs, CvxNet with $5$ primitives fail to accurately capture the object's
geometry (see \figref{fig:shapenet_results_supp_chairs} third column). For
example, $5$ primitives are not enough for representing the legs of the chair.
Instead our model with the same number of parts consistently captures the 3D
geometry. On the contrary, CvxNet with $25$ primitives yield more geometrically
accurate reconstructions, however, the reconstructed primitives are not as 
semantically meaningful.
\begin{table}
    \centering
    \begin{tabular}{l||c|ccc}
        \toprule
        & OccNet & CvxNet - 5 & CvxNet - 25 & Ours \\
        \midrule
        IoU & 0.339 & 0.246 & 0.278 &  \bf{0.318} \\
        Chamfer-$L_1$ & 0.514 & 0.613 & 0.537  & \bf{0.347} \\
        \bottomrule
    \end{tabular}
    \caption{{\bf Single Image Reconstruction on ShapeNet lamps.} Quantitative evaluation of our method against OccNet
    \cite{Mescheder2019CVPR} and CvxNet \cite{Deng2020CVPR} with 5 and 25 primitives.}
    \label{tab:shapenet_results_supp_lamps}
    \vspace{-0.4em}
\end{table}

\begin{table}
    \centering
    \begin{tabular}{l||c|ccc}
        \toprule
        & OccNet & CvxNet - 5 & CvxNet - 25 & Ours \\
        \midrule
        IoU & 0.432 & 0.364 & 0.392 &  \bf{0.412} \\
        Chamfer-$L_1$ & 0.312 & 0.587 & 0.557  & \bf{0.337} \\
        \bottomrule
    \end{tabular}
    \caption{{\bf Single Image Reconstruction on ShapeNet chairs.} Quantitative evaluation of our method against OccNet
    \cite{Mescheder2019CVPR} and CvxNet \cite{Deng2020CVPR} with 5 and 25 primitives.}
    \label{tab:shapenet_results_supp_chairs}
    \vspace{-0.4em}
\end{table}

\section{Semantically Consistent Abstractions}
\label{sec:consistency}

In this section, we provide additional information regarding our Semantic
Consistency experiment from Section 4.3 in our main submission. Furthermore,
since D-FAUST contains 4D scans of humans in motion, we also evaluate the
temporal consistency of the predicted primitives on various actions.

\begin{figure}[H]
        \centering
        \includegraphics[width=1.0\textwidth]{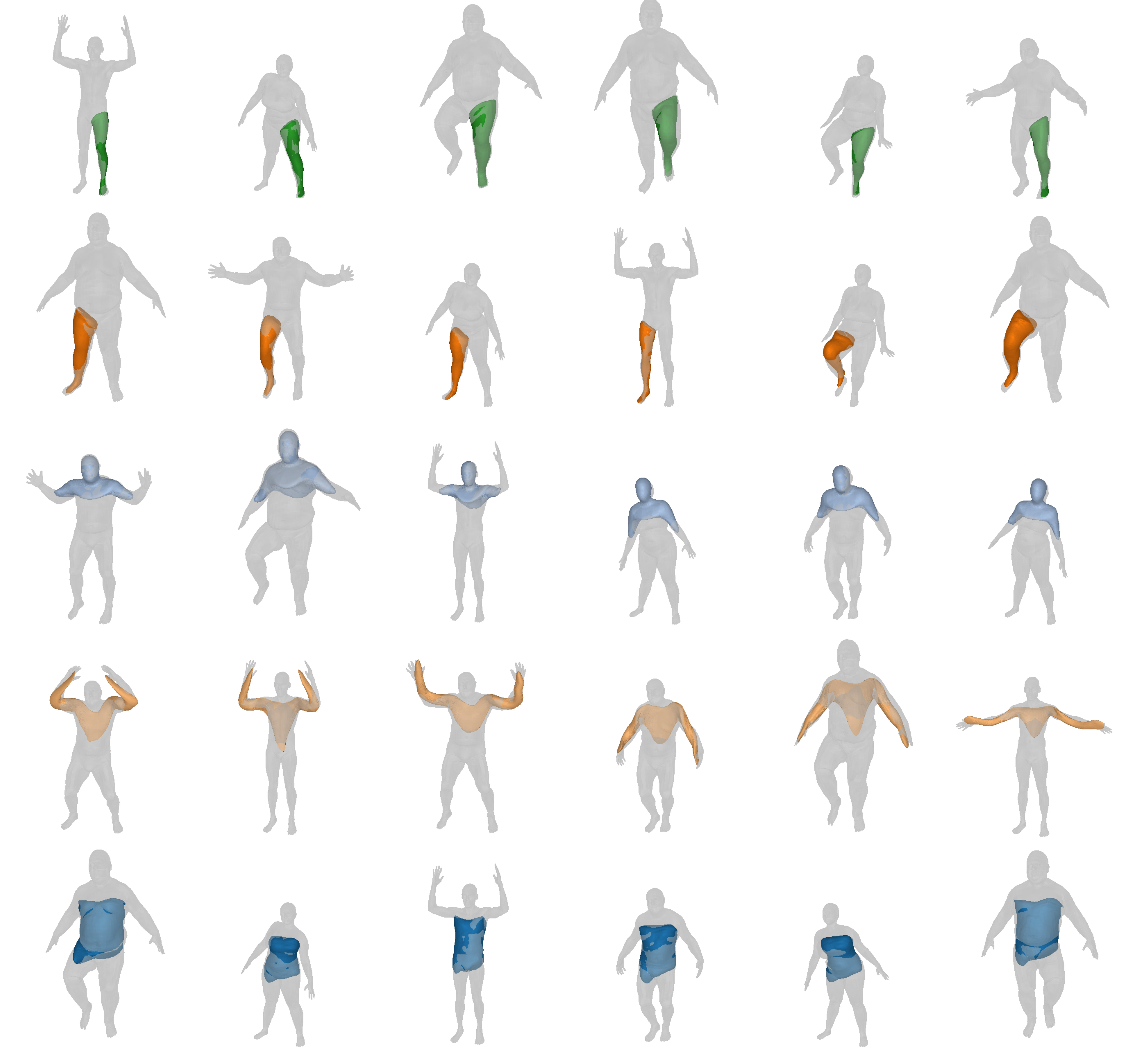}
    \caption{{\bf Representation Consistency.} Our predicted primitives
are consistently used for representing the same human part for different humans.}
    \label{fig:semantic_consistency_supp}
\end{figure}

\subsection{Temporal Consistency}

\begin{figure}[H]
    \begin{subfigure}[t]{\textwidth}
        \centering
        \includegraphics[trim=100 0 100 0,clip,width=0.1\textwidth]{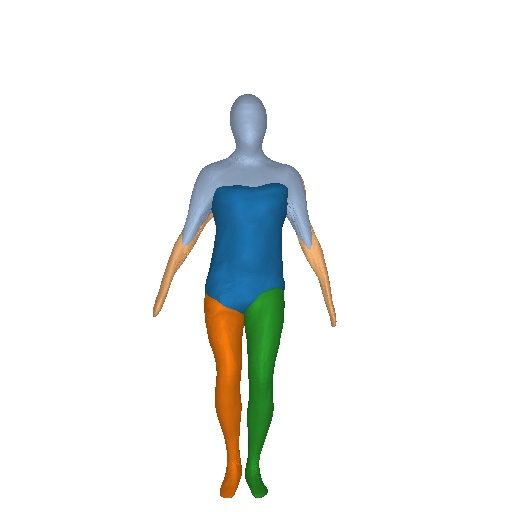}%
        \includegraphics[trim=100 0 100 0,clip,width=0.1\textwidth]{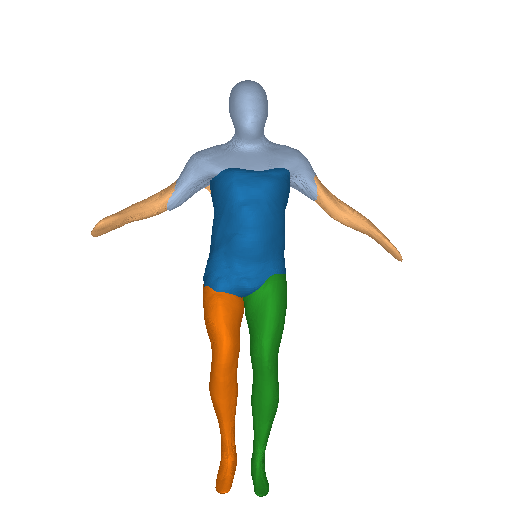}%
        \includegraphics[trim=100 0 100 0,clip,width=0.1\textwidth]{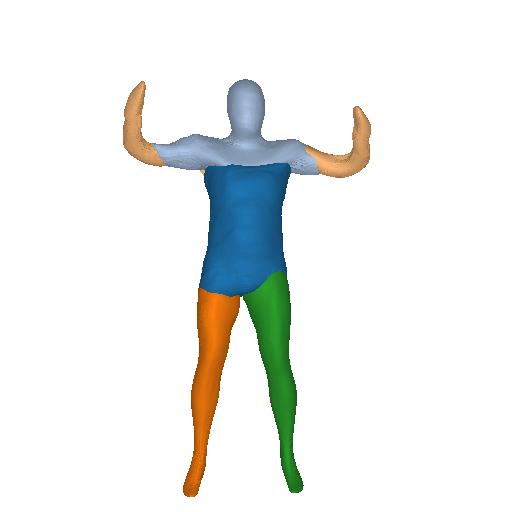}%
        \includegraphics[trim=100 0 100 0,clip,width=0.1\textwidth]{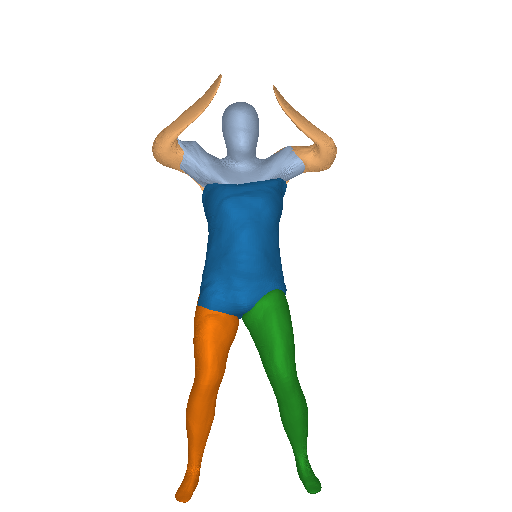}%
        \includegraphics[trim=100 0 100 0,clip,width=0.1\textwidth]{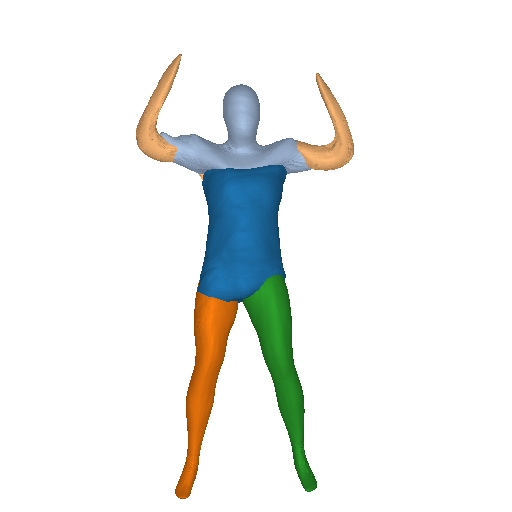}%
        \includegraphics[trim=100 0 100 0,clip,width=0.1\textwidth]{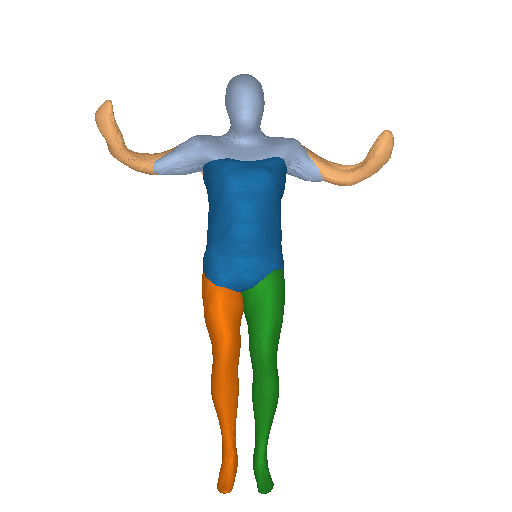}%
        \includegraphics[trim=100 0 100 0,clip,width=0.1\textwidth]{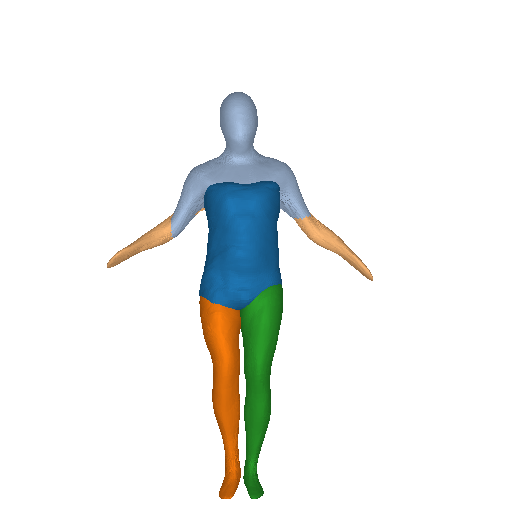}%
        \includegraphics[trim=100 0 100 0,clip,width=0.1\textwidth]{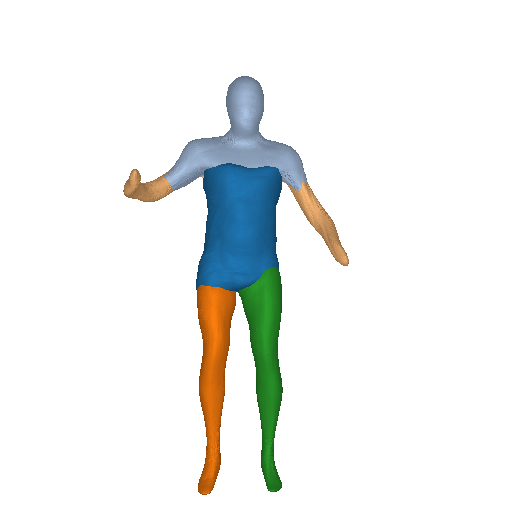}%
        \includegraphics[trim=100 0 100 0,clip,width=0.1\textwidth]{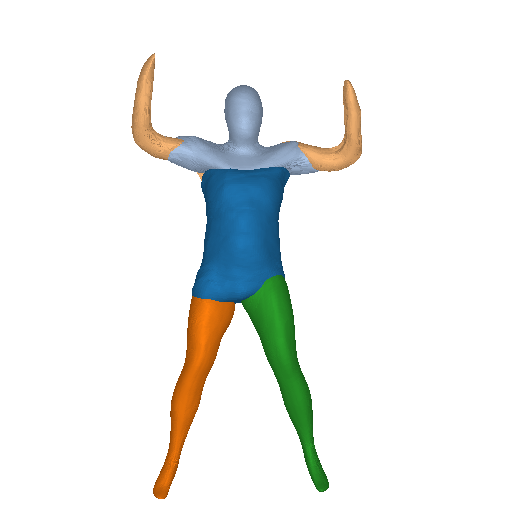}%
        \includegraphics[trim=100 0 100 0,clip,width=0.1\textwidth]{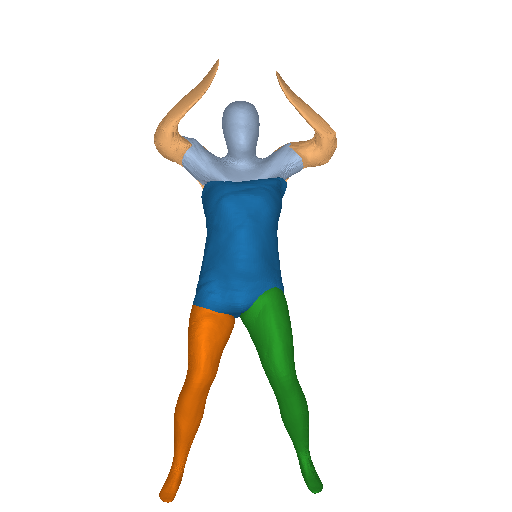}
    \end{subfigure}
    \begin{subfigure}[t]{\textwidth}
        \centering
        \includegraphics[trim=100 0 100 0,clip,width=0.1\textwidth]{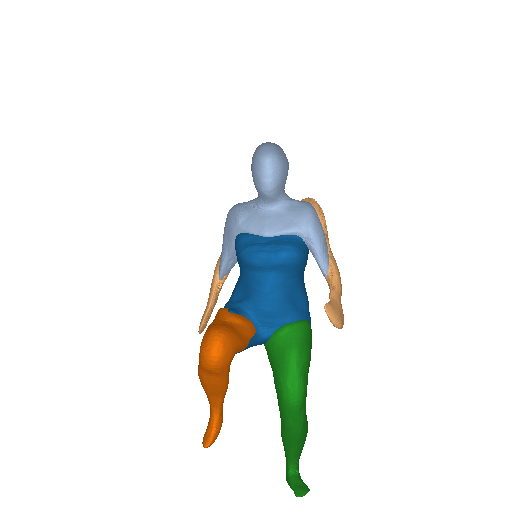}%
        \includegraphics[trim=100 0 100 0,clip,width=0.1\textwidth]{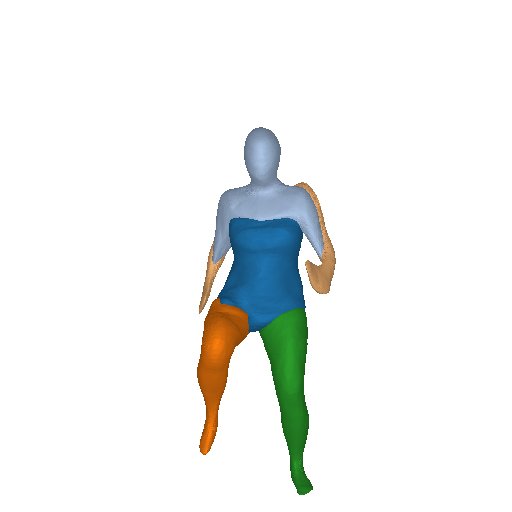}%
        \includegraphics[trim=100 0 100 0,clip,width=0.1\textwidth]{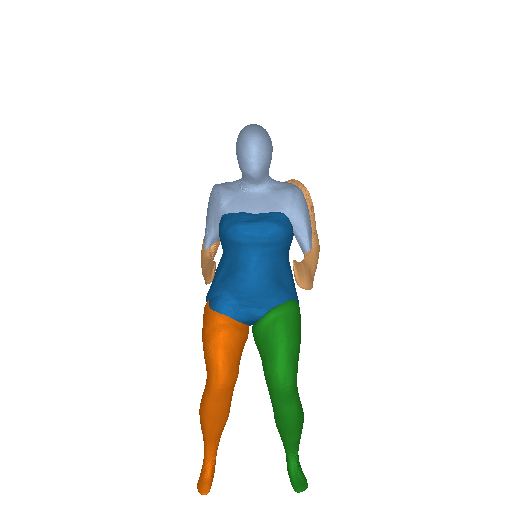}%
        \includegraphics[trim=100 0 100 0,clip,width=0.1\textwidth]{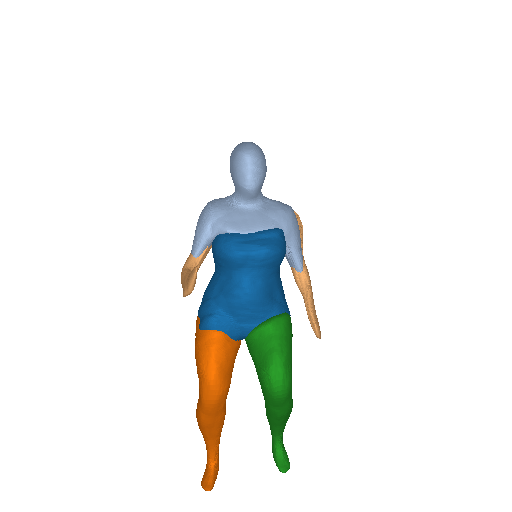}%
        \includegraphics[trim=100 0 100 0,clip,width=0.1\textwidth]{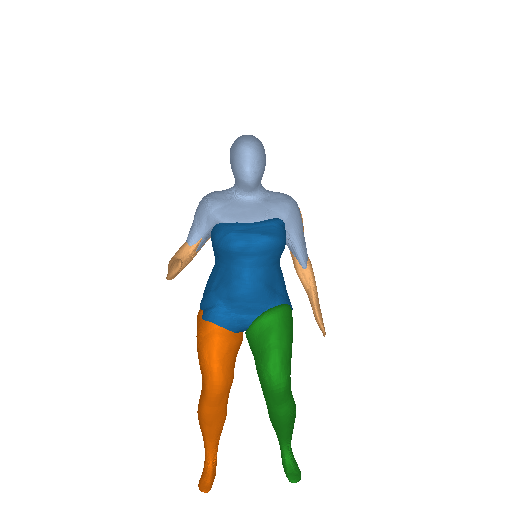}%
        \includegraphics[trim=100 0 100 0,clip,width=0.1\textwidth]{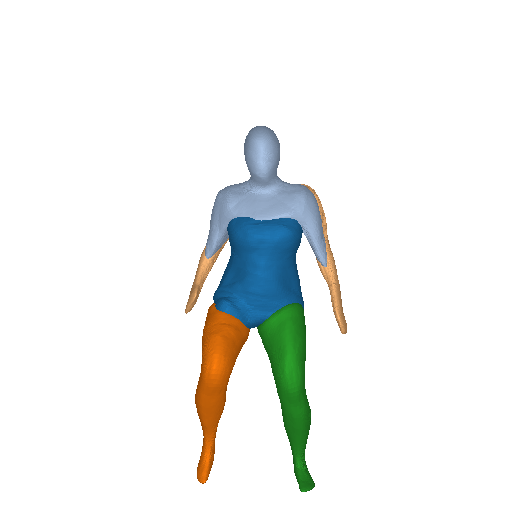}%
        \includegraphics[trim=100 0 100 0,clip,width=0.1\textwidth]{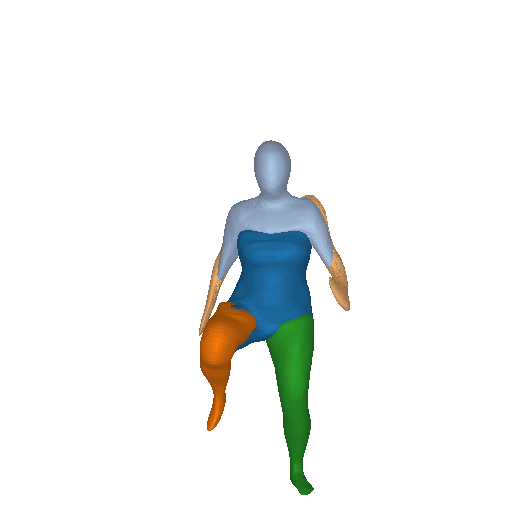}%
        \includegraphics[trim=100 0 100 0,clip,width=0.1\textwidth]{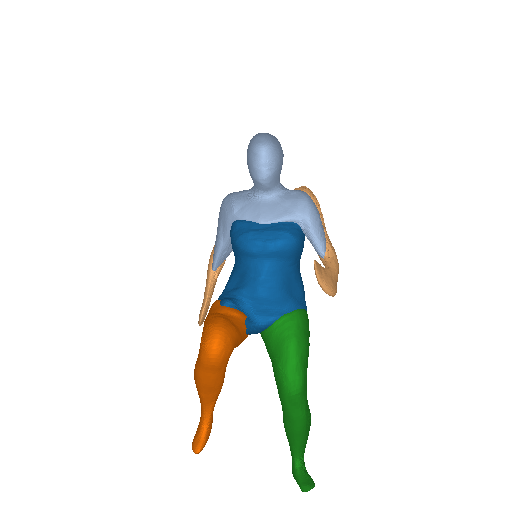}%
        \includegraphics[trim=100 0 100 0,clip,width=0.1\textwidth]{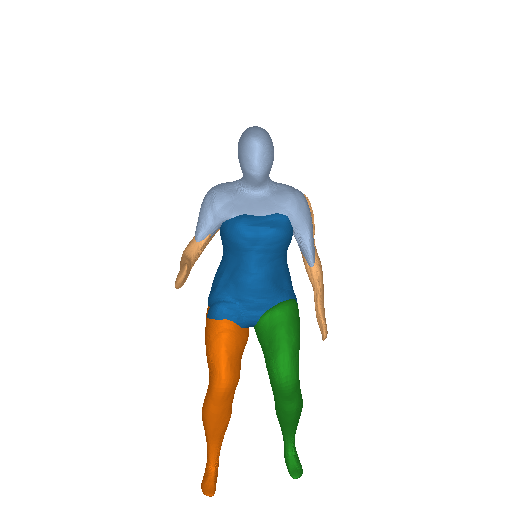}%
        \includegraphics[trim=100 0 100 0,clip,width=0.1\textwidth]{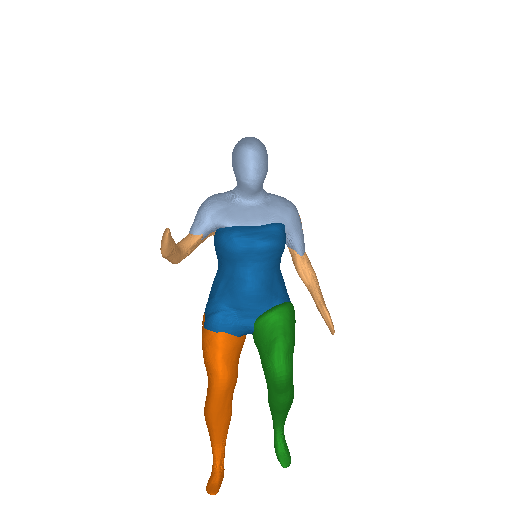}
    \end{subfigure}
    \begin{subfigure}[t]{\textwidth}
        \centering
        \includegraphics[trim=100 0 100 0,clip,width=0.1\textwidth]{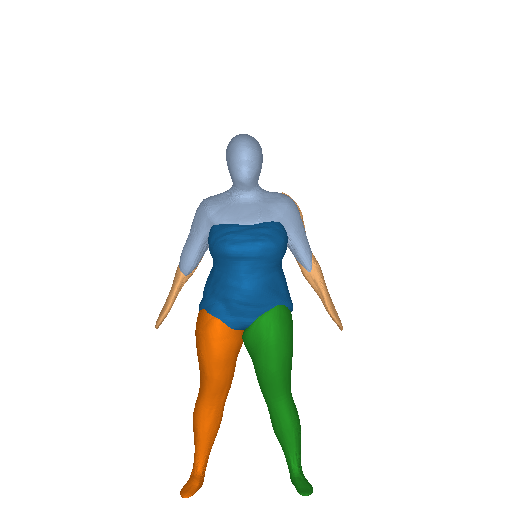}%
        \includegraphics[trim=100 0 100 0,clip,width=0.1\textwidth]{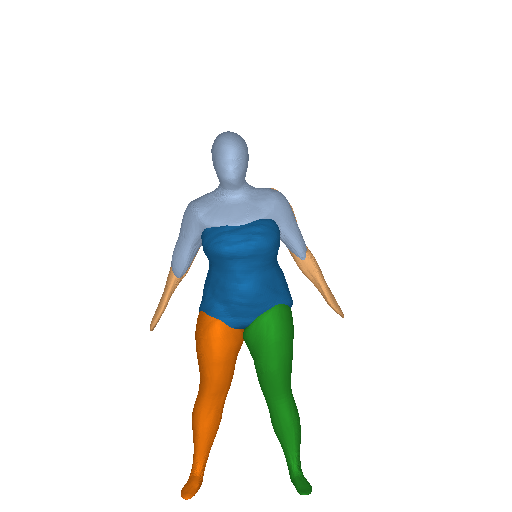}%
        \includegraphics[trim=100 0 100 0,clip,width=0.1\textwidth]{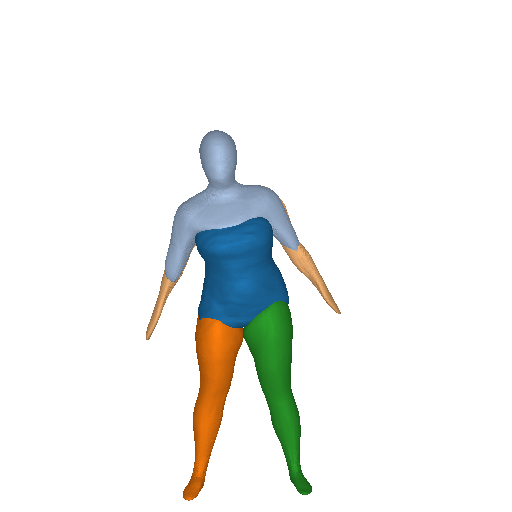}%
        \includegraphics[trim=100 0 100 0,clip,width=0.1\textwidth]{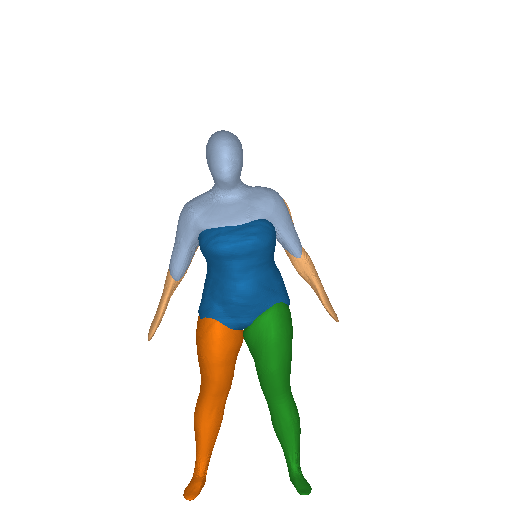}%
        \includegraphics[trim=100 0 100 0,clip,width=0.1\textwidth]{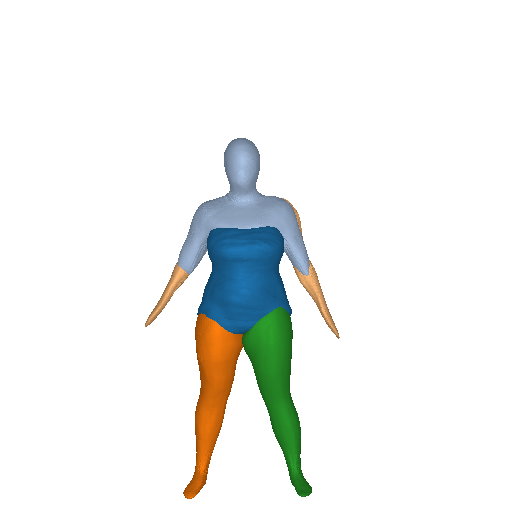}%
        \includegraphics[trim=100 0 100 0,clip,width=0.1\textwidth]{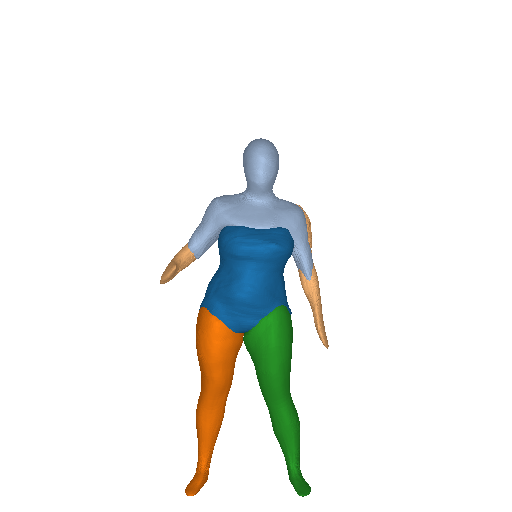}%
        \includegraphics[trim=100 0 100 0,clip,width=0.1\textwidth]{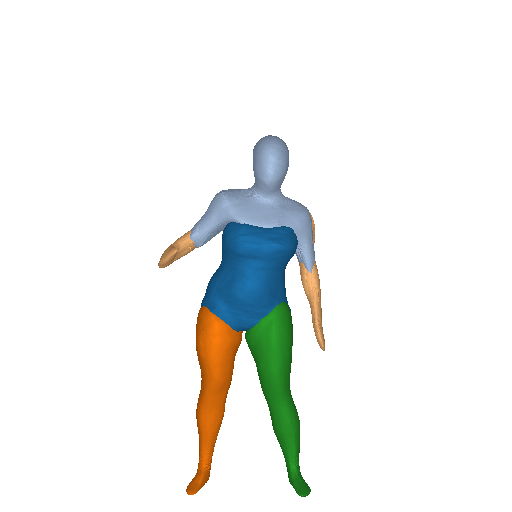}%
        \includegraphics[trim=100 0 100 0,clip,width=0.1\textwidth]{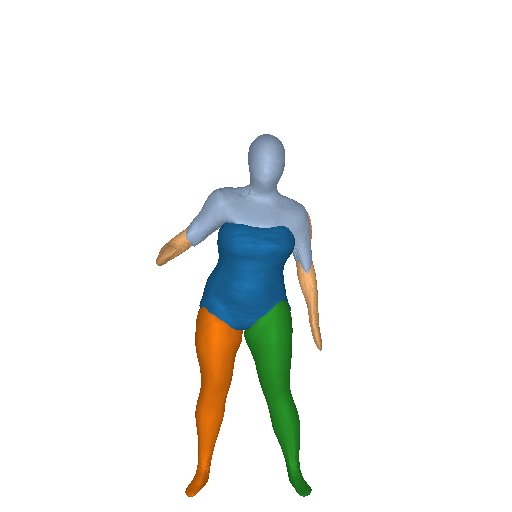}%
        \includegraphics[trim=100 0 100 0,clip,width=0.1\textwidth]{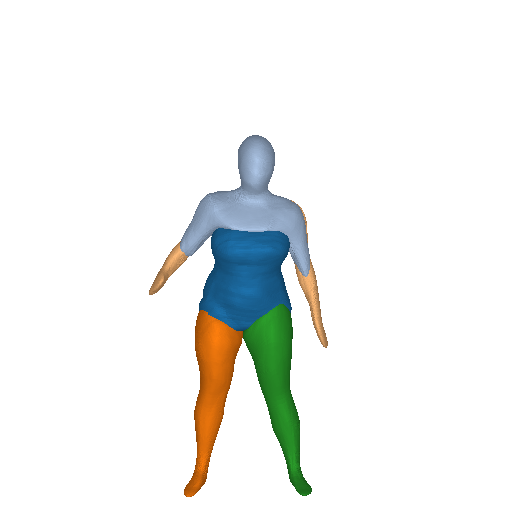}%
        \includegraphics[trim=100 0 100 0,clip,width=0.1\textwidth]{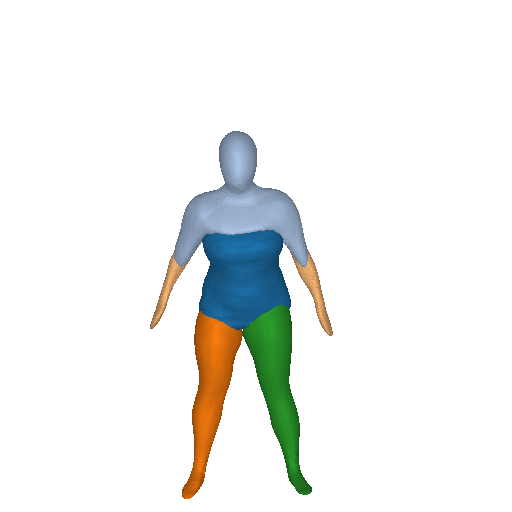}
    \end{subfigure}
    \begin{subfigure}[t]{\textwidth}
        \centering
        \includegraphics[trim=100 0 100 0,clip,width=0.1\textwidth]{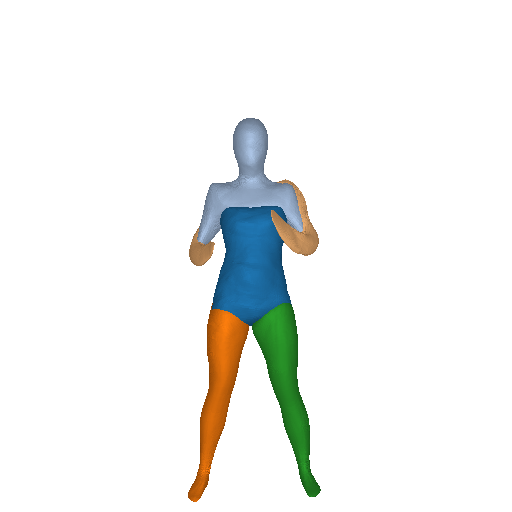}%
        \includegraphics[trim=100 0 100 0,clip,width=0.1\textwidth]{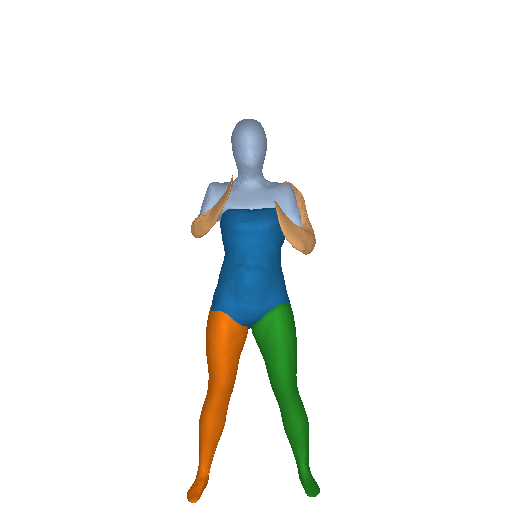}%
        \includegraphics[trim=100 0 100 0,clip,width=0.1\textwidth]{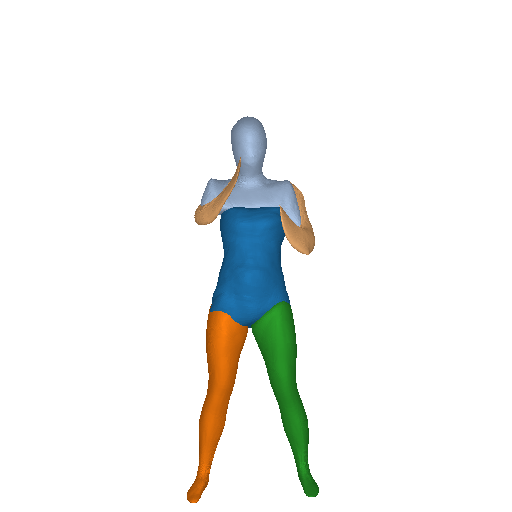}%
        \includegraphics[trim=100 0 100 0,clip,width=0.1\textwidth]{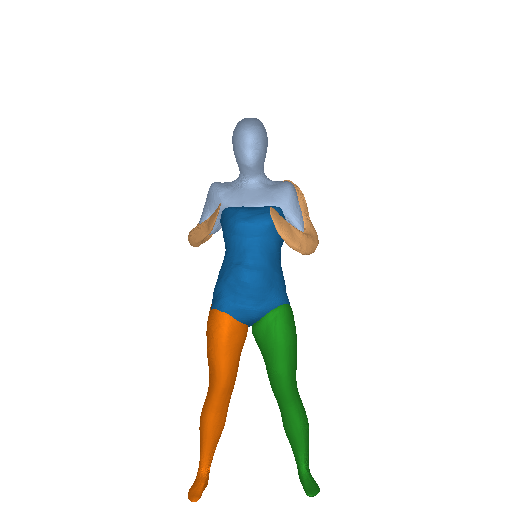}%
        \includegraphics[trim=100 0 100 0,clip,width=0.1\textwidth]{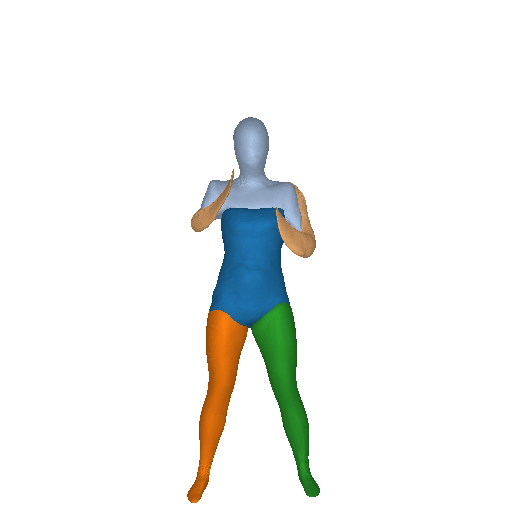}%
        \includegraphics[trim=100 0 100 0,clip,width=0.1\textwidth]{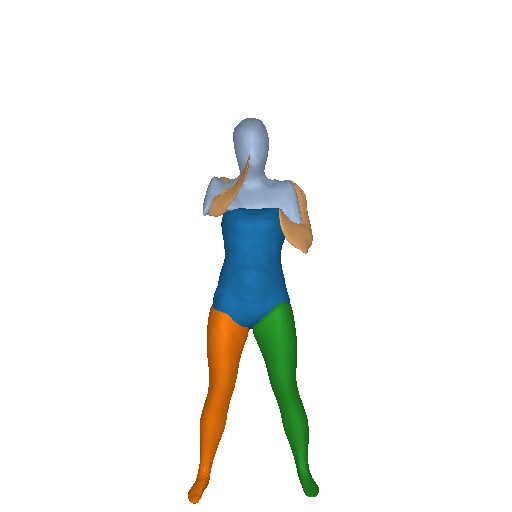}%
        \includegraphics[trim=100 0 100 0,clip,width=0.1\textwidth]{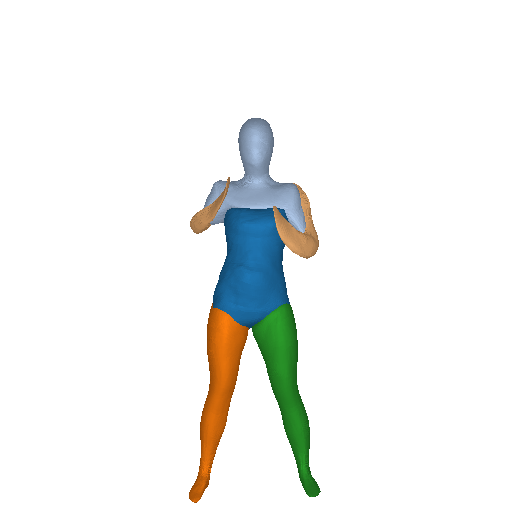}%
        \includegraphics[trim=100 0 100 0,clip,width=0.1\textwidth]{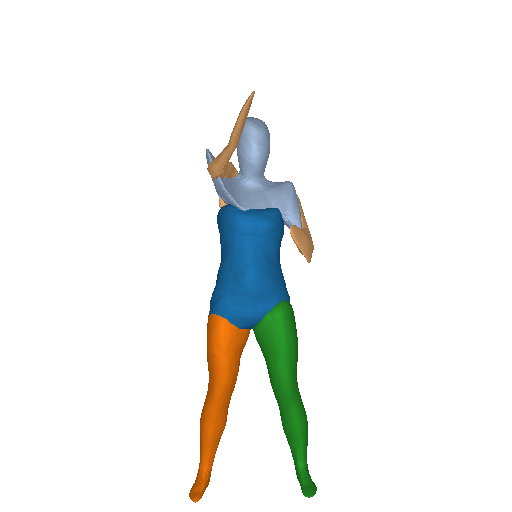}%
        \includegraphics[trim=100 0 100 0,clip,width=0.1\textwidth]{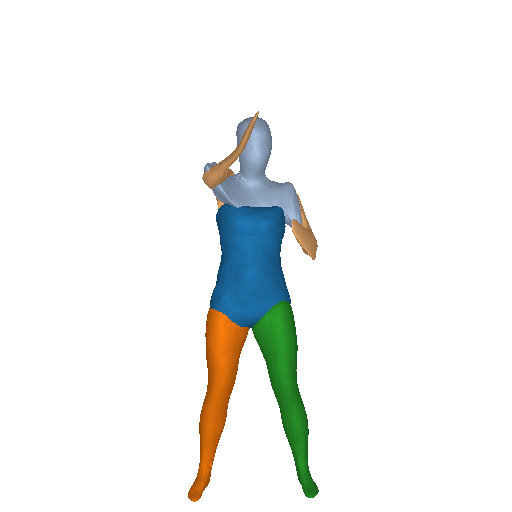}%
        \includegraphics[trim=100 0 100 0,clip,width=0.1\textwidth]{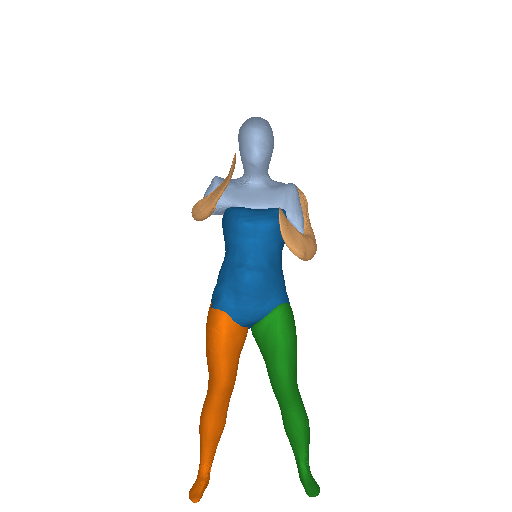}
    \end{subfigure}\\[0.3em]
    \includegraphics[width=1.0\textwidth]{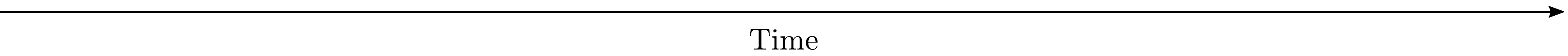}
    \caption{{\bf Temporal Consistency of Predicted Primitives.} We note that Neural Parts yield primitives that preserve their semantic identity while different humans perform various actions.}
    \label{fig:temporal_consistency}
\end{figure}

In this section, we provide a qualitative analysis on the temporal consistency
of the predicted primitives. Results are summarized in
\figref{fig:temporal_consistency}. We observe that Neural Parts consistently
use the same primitive for representing the same object part regardless of the
breadth of the part's motion. Notably, this temporal consistency is an emergent
property of our method and not one that is enforced with any kind of loss.
Additional visualizations are provided in the supplementary video.

\begin{figure}[H]
    \begin{subfigure}[t]{0.33\textwidth}
        \centering
        \includegraphics[width=0.6\textwidth]{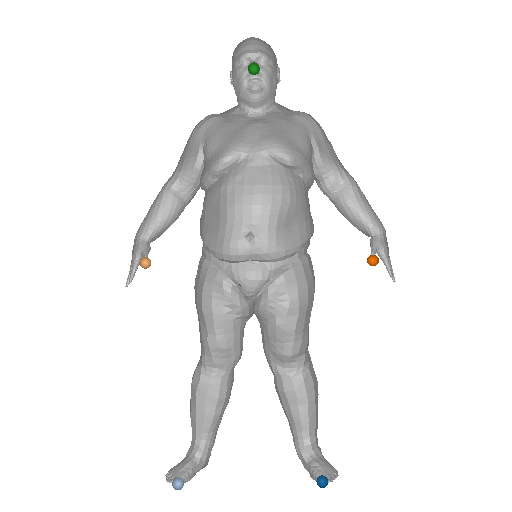}
    \end{subfigure}%
    \begin{subfigure}[t]{0.33\textwidth}
        \centering
        \includegraphics[width=0.6\textwidth]{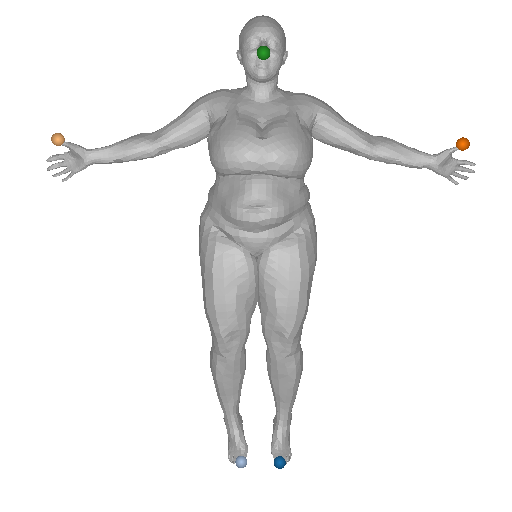}
    \end{subfigure}
    \begin{subfigure}[t]{0.33\textwidth}
        \centering
        \includegraphics[width=0.6\textwidth]{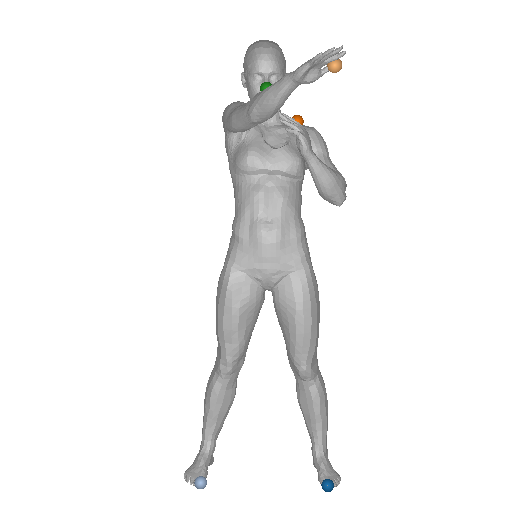}
    \end{subfigure}
    \caption{{\bf Semantic vertices provided by SMPL-X.} We highlight the 5 vertex indices that we use to identify left thumb (L-thumb), right thumb (R-thumb), left toe (L-toe), right toe (R-toe) and nose.}
    \label{fig:semantic_vertices}
\end{figure}

\subsection{Semantic Consistency}

In this experiment, we provide additional information regarding the Semantic
Consistency experiment in our main submission (see Fig. $10$). In particular, we evaluate whether a
specific human part is represented consistently by the same primitive (see
\figref{fig:semantic_consistency_supp}). To measure this quantitatively, we
select 5 representative vertices on each target mesh (the vertex indices are
provided by SMPL-X \cite{Pavlakos2019CVPR}) and measure the classification
accuracy of those points when using the label of the closest primitive. A
visualization of the $5$ vertices on the human body is given in
\figref{fig:semantic_vertices}. Note that the vertex indices are consistent
across all subjects because D-FAUST meshes are generated using
SMPL~\cite{Loper2015SIGGRAPH}.

\end{document}